\documentclass[10pt,twocolumn,letterpaper]{article}

\usepackage{arxiv}

\usepackage[dvipsnames]{xcolor}

\newcommand{\PAR}[1]{\vskip2pt \noindent{\bf #1~}}

\usepackage{makecell}
\usepackage{dsfont}
\usepackage{pifont}
\newcommand{\cmark}{\ding{51}}
\newcommand{\xmark}{\ding{55}}
\usepackage{mathtools}
\usepackage{multirow}
\usepackage{tikz}
\usetikzlibrary{calc, math}

\usepackage{algorithm}
\usepackage{algpseudocode}

\definecolor{arxivblue}{rgb}{0.21,0.49,0.74}
\usepackage[pagebackref,breaklinks,colorlinks,allcolors=arxivblue]{hyperref}

\usepackage[accsupp]{axessibility}

\makeatletter
\robustify\@latex@warning@no@line
\makeatother
\usepackage{authblk}
\makeatletter
\renewcommand\AB@affilsepx{, \protect\Affilfont}
\makeatother

\title{PBR-NeRF: Inverse Rendering with Physics-Based Neural Fields}

\author[1]{Sean Wu}
\author[2]{Shamik Basu}
\author[1]{Tim Br{\"o}dermann}
\author[1,3]{Luc Van Gool}
\author[1]{Christos Sakaridis}
\affil[1]{ETH Z\"urich}
\affil[2]{University Of Bologna}
\affil[3]{INSAIT, Sofia University St.~Kliment Ohridski}

\begin{document}
\maketitle

\begin{abstract}
We tackle the ill-posed inverse rendering problem in 3D reconstruction with a Neural Radiance Field (NeRF) approach informed by Physics-Based Rendering (PBR) theory, named PBR-NeRF.
Our method addresses a key limitation in most NeRF and 3D Gaussian Splatting approaches: they estimate view-dependent appearance without modeling scene materials and illumination.
To address this limitation, we present an inverse rendering (IR) model capable of jointly estimating scene geometry, materials, and illumination.
Our model builds upon recent NeRF-based IR approaches, but crucially introduces two novel physics-based priors that better constrain the IR estimation.
Our priors are rigorously formulated as intuitive loss terms and achieve state-of-the-art material estimation without compromising novel view synthesis quality.
Our method is easily adaptable to other inverse rendering and 3D reconstruction frameworks that require material estimation.
We demonstrate the importance of extending current neural rendering approaches to fully model scene properties beyond geometry and view-dependent appearance.
Code is publicly available at: \url{https://github.com/s3anwu/pbrnerf}.
\end{abstract}

\section{Introduction}
\label{sec:intro}

\begin{figure}[tb]
    \centering
    \includegraphics[width=\linewidth]{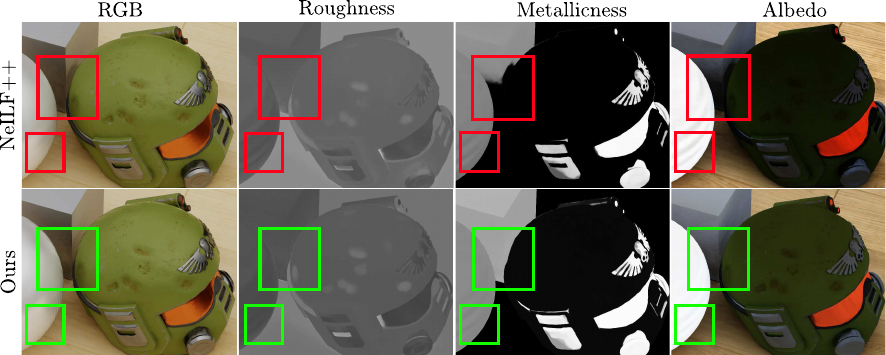}
    \vspace{-6mm}
    \caption{\textbf{Improving inverse rendering with physics-based priors.} The proposed PBR-NeRF significantly outperforms the NeILF++ \cite{zhang2023neilf++} baseline simply by using our novel Conservation of Energy and NDF-weighted Specular Losses. Our physics-based losses correct ``baked-in'' specular highlights misrepresented in the diffuse lobe (highlighted areas) by (1) enforcing energy conservation and (2) accurately separating specular and diffuse reflections. The result is a more realistic, physically consistent material and lighting estimation.}
    \label{fig:fun_teaser}
    \vspace{-7mm}
\end{figure}

Recent advances in neural rendering and 3D reconstruction using Neural Radiance Fields (NeRF)~\cite{mildenhall2020nerf} and 3D Gaussian Splatting (3DGS)~\cite{kerbl3Dgaussians} have driven massive research interest.
However, reconstructing accurate scene properties from posed multi-view images remains a challenging, open problem due to inherent ambiguities.
The material-lighting ambiguity makes the inverse rendering problem fundamentally ill-posed, with identical images being explainable by infinitely many tuples of scene materials, illumination, and geometry.
Addressing this ill-posed inverse problem requires strong priors to narrow the possible solution space.

To overcome this issue, we leverage Physics-Based Rendering (PBR) theory~\cite{kajiya1986rendering,pharr2016physically} from computer graphics.
PBR provides a framework for generating physically accurate images given known scene materials, illumination, and geometry.
By reversing this process, we derive physics-based priors to better constrain neural forward and inverse rendering and enhance estimation robustness.
More specifically, while NeRF, 3DGS, and their neural field derivatives achieve state-of-the-art performance in modeling scene geometry and view-dependent appearance, they do so by treating scenes as ``black boxes'' that ignore the underlying physics of light transport.
These methods represent light by volume-rendering millions of translucent particles, conditioning each particle's emitted radiance on its position and viewing direction.
This formulation significantly improves expressiveness over traditional methods \cite{agarwal2011building,schoenberger2016sfm}, but it does not guarantee physically accurate results.
For example, reflective surfaces may be improperly modeled, causing artifacts such as ``baked-in'' specular highlights, where particles emit more light in outgoing directions than physically possible.
Ultimately, the performance of NeRF and 3DGS in 3D reconstruction is inherently limited by their inability to accurately and consistently model diverse light interactions.

In this paper, we solve the complete inverse rendering problem by jointly estimating the geometry, illumination, and materials to accurately model view-dependent appearance.
We extend the NeILF/NeILF++~\cite{yao2022neilf,zhang2023neilf++} neural fields framework, combining the expressiveness of deep neural networks with the theoretical guarantees of PBR theory.
Specifically, our PBR-NeRF approach improves NeILF/NeILF++'s reflection model by leveraging its underlying
Disney Bidirectional Reflectance Distribution Function (BRDF) model~\cite{burley2012physically}.
Our method enforces energy conservation and promotes the disentanglement of the diffuse and specular lobes of the BRDF, mitigating issues such as highlights ``baked in'' the diffuse albedo that are commonly observed in NeRF and 3DGS reconstructions.

In particular, we achieve these goals through two novel physics-based losses applied directly to the Disney BRDF.
Our first loss enforces energy conservation by penalizing physically invalid BRDFs that reflect more energy than received.
Our second loss promotes the separation of specular and diffuse BRDF lobes, which is crucial for accurately modeling highly specular surfaces. Without constraints, the diffuse lobe (which defines the albedo) tends to overcompensate near symmetric reflection angles, where both lobes contribute to the aggregate BRDF.
By penalizing the diffuse lobe's magnitude, weighted by the normal distribution function to target specular angles, this loss indirectly encourages the \emph{specular lobe} to expand and fully ``explain'' specular highlights effectively.

We thoroughly validate PBR-NeRF on two major benchmarks for inverse rendering and novel view synthesis: the NeILF++~\cite{zhang2023neilf++} and DTU~\cite{jensen2014dtu} datasets. Our experiments show that PBR-NeRF sets the new state of the art for material estimation in inverse rendering, while maintaining or surpassing the competitive novel view synthesis performance of its baseline. Crucially, our model delivers faithful albedo estimates that remain consistent for a given scene across diverse illuminations, thanks to its effective diffuse-specular decomposition.
This decomposition also improves metallicness and roughness estimates as an additional benefit, further highlighting the effectiveness of our approach.

\section{Related Work}
\label{sec:formatting}

\PAR{Neural fields for novel view synthesis} have achieved impressive progress by representing 3D scenes as continuous volumetric fields~\cite{mildenhall2020nerf}.
This family of methods achieves photorealistic synthesis from unseen viewpoints through differentiable volume rendering, which estimates outgoing radiance by alpha-compositing color over sampled points along each camera ray. We concisely review representative works here and refer the reader to~\cite{xie2022neural} for a comprehensive overview of the associated literature.
Scene geometry is commonly represented using density fields~\cite{mildenhall2020nerf,zhang2020nerf++,barron2021mip,barron2022mip,muller2022instant,barron2023zip,xu2024murf,rebain2021derf}, 3D Gaussians~\cite{kerbl3Dgaussians,gao2023relightable,wang2024specgaussian}, or signed distance fields (SDFs)~\cite{yariv2021volsdf,yariv2020multiview,wang2021neus,zhang2021learning,zhang2022critical} for images and other optical modalities~\cite{borts2024radar,huang2023neural,rudnev2023eventnerf}.
While the density fields in NeRF and 3D Gaussians can model arbitrary complex scene geometries, they fundamentally lack well-defined surfaces and normals.
These particle-centric representations result in ambiguous surface definitions, which hinders their applicability to PBR.
In contrast, SDFs explicitly define surfaces as the zero-level set of a neural network, providing accurate surface and normal estimation.
This surface-centric representation makes SDFs inherently more compatible with inverse rendering tasks which rely on PBR and require precise surface definitions.

\begin{figure*}[tb]
    \centering
    \includegraphics[width=\textwidth]{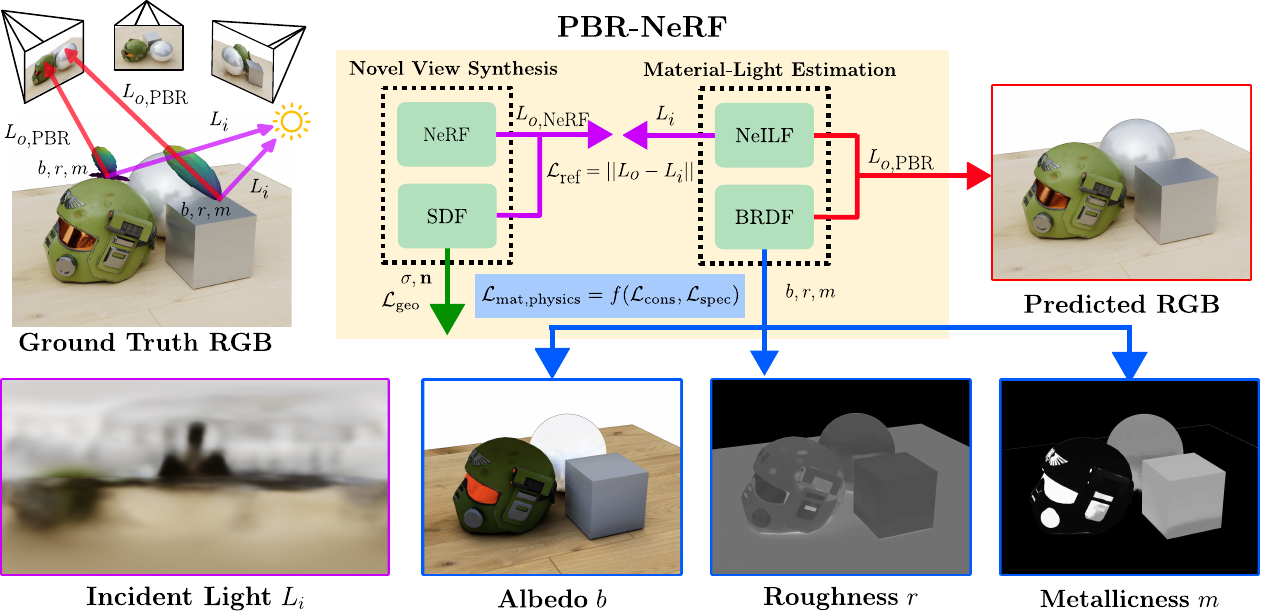}
    \vspace{-7mm}
    \caption{\textbf{Overview of our PBR-NeRF architecture for neural inverse rendering.} Our two novel physics-based material losses, i.e.\ $\mathcal{L}_{\text{cons}}$ for conservation of energy and $\mathcal{L}_{\text{spec}}$ for disentangling the diffuse and specular BRDF components, are derived in Sec.~\ref{sec:pbr_losses}. The complete PBR-NeRF model comprises multiple neural fields, which are optimized in a stage-wise fashion: a standard NeRF+SDF modeling radiance and geometry, a neural incident light field (NeILF) modeling spatially-varying illumination, and a BRDF field modeling materials via the Disney BRDF~\cite{burley2012physically}. Our physics-based losses provide the BRDF field with valuable inductive biases for improved material estimation, which help resolve to a large extent the inherent material-lighting ambiguity in inverse rendering and thus benefit the incident light field as well. Consequently, novel scene views synthesized with our model enjoy state-of-the-art quality.}
    \label{fig:pbr-nerf-arch-diagram}
    \vspace{-5mm}
\end{figure*}

\PAR{Neural inverse rendering} predates neural fields~\cite{sengupta2019neural,li2020inverse,wang2021learning,yu2019inverserendernet,sang2020single,lyu2023diffusion} and learns to predict 3D geometry, materials, and illumination from 2D images. Decomposed neural fields~\cite{verbin2022ref,wang2023neural,ye2023intrinsicnerf,ramazzina2023scatternerf} similarly extend conventional neural fields with inverse rendering or intrinsic decomposition to estimate physical scene properties such as materials and illumination.
Compared to NeRF and 3DGS, the above additional material and lighting estimation enables complex downstream tasks such as relighting~\cite{xu2023renerf,rudnev2022nerf,lyu2022neural}, material editing, and appearance manipulation.
The decomposition of reflected radiance into materials and illumination can also address issues such as ``baked-in'' highlights in predicted colors by properly modeling view-dependent effects.
However, inverse rendering introduces significant complexity due to the material-lighting-geometry ambiguity, which is harder to lift than the geometry-radiance ambiguity of standard novel view synthesis.
Many approaches simplify this task by assuming known lighting, environment map lighting~\cite{zhang2021physg,zhang2021nerfactor,Munkberg2022nvdiffrec,boss2021nerd,boss2021neural}, known geometry~\cite{yao2022neilf}, or basic Lambertian material models.
State-of-the-art inverse rendering approaches jointly model (1) direct and indirect lighting from near and far sources \cite{yao2022neilf,zhang2023neilf++,wu2023nefii}, and (2) spatially-varying microfacet BRDFs~\cite{srinivasan2021nerv,yao2022neilf,zhang2023neilf++}.
For computational efficiency, techniques such as light caching~\cite{zhang2022modeling,yao2022neilf,liu2023nero,zhang2023neilf++,jin2023tensoir,attal2024flash} and incident light sampling~\cite{yao2022neilf,zhang2023neilf++,attal2024flash} are crucial to avoid costly recursive light tracing.

NeILF~\cite{yao2022neilf} proposed joint incident light and BRDF fields, improving on environment-map-based approaches such as NeRV~\cite{srinivasan2021nerv} by handling near-field lights and mixed lighting settings in addition to indirect lighting and occlusions.
Building on this, NeILF++~\cite{zhang2023neilf++} introduced inter-reflectable light fields, combining the NeILF and BRDF fields with an SDF for geometry estimation.
This approach uses the SDF's volume-rendered outgoing radiance to supervise the incident radiance predicted by the NeILF.
The reflection loss of~\cite{zhang2023neilf++} harnesses the SDF's strong novel view synthesis performance to steer the ill-posed inverse rendering, achieving state-of-the-art material estimation among neural inverse rendering works.
While largely effective, current material estimation methods including~\cite{yao2022neilf,zhang2023neilf++} lack informative material priors for disambiguating specular from diffuse properties of scene surfaces.
This typically results in incorrectly attributing specular highlights to alleged changes in diffuse albedo, even though such errors do not affect view synthesis results.

\PAR{Material priors for inverse rendering} are crucial for constraining material estimation and reducing material-lighting ambiguities.
BRDF smoothness priors ~\cite{yao2022neilf,zhang2023neilf++,jin2023tensoir,attal2024flash} can discourage abrupt spatial changes in material properties.
NeILF~\cite{yao2022neilf} and NeILF++~\cite{zhang2023neilf++} use a Lambertian prior to regularize the Disney BRDF's metallicness and roughness parameters.
However, the Lambertian prior induces a strong assumption of perfectly rough, non-metallic materials that limits BRDF expressiveness.
For instance, specular effects often appear in the predicted BRDF's diffuse lobe due to the Lambertian prior suppressing the specular lobe.
We address these limitations by proposing two novel physics-based losses that enhance the physical validity and accuracy of estimated BRDFs without sacrificing expressiveness.

\section{Method}
We present an overview of our PBR-NeRF method with our novel Conservation of Energy and NDF-weighted Specular losses in Fig.~\ref{fig:pbr-nerf-arch-diagram}.

\subsection{Background}
\label{sec:idr}

We build upon the NeILF++~\cite{zhang2023neilf++} implicit differential renderer (IDR) as the foundation for our physics-based contributions.
The IDR estimates outgoing radiance $L_o(\mathbf{x}, \boldsymbol{\omega}_o)$ by solving the Rendering Equation~\cite{kajiya1986rendering}:
\begin{equation}
    \label{eq:rendering_eqn}
    L_o(\mathbf{x}, \boldsymbol{\omega}_o) = \int_\Omega f_r(\mathbf{x}, \boldsymbol{\omega}_i, \boldsymbol{\omega}_o) L_i(\mathbf{x}, \boldsymbol{\omega}_i) (\boldsymbol{\omega}_i \cdot \mathbf{n}) d\boldsymbol{\omega}_i,
\end{equation}
where $L_o(\mathbf{x}, \boldsymbol{\omega}_o)$ is the outgoing radiance from a surface point $\mathbf{x}$ in direction $\boldsymbol{\omega}_o$, $f_r(\mathbf{x}, \boldsymbol{\omega}_i, \boldsymbol{\omega}_o)$ is the BRDF, and integration is performed over incident directions $\boldsymbol{\omega}_i$ in the positive hemisphere $\Omega$.

To approximate the BRDF in~\eqref{eq:rendering_eqn}, we use the simplified Disney BRDF model~\cite{burley2012physically} with three spatially varying parameters: albedo (base color) $b \in [0,1]^3$, metallicness $m \in [0,1]$, and roughness $r \in [0,1]$. The Disney BRDF is decomposed into a diffuse lobe $f_d(\mathbf{x})$ and a specular lobe $f_s(\mathbf{x}, \boldsymbol{\omega}_i, \boldsymbol{\omega}_o)$:
\begin{equation}
    \label{eq:brdf}
    f_r(\mathbf{x}, \boldsymbol{\omega}_i, \boldsymbol{\omega}_o) = f_d(\mathbf{x}) + f_s(\mathbf{x}, \boldsymbol{\omega}_i, \boldsymbol{\omega}_o).
\end{equation}
The diffuse lobe $f_d$ models the view-independent appearance based on the albedo $b$ and metallicness $m$ as
\begin{equation}
    \label{eq:diffuse_brdf}
    f_d(\mathbf{x}) = \frac{1-m(\mathbf{x})}{\pi} b(\mathbf{x}).
\end{equation}
The specular lobe $f_s$ models light $L_o$ reflected in direction $\boldsymbol{\omega}_o$ due to incident radiance $L_i$ from direction $\boldsymbol{\omega}_i$ being reflected across the halfway vector $\boldsymbol{\omega}_h = \frac{\boldsymbol{\omega}_o + \boldsymbol{\omega}_i}{||\boldsymbol{\omega}_o + \boldsymbol{\omega}_i||}$.
The Disney BRDF approximates rough surface effects in the specular lobe using the microfacet BRDF model~\cite{burley2012physically}.
This specular lobe of the BRDF is expressed as
\begin{equation}
    \label{eq:specular_brdf}
    f_s(\mathbf{x}, \boldsymbol{\omega}_o, \boldsymbol{\omega}_i) = \frac{D(\boldsymbol{\omega}_h) F(\boldsymbol{\omega}_o, \boldsymbol{\omega}_h) G(\boldsymbol{\omega}_i, \boldsymbol{\omega}_o, \mathbf{n})}{4(\mathbf{n}\cdot \boldsymbol{\omega}_i)(\mathbf{n}\cdot \boldsymbol{\omega}_o)},
\end{equation}
where the normal distribution function (NDF) $D$ models the distribution of microfacet orientations, the Fresnel term $F$ models Fresnel reflections, and $G$ models geometric occlusion.
We adopt the Spherical Gaussian approximation from~\cite{zhang2021physg,yao2022neilf,zhang2023neilf++}, in which roughness $r$ controls the sharpness of the NDF through
\begin{equation}
    D(\boldsymbol{\omega}_h) = \frac{1}{\pi r^4}\exp\left(\frac{2}{r^4}(\boldsymbol{\omega}_h\cdot \mathbf{n} -1)\right).
\end{equation}

Following NeILF++~\cite{zhang2023neilf++}, our IDR framework uses three implicit networks (cf.\ Fig.~\ref{fig:pbr-nerf-arch-diagram}):
\begin{enumerate}
    \item \textbf{BRDF MLP}: predicts the Disney BRDF parameters $b$, $r$, and $m$ to compute the BRDF $f_r(\mathbf{x}, \boldsymbol{\omega}_o, \boldsymbol{\omega}_i)$.
    \item \textbf{NeILF MLP}: predicts incident radiance $L_i(\mathbf{x}, \boldsymbol{\omega}_i)$.
    \item \textbf{NeRF SDF}: predicts density $\sigma$ (geometry) and color $c$, where $c$ estimates the outgoing radiance $L_o(\mathbf{x}, \boldsymbol{\omega}_o)$.
\end{enumerate}
We approximate the rendering integral in~\eqref{eq:rendering_eqn} according to NeILF/NeILF++~\cite{yao2022neilf,zhang2023neilf++} using a fixed set $S_L$ of incident directions $\boldsymbol{\omega}_i$ generated by Fibonacci sampling:
\begin{equation}
    \label{eq:neilfpp_rendering_eqn}
    L_o(\mathbf{x},\boldsymbol{\omega}_o) = \frac{2\pi}{|S_L|} \sum_{\boldsymbol{\omega}_i \in S_L} f_r(\mathbf{x}, \boldsymbol{\omega}_i, \boldsymbol{\omega}_o) L_i(\mathbf{x}, \boldsymbol{\omega}_i) (\boldsymbol{\omega}_i \cdot \mathbf{n}).
\end{equation}
To simplify notation in the following sections, we omit the arguments $\mathbf{x}$, $\boldsymbol{\omega}_i$, and $\boldsymbol{\omega}_o$ from $f_r$, $f_d$, $f_s$, and $L_i$.

\subsection{Physics-Based Losses}
\label{sec:pbr_losses}
\begin{figure}
    \centering
    \includegraphics[width=\linewidth]{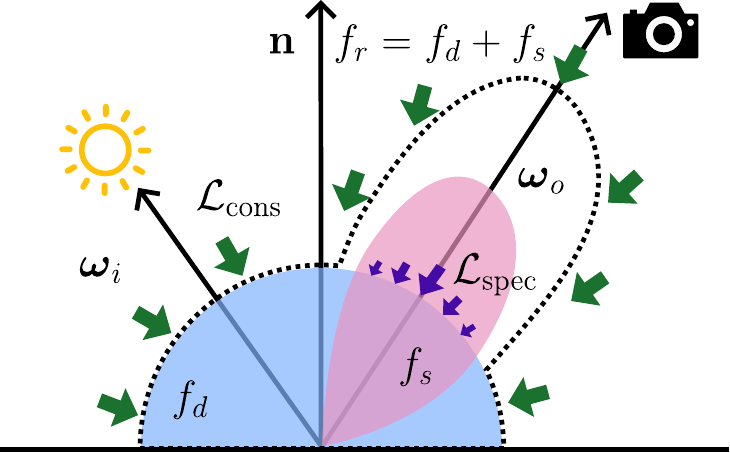}
    \vspace{-6mm}
    \caption{\textbf{Illustration of our physics-based losses}. We constrain our Disney BRDF~\cite{burley2012physically} material and NeILF~\cite{yao2022neilf,zhang2023neilf++} incident light estimation with two novel physics-based losses: (1) the Conservation of Energy Loss $\mathcal{L}_\text{cons}$ to supervise the complete BRDF $f_r = f_s + f_d$ denoted by the dotted envelope, and (2) the NDF-weighted Specular Loss $\mathcal{L}_\text{spec}$ to adjust the relative magnitudes of the specular $f_s$ (red) and diffuse $f_d$ (blue) BRDF lobes.}
    \label{fig:pbrnerf_novel_losses}
    \vspace{-5mm}
\end{figure}

Since inverse rendering is ill-posed, we propose two novel losses inspired by physical priors to constrain both material and lighting estimation, which are illustrated in Fig.~\ref{fig:pbrnerf_novel_losses}.

\PAR{Conservation of Energy Loss.}
Our first contribution addresses the non-energy-conserving behavior inherent in the Disney BRDF~\cite{burley2012physically} and other microfacet models.
Despite their widespread use, these models fail to conserve energy, as demonstrated in previous works~\cite{heitz2014understanding,burley2015extending}.
This limitation poses a critical challenge for inverse rendering, as it permits a material to create energy (reflect more light than received) or destroy energy (reflect too little light).
Such inaccuracies skew material estimates and disrupt lighting estimation, causing the estimated illumination to overcompensate and hence appear too bright or dark.
These effects cascade to downstream tasks, such as relighting and object insertion, and pose a risk for them.

To mitigate energy creation issues, we enforce the energy conservation property for the BRDF by requiring ${\int_\Omega f_r(\boldsymbol{\omega}_i \cdot \mathbf{n}) d\boldsymbol{\omega}_i \leq 1}$.
This constraint, derived from the rendering equation~\eqref{eq:rendering_eqn}, requires that the sum of reflected radiance weights $f_r(\boldsymbol{\omega}_i \cdot \mathbf{n})$ across all incident directions in the hemisphere $\Omega$ must not exceed unity.
By enforcing this intrinsic physical property, independent of incident lighting, we ensure physically correct materials.

We reformulate this constraint in our discretized setting as the Conservation of Energy Loss:
\begin{equation}
    \label{eq:cons_loss}
    \mathcal{L}_\text{cons} = \max\bigg\{\bigg(\frac{2\pi}{|S_L|}\sum_{\boldsymbol{\omega}_i \in S_L} f_r (\boldsymbol{\omega}_i \cdot \mathbf{n})\bigg) -1, 0\bigg\}.
\end{equation}
This ReLU-style formulation prevents overestimation of reflected light, allowing reflected irradiance to vary but without exceeding the incident irradiance.
We only penalize the model when the sum of weights $f_r (\boldsymbol{\omega}_i \cdot \mathbf{n})$ exceeds 1, but any value below can freely vary.
As seen in Fig.~\ref{fig:pbrnerf_novel_losses}, the Conservation of Energy Loss affects both specular and diffuse lobes, ensuring that the overall BRDF $f_r$ conserves energy.

\PAR{NDF-weighted Specular Loss.}
Our second physics-based loss targets the imbalance between the diffuse and specular lobes frequently observed in inverse rendering methods, such as NeILF++~\cite{zhang2023neilf++}, which often assume Lambertian reflection.
Real-world materials, however, often violate idealized Lambertian behavior, resulting in ``baked-in'' specular highlights where the predicted diffuse lobe $f_d$ overcompensates for insufficient specular reflection.
For these ``baked-in'' specular highlights, the aggregate BRDF value $f_r$ may be accurate for specular directions, but the imbalance between $f_d$ and $f_s$ causes incorrect diffuse behavior at non-specular angles, degrading material estimation quality.

To correct this imbalance, we propose the NDF-weighted Specular Loss to penalize excessive diffuse reflection in specular regions and we define our loss as
\begin{align}
    \label{eq:spec_loss}
    &\mathcal{L}_\text{spec} = \notag \\
    &\frac{1}{|S_L|}\sum_{\boldsymbol{\omega}_i \in S_L} \operatorname{softmax}\left(\frac{\operatorname{detach}\left(D\left(\frac{\boldsymbol{\omega}_o + \boldsymbol{\omega}_i}{||\boldsymbol{\omega}_o + \boldsymbol{\omega}_i||}\right)\right)}{T_\text{spec}}\right) f_d,
\end{align}
where $D$ is the NDF evaluated using the halfway vector $\frac{\boldsymbol{\omega}_o + \boldsymbol{\omega}_i}{||\boldsymbol{\omega}_o + \boldsymbol{\omega}_i||} = \boldsymbol{\omega}_h$.
By using a softmax-weighted NDF with temperature $T_\text{spec}$, we selectively penalize diffuse reflections in regions where specular effects dominate.
We detach the NDF term from the computation graph to prevent gradient flow through $D(\boldsymbol{\omega}_h)$ during backpropagation, ensuring that only diffuse contributions to $\mathcal{L}_{\text{spec}}$ result in weight updates.

Our NDF-weighted Specular Loss complements the standard RGB rendering loss by pushing the specular lobe $f_s$ to compensate for the penalized diffuse lobe $f_d$, creating a stronger separation between diffuse and specular reflections.
A proper weighting of the opposing RGB rendering loss and other regularization terms ensures that the diffuse lobe is not arbitrarily suppressed by our specular loss.
As with $\mathcal{L}_{\text{cons}}$, the specular loss also benefits lighting estimation.
An oversized diffuse lobe can lead to underestimated light intensities, and excessive diffuse reflection can also force the estimated incident light to incorrectly reproduce complex, view-dependent specular effects.
Together, our two losses $\mathcal{L}_{\text{cons}}$ and $\mathcal{L}_{\text{spec}}$ significantly improve the accuracy of material and lighting estimation, resulting in more realistic scene reconstructions, as we detail in Sec.~\ref{sec:exp}.

\subsection{Joint Optimization}
\label{sec:joint_mat_illum_geo_opt}
Following NeILF++~\cite{zhang2023neilf++}, we optimize the scene reconstruction in three phases.
We briefly summarize the three phases here (See Appendix Sec.~\ref{sec:supp:joint_optimization} for full details):
\begin{enumerate}
    \item \textbf{Geometry}: only train NeRF SDF to initialize estimated geometry using a geometry-based loss $\mathcal{L}_\text{geo}$.
    \item \textbf{Material}: train NeILF and BRDF MLPs while freezing the NeRF SDF to initialize estimated illumination and materials with a material-based loss $\mathcal{L}_\text{mat}$.
    \item \textbf{Joint Optimization}: train all fields (NeRF SDF, NeILF, BRDF) with all losses $\mathcal{L} = \mathcal{L}_\text{geo} + \mathcal{L}_\text{mat}$.
\end{enumerate}
Each phase uses different losses that encode specific priors or constraints relevant to that phase.
The full material-based loss including our complete physics-based loss $\mathcal{L}_\text{mat,physics}$ is
\vspace{-3mm}
\begin{equation}
    \label{eq:mat_loss}
    \mathcal{L}_\text{mat} = \lambda_\text{pbr}\mathcal{L}_\text{pbr} + \lambda_\text{ref}\mathcal{L}_\text{ref} + \lambda_\text{smth}\mathcal{L}_\text{smth} + \mathcal{L}_\text{mat,physics},
\end{equation}
where $\mathcal{L}_\text{mat,physics} = \lambda_\text{cons}\mathcal{L}_\text{cons} + \lambda_\text{spec}\mathcal{L}_\text{spec}$ includes our Conservation of Energy Loss and our NDF-weighted Specular Loss, $\mathcal{L}_\text{pbr}$ is a standard RGB rendering loss supervising the estimated outgoing radiance from~\eqref{eq:neilfpp_rendering_eqn}, $\mathcal{L}_\text{ref}$ is the NeILF++ reflection loss \cite{zhang2023neilf++}, $\mathcal{L}_\text{smth}$ is the NeILF/NeILF++ BRDF smoothness loss~\cite{yao2022neilf,zhang2023neilf++}, and $\lambda_\text{pbr},\lambda_\text{ref},\lambda_\text{smth},\lambda_\text{cons},\lambda_\text{spec}$ are positive weights.

\section{Experiments}
\label{sec:exp}

\begin{table*}[h]
    \centering
    \caption{\textbf{Comparison against state-of-the-art methods on the NeILF++ dataset~\cite{zhang2023neilf++}.} Material estimation and novel view synthesis are compared across different scenes (City, Studio, and Castel) and types of illumination (Env: global environment map, Mix: mixed lighting with environment map, point, and area light sources) using PSNR and SSIM. \dag: our reproduced results. N/A: no official reported results.}
    \label{tab:neilfpp_dataset_sota}
    \vspace{-3mm}
    \resizebox{\textwidth}{!}{
    \begin{tabular}{l|l|cccccc|cccccc|cc}
        \specialrule{.2em}{.1em}{.1em}
        \multirow{2}{*}{\shortstack[l]{Predicted \\ Quantity}} & \multirow{2}{*}{Method} & \multicolumn{2}{c}{Env-City} & \multicolumn{2}{c}{Env-Studio} & \multicolumn{2}{c|}{Env-Castel} & \multicolumn{2}{c}{Mix-City} & \multicolumn{2}{c}{Mix-Studio} & \multicolumn{2}{c|}{Mix-Castel} & \multicolumn{2}{c}{Mean} \\
        & & PSNR & SSIM & PSNR & SSIM & PSNR & SSIM & PSNR & SSIM & PSNR & SSIM & PSNR & SSIM & PSNR & SSIM \\
        \hline
        \multirow{2}{*}{RGB}
        & PhySG~\cite{zhang2021physg} & 24.82 & N/A & 25.65 & N/A & 27.24 & N/A & 24.38 & N/A & 24.04 & N/A & 25.81 & N/A & 25.32 & N/A \\
        & SG-ENV~\cite{zhang2021physg} & 31.01 & N/A & 29.46 & N/A & 32.34 & N/A & 27.20 & N/A & 25.88 & N/A & 27.70 & N/A & 28.93 & N/A \\
        & NeILF~\cite{yao2022neilf} \dag & \textbf{33.23} & 92.29 & \textbf{31.12} & \textbf{83.55} & \textbf{36.21} & \textbf{94.13} & 30.00 & 86.95 & 27.56 & 74.20 & 30.91 & 89.83 & \textbf{31.50} & 86.83 \\
        & NeILF++~~\cite{zhang2023neilf++}\dag & 31.53 & 91.51 & 29.77 & 80.15 & 33.87 & 93.26 & 29.64 & 88.83 & 27.50 & 74.70 & 30.74 & 91.07 & 30.51 & 86.59 \\
        & PBR-NeRF (ours) & 32.60 & \textbf{92.33} & 30.96 & 80.77 & 34.88 & 93.66 & \textbf{30.03} & \textbf{89.17} & \textbf{27.84} & \textbf{75.01} & \textbf{31.34} & \textbf{91.39} & 31.27 & \textbf{87.05} \\
        \hline
        \multirow{2}{*}{Roughness}
        & PhySG~\cite{zhang2021physg} & 6.62 & N/A & 11.29 & N/A & 6.22 & N/A & 6.27 & N/A & 6.83 & N/A & 6.14 & N/A & 7.23 & N/A \\
        & SG-ENV~\cite{zhang2021physg} & 9.61 & N/A & 17.64 & N/A & 9.74 & N/A & 8.77 & N/A & 12.58 & N/A & 9.14 & N/A & 11.25 & N/A \\
        & NeILF~\cite{yao2022neilf}\dag & 16.71 & 86.77 & 17.72 & 86.48 & 18.05 & 89.56 & 15.11 & 81.42 & 14.54 & 78.53 & 15.04 & 81.28 & 16.19 & 84.01 \\
        & NeILF++~~\cite{zhang2023neilf++}\dag & 21.22 & 91.27 & 22.01 & 91.89 & 19.91 & 90.63 & \textbf{22.19} & 91.57 & \textbf{23.43} & 92.37 & \textbf{22.20} & 91.61 & 21.83 & 91.56 \\
        & PBR-NeRF (ours) & \textbf{22.19} & \textbf{92.32} & \textbf{23.45} & \textbf{92.74} & \textbf{22.23} & \textbf{92.07} & 22.15 & \textbf{92.04} & 22.63 & \textbf{92.57} & 22.18 & \textbf{92.10} & \textbf{22.47} & \textbf{92.31}\\
        \hline
        \multirow{2}{*}{Metallicness}
        & PhySG~\cite{zhang2021physg} & 8.72 & N/A & 7.97 & N/A & 8.35 & N/A & 8.67 & N/A & 8.95 & N/A & 8.76 & N/A & 8.57 & N/A \\
        & SG-ENV~\cite{zhang2021physg} & 17.01 & N/A & 16.40 & N/A & 16.39 & N/A & 15.44 & N/A & 14.25 & N/A & 14.49 & N/A & 15.66 & N/A \\
        & NeILF~\cite{yao2022neilf}\dag & 19.48 & 64.43 & 18.75 & 92.56 & 18.21 & \textbf{92.58} & 21.03 & 65.53 & \textbf{19.45} & \textbf{89.85} & 19.60 & 76.44 & 19.42 & 80.23 \\
        & NeILF++~~\cite{zhang2023neilf++}\dag & 19.49 & \textbf{86.50} & 18.17 & \textbf{92.99} & 17.41 & 91.92 & 20.80 & \textbf{80.48} & 19.18 & 68.74 & 17.98 & \textbf{80.14} & 18.84 & \textbf{83.46} \\
        & PBR-NeRF (ours) & \textbf{21.73} & 72.66 & \textbf{22.26} & 73.26 & \textbf{20.88} & 88.65 & \textbf{22.45} & 68.40 & 19.24 & 63.25 & \textbf{23.15} & 78.28 & \textbf{21.62} & 74.08 \\
        \hline
        \multirow{2}{*}{Albedo}
        & PhySG~\cite{zhang2021physg} & 15.01 & N/A & 16.96 & N/A & 16.13 & N/A & 14.16 & N/A & 12.43 & N/A & 14.29 & N/A & 14.83 & N/A \\
        & SG-ENV~\cite{zhang2021physg} & \textbf{22.38} & N/A & \textbf{20.74} & N/A & \textbf{22.21} & N/A & 16.92 & N/A & 13.16 & N/A & 16.69 & N/A & 18.68 & N/A \\
        & NeILF~\cite{yao2022neilf}\dag & 17.73 & 90.32 & 21.11 & \textbf{90.21} & 18.24 & 90.18 & 17.01 & 77.14 & 18.52 & \textbf{81.01} & 16.32 & 78.33 & 18.15 & 84.53 \\
        & NeILF++~\cite{zhang2023neilf++}\dag & 18.21 & 82.62 & 19.50 & 79.21 & 17.73 & 82.48 & 16.37 & 70.13 & 14.48 & 64.61 & 17.43 & 76.20 & 17.29 & 75.87 \\
        & PBR-NeRF (ours) & 19.59 & \textbf{90.60} & 20.24 & 88.33 & 19.88 & \textbf{91.45} & \textbf{20.18} & \textbf{85.53} & \textbf{19.44} & 77.86 & \textbf{21.14} & \textbf{87.13} & \textbf{20.08} & \textbf{86.82} \\
        \specialrule{.2em}{.1em}{.1em}
    \end{tabular}
    }
    \vspace{-3mm}
\end{table*}

\subsection{Experimental Setup}
\label{sec:exp_setup}

\PAR{Implementation Details.}
We implement PBR-NeRF on top of NeILF++~\cite{zhang2023neilf++} using PyTorch~\cite{Ansel_PyTorch_2_Faster_2024} and keep all hyperparameters from NeILF++ the same, except for $|S_L|=256$ and a training batch size of 8192 rays.
The geometry, material, and joint phases last 5K, 1K, and 30K iterations, respectively.
We train with a learning rate of 0.002, which is fixed for the geometry and material phase, but decreases by a factor of 5 every 10K iterations during the joint phase.
Training runs on a single NVIDIA A6000, with runtimes ranging from 3 to 7.5 hours.
Using grid search, we identify two sets of weights for our physics-based losses: $\lambda_\text{cons} = 0.01$ and $\lambda_\text{spec}=0.5$ for the NeILF++ dataset, and $\lambda_\text{cons} = 0.01$ and $\lambda_\text{spec}=0.01$ for DTU.
We weigh $\lambda_\text{spec}$ lower than the RGB loss, $\lambda_\text{pbr}=1$, so that $\mathcal{L}_\text{spec}$ has a small yet meaningful weighting.
More hyperparameter details are provided in Appendix Sec.~\ref{sec:supp:hyperparam_sweep}.

\PAR{Datasets.}
We evaluate our method on the NeILF++~\cite{zhang2023neilf++} and DTU~\cite{jensen2014dtu} datasets. We use the NeILF++ synthetic dataset to evaluate our material-illumination estimation, assuming known geometry.
This dataset provides ground-truth geometry, RGB, albedo, metallicness, and roughness for a scene rendered under 6 static illumination patterns, including three environment maps (Env) and three mixed-lighting scenarios with additional point and area light sources (Mix).
Each pattern includes 96 RGB HDR $1600 \times 1200$ images, split into 87 training and 9 validation images.
Ground-truth geometry and poses are used in training, while ground-truth color, albedo, metallicness, and roughness are only used for evaluation.
DTU captures 15 real-world objects in a controlled lab setting.
Each scene includes 49 RGB LDR $1600 \times 1200$ images, split into 44 training and 5 validation images.
We evaluate the predicted geometry by computing the Chamfer Distance with the provided reference point cloud generated by structured light scanners.
We only qualitatively evaluate material estimation on DTU because it uses LDR images and does not provide ground-truth materials.
We do not evaluate lighting because the DTU dataset is unsuitable for evaluating predicted lighting without HDR images and ground-truth environment maps.
Furthermore, the predicting lighting is not meaningful because lighting cannot be reliably estimated from the LDR DTU images due to clipping and unknown tone mapping.

\begin{figure*}[tb]
    \renewcommand{\arraystretch}{1.5}
    \centering
    \small
    \resizebox{\textwidth}{!}{
                \begin{tabular}{ccccccc}
            & & Lighting \dag & RGB & Roughness & Metallicness & Albedo \\
            \multirow{3}{*}[0.5in]{\raisebox{-1.2in}{\rotatebox[origin=c]{90}{City Mix}}}
                        & \multirow{1}{*}[0.5in]{\rotatebox[origin=c]{90}{NeILF++}}
                        & \includegraphics[width=0.2\linewidth]{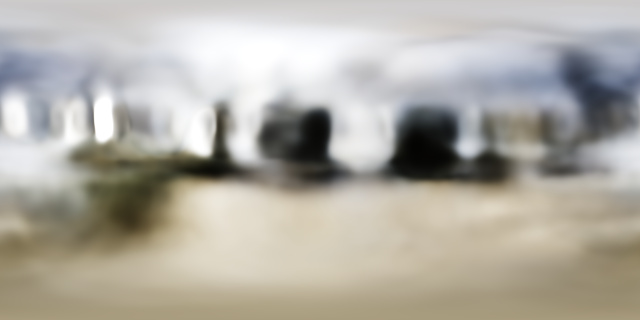}
                        & \includegraphics[width=0.2\linewidth]{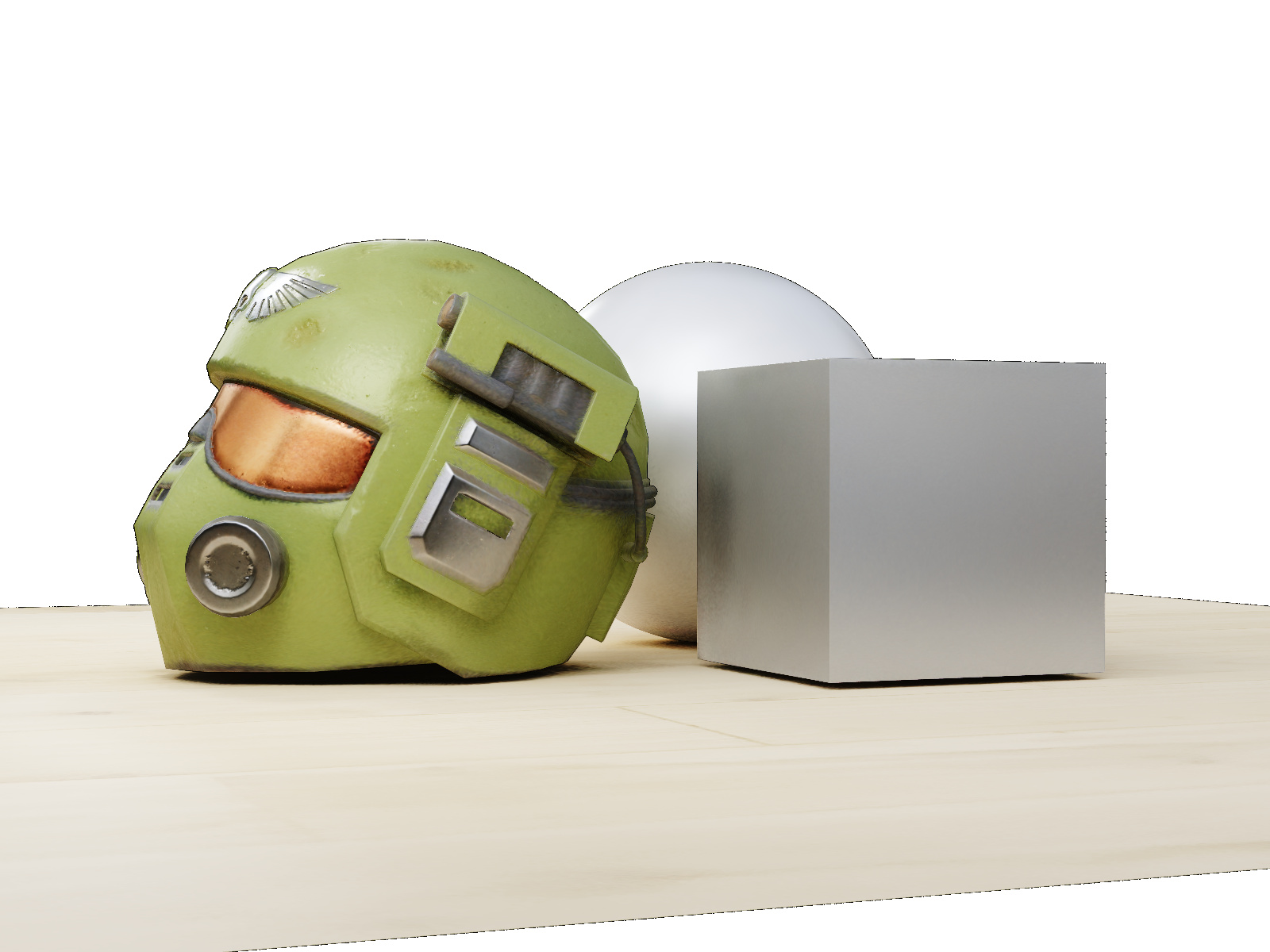}
                        & \includegraphics[width=0.2\linewidth]{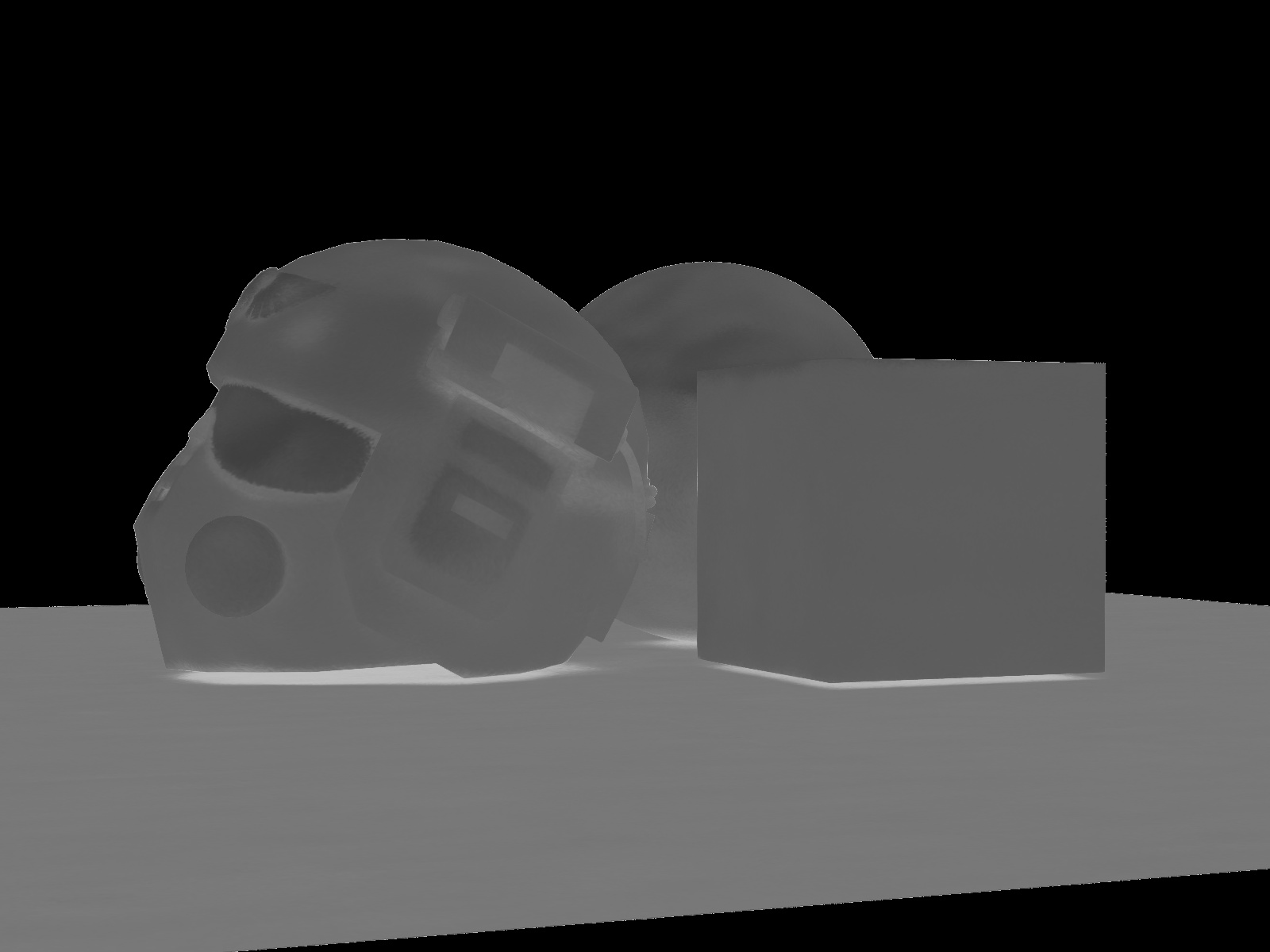}
                        & \includegraphics[width=0.2\linewidth]{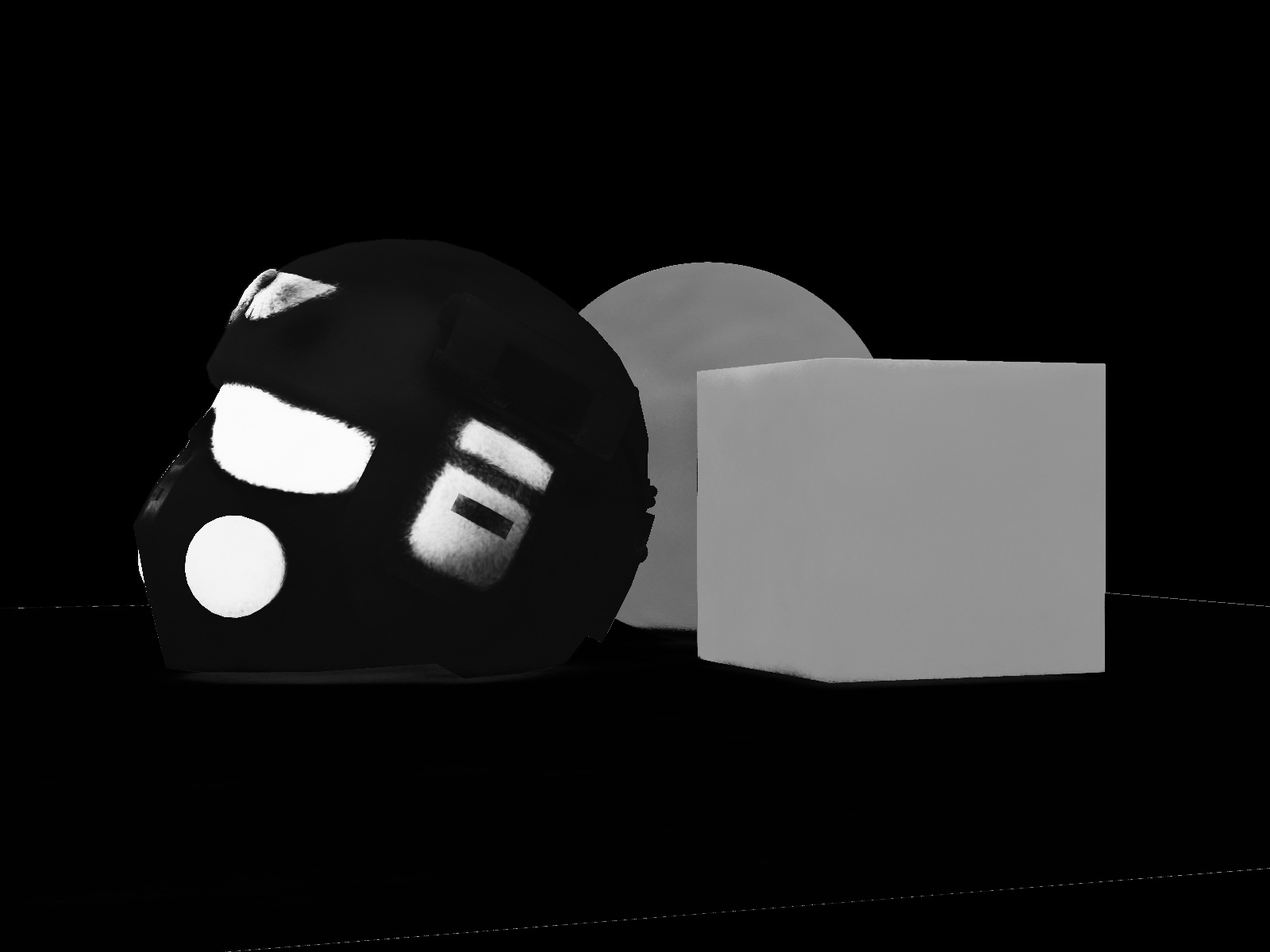}
                        & \includegraphics[width=0.2\linewidth]{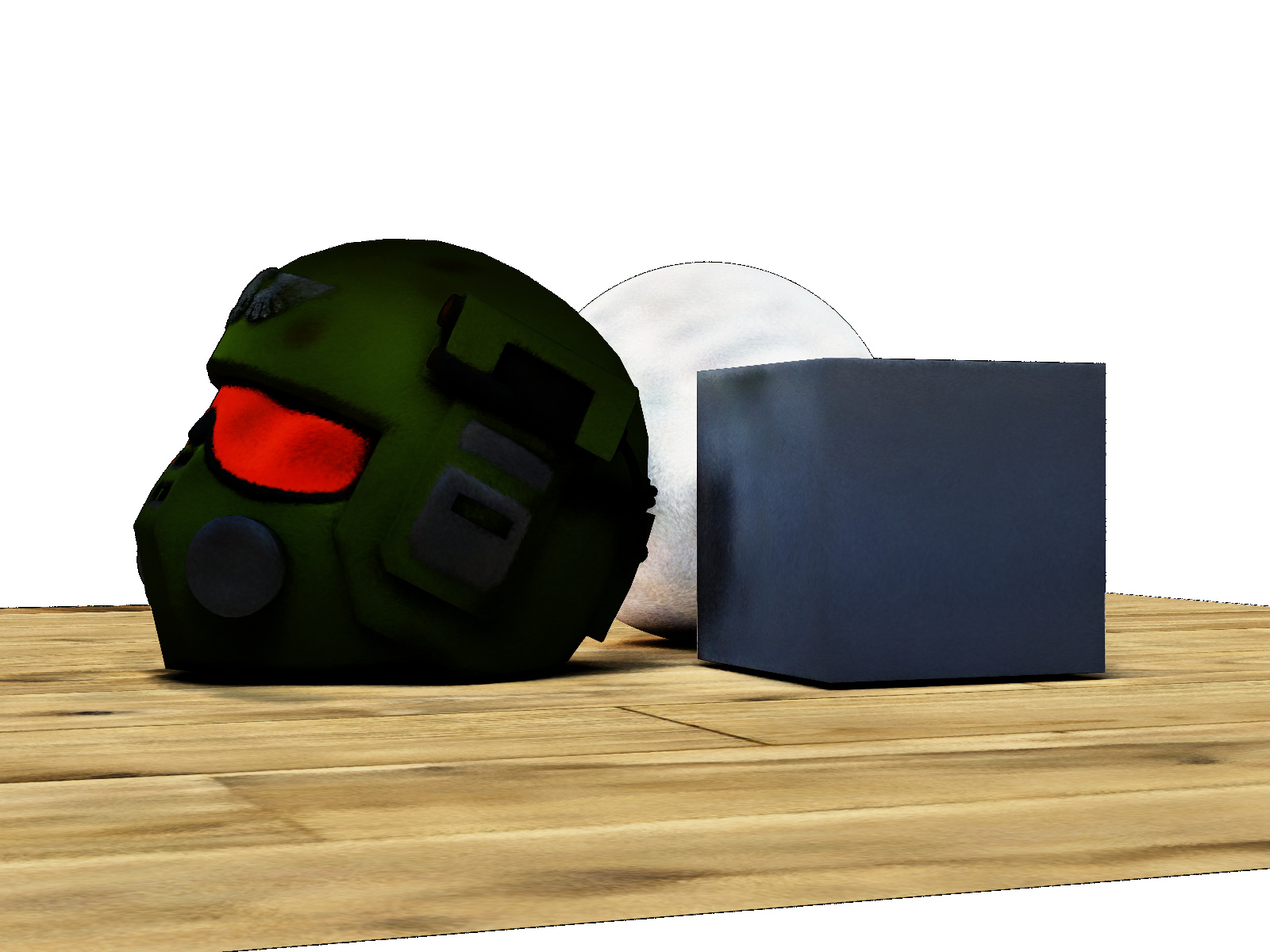} \\
                        & \multirow{1}{*}[0.5in]{\rotatebox[origin=c]{90}{Ours}}
                        & \includegraphics[width=0.2\linewidth]{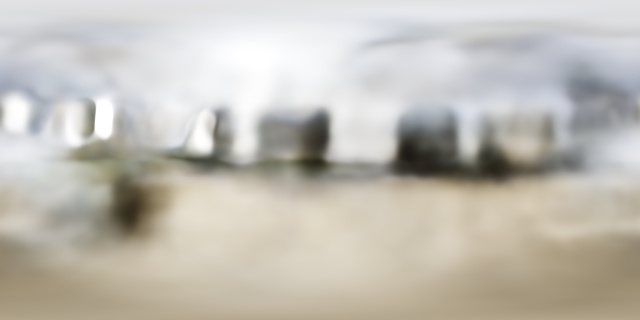}
                        & \includegraphics[width=0.2\linewidth]{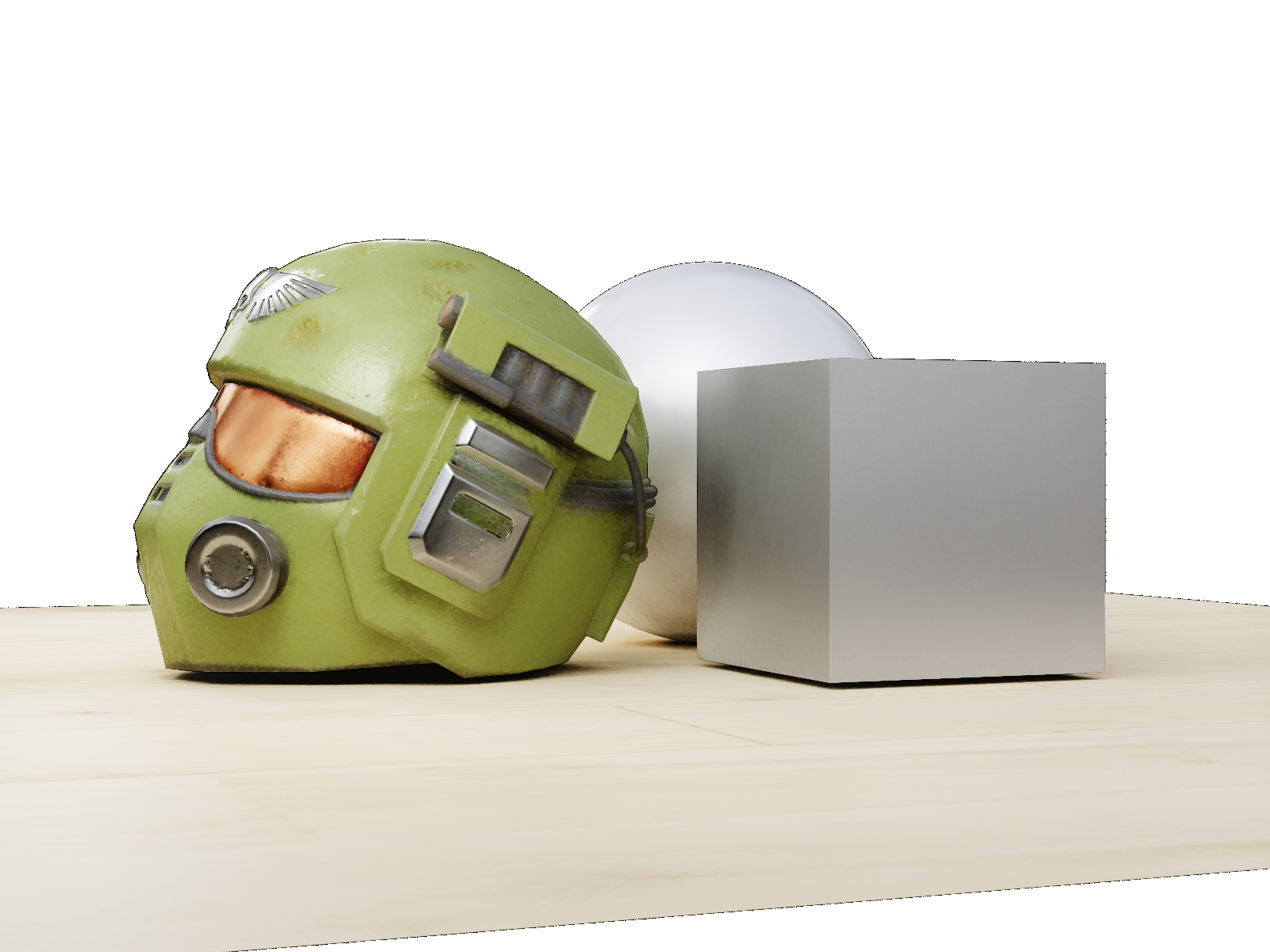}
                        & \includegraphics[width=0.2\linewidth]{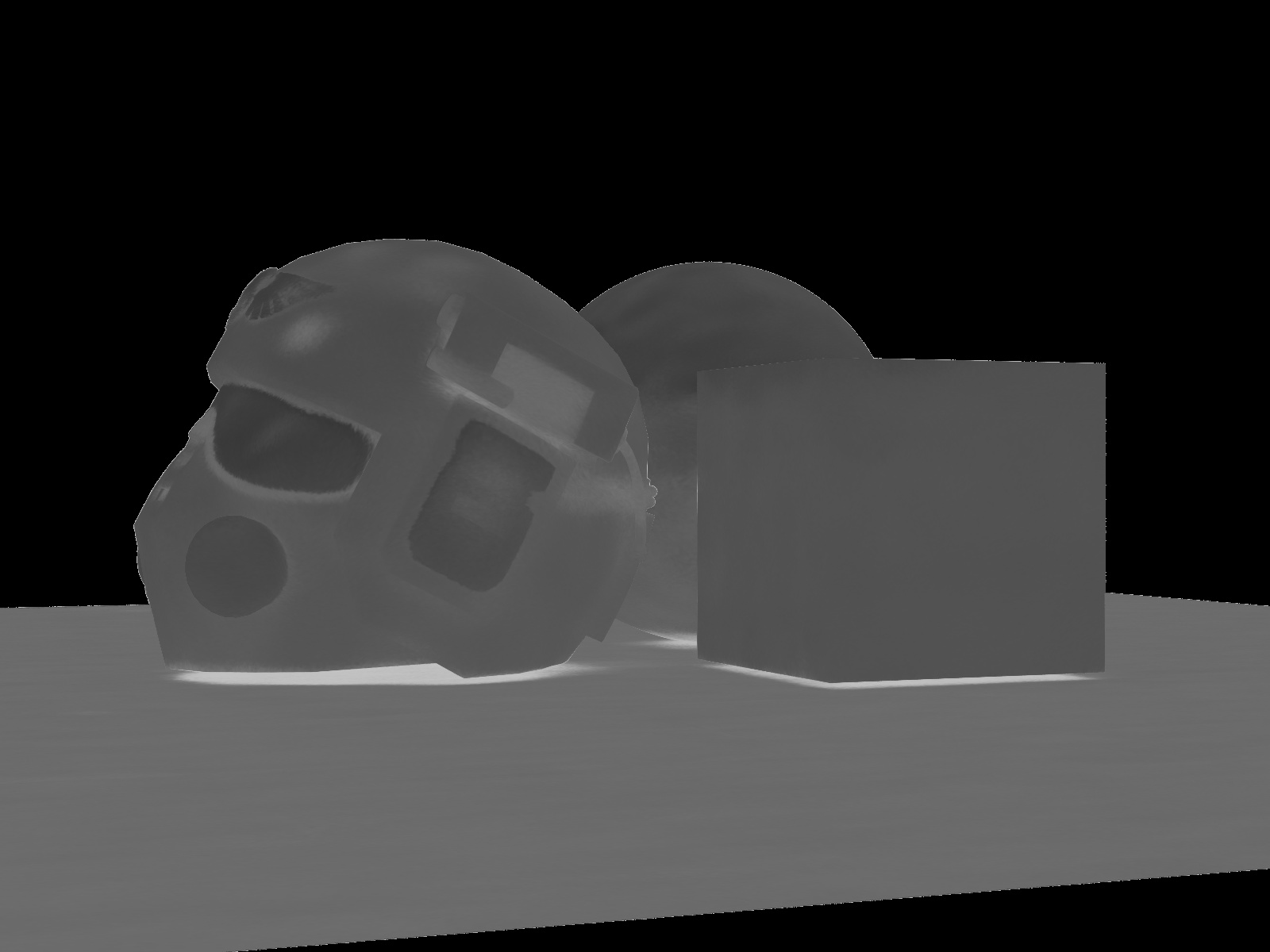}
                        & \includegraphics[width=0.2\linewidth]{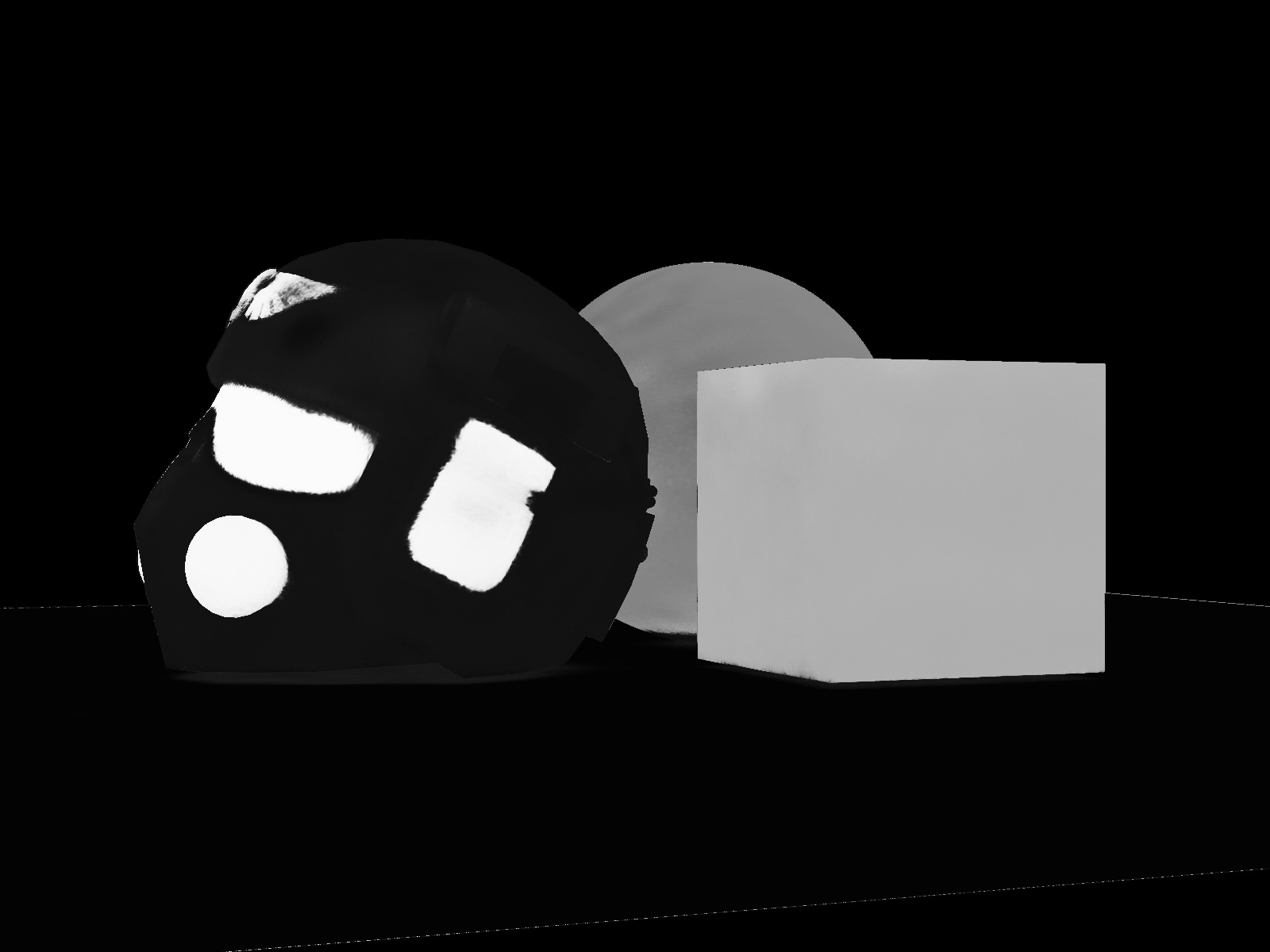}
                        & \includegraphics[width=0.2\linewidth]{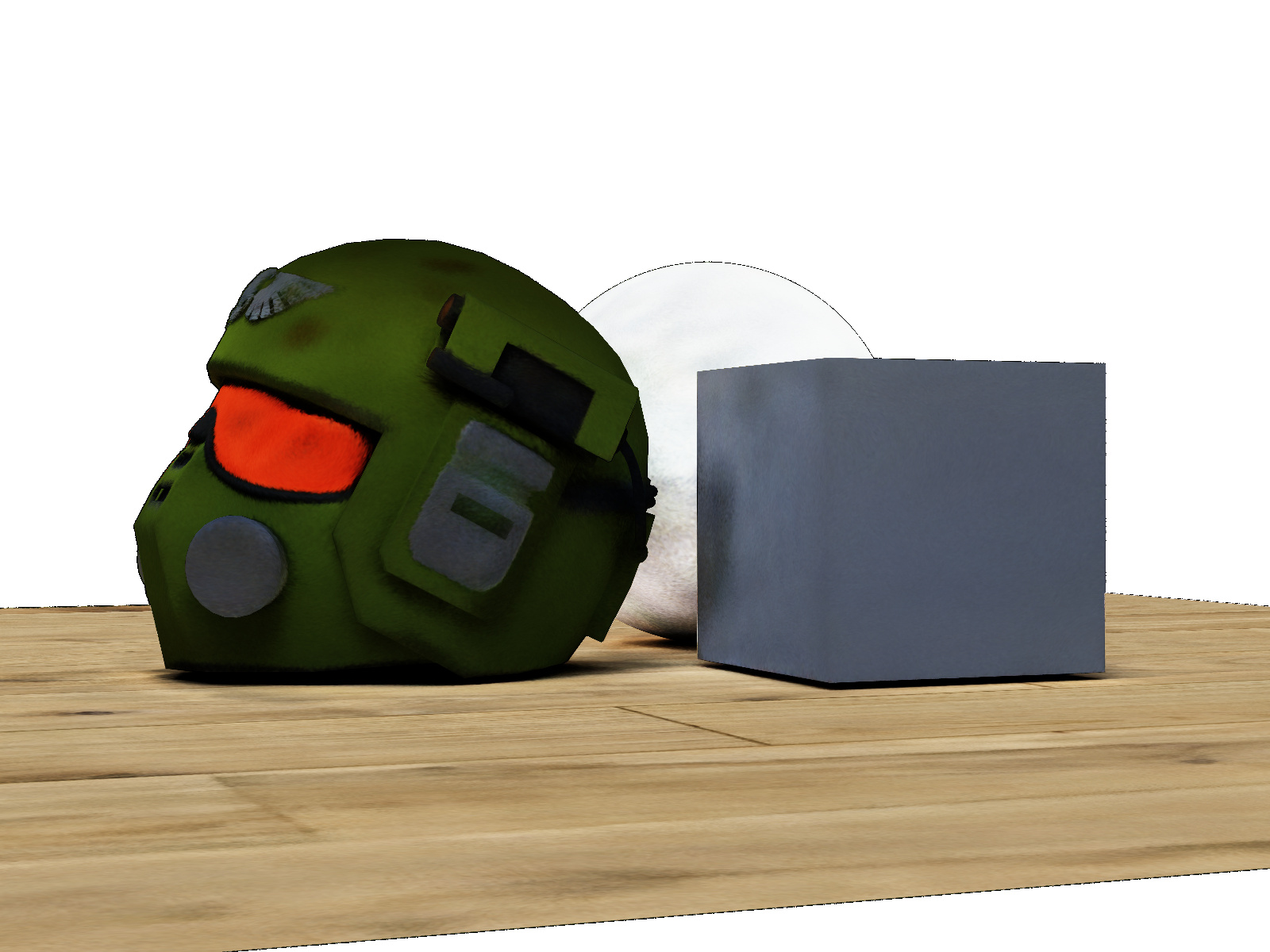} \\
                        & \multirow{1}{*}[0.5in]{\rotatebox[origin=c]{90}{Ground Truth}}
                        &
                        & \includegraphics[width=0.2\linewidth]{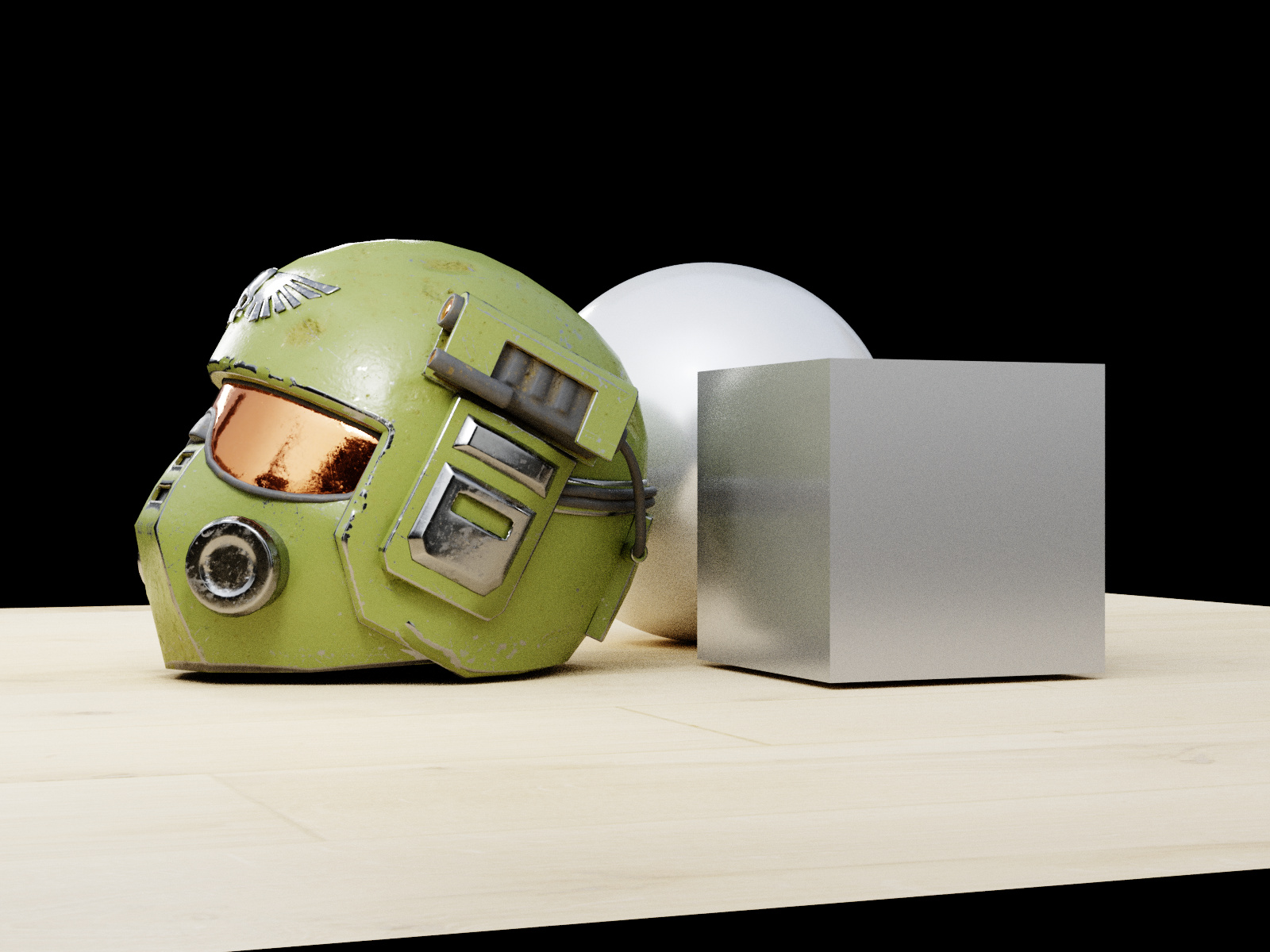}
                        & \includegraphics[width=0.2\linewidth]{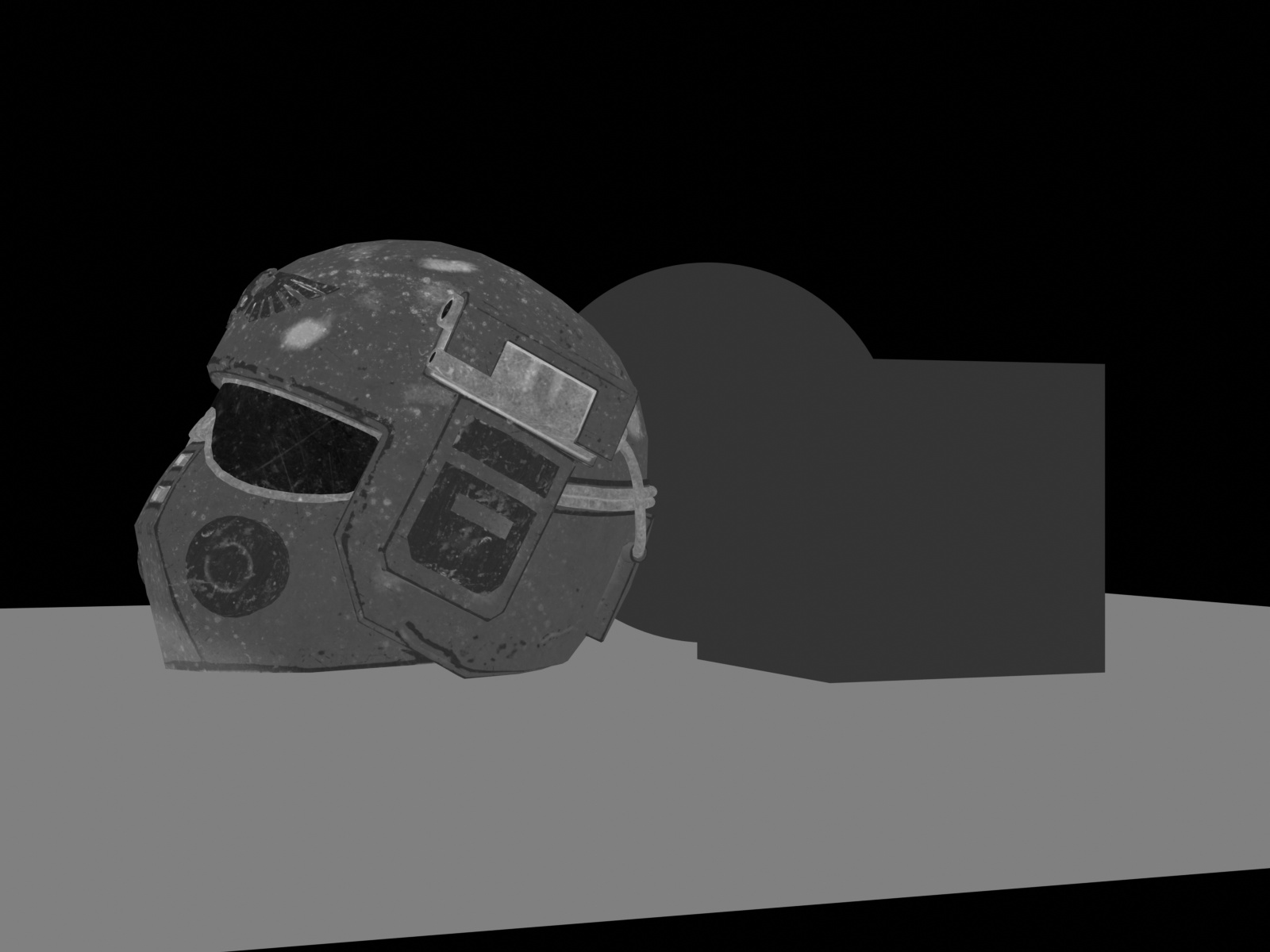}
                        & \includegraphics[width=0.2\linewidth]{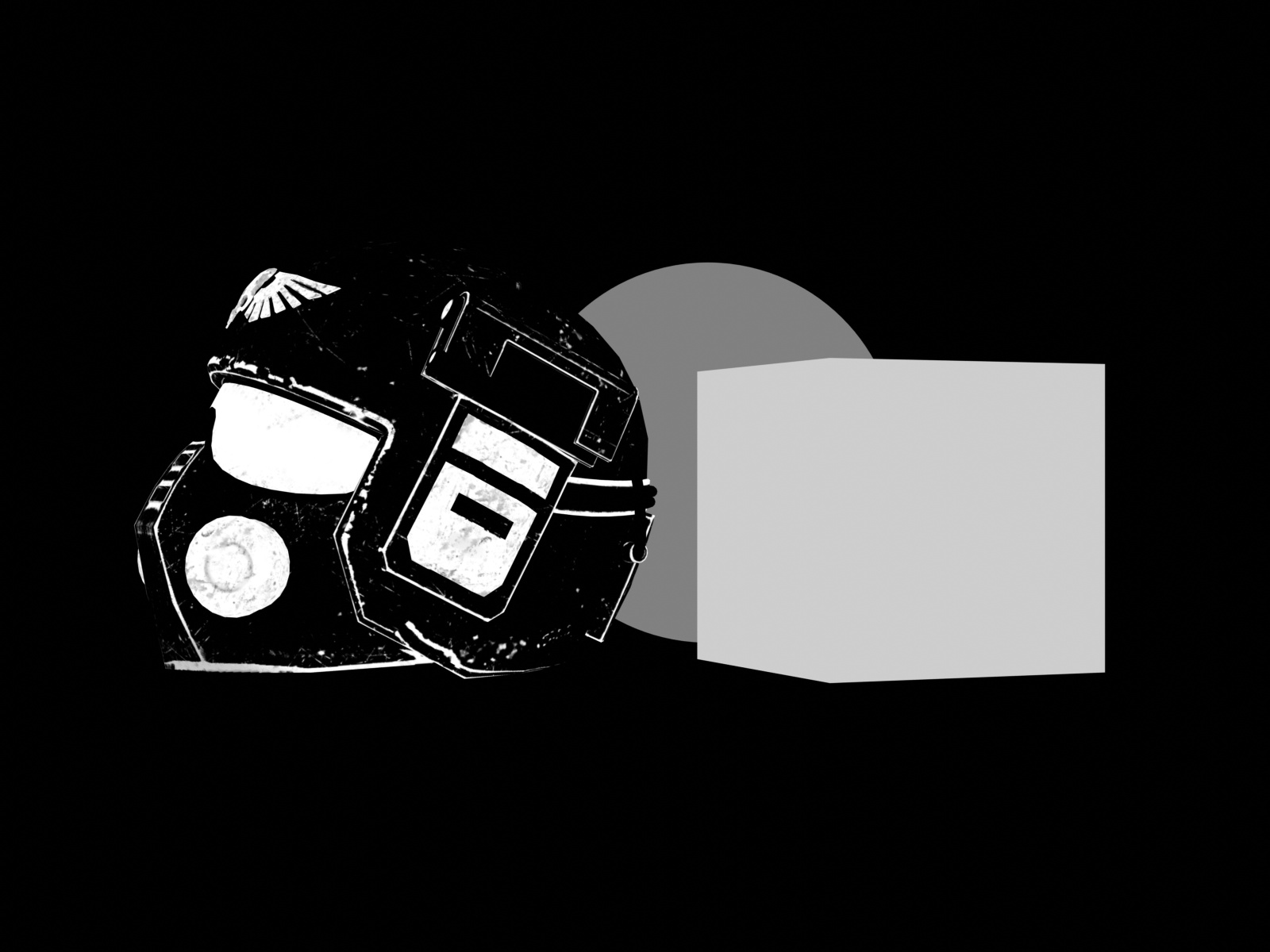}
                        & \includegraphics[width=0.2\linewidth]{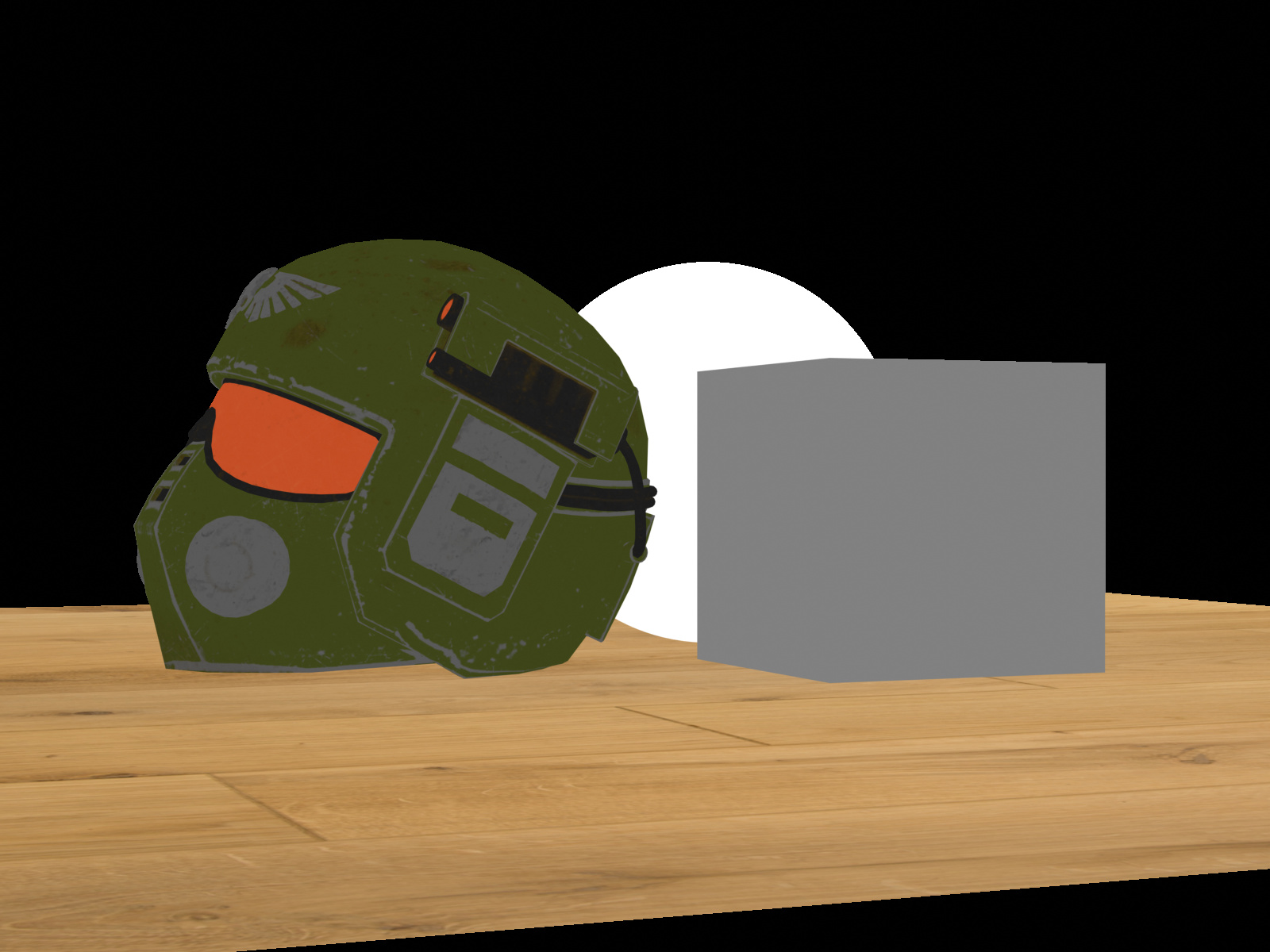} \\
            \hline & \\[-1.0em]
            \multirow{3}{*}[0.5in]{\raisebox{-1.2in}{\rotatebox[origin=c]{90}{Studio Mix}}}
                        & \multirow{1}{*}[0.5in]{\rotatebox[origin=c]{90}{NeILF++}}
                        & \includegraphics[width=0.2\linewidth]{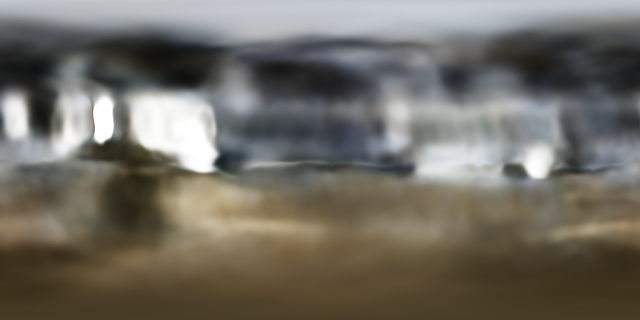}
                        & \includegraphics[width=0.2\linewidth]{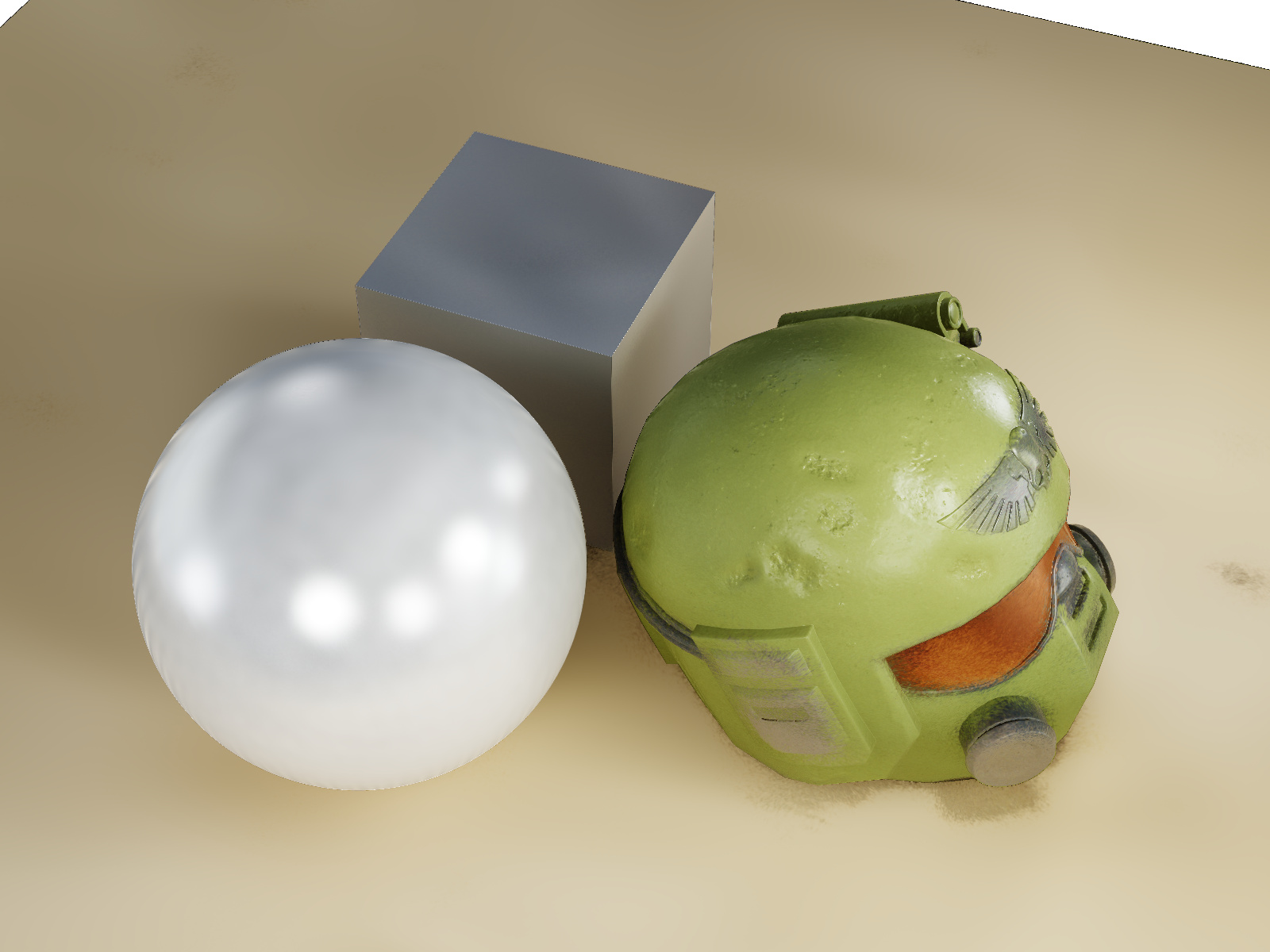}
                        & \includegraphics[width=0.2\linewidth]{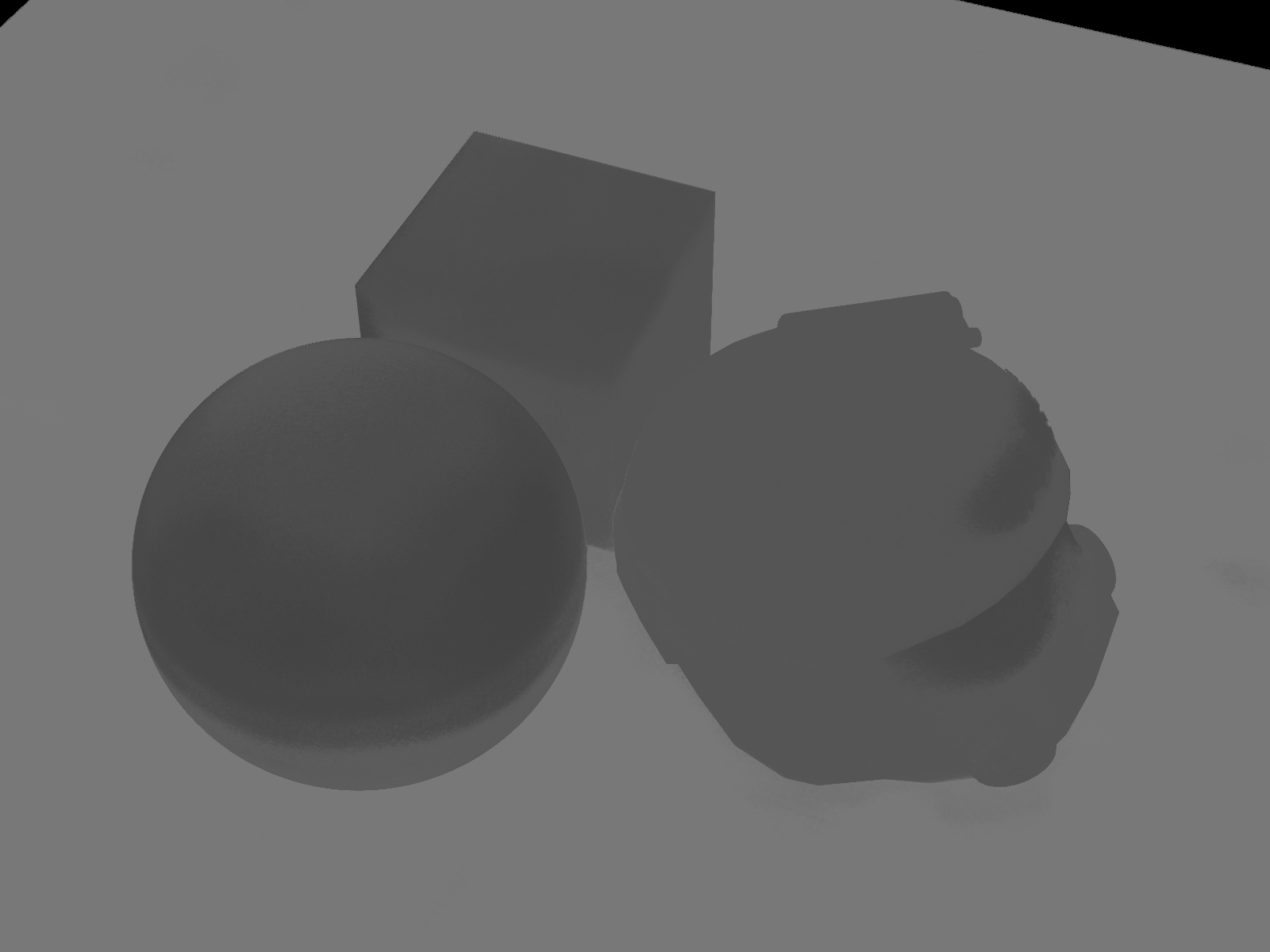}
                        & \includegraphics[width=0.2\linewidth]{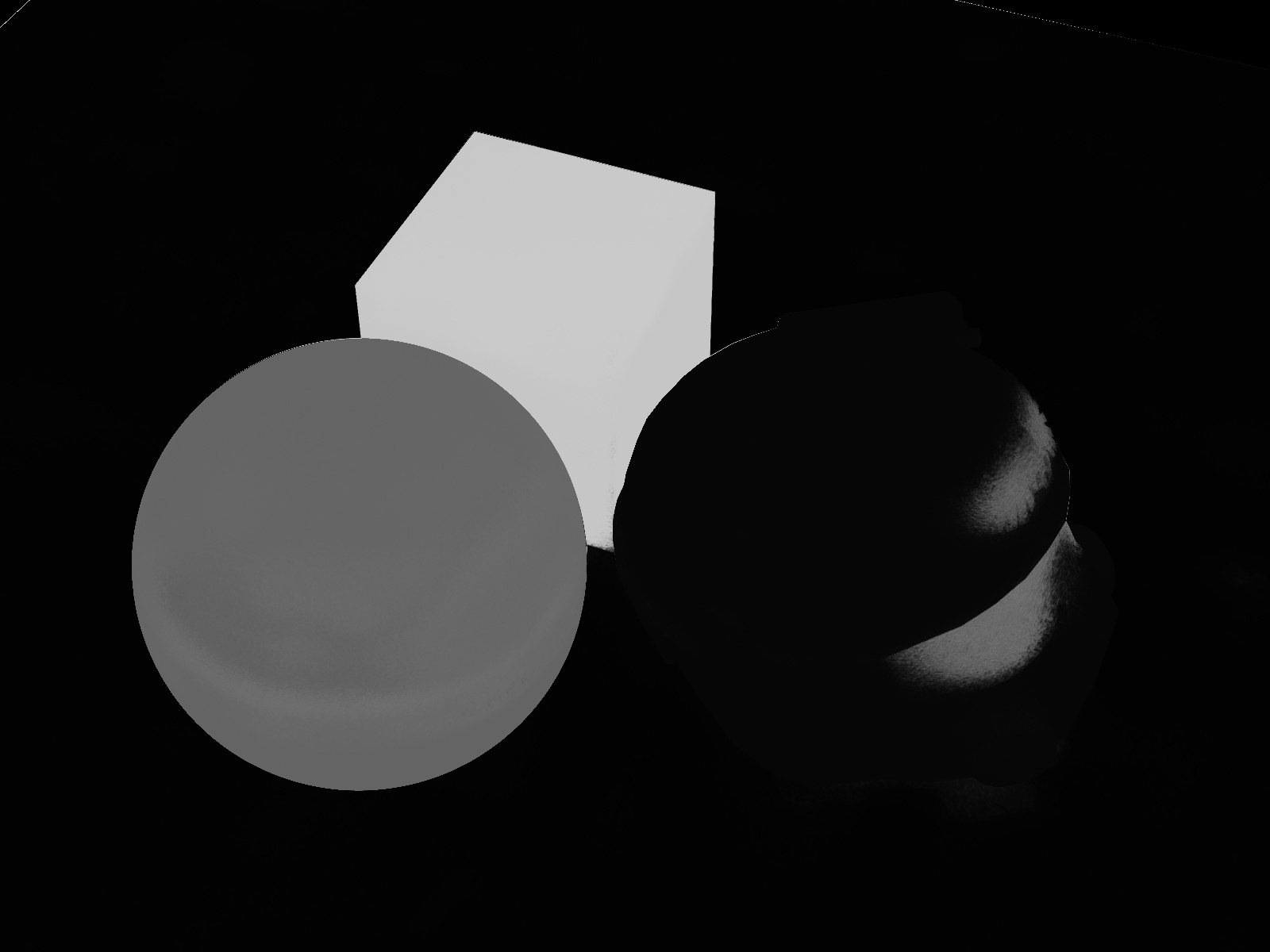}
                        & \includegraphics[width=0.2\linewidth]{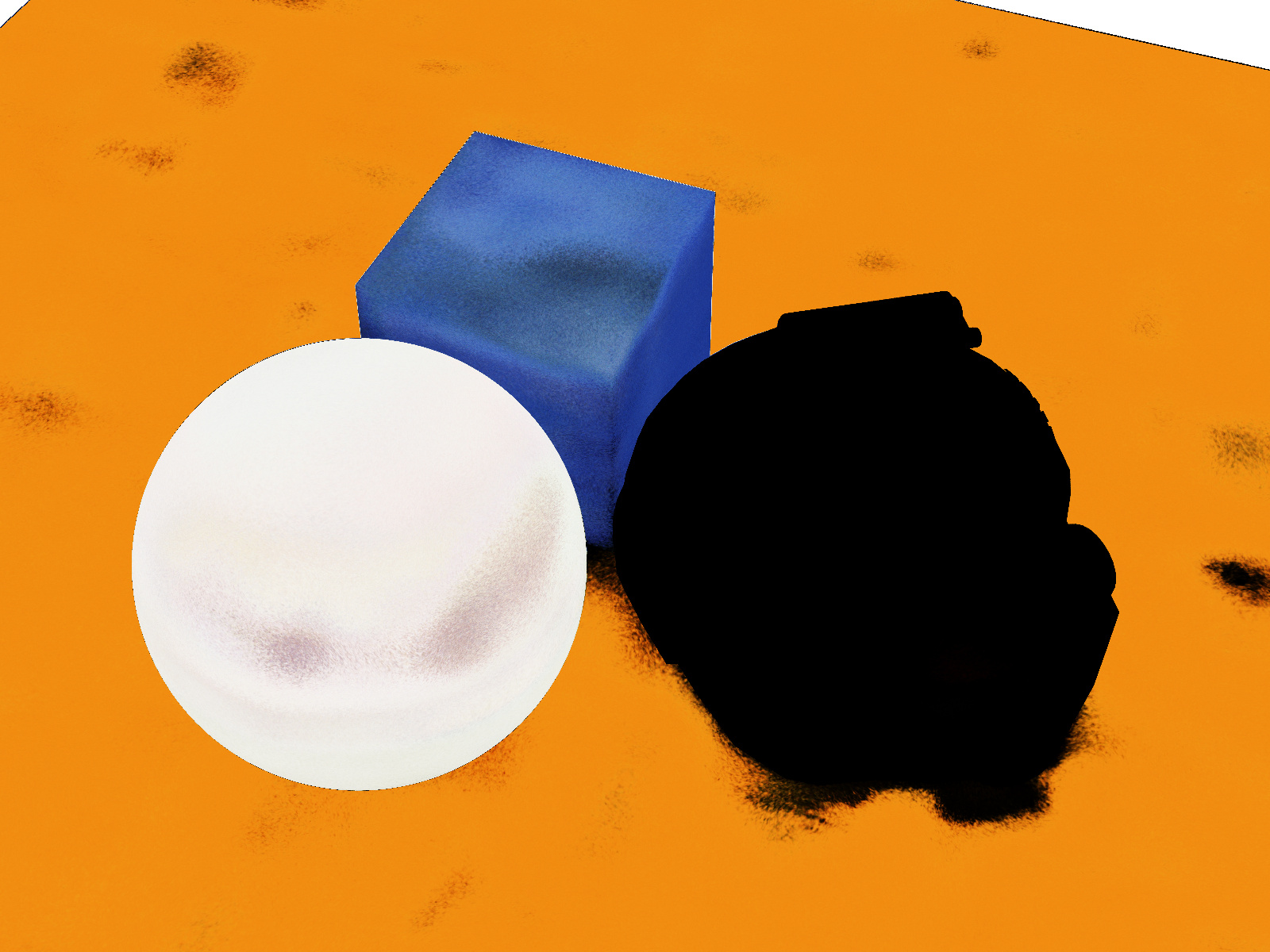} \\
                        & \multirow{1}{*}[0.5in]{\rotatebox[origin=c]{90}{Ours}}
                        & \includegraphics[width=0.2\linewidth]{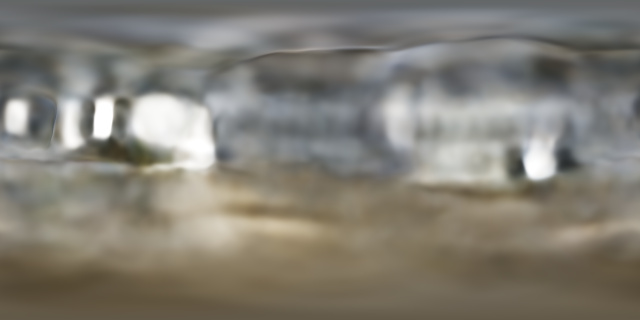}
                        & \includegraphics[width=0.2\linewidth]{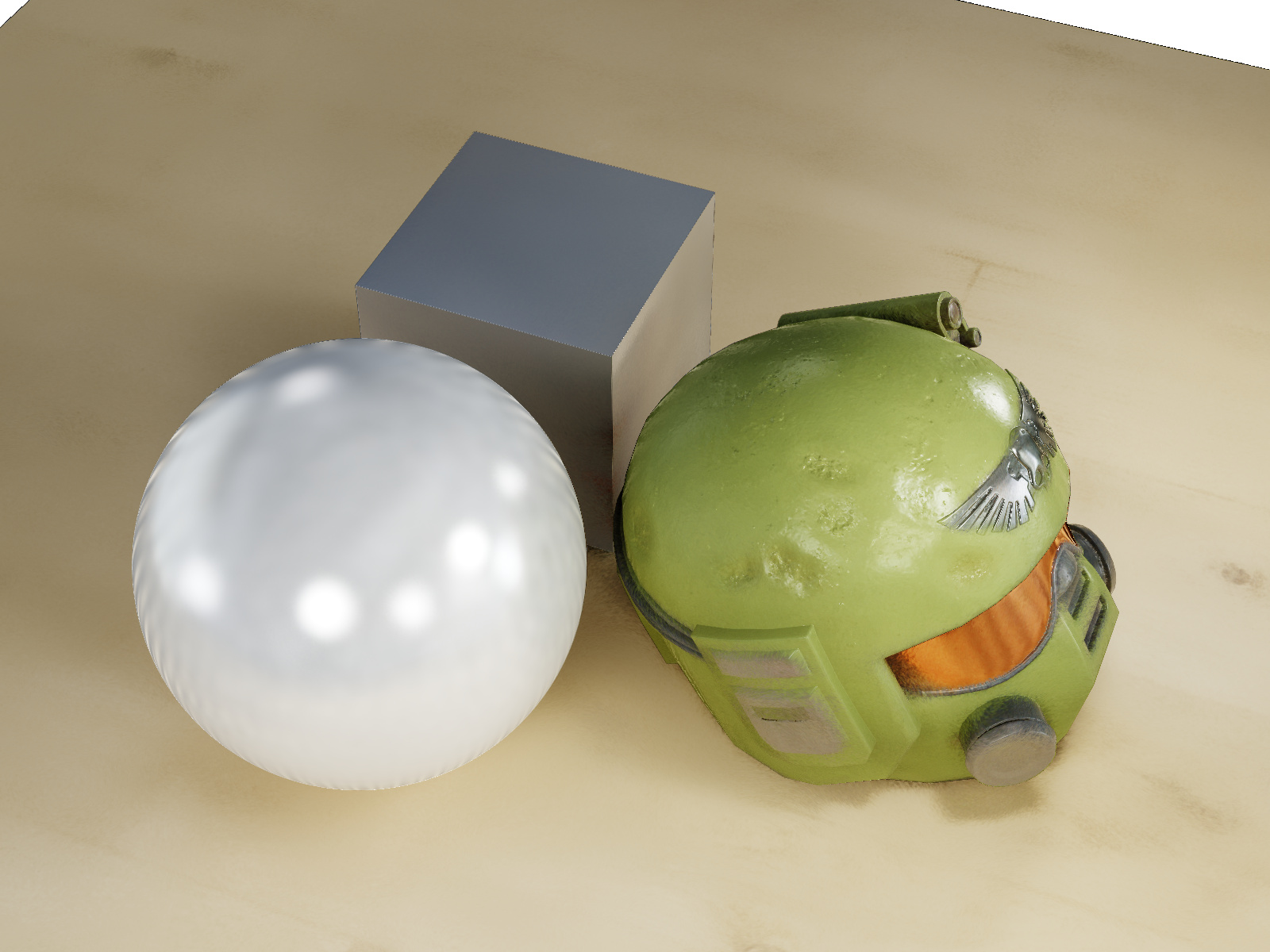}
                        & \includegraphics[width=0.2\linewidth]{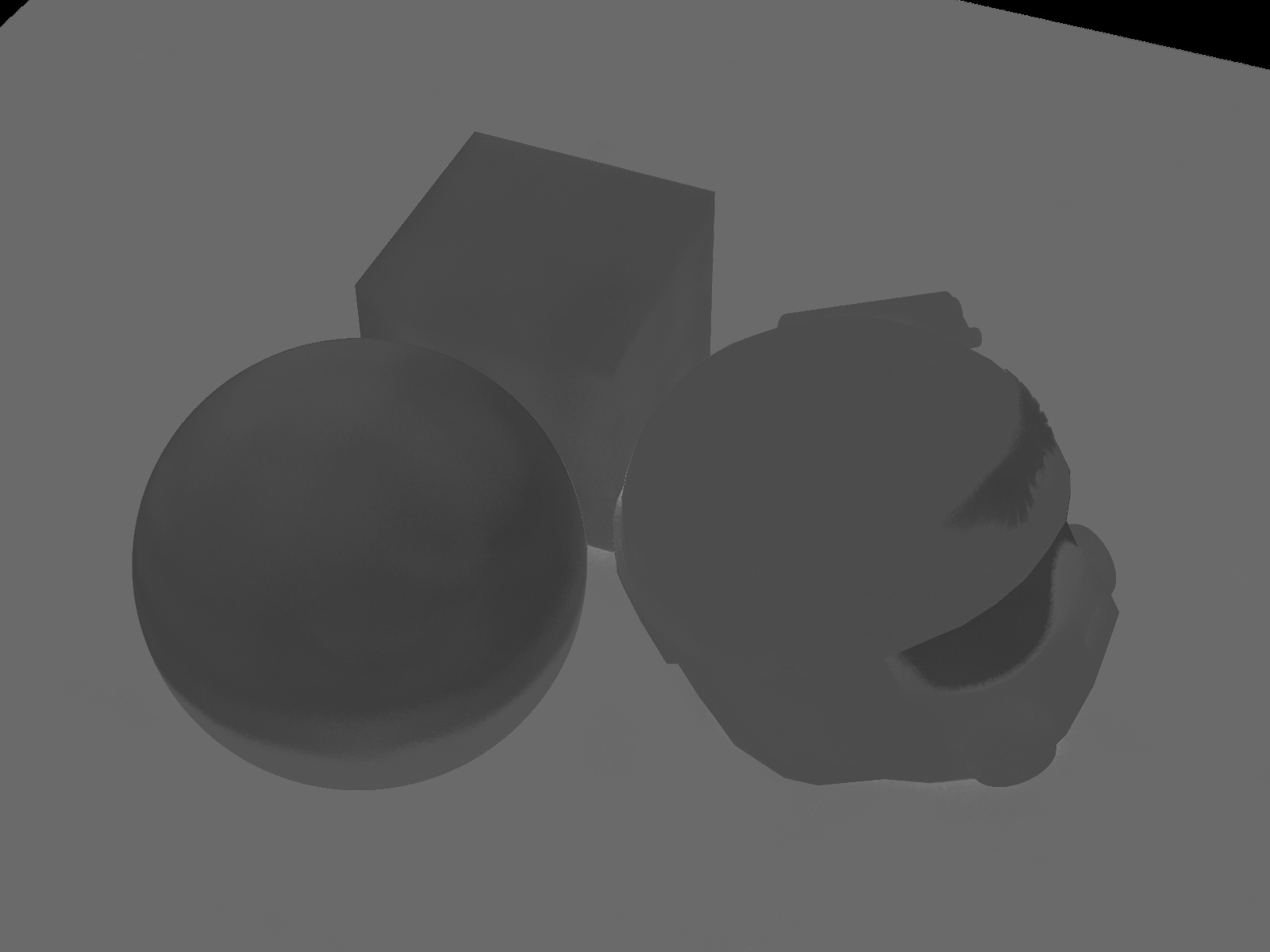}
                        & \includegraphics[width=0.2\linewidth]{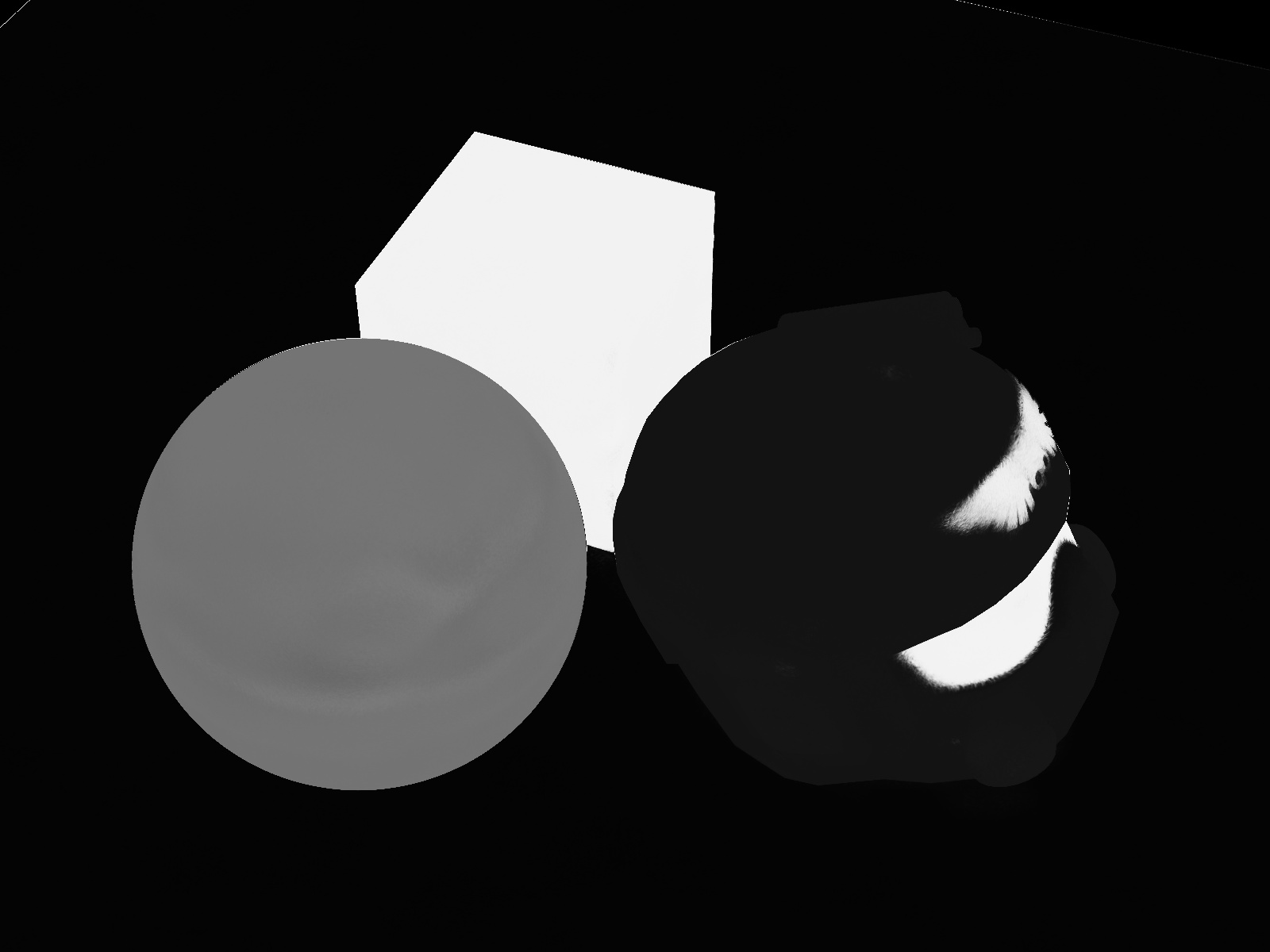}
                        & \includegraphics[width=0.2\linewidth]{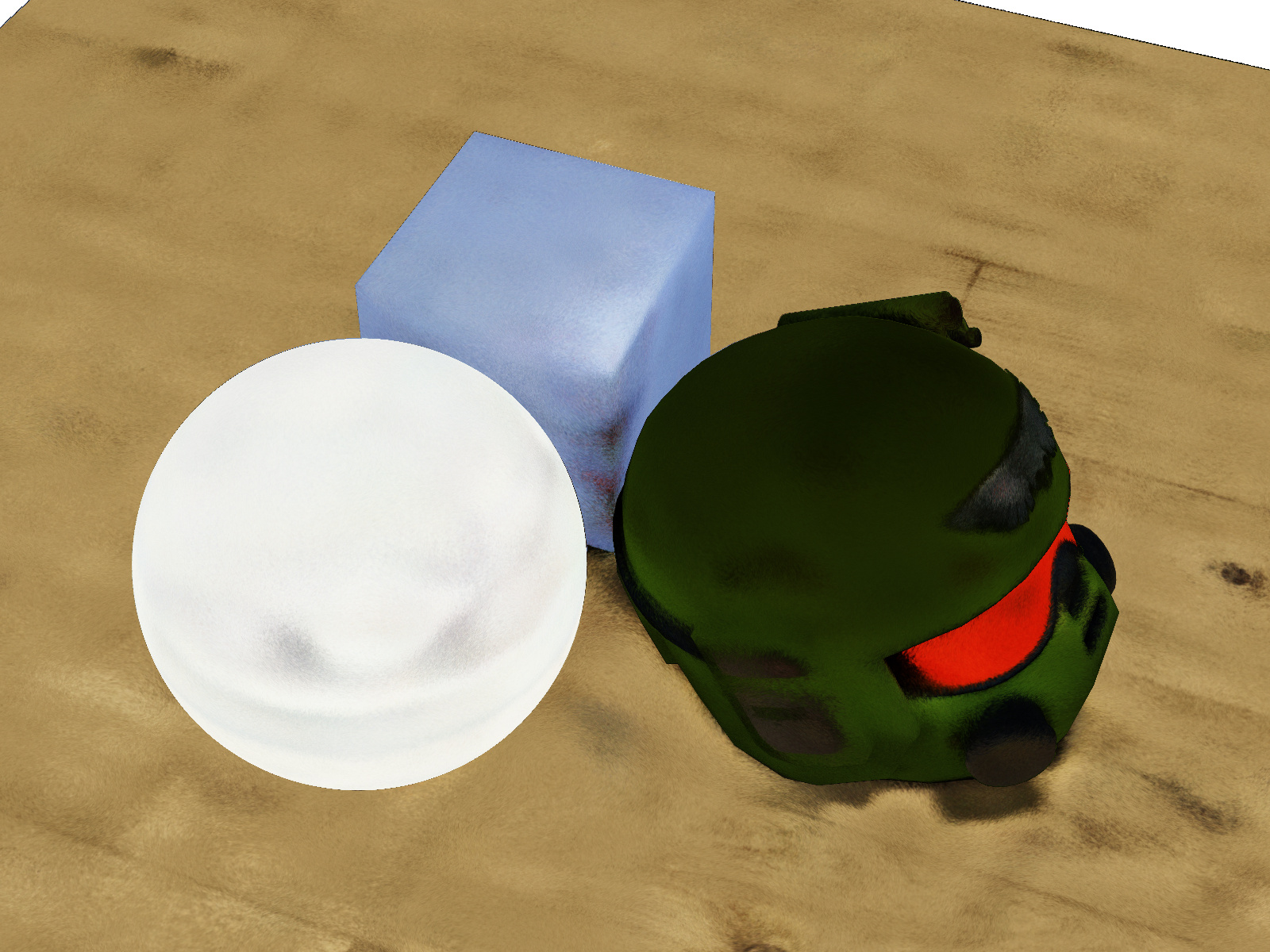} \\
                        & \multirow{1}{*}[0.5in]{\rotatebox[origin=c]{90}{Ground Truth}}
                        &
                        & \includegraphics[width=0.2\linewidth]{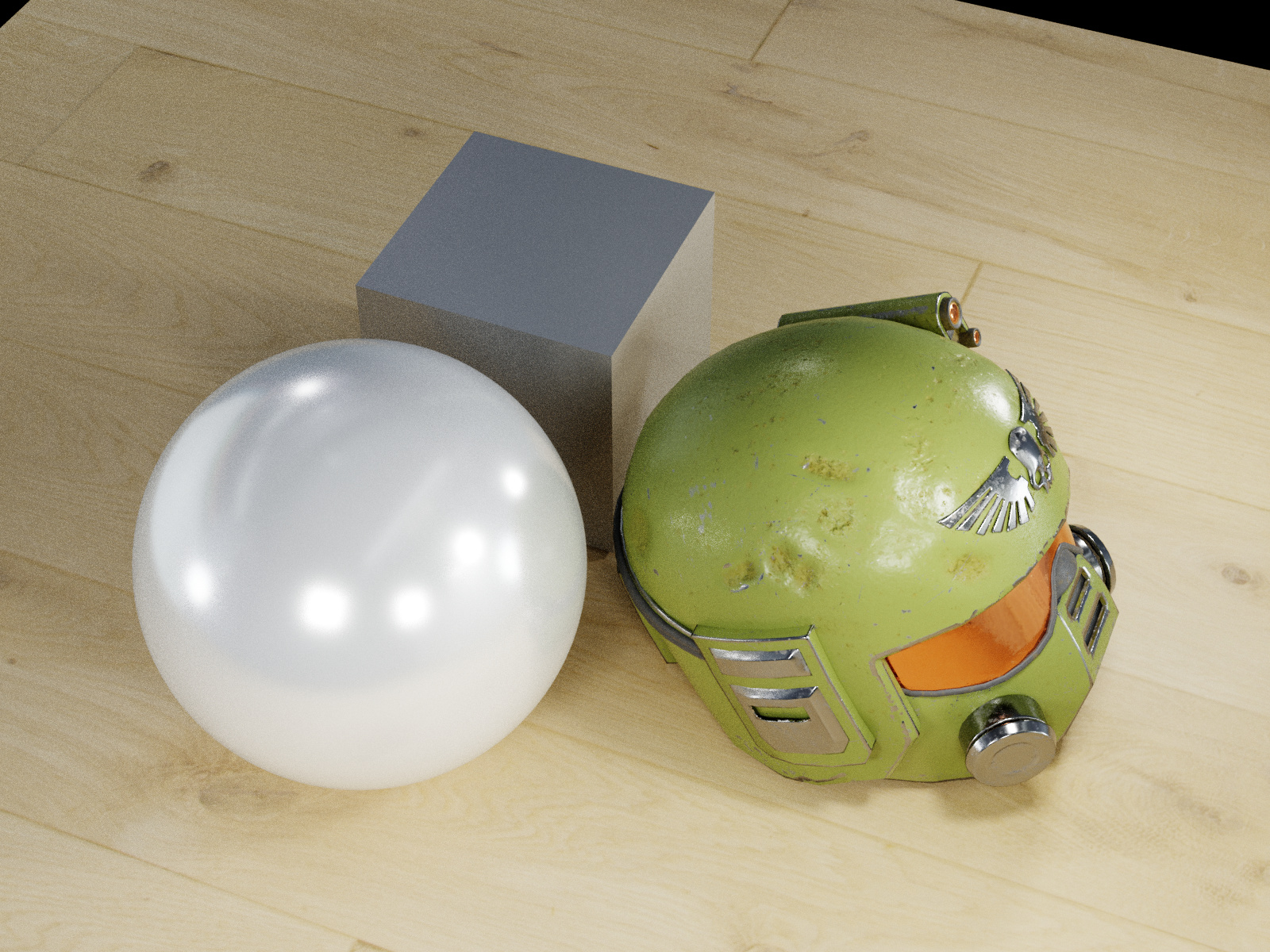}
                        & \includegraphics[width=0.2\linewidth]{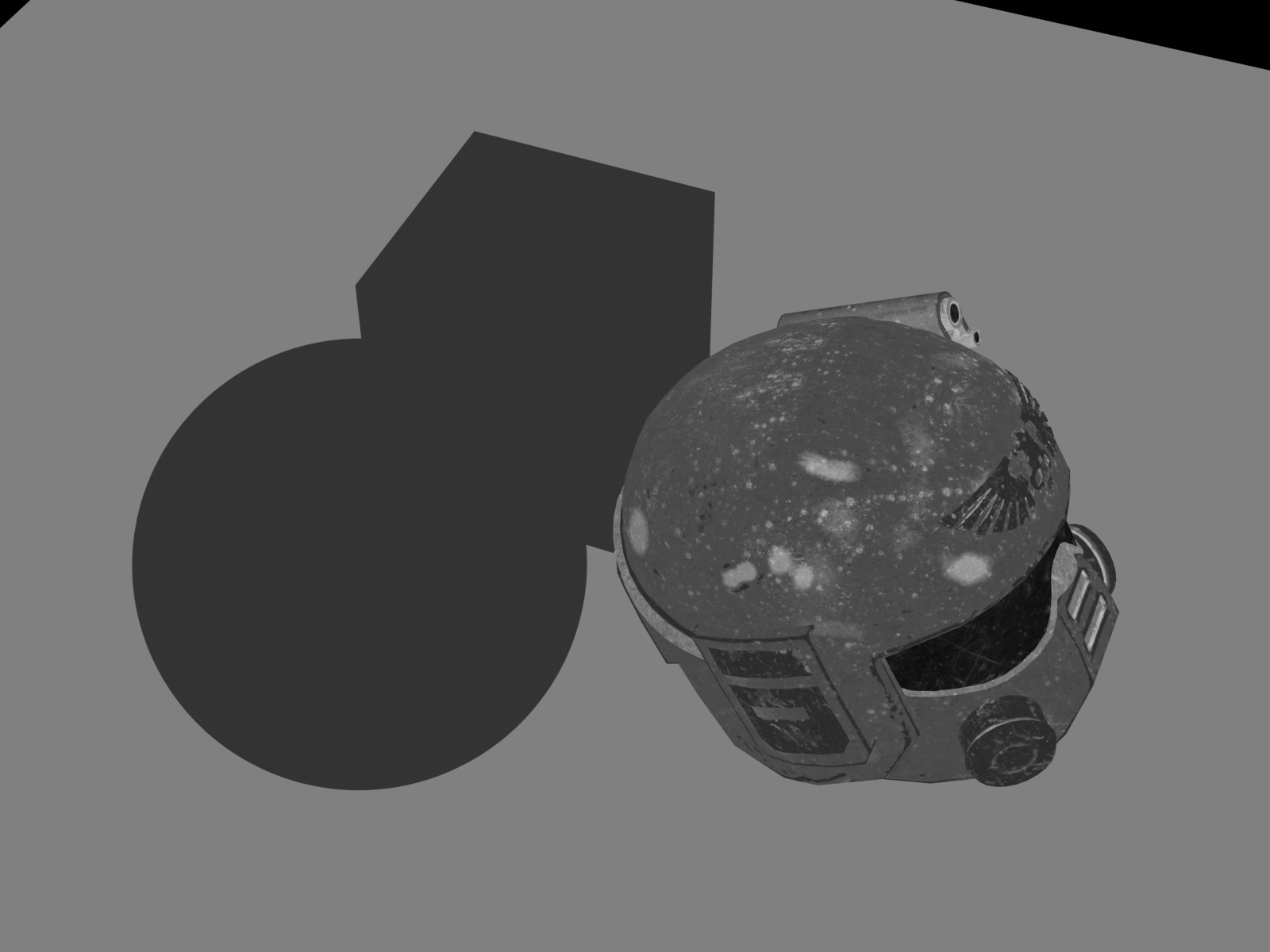}
                        & \includegraphics[width=0.2\linewidth]{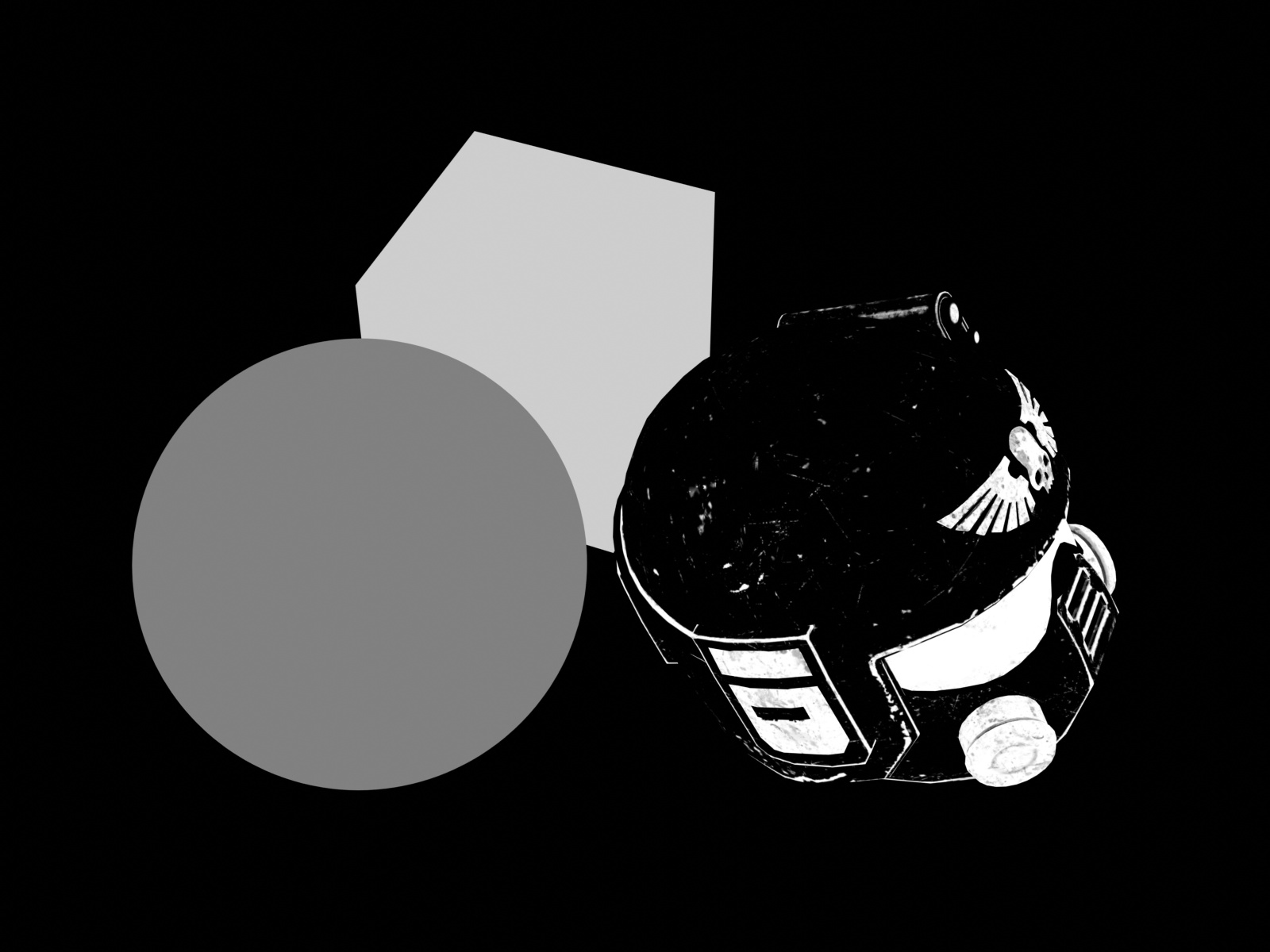}
                        & \includegraphics[width=0.2\linewidth]{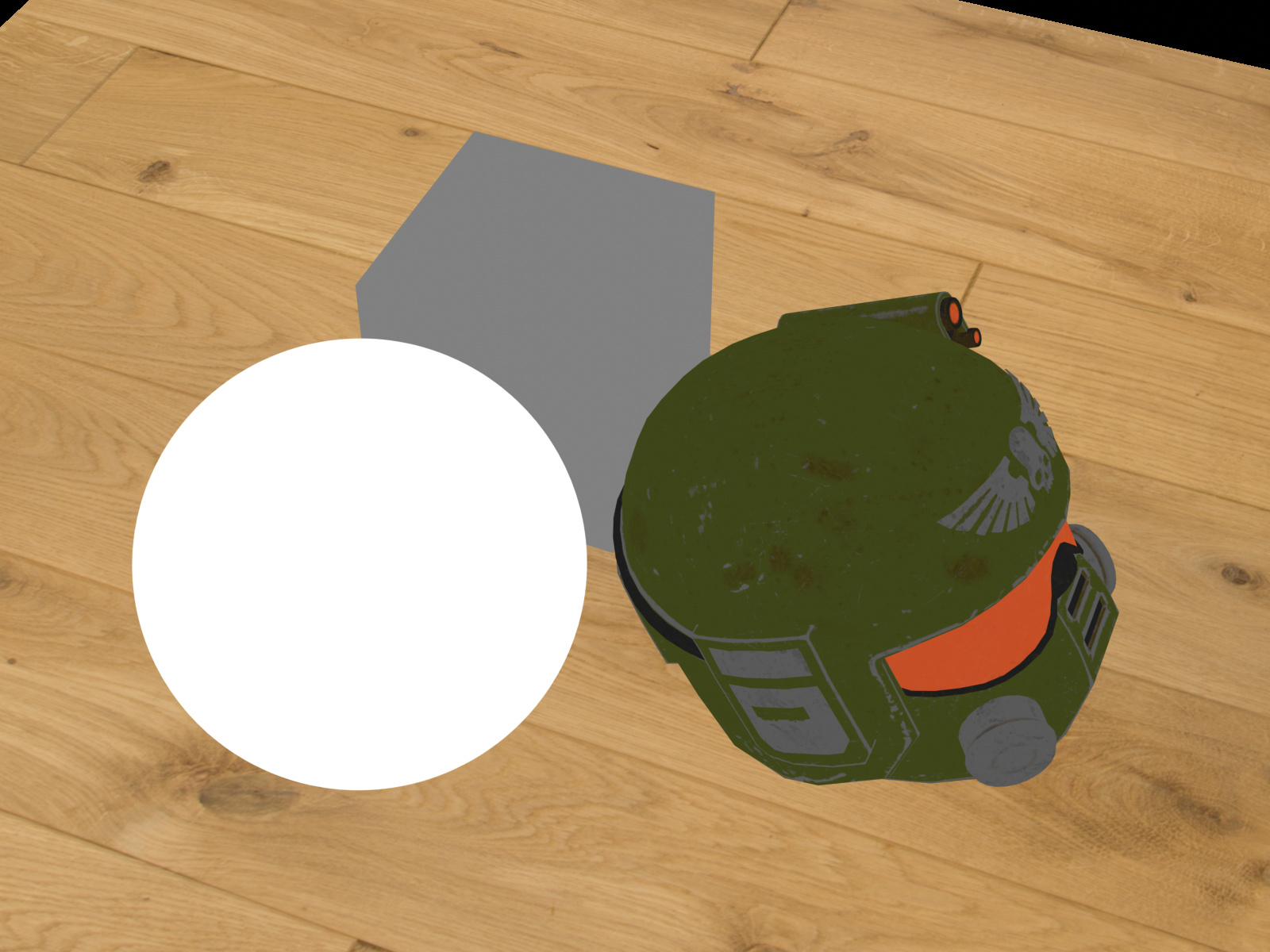} \\
            \hline & \\[-1.0em]
            \multirow{3}{*}[0.5in]{\raisebox{-1.2in}{\rotatebox[origin=c]{90}{Castel Mix}}}
                        & \multirow{1}{*}[0.5in]{\rotatebox[origin=c]{90}{NeILF++}}
                        & \includegraphics[width=0.2\linewidth]{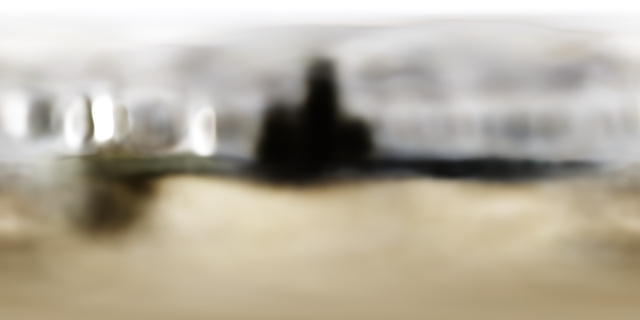}
                        & \includegraphics[width=0.2\linewidth]{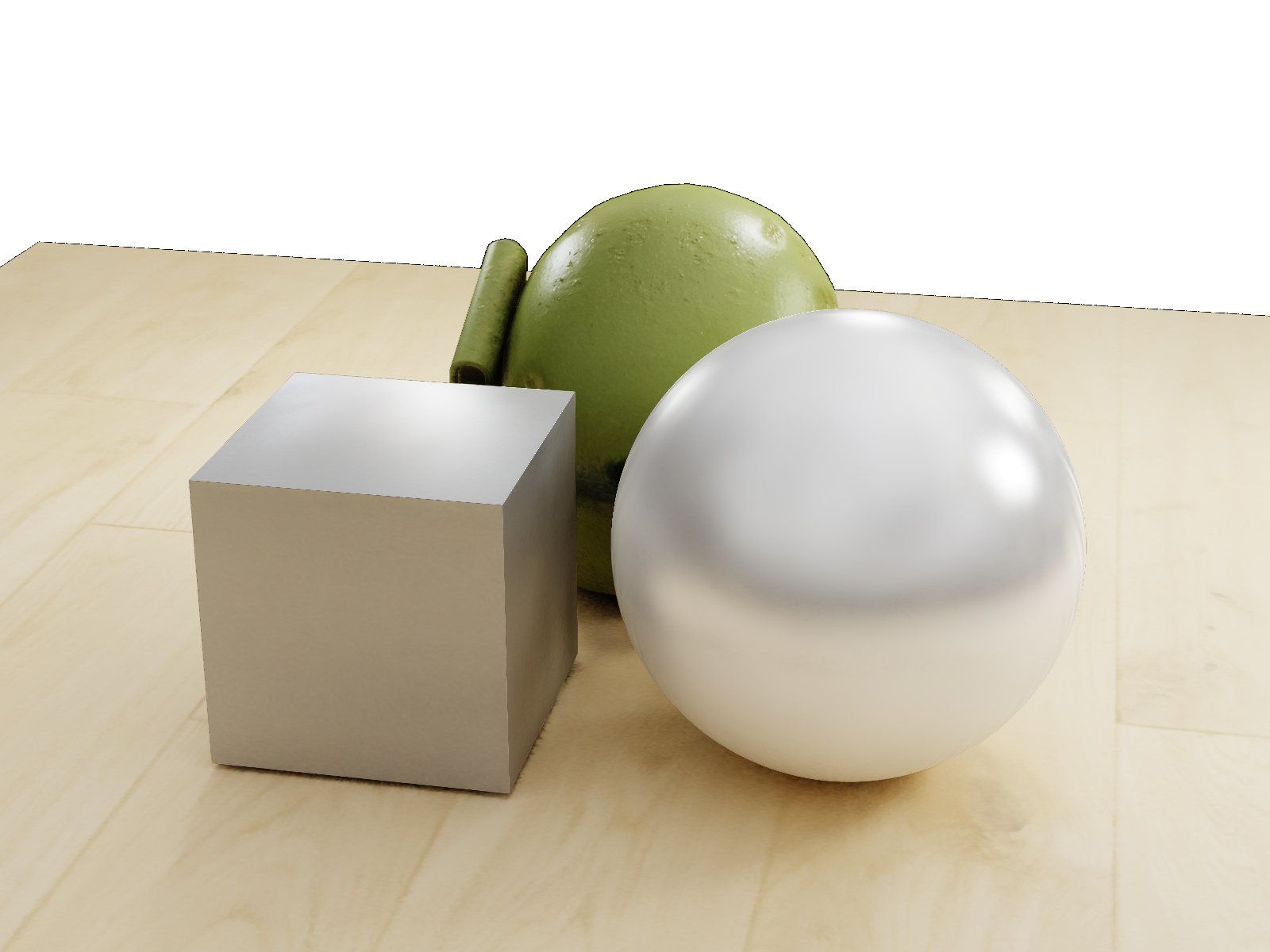}
                        & \includegraphics[width=0.2\linewidth]{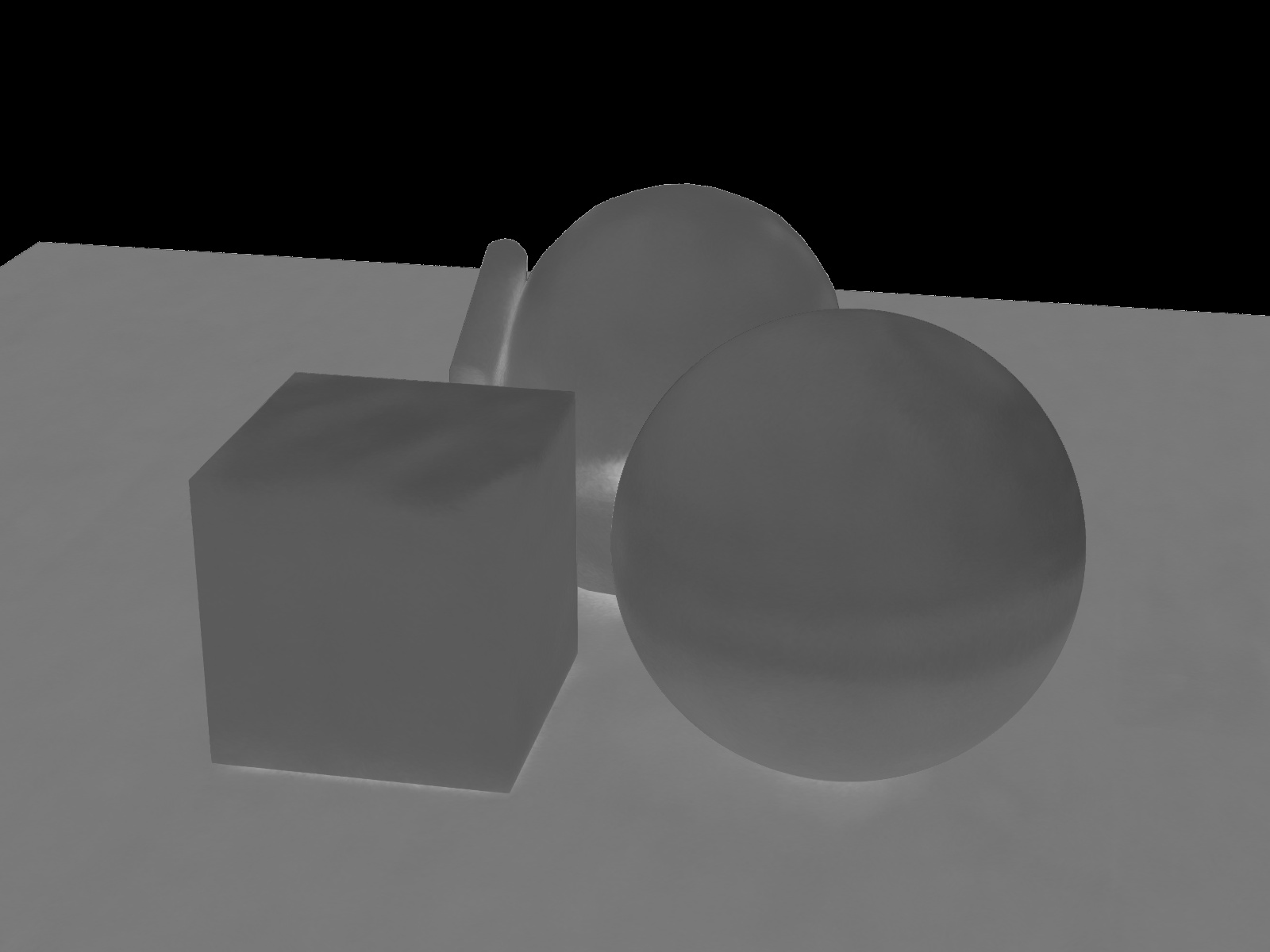}
                        & \includegraphics[width=0.2\linewidth]{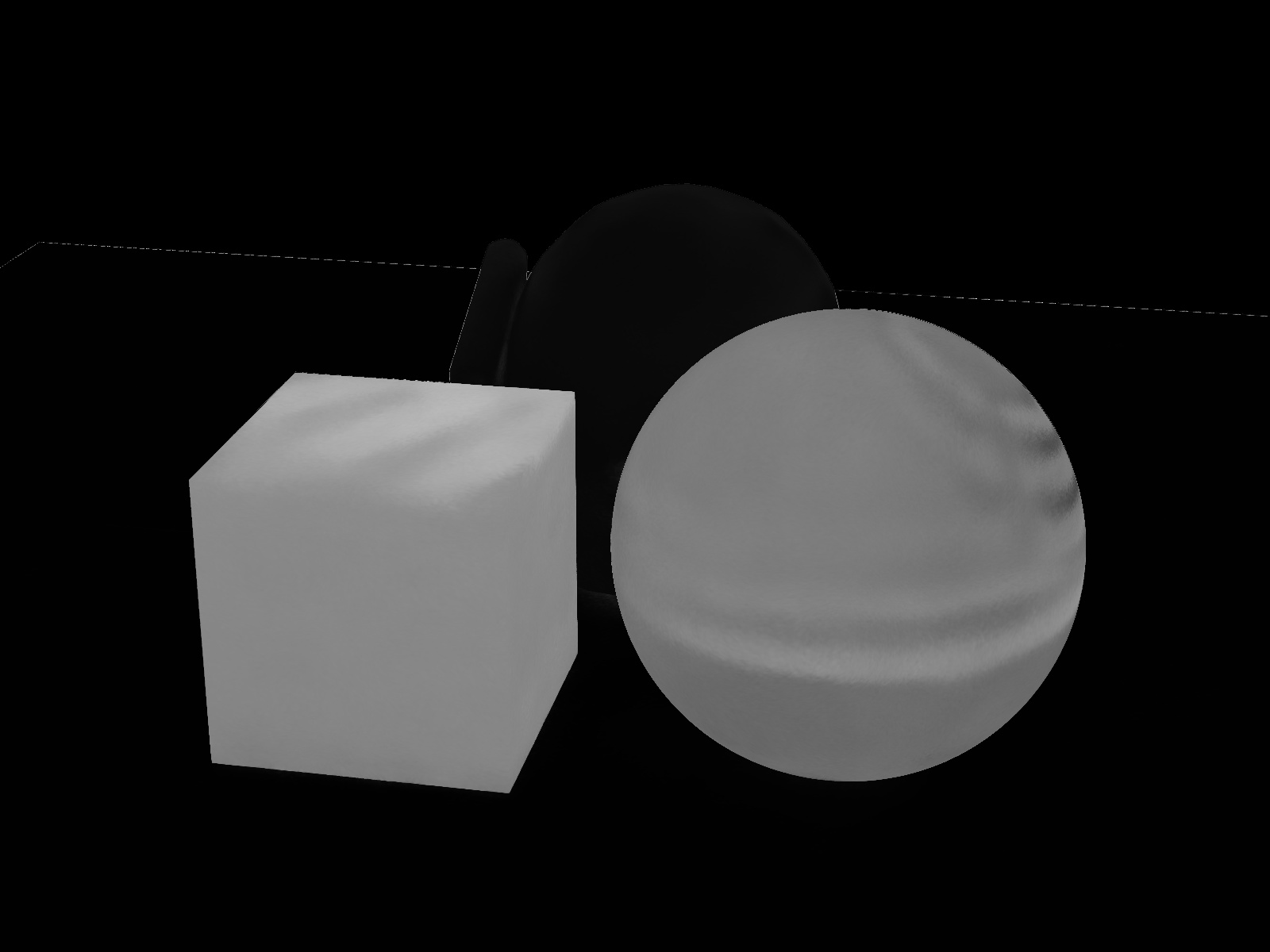}
                        & \includegraphics[width=0.2\linewidth]{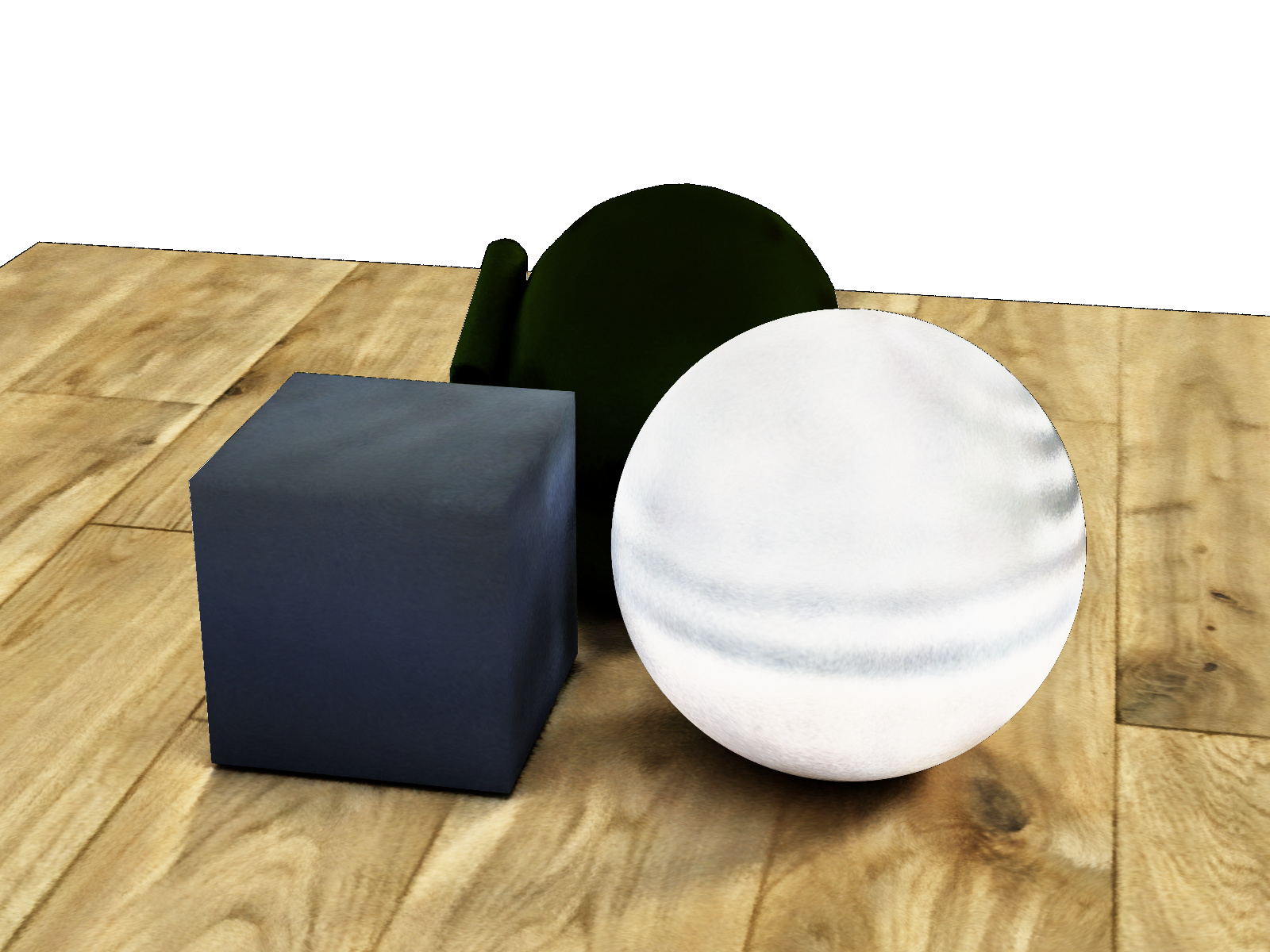} \\
                        & \multirow{1}{*}[0.5in]{\rotatebox[origin=c]{90}{Ours}}
                        & \includegraphics[width=0.2\linewidth]{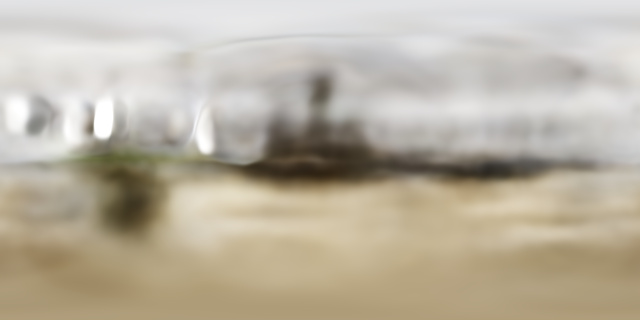}
                        & \includegraphics[width=0.2\linewidth]{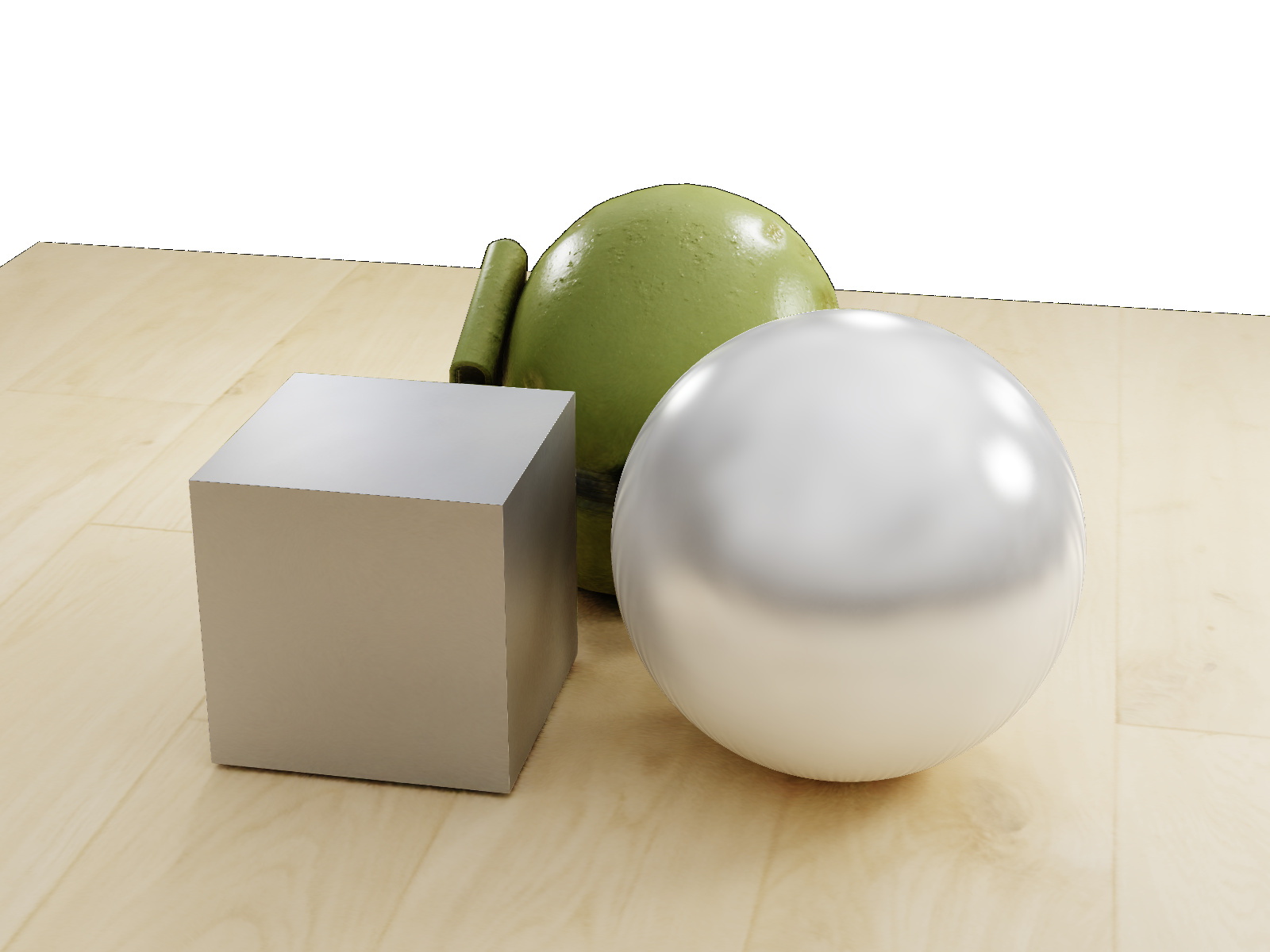}
                        & \includegraphics[width=0.2\linewidth]{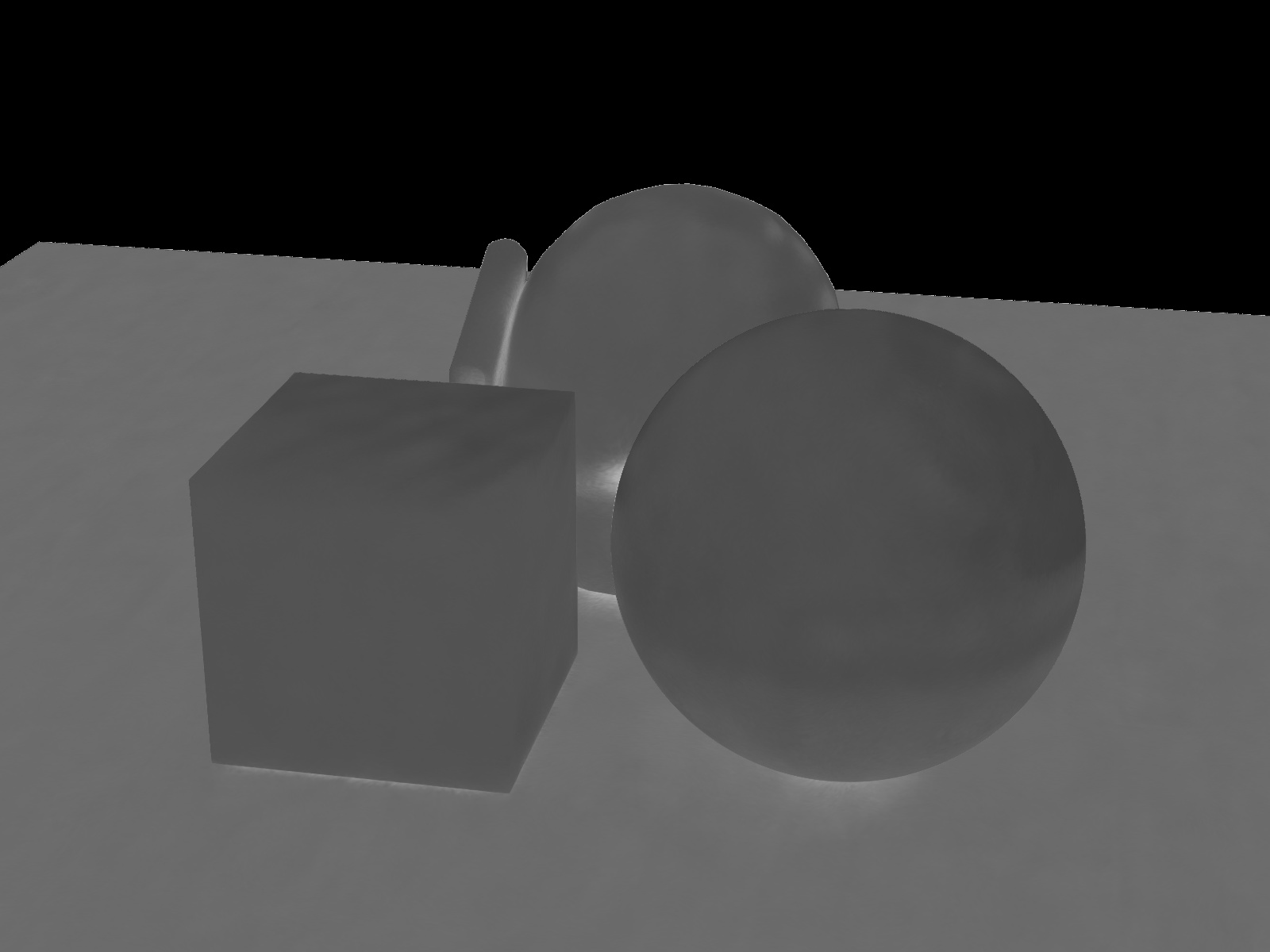}
                        & \includegraphics[width=0.2\linewidth]{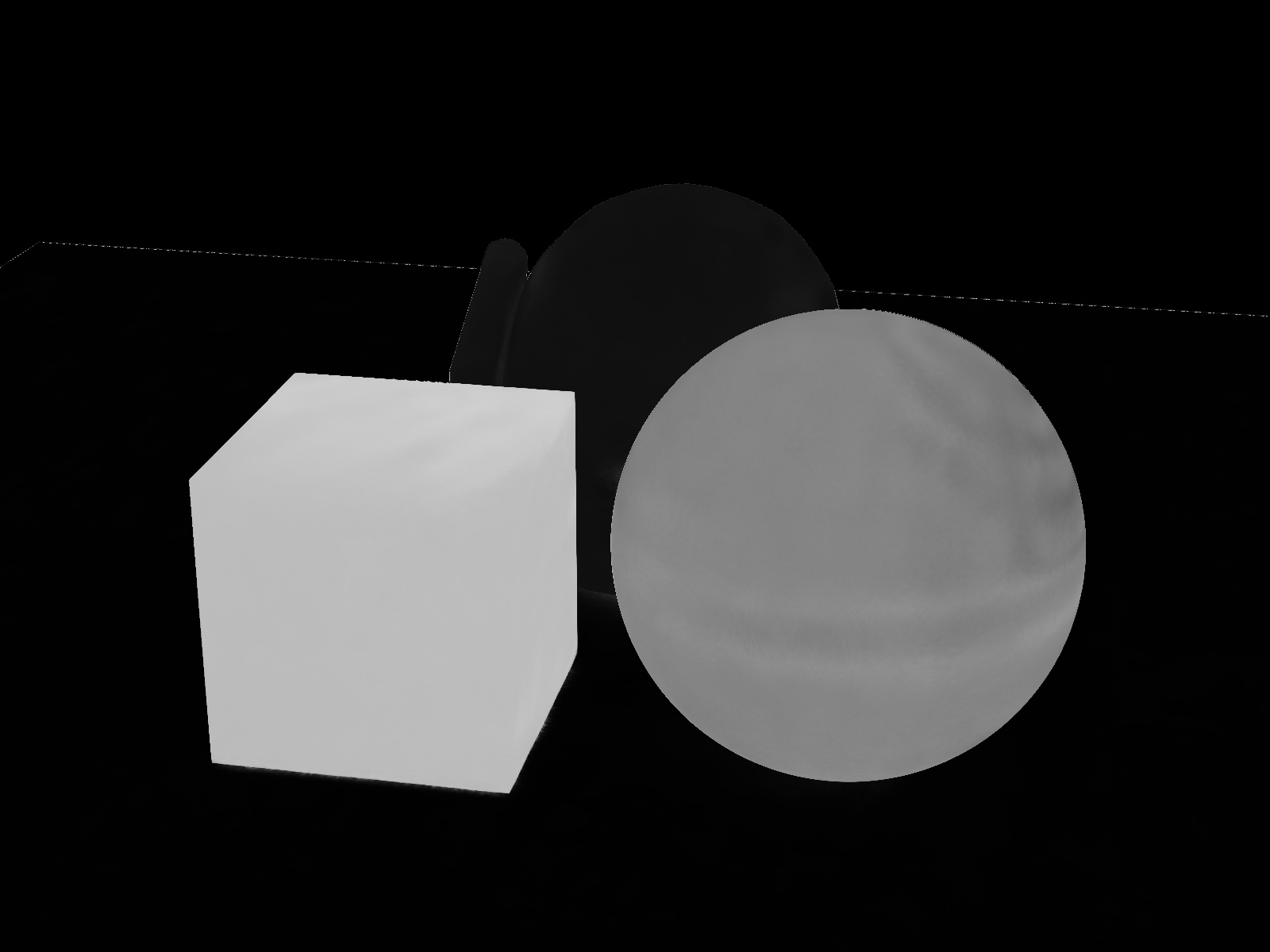}
                        & \includegraphics[width=0.2\linewidth]{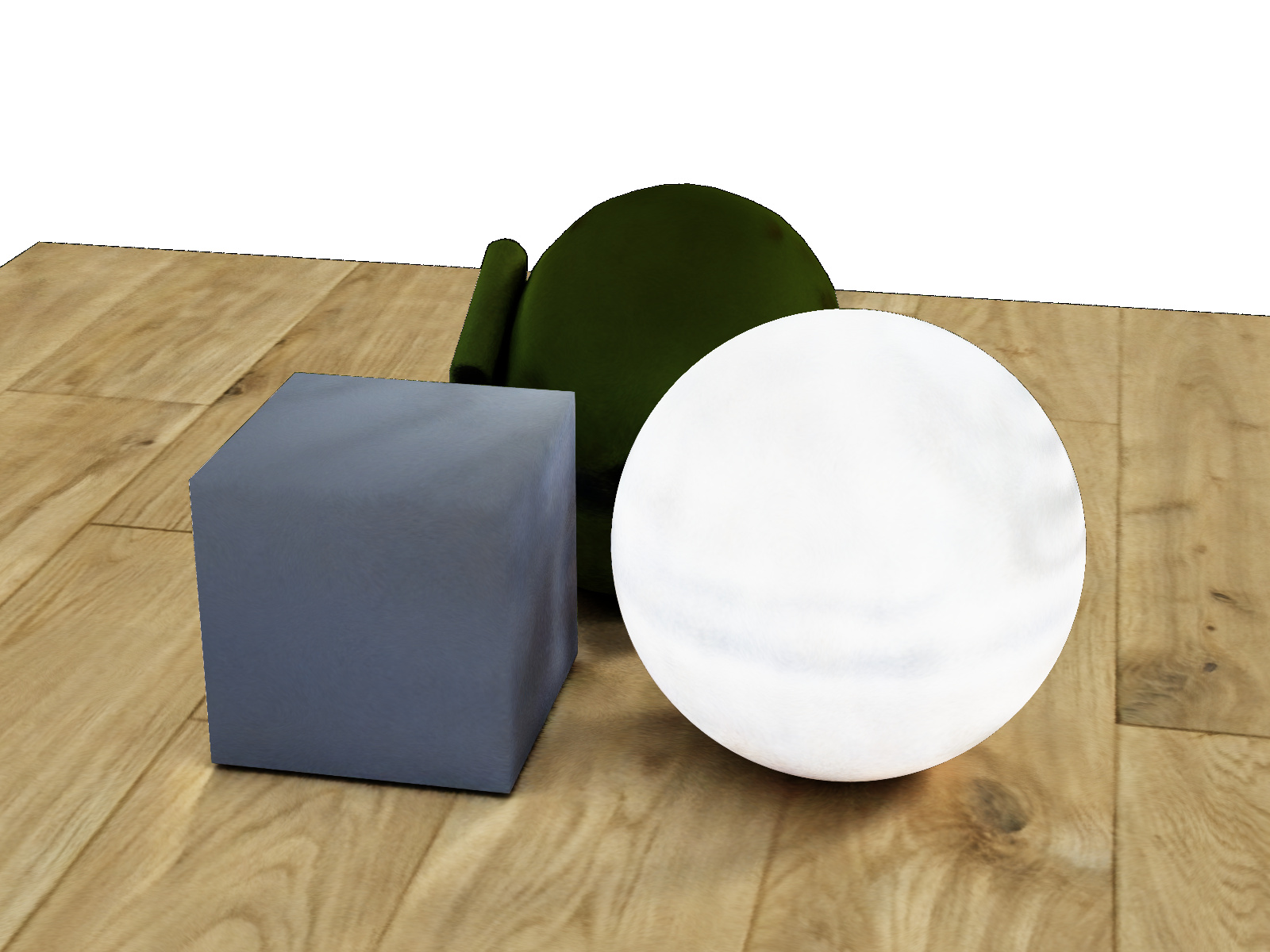} \\
                        & \multirow{1}{*}[0.5in]{\rotatebox[origin=c]{90}{Ground Truth}}
                        &
                        & \includegraphics[width=0.2\linewidth]{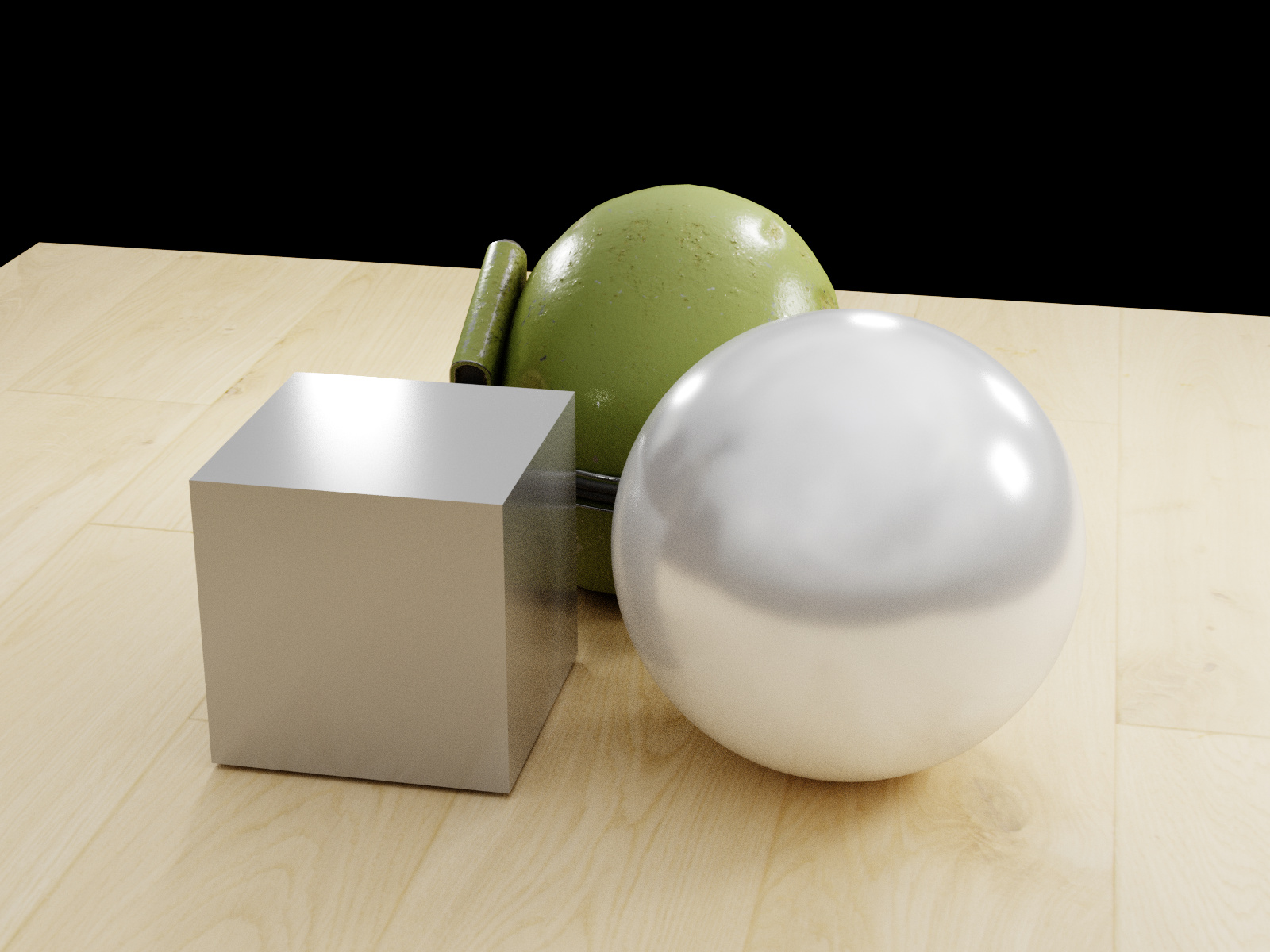}
                        & \includegraphics[width=0.2\linewidth]{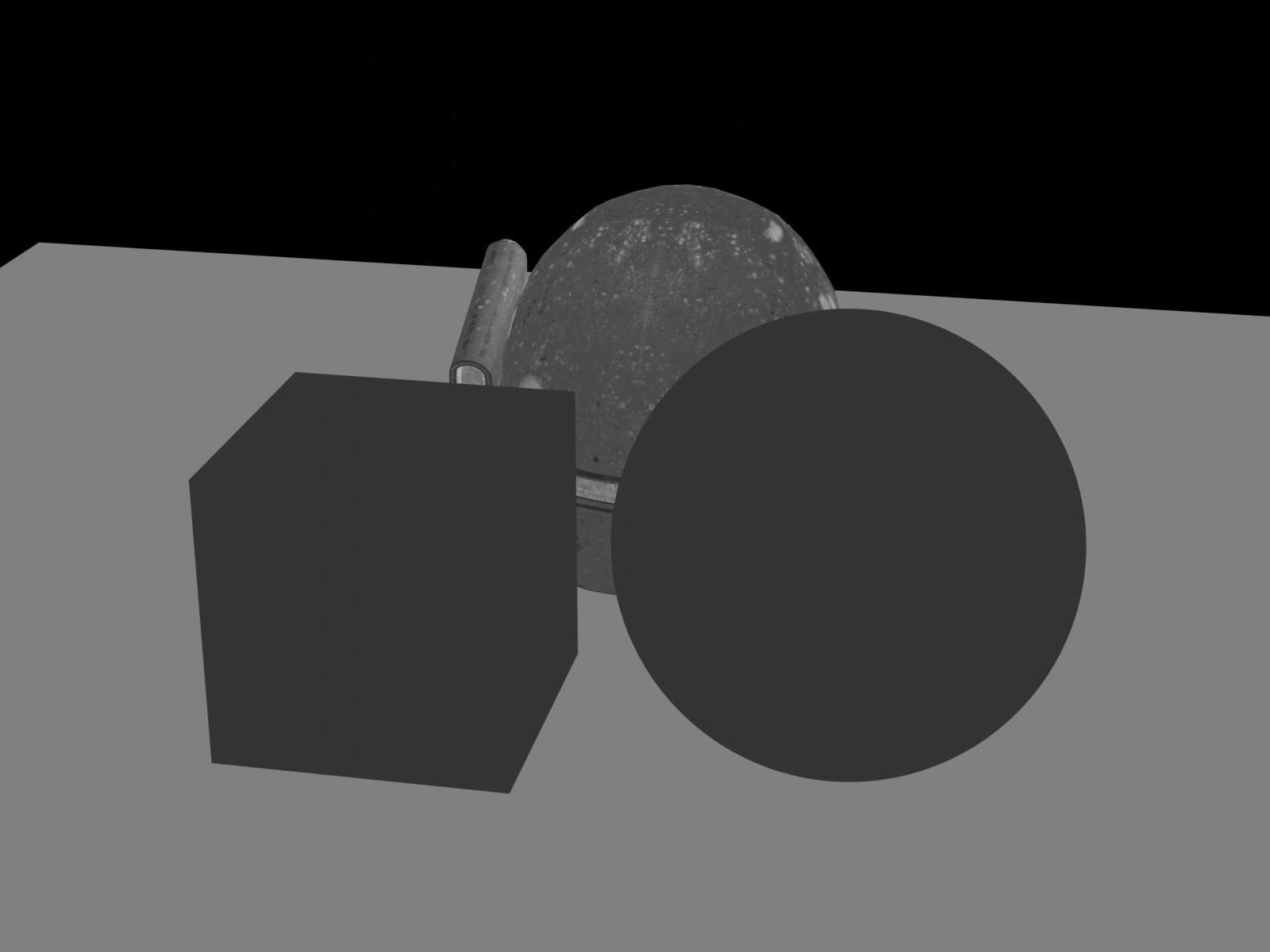}
                        & \includegraphics[width=0.2\linewidth]{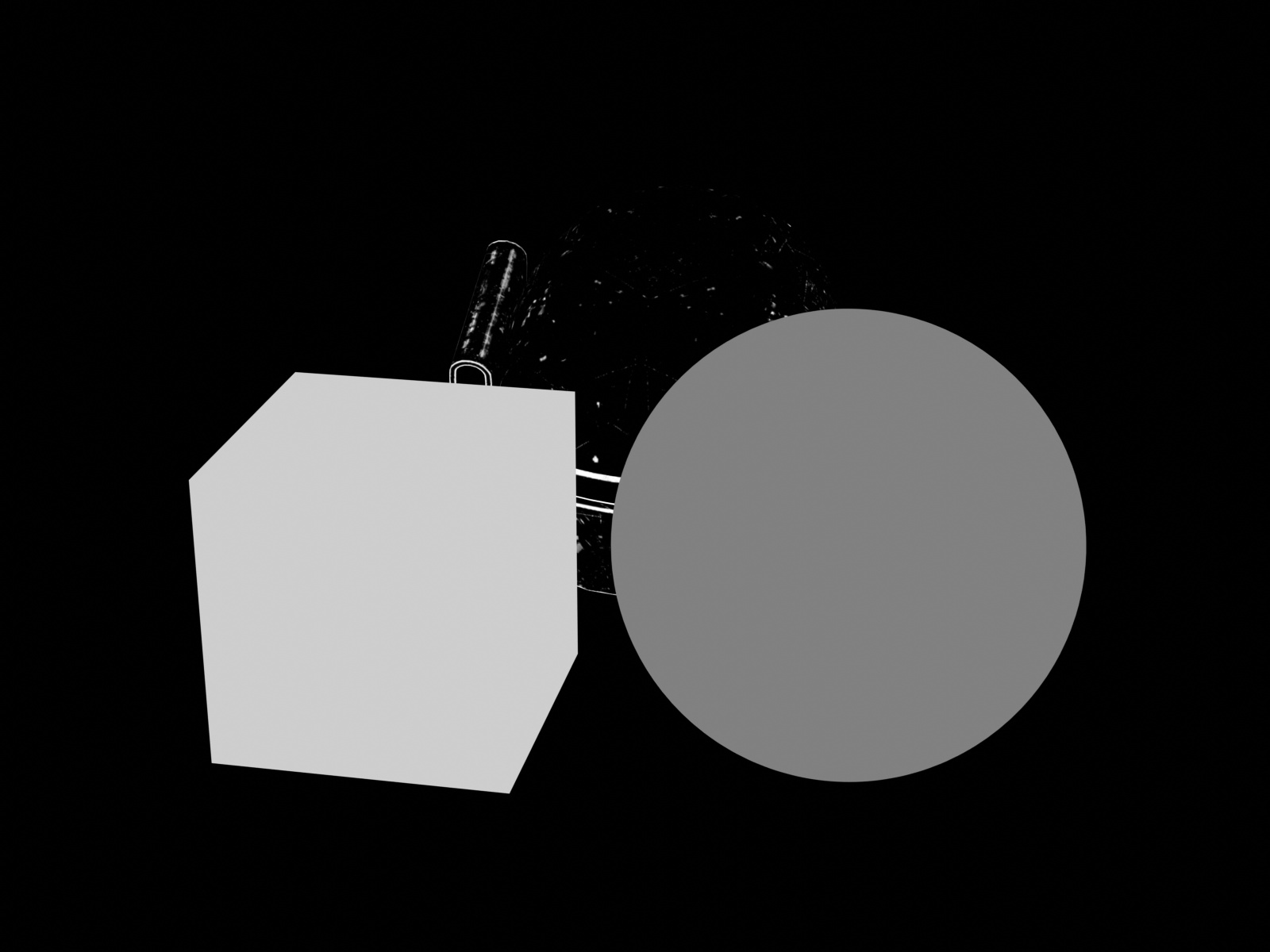}
                        & \includegraphics[width=0.2\linewidth]{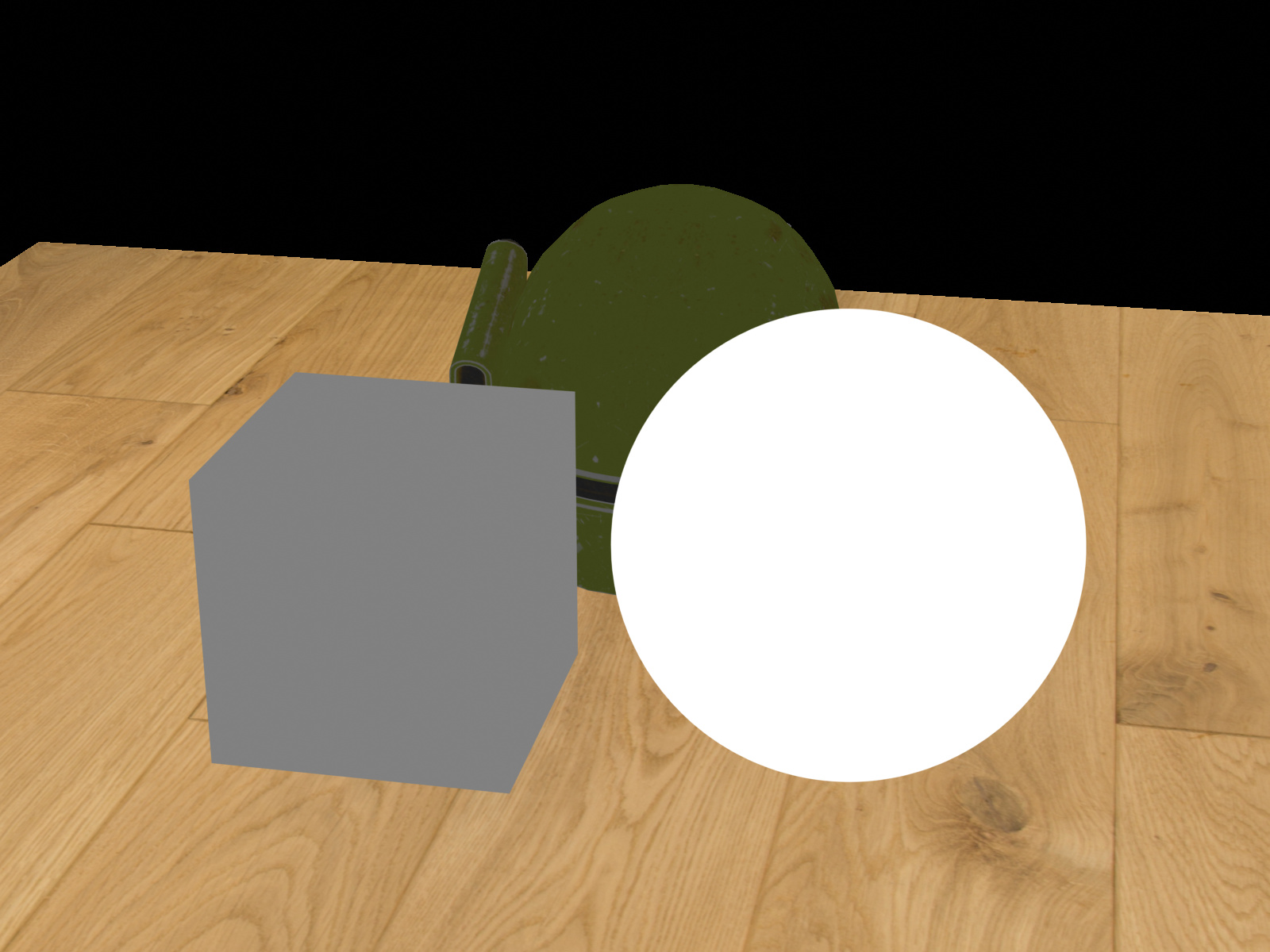} \\
            \end{tabular}
    }
    \vspace{-8pt}
    \caption{\textbf{Qualitative comparisons on the NeILF++ dataset~\cite{zhang2023neilf++}.} \dag: no ground-truth environment maps are provided with the dataset.}
    \label{fig:neilfpp_dataset_qualitative}
\end{figure*}

\begin{table*}[tb]
	\caption{\textbf{Comparison for novel view synthesis against state-of-the-art methods on DTU~\cite{jensen2014dtu}.} Performance is compared across different scenes using the RGB PSNR and Chamfer Distance metrics. \dag: our reproduced results. $\ddagger$: mean result over 3 runs. *: only 1 of either RGB PSNR or Chamfer Distance results is publicly available.}
	\label{tab:dtu_dataset_sota}
    \vspace{-3mm}
	\resizebox{\textwidth}{!}{
		\begin{tabular}{l|l|ccccccccccccccc|c}
			\specialrule{.2em}{.1em}{.1em}
			& Method \textbackslash{} DTU scan        & 24    & 37    & 40    & 55    & 63    & 65    & 69    & 83    & 97   & 105   & 106   & 110   & 114   & 118   & 122   & Mean$\pm\sigma$  \\ \hline
			\multirow{4}{*}{\begin{tabular}[c]{@{}l@{}}RGB \\ PSNR\end{tabular}} & NeRFactor~\cite{zhang2021nerfactor}*      & 23.24 & 21.91 & 23.33 & 26.86 & 22.70 & 24.71 & \textbf{27.59} & 22.56 & 20.45 & 25.08 & 26.30 & 25.14 & 21.35 & 26.44 & 26.53 & 24.28 \\
            & PhySG~\cite{zhang2021physg}* & 17.38 & 15.11 & 20.65 & 18.71 & 18.89 & 18.47 & 18.08 & 21.98 & 17.31 & 20.67 & 18.75 & 17.55 & 21.20 & 18.78 & 23.16 & 19.11 \\
            & NeILF++~\cite{zhang2023neilf++}\dag$\ddagger$ & 25.85 & 21.93 & 25.42 & 27.99 & 29.76 & 27.45 & 26.54 & 28.37 & 23.73 & 28.28 & 31.56 & 30.45 & 26.86 & 31.32 & 32.19 & 27.85$\pm$1.04  \\
            & PBR-NeRF (ours)$\ddagger$ & \textbf{26.11} & \textbf{24.49} & \textbf{27.56} & \textbf{28.30} & \textbf{30.13} & \textbf{27.57} & 27.07 & \textbf{29.89} & \textbf{24.76} & \textbf{28.71} & \textbf{31.83} & \textbf{30.68} & \textbf{27.24} & \textbf{31.64} & \textbf{33.48} & \textbf{28.63} $\pm$ 0.29 \\
            \hline
            \multirow{4}{*}{\begin{tabular}[c]{@{}l@{}}Chamfer \\ Distance \\ {[mm]} \end{tabular}}
            & VolSDF~\cite{yariv2021volsdf}* & 0.810 & 1.140 & 0.490 & 1.250 & 0.700 & 0.720 & 1.290 & 1.180 & \textbf{1.260} & \textbf{0.700} & 0.660 & 1.080 & 0.420 & 0.610 & 0.550 & 0.857 \\
            & NeuS~\cite{wang2021neus}* & 0.930 & \textbf{1.000} & \textbf{0.430} & 1.100 & \textbf{0.650} & \textbf{0.570} & 1.480 & \textbf{1.090} & 1.370 & 0.830 & \textbf{0.520} & 1.200 & \textbf{0.350} & \textbf{0.490} & 0.540 & 0.837 \\
            & NeILF++~\cite{zhang2023neilf++}\dag$\ddagger$ & \textbf{0.748} & 3.028 & 1.285 & \textbf{0.448} & 1.627 & 0.640 & 1.287 & 1.649 & 2.697 & 1.008 & 0.604 & 0.745 & 0.397 & 0.513 & 0.824 & 1.167$\pm$0.68 \\
            & PBR-NeRF (ours)$\ddagger$ & 0.816 & 1.343 & 0.486 & 0.511 & 1.665 & 0.688 & \textbf{0.647} & 1.174 & 1.416 & 1.038 & 0.590 & \textbf{0.676} & 0.398 & 0.502 & \textbf{0.468} & \textbf{0.828}$\pm$0.09 \\
			\specialrule{.2em}{.1em}{.1em}
        \end{tabular}
    }
    \vspace{-6mm}
\end{table*}

\subsection{Comparison with The State of The Art}

\PAR{Quantitative Results.}
Table~\ref{tab:neilfpp_dataset_sota} shows that PBR-NeRF achieves state-of-the-art material estimation on the NeILF++ dataset, outperforming previous methods by 1–3 PSNR on average across albedo, metallicness, and roughness.
We outperform all baselines in roughness and metallicness PSNR in most scenes across all illuminations.
For albedo, we observe substantial gains in PSNR and SSIM over all competitors both on overall average and in Mix lighting scenarios, which are generally more complex due to their combined lighting setups.
Our closest competitor for albedo estimation is SG-Env, which combines the PhySG~\cite{zhang2021physg} Spherical Gaussian (SG) lighting representation with a spatially-varying microfacet BRDF.
SG-Env estimates albedo better on the Env scenes since its explicit SG lighting models environment lighting better than PBR-NeRF's under-constrained implicit lighting MLP.
However, PBR-NeRF performs over 3 albedo PSNR better than SG-Env on the more challenging Mix scenes, as SGs cannot model spatially varying illumination such as mixed light sources,
indirect lighting, and occlusions.
While SG-Env struggles with mixed, SV lighting, PBR-NeRF estimates albedo equally well \emph{across} environment and mixed lighting.
Despite performing slightly worse than NeILF~\cite{yao2022neilf} and NeILF++~\cite{zhang2023neilf++} in metallicness SSIM, we outperform all competitors in all other material estimation metrics.

Among inverse rendering methods, PBR-NeRF achieves state-of-the-art results in novel view synthesis and geometry estimation on DTU, as shown in Table~\ref{tab:dtu_dataset_sota}. We report the average results and standard deviation over three training runs both for our approach and NeILF++~\cite{zhang2023neilf++}, mitigating the impact of training variability. Our method achieves an improvement of 0.78 in mean PSNR and 0.339 mm in Chamfer Distance over NeILF++ with lower variance, showing that our proposed losses positively impact geometry estimation too.
Additionally, PBR-NeRF ranks first in 14 scenes for RGB PNSR, with PSNR gains ranging from 0.12 to 2.56 over NeILF++, demonstrating consistent improvements over our closest baseline, and it substantially outperforms all other methods.

\begin{figure*}[tb]
    \renewcommand{\arraystretch}{1.5}
    \centering
    \small
    \resizebox{\textwidth}{!}{
        \begin{tabular}{cccc|ccc|ccc}
            & \multicolumn{3}{c|}{DTU Scan 37} & \multicolumn{3}{c|}{DTU Scan 63} & \multicolumn{3}{c}{DTU Scan 105} \\
            \rotatebox{90}{RGB}
            & \includegraphics[width=0.12\linewidth]{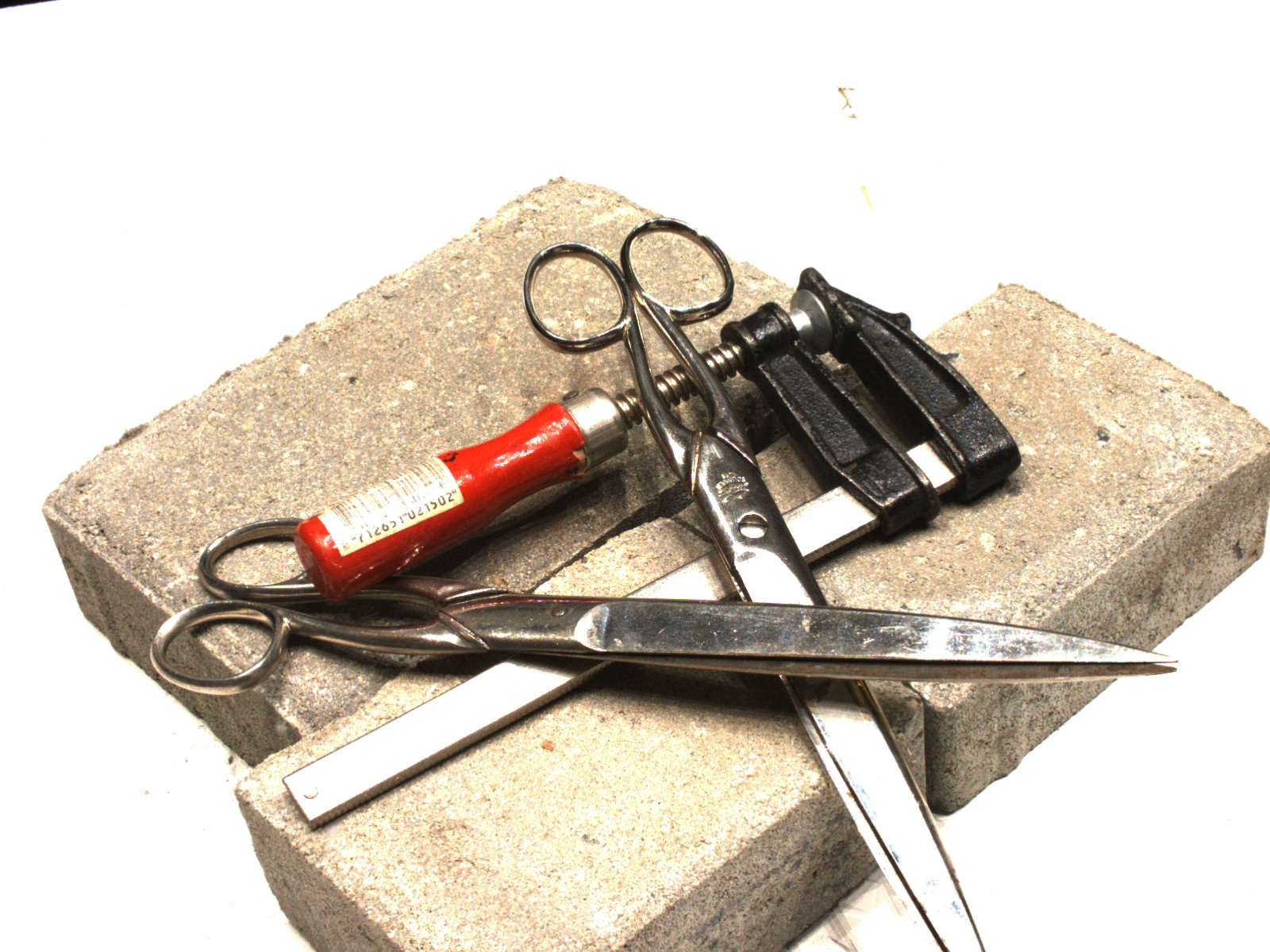}
            & \includegraphics[width=0.12\linewidth]{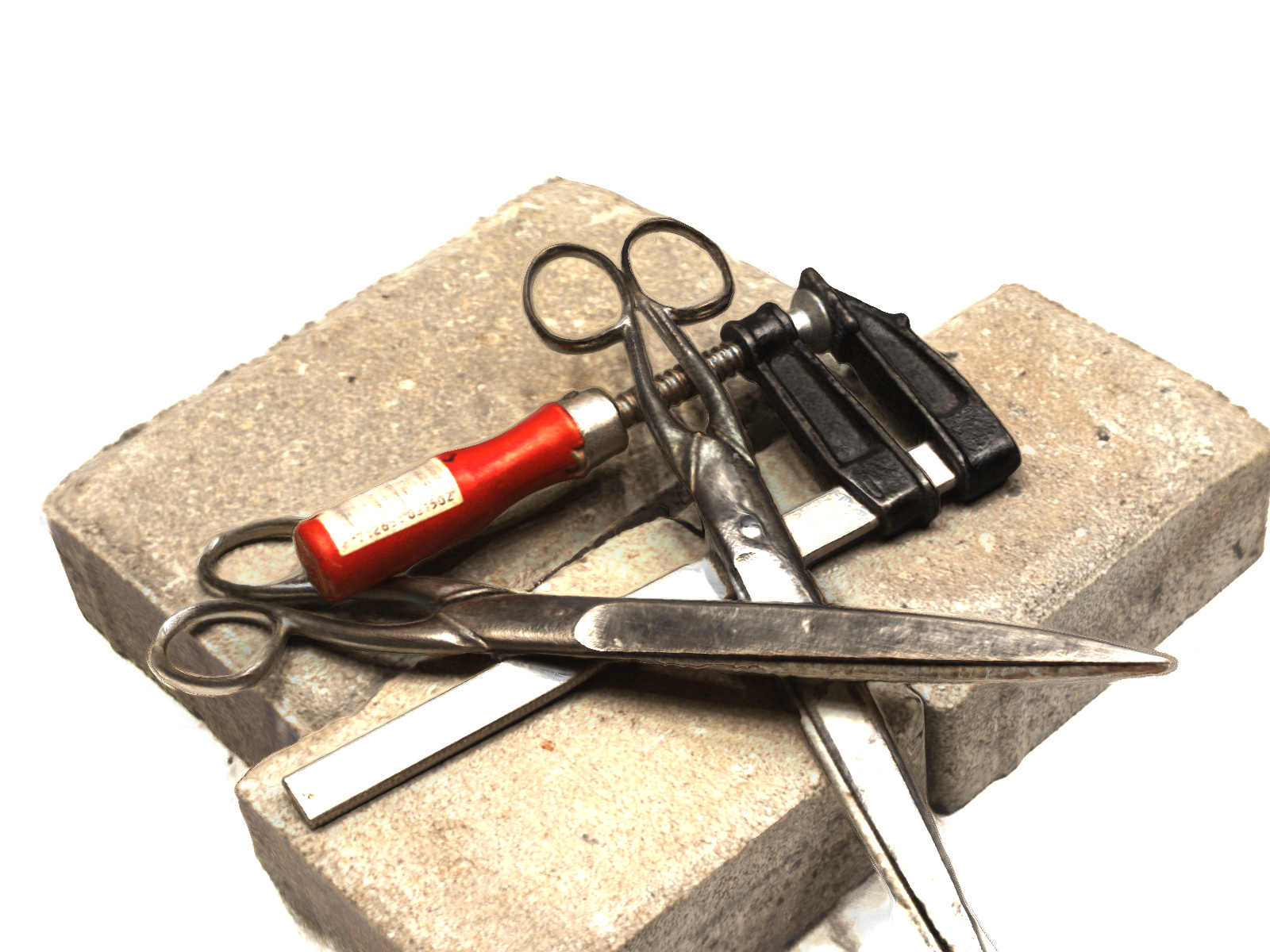}
            & \includegraphics[width=0.12\linewidth]{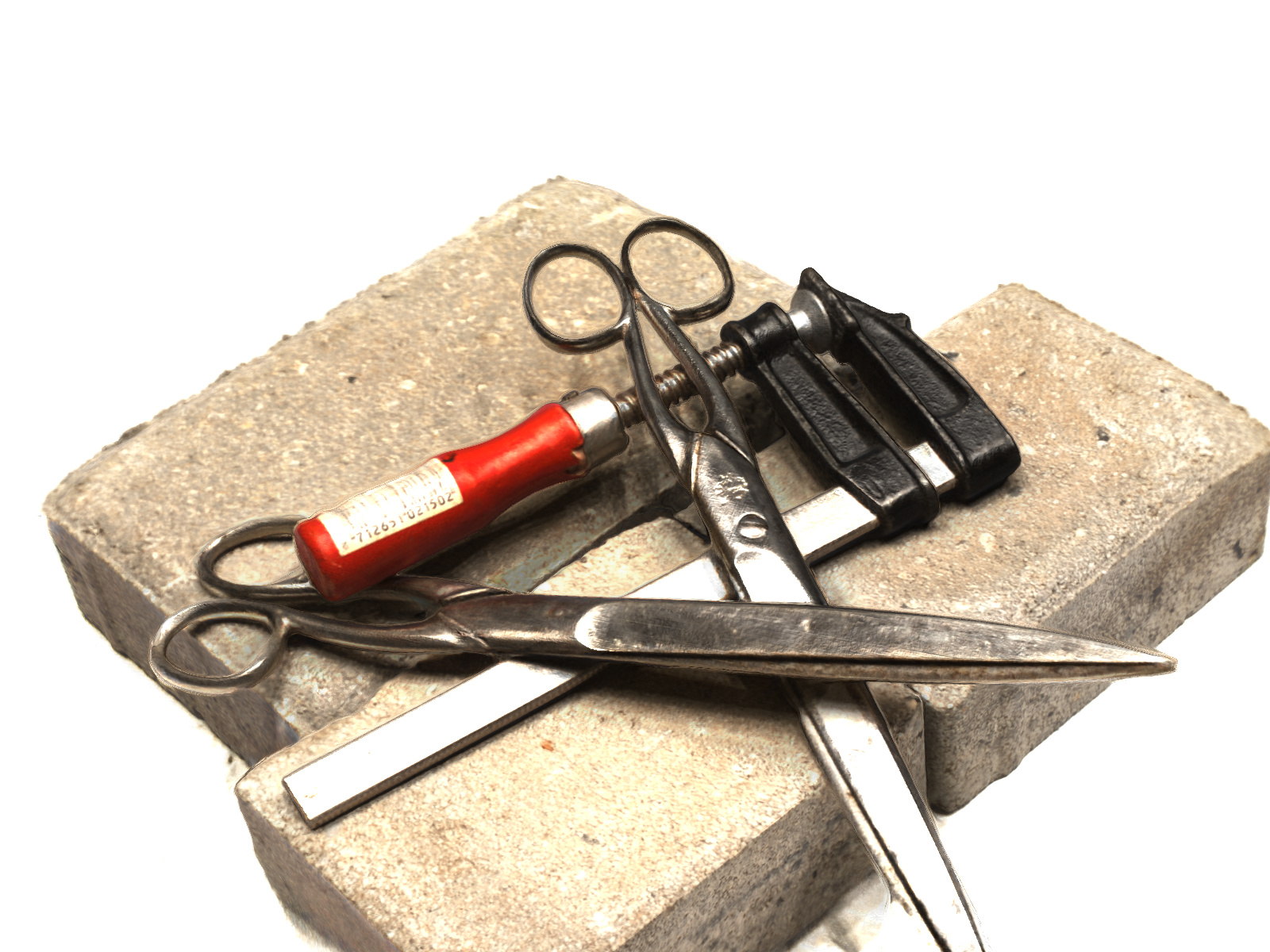}
            & \includegraphics[width=0.12\linewidth]{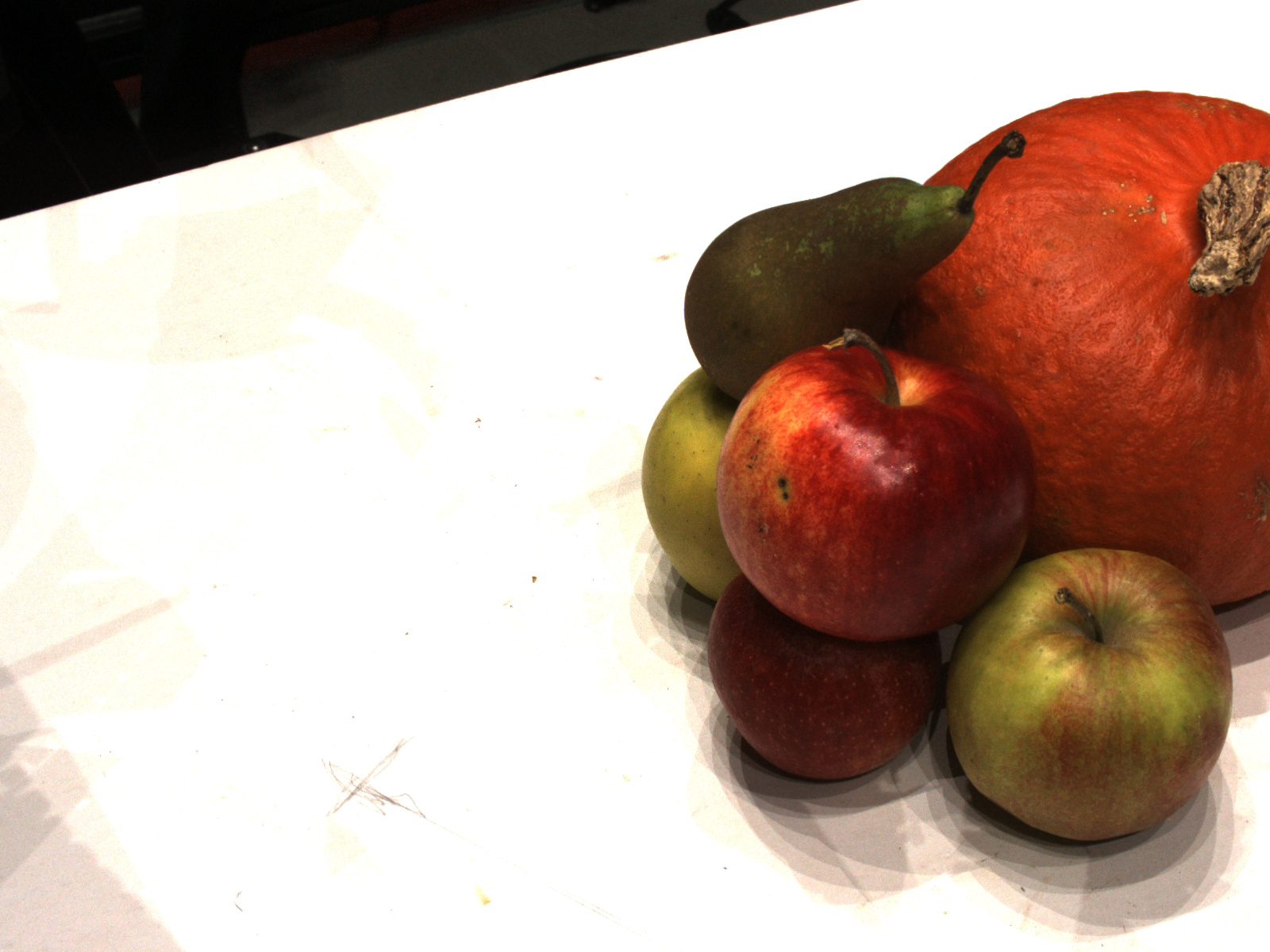}
            & \includegraphics[width=0.12\linewidth]{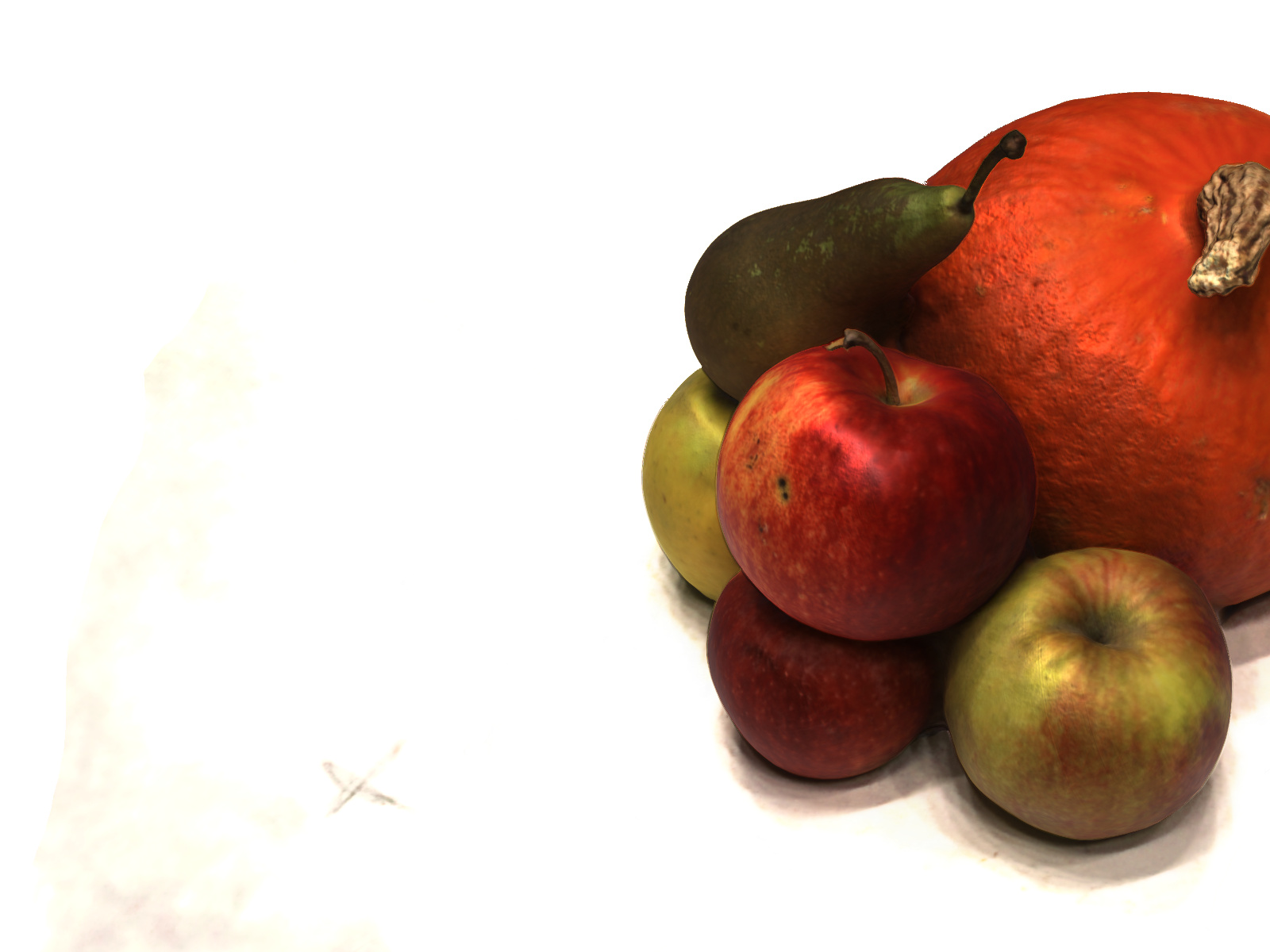}
            & \includegraphics[width=0.12\linewidth]{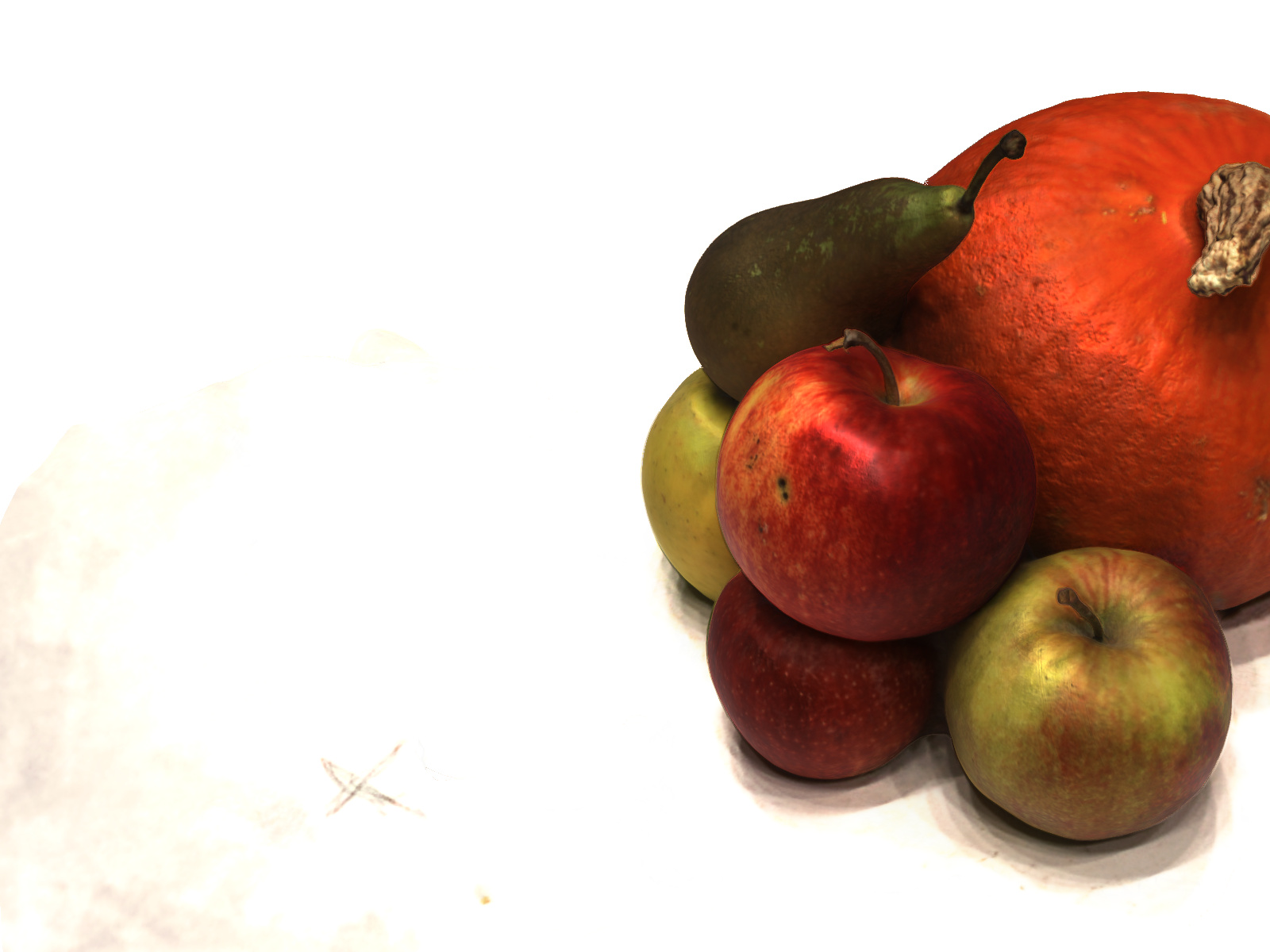}
            & \includegraphics[width=0.12\linewidth]{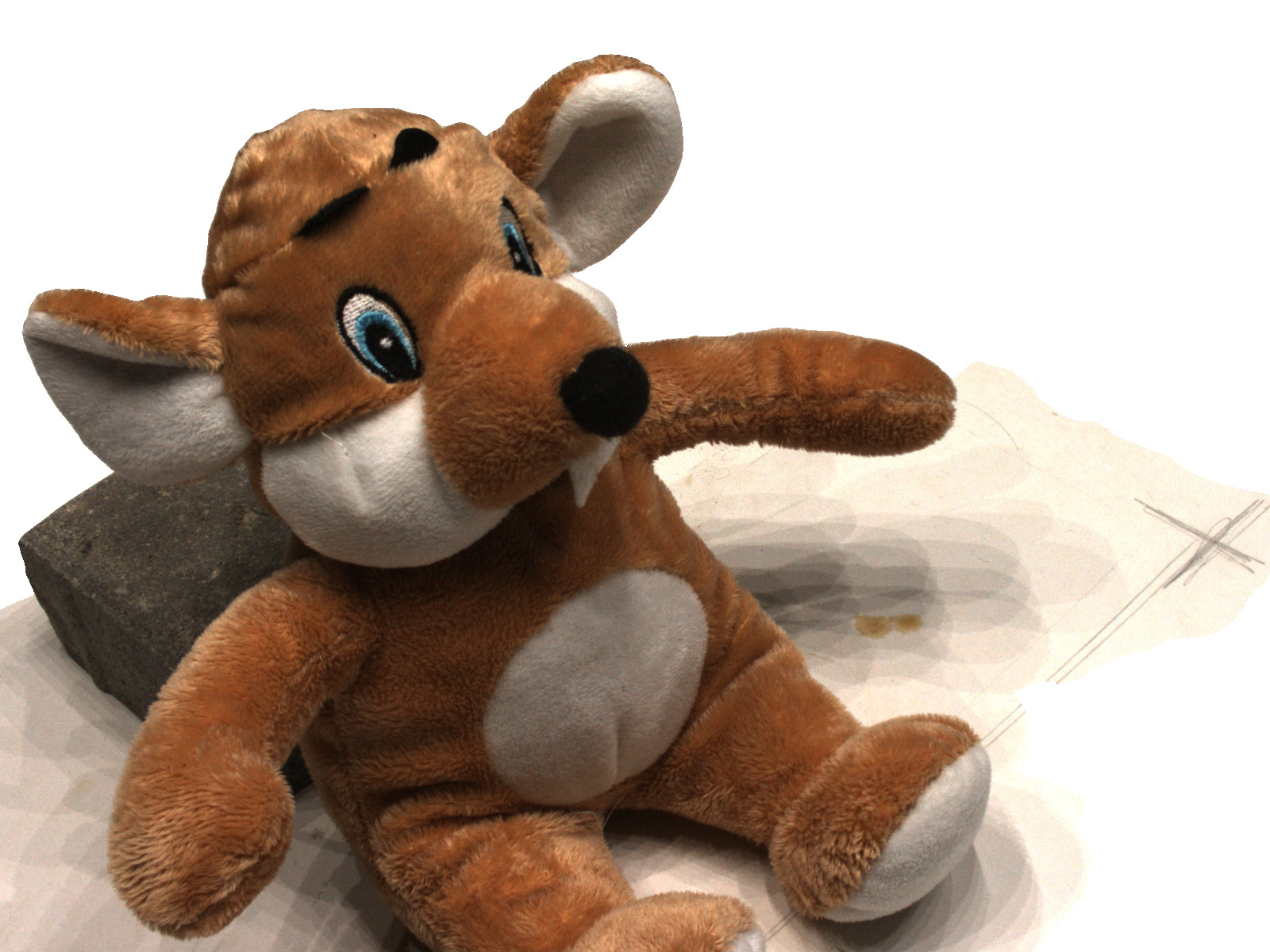}
            & \includegraphics[width=0.12\linewidth]{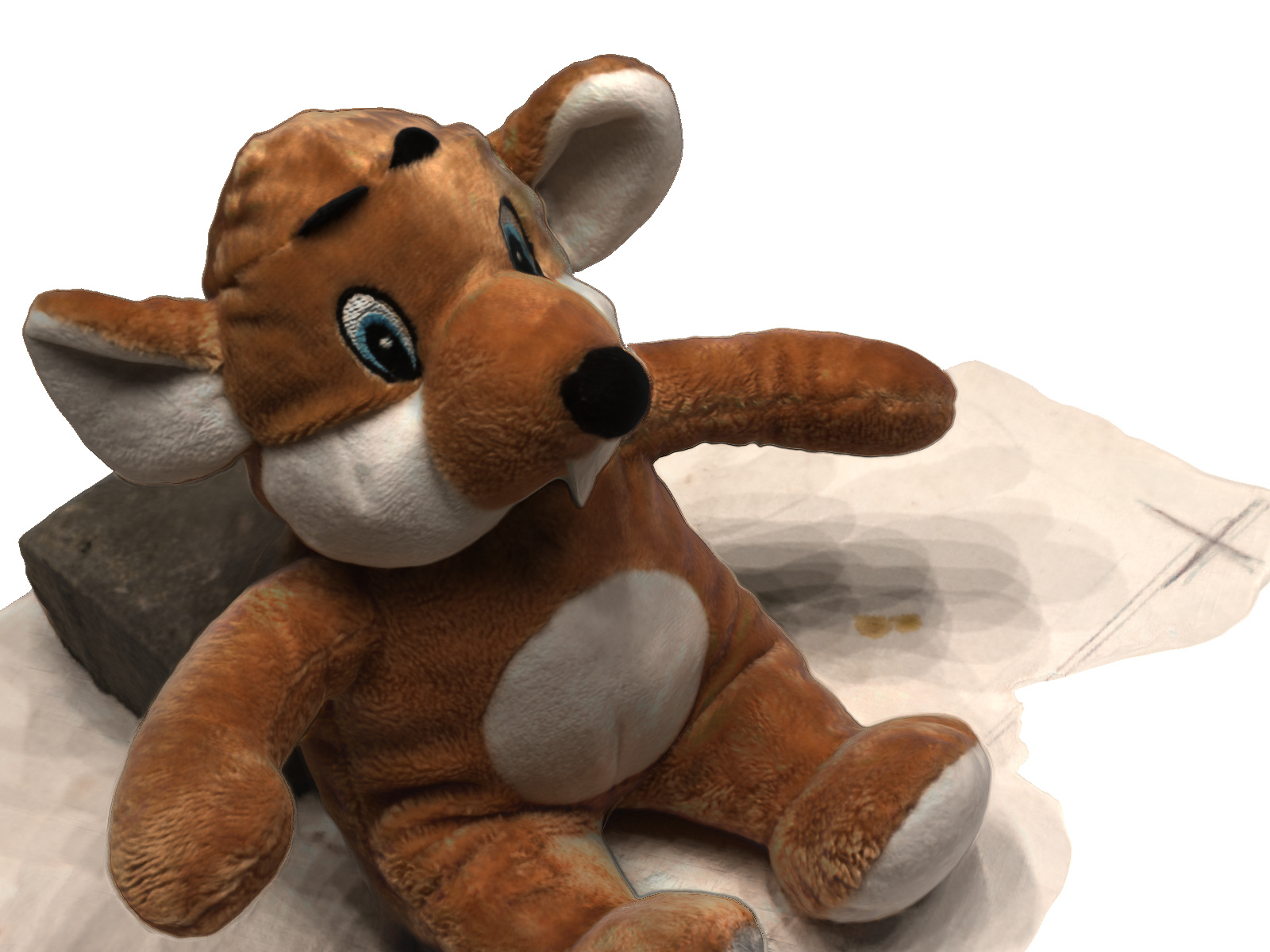}
            & \includegraphics[width=0.12\linewidth]{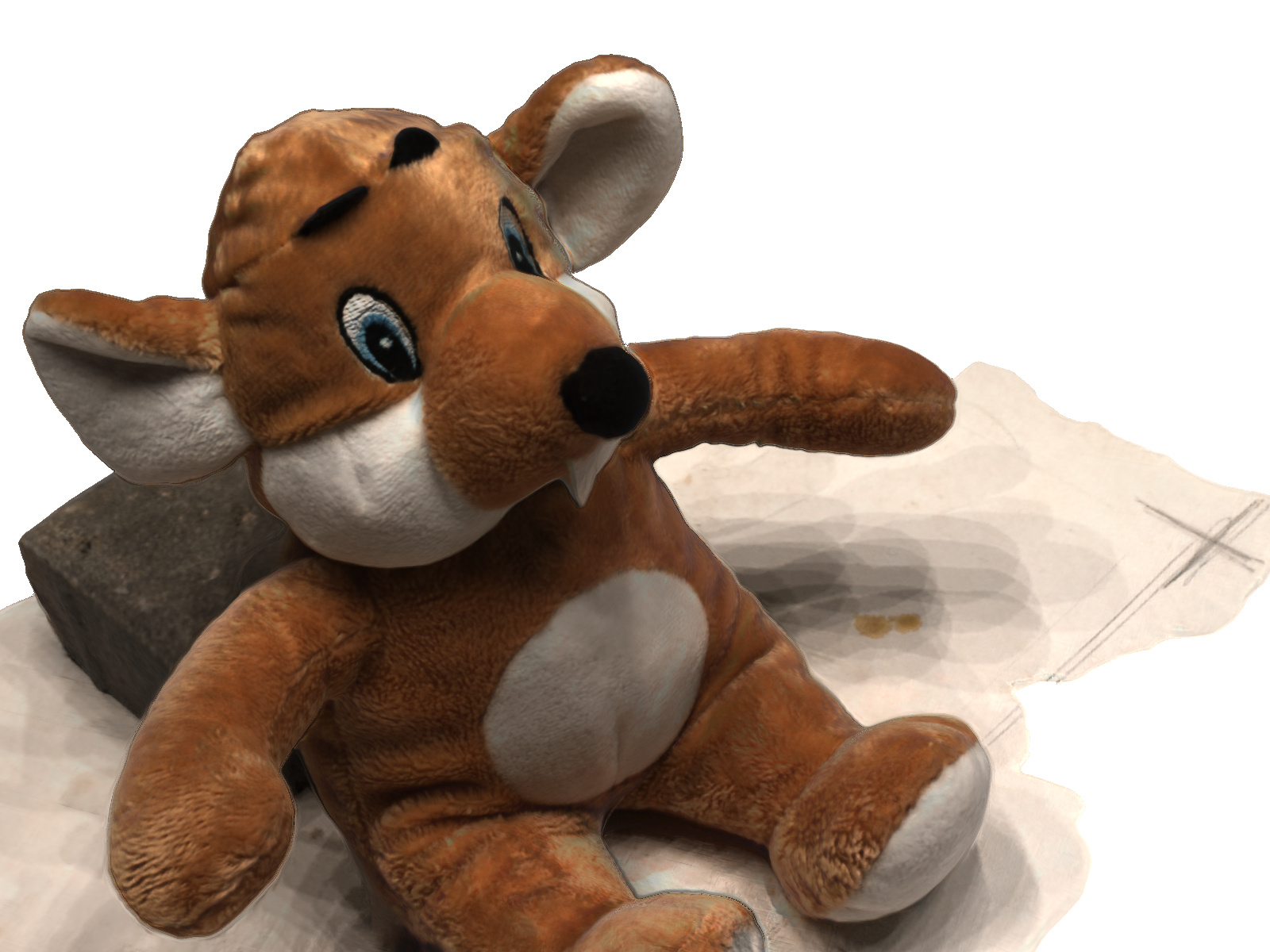} \\
            \rotatebox{90}{Normals}
            & \includegraphics[width=0.12\linewidth]{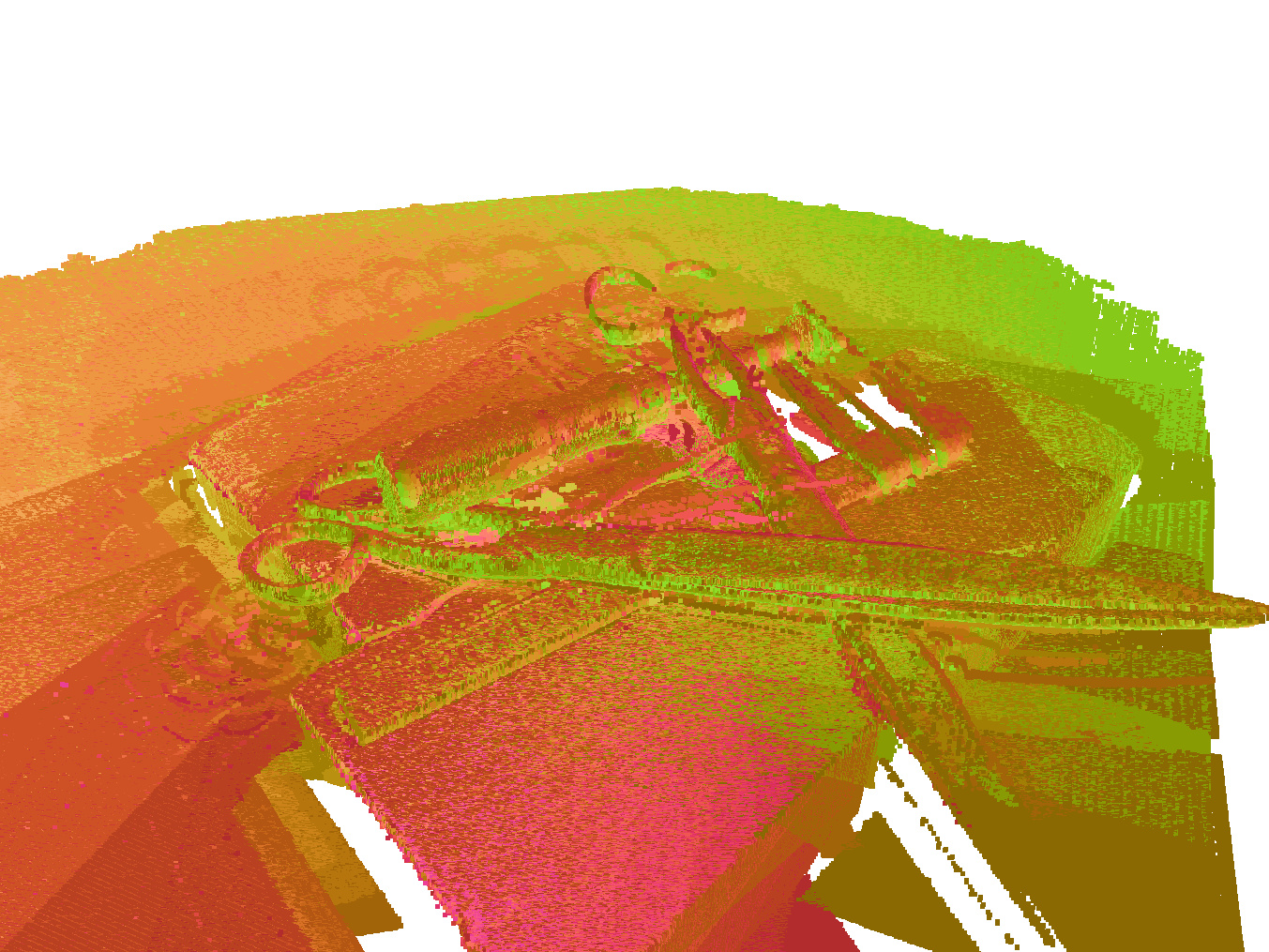}
            & \includegraphics[width=0.12\linewidth]{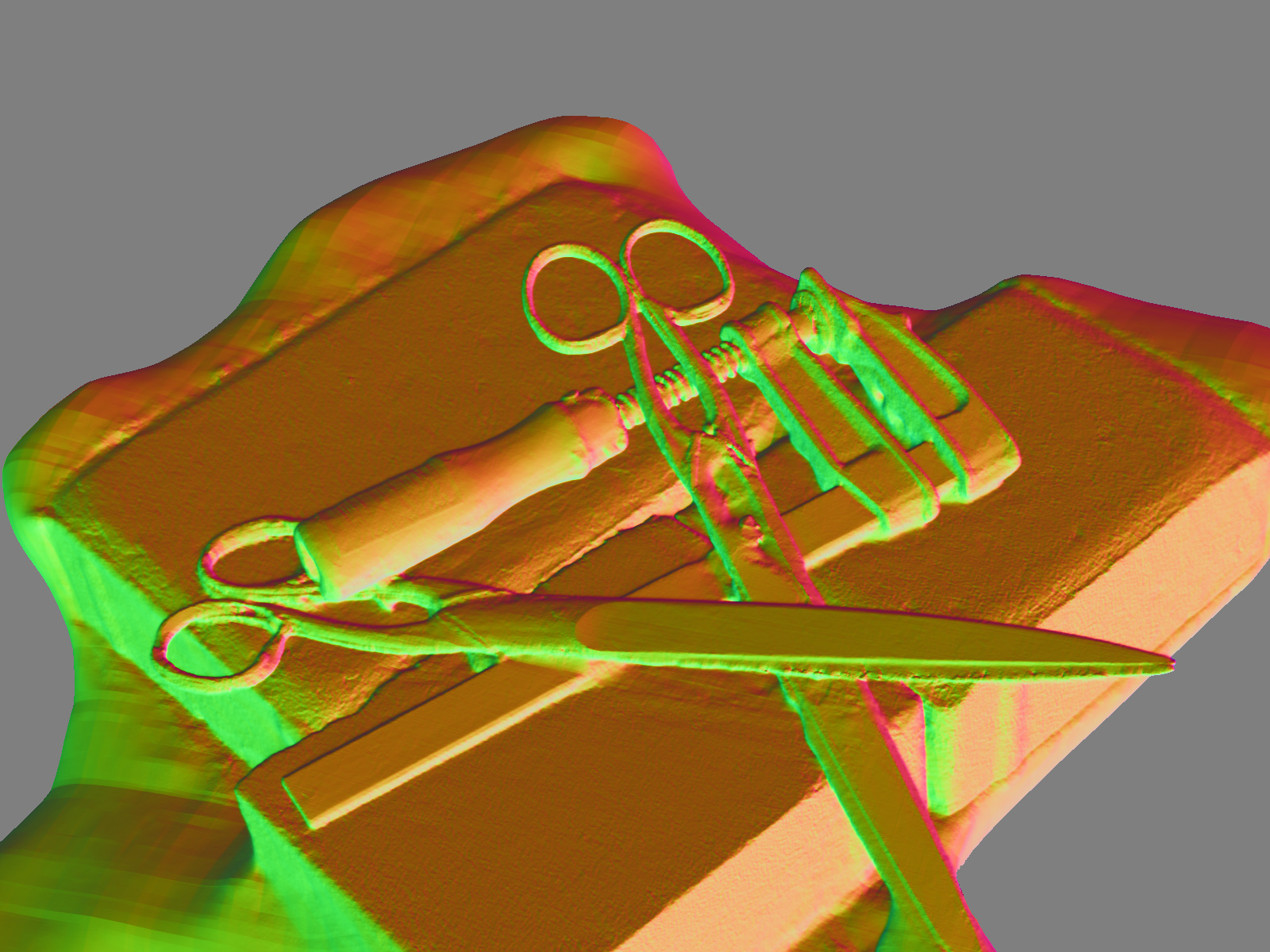}
            & \includegraphics[width=0.12\linewidth]{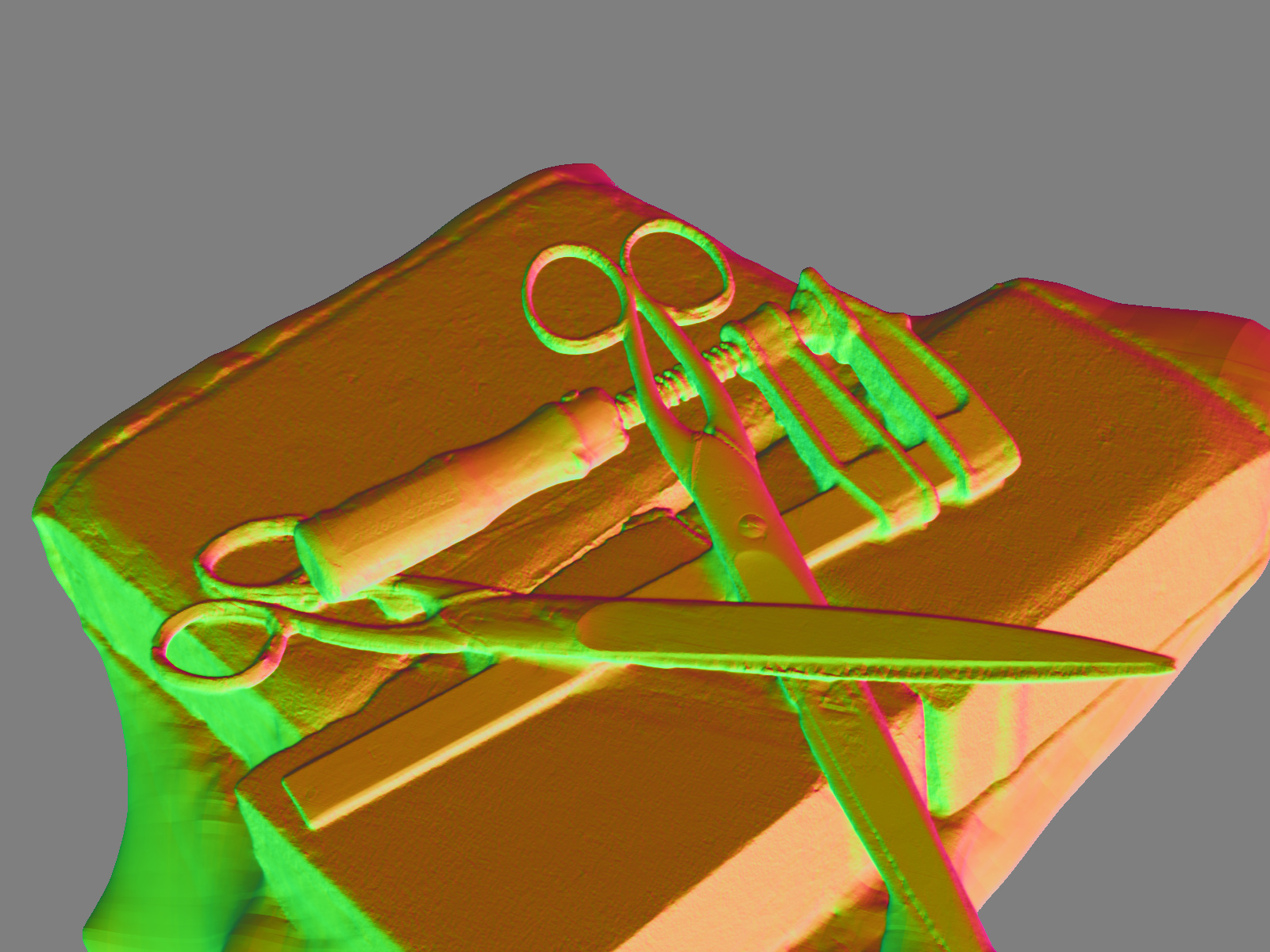}
            & \includegraphics[width=0.12\linewidth]{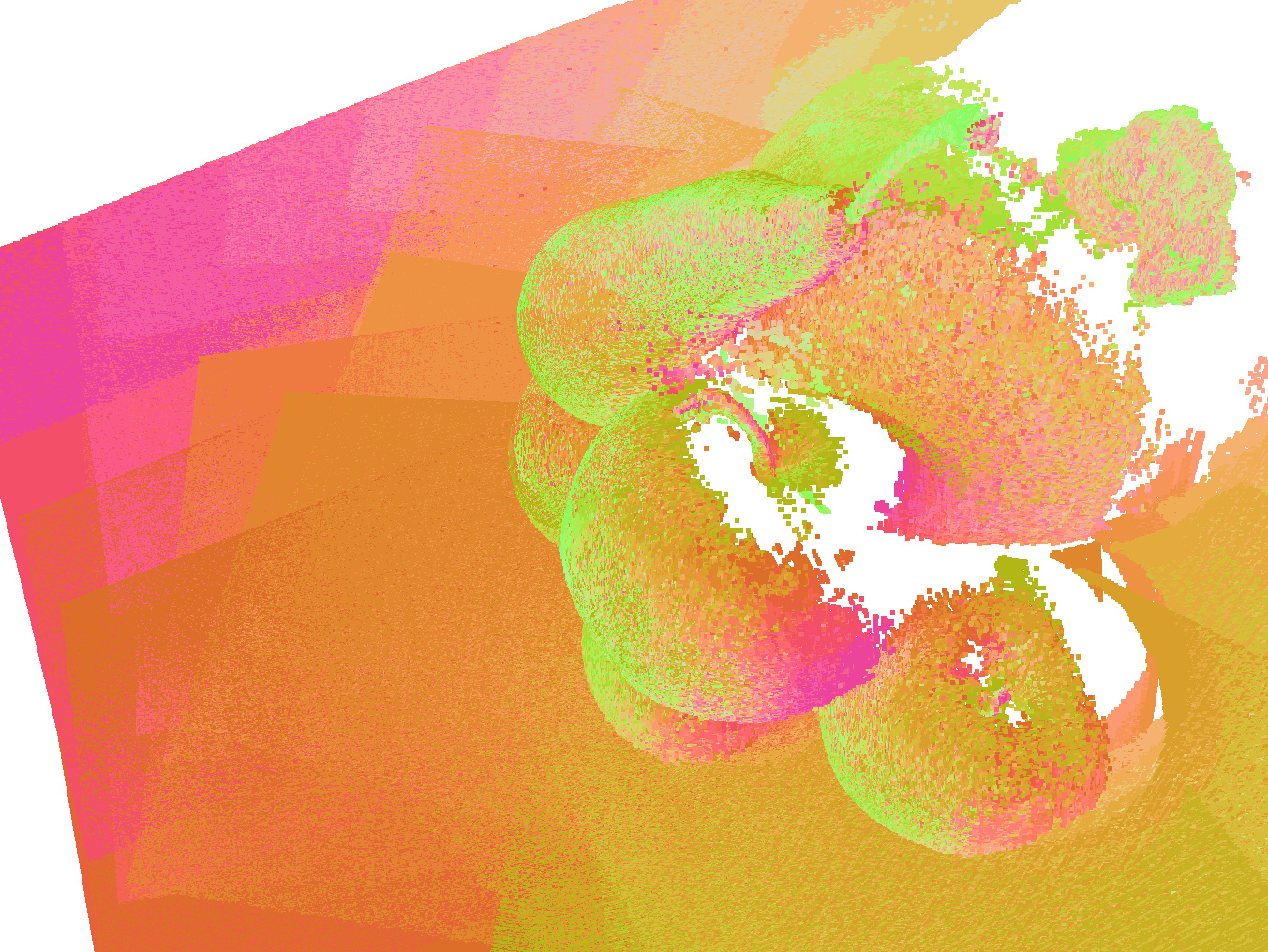}
            & \includegraphics[width=0.12\linewidth]{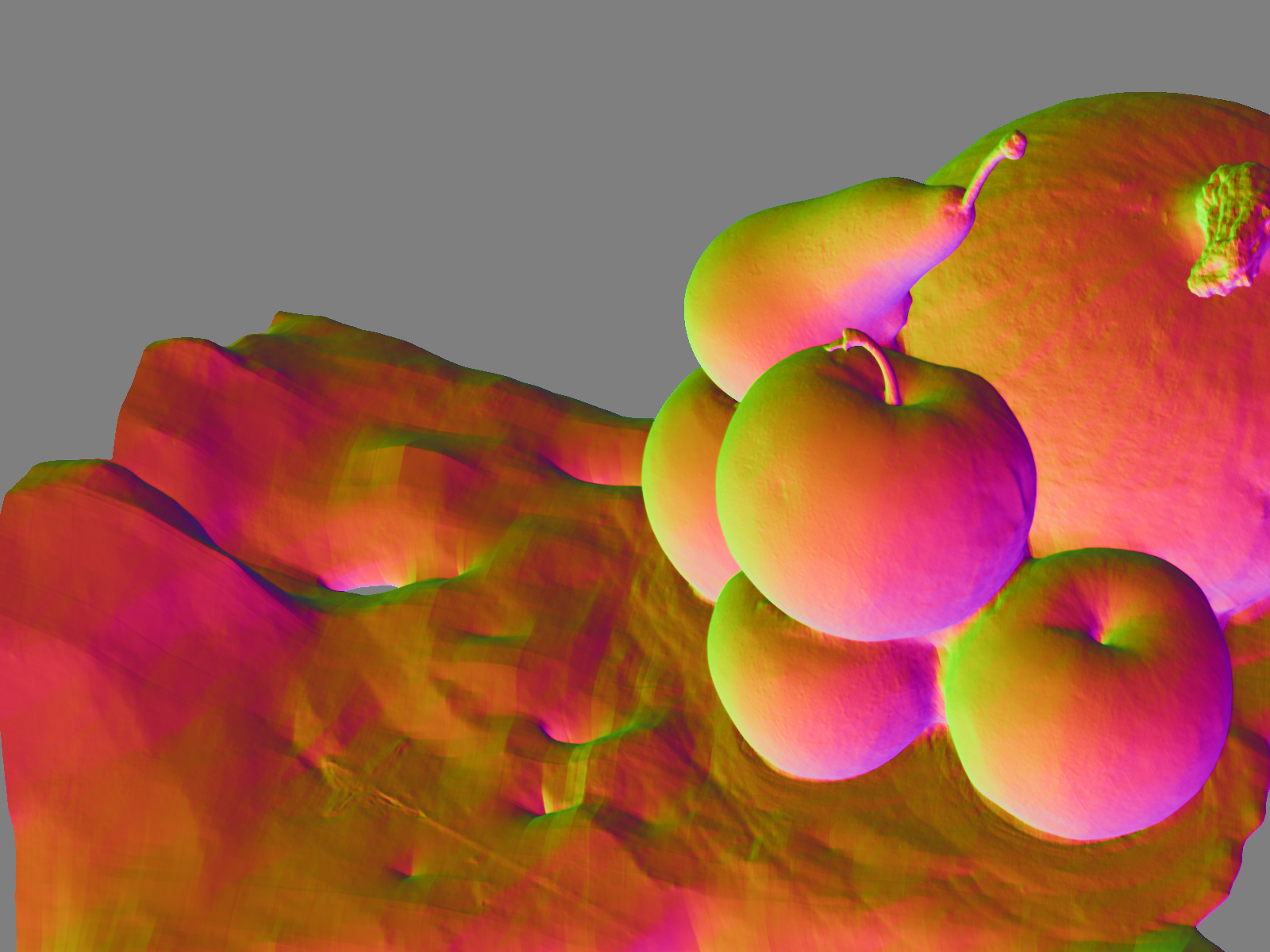}
            & \includegraphics[width=0.12\linewidth]{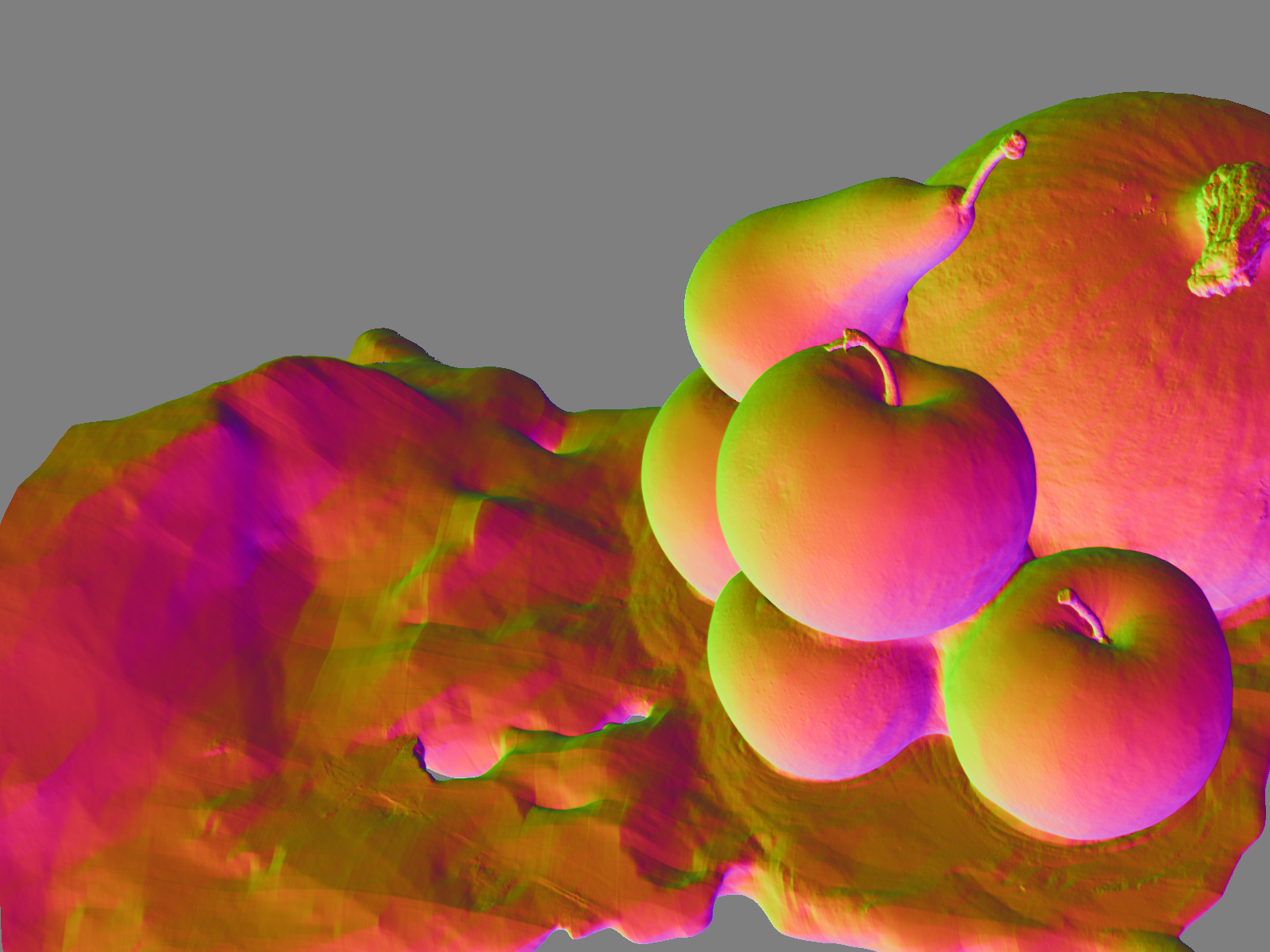}
            & \includegraphics[width=0.12\linewidth]{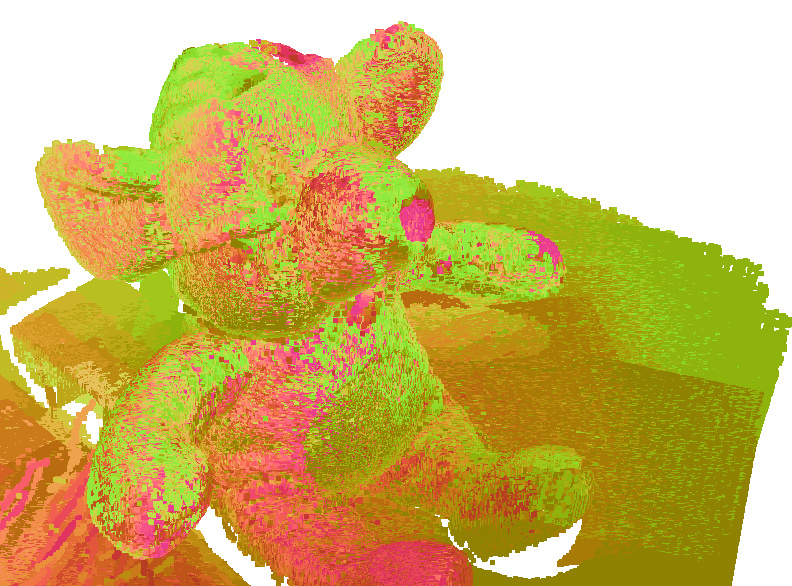}
            & \includegraphics[width=0.12\linewidth]{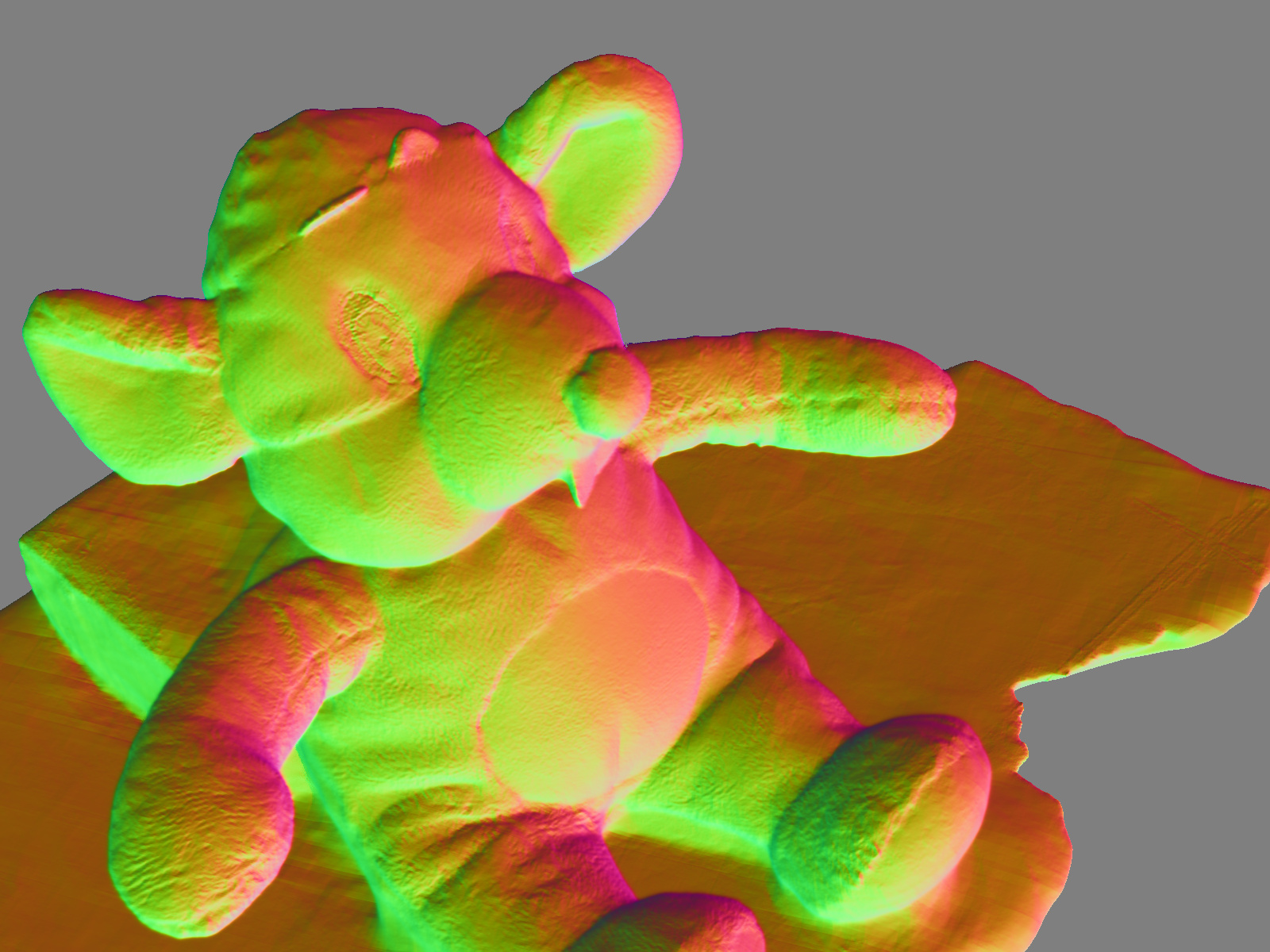}
            & \includegraphics[width=0.12\linewidth]{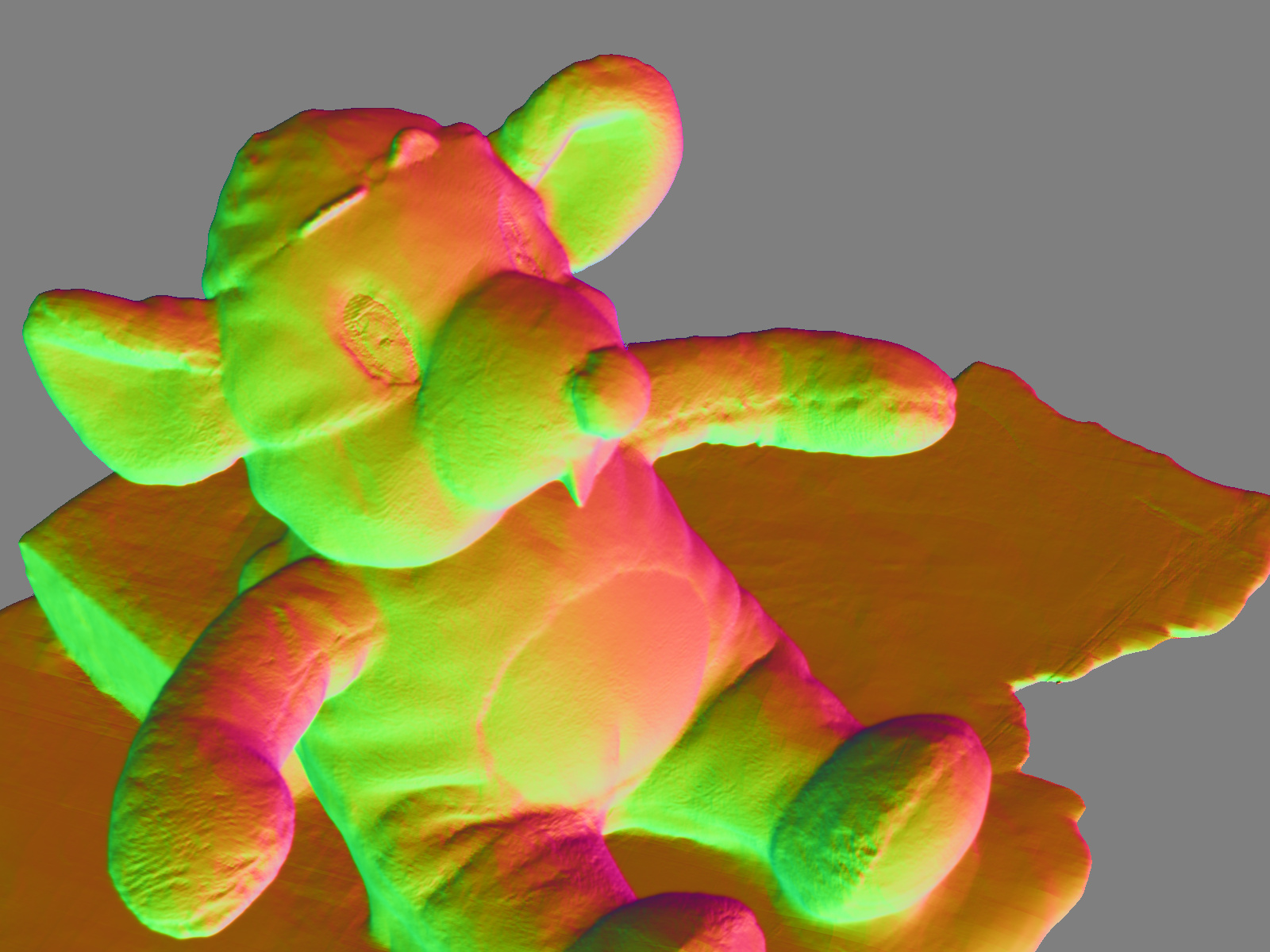} \\
            \rotatebox{90}{Roughness}
            &
            & \includegraphics[width=0.12\linewidth]{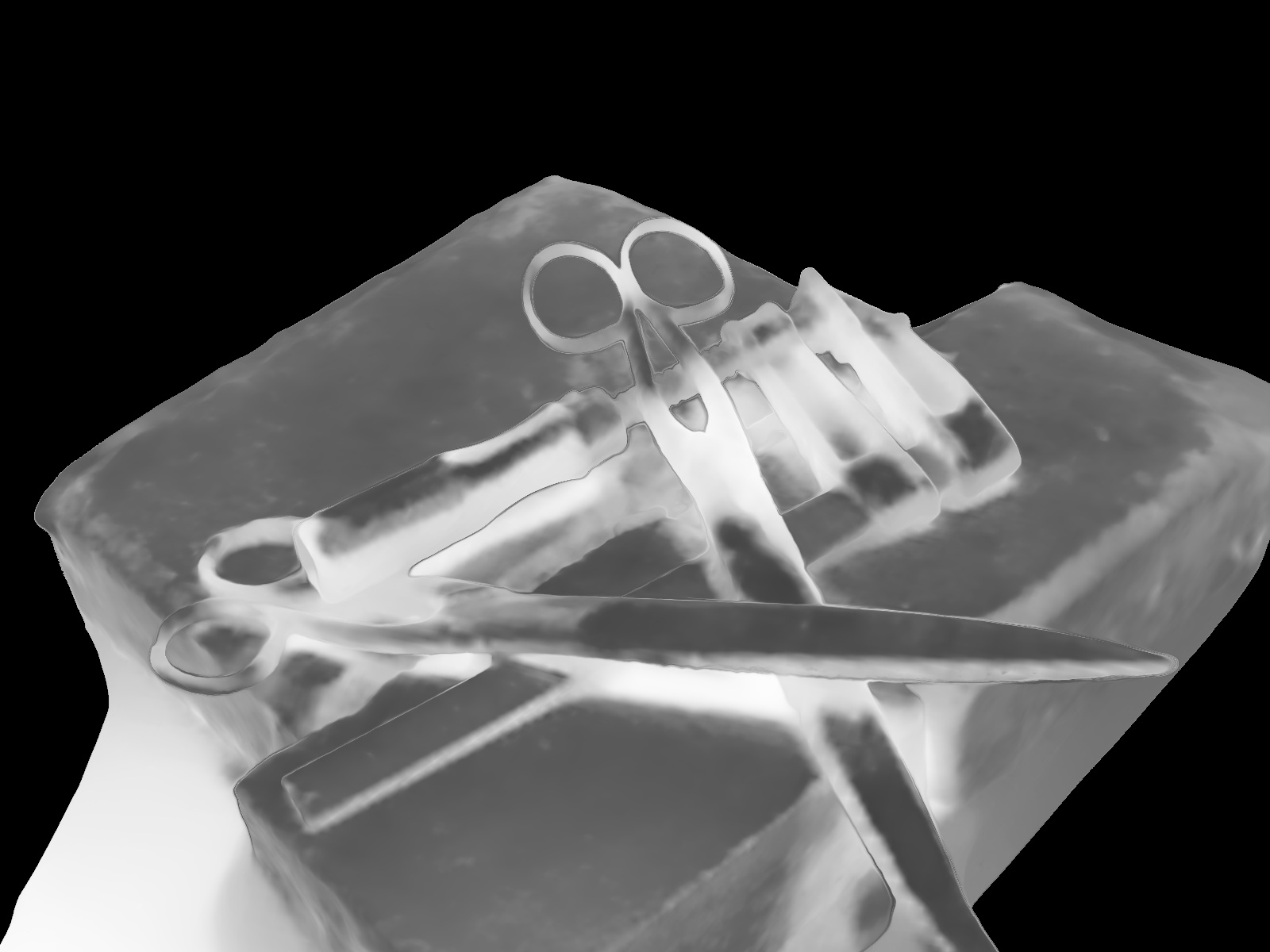}
            & \includegraphics[width=0.12\linewidth]{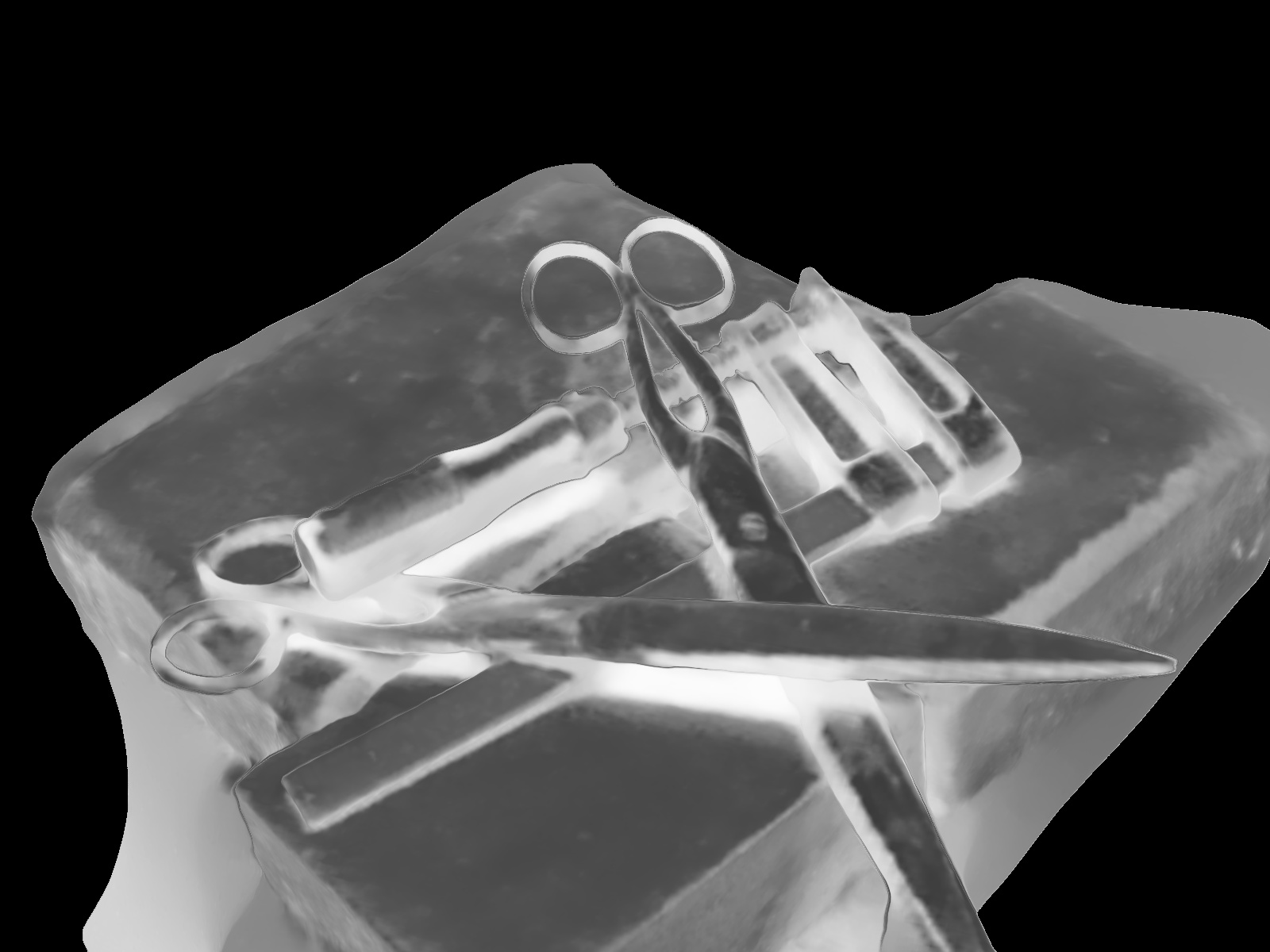}
            &
            & \includegraphics[width=0.12\linewidth]{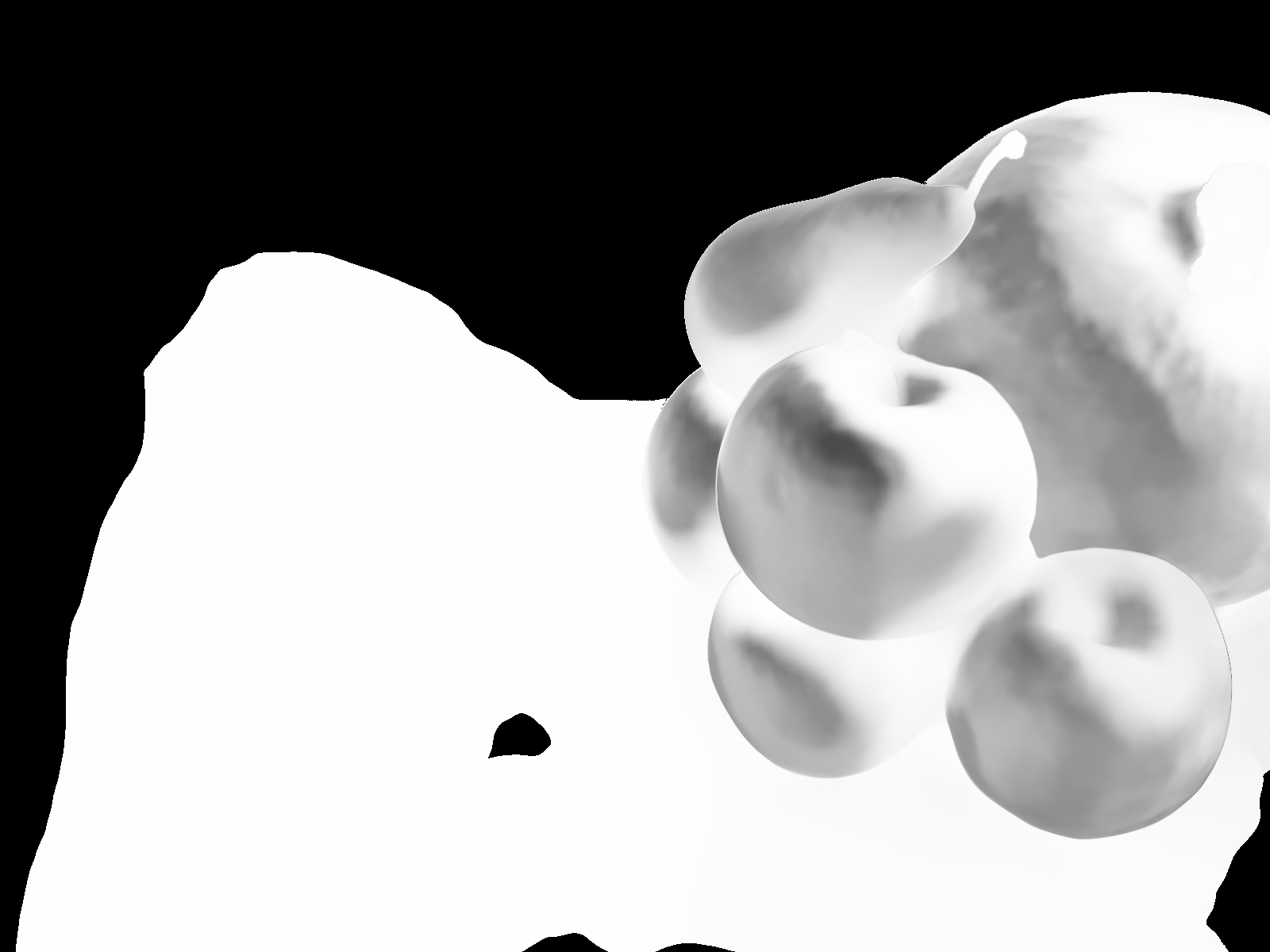}
            & \includegraphics[width=0.12\linewidth]{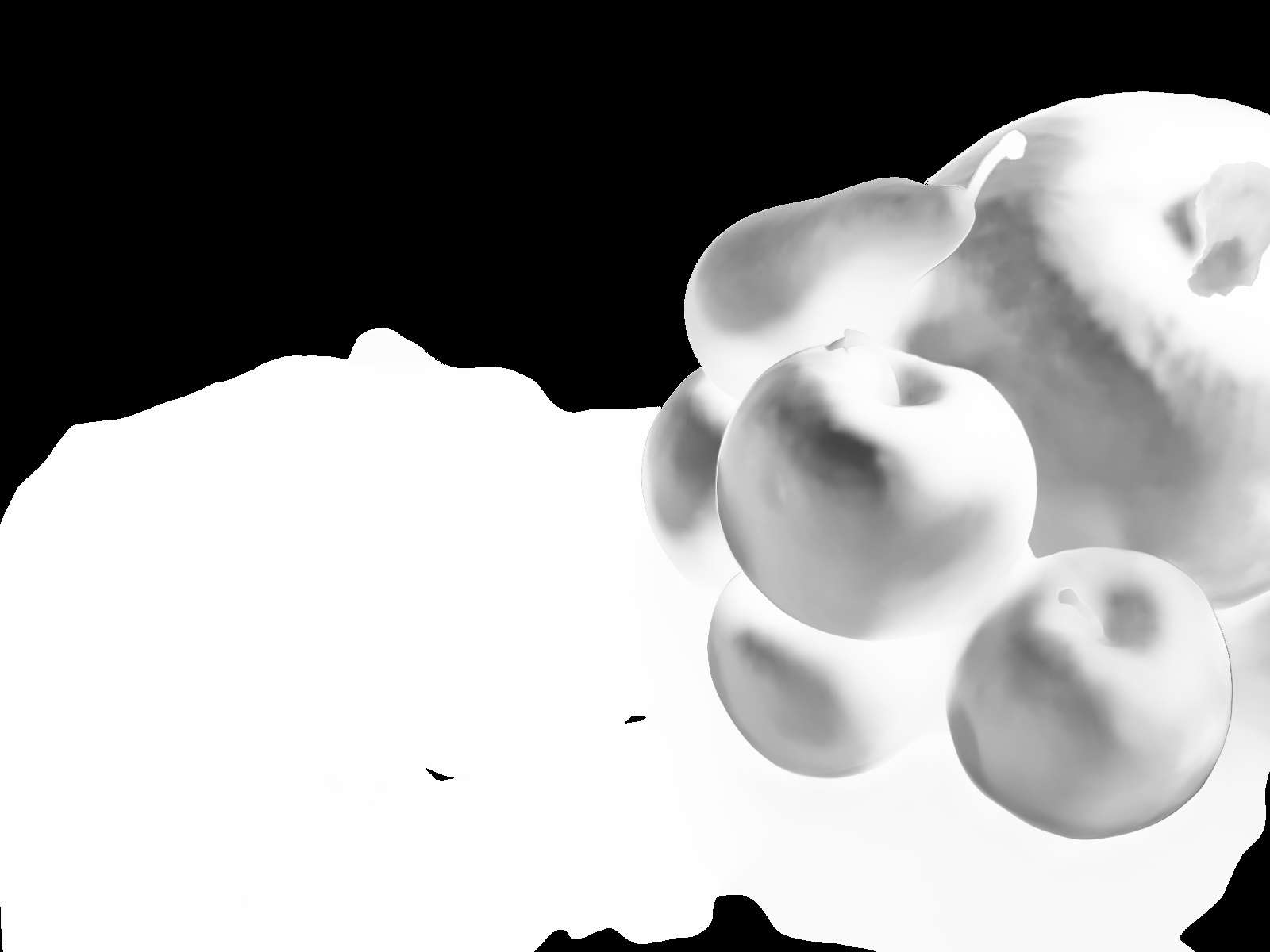}
            &
            & \includegraphics[width=0.12\linewidth]{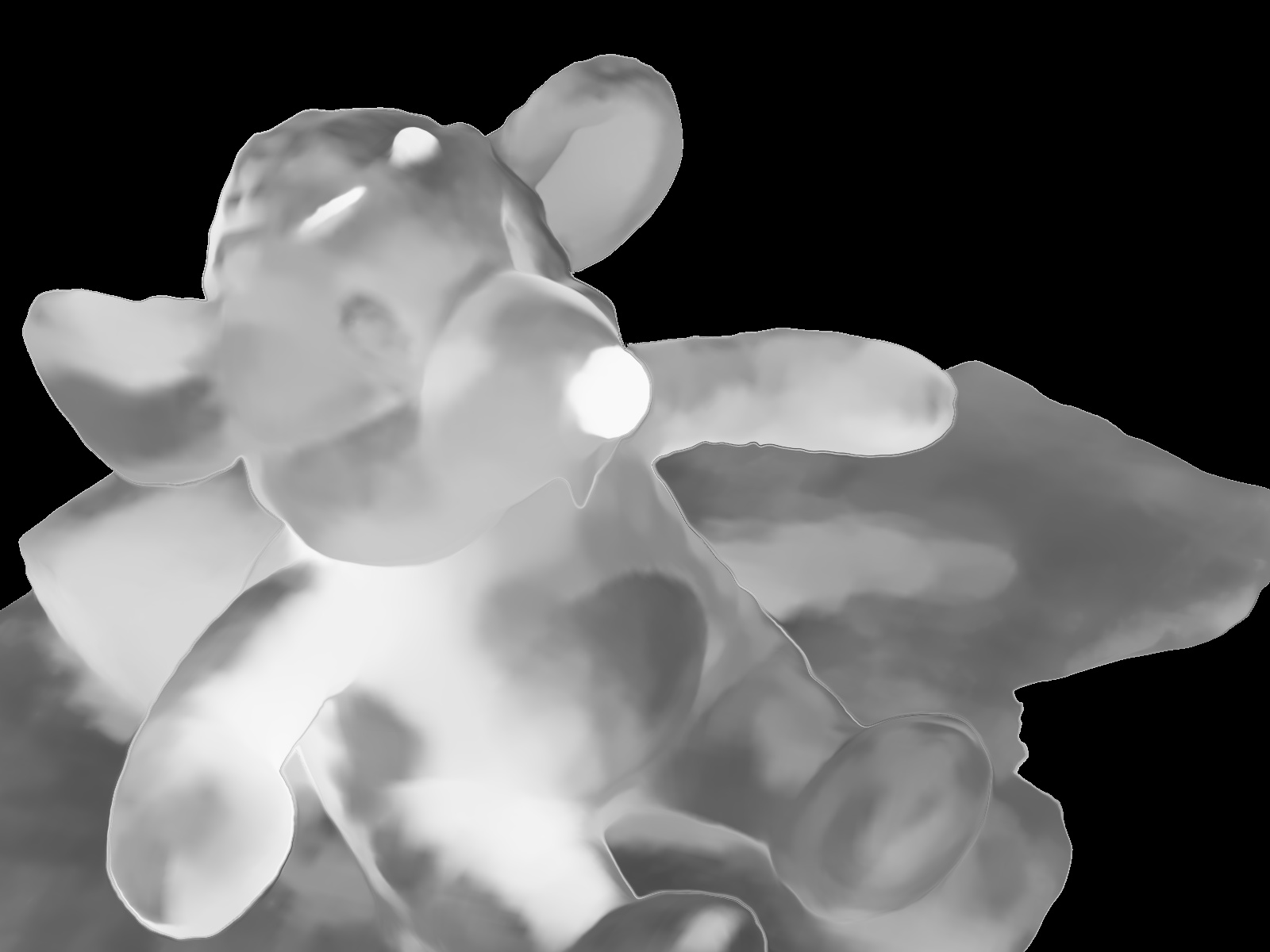}
            & \includegraphics[width=0.12\linewidth]{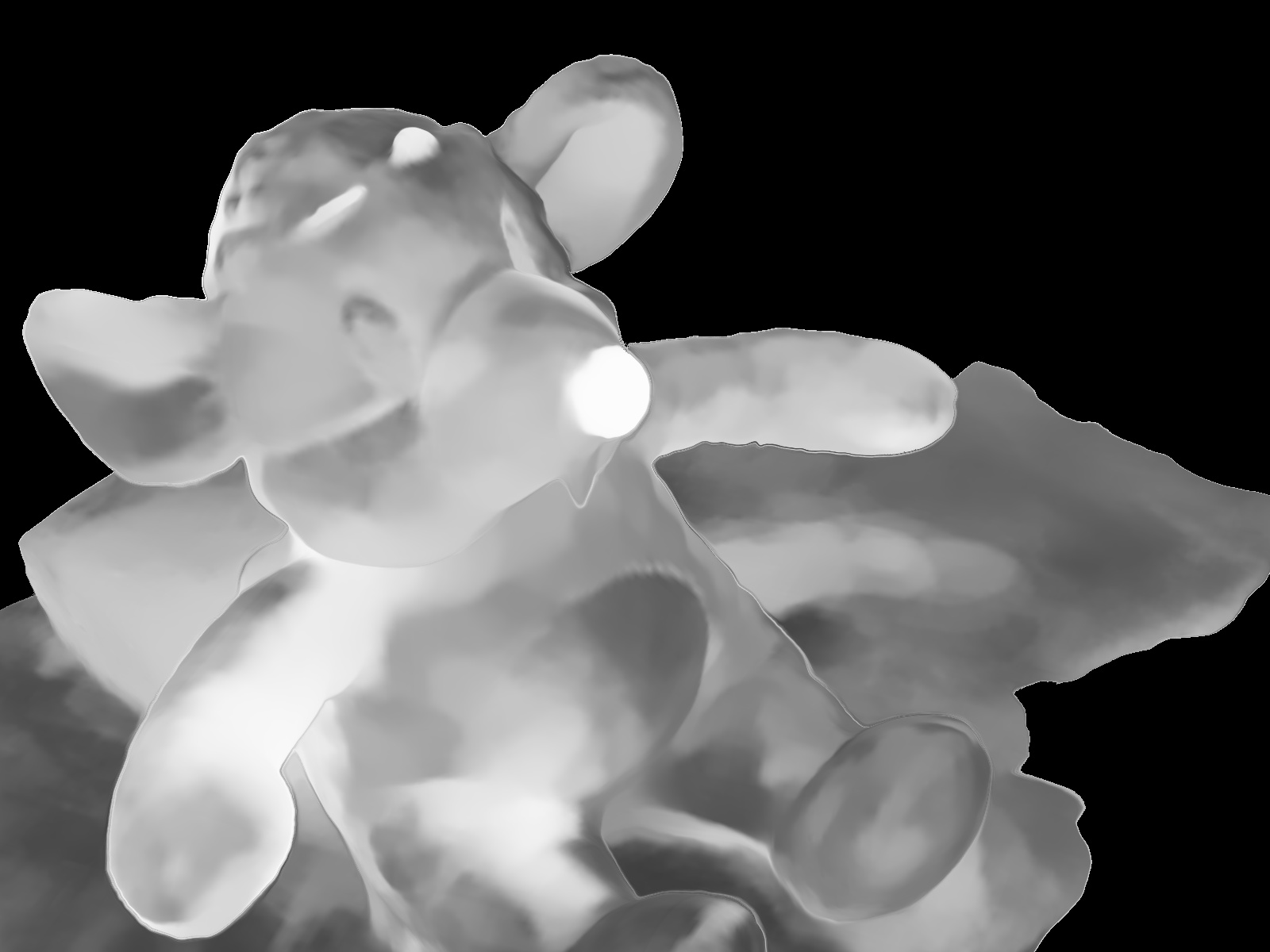}\\
            \rotatebox{90}{Metallicness}
            & & \includegraphics[width=0.12\linewidth]{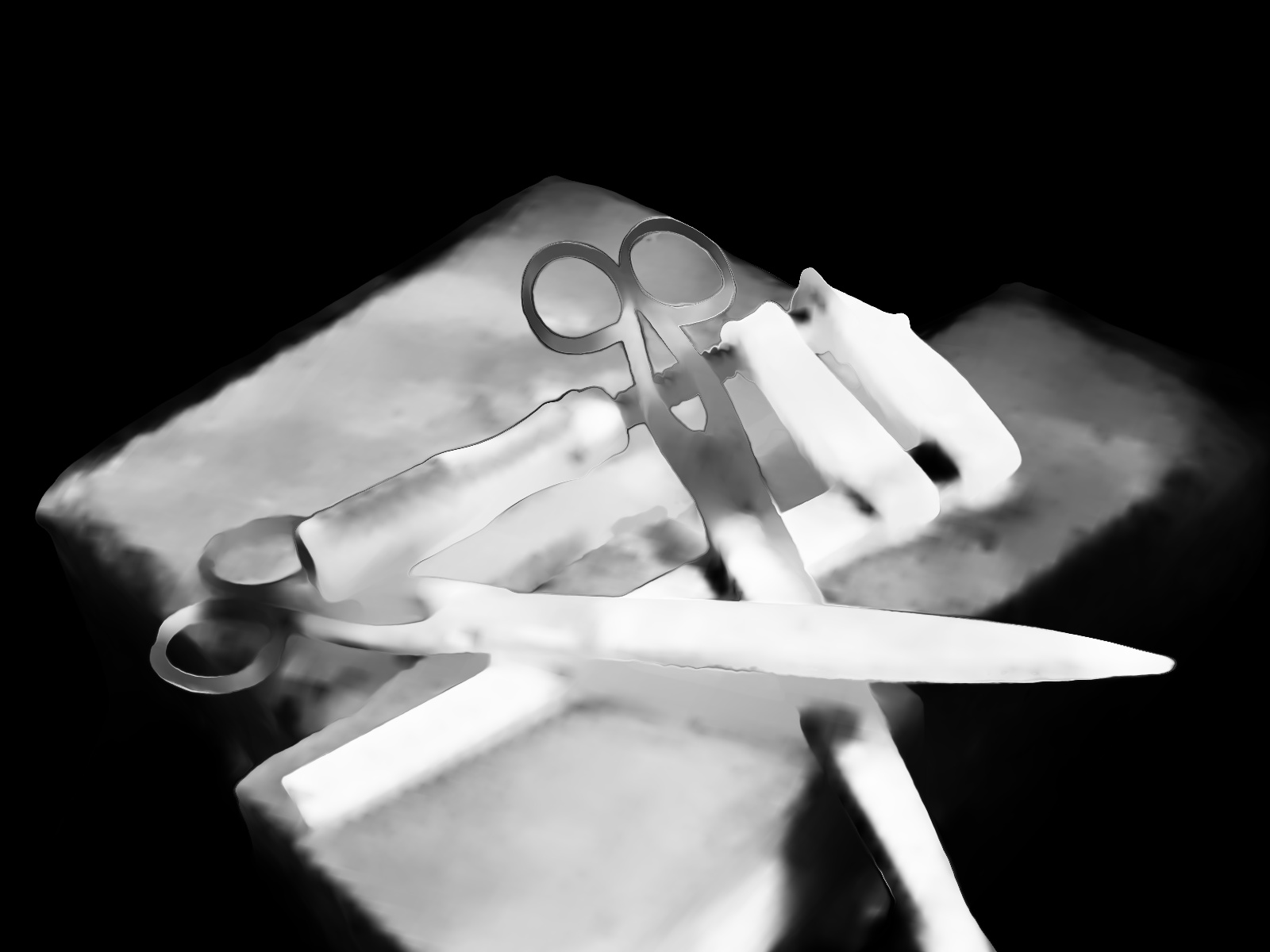}
            & \includegraphics[width=0.12\linewidth]{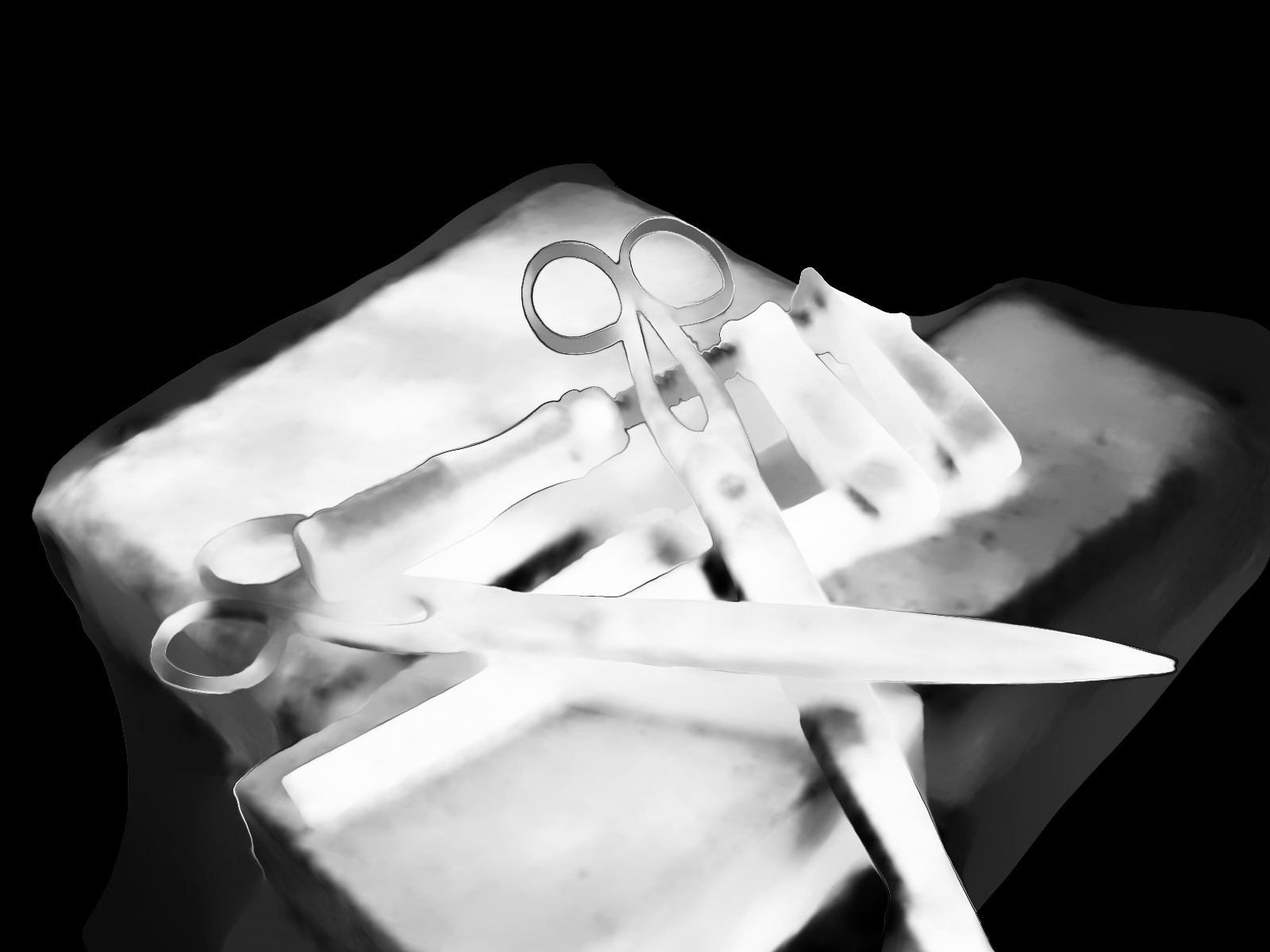}
            & & \includegraphics[width=0.12\linewidth]{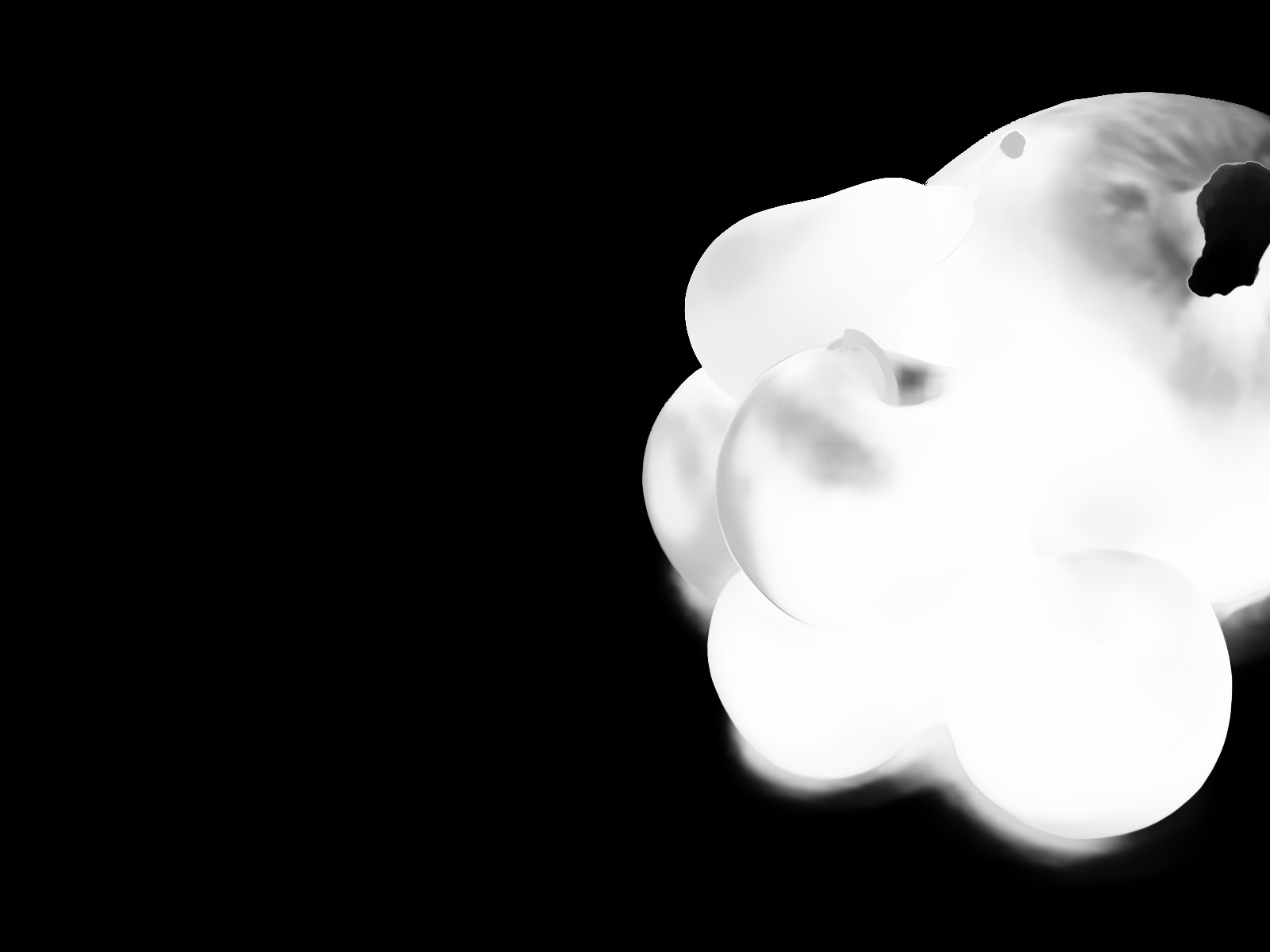}
            & \includegraphics[width=0.12\linewidth]{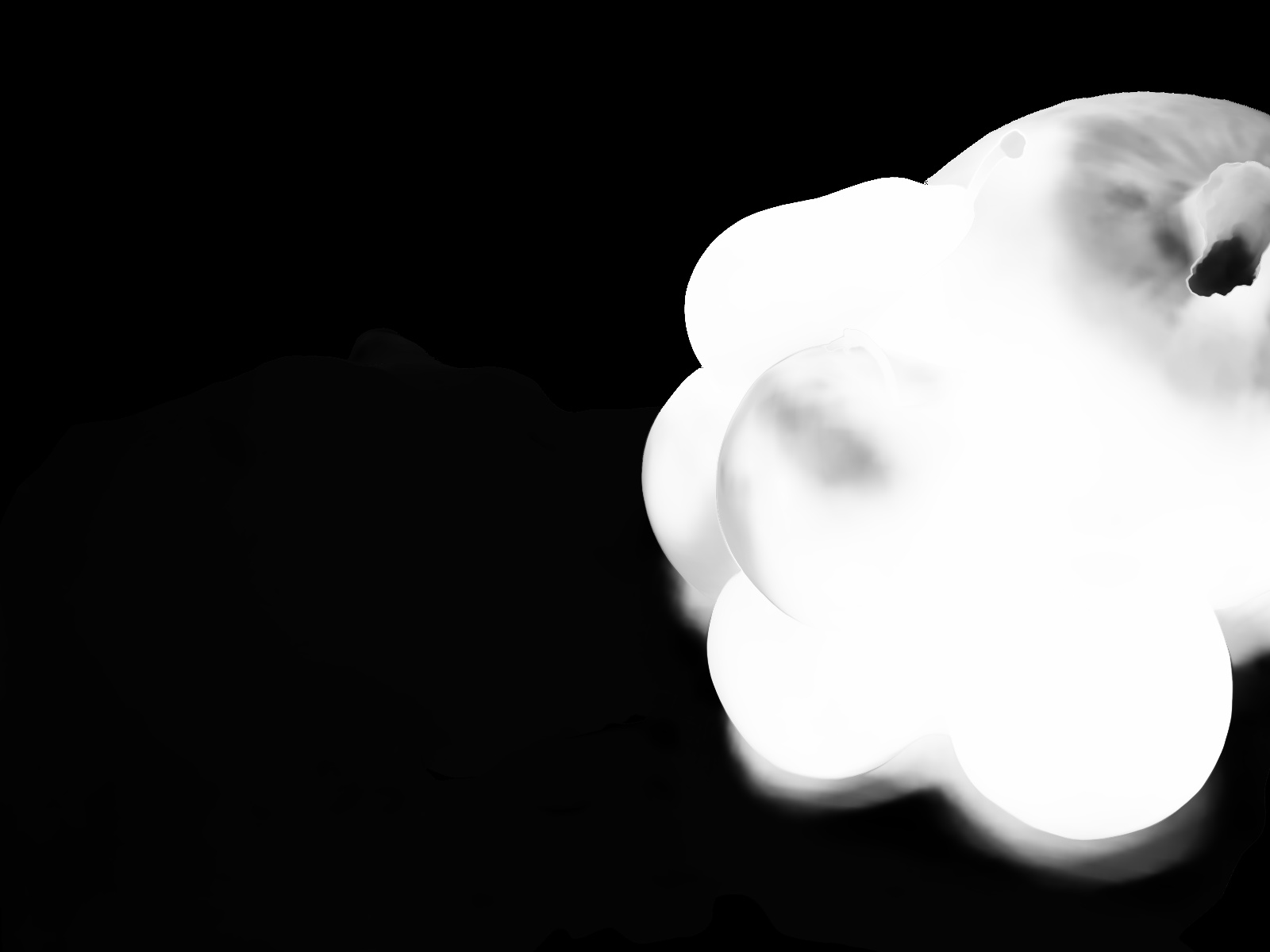}
            & & \includegraphics[width=0.12\linewidth]{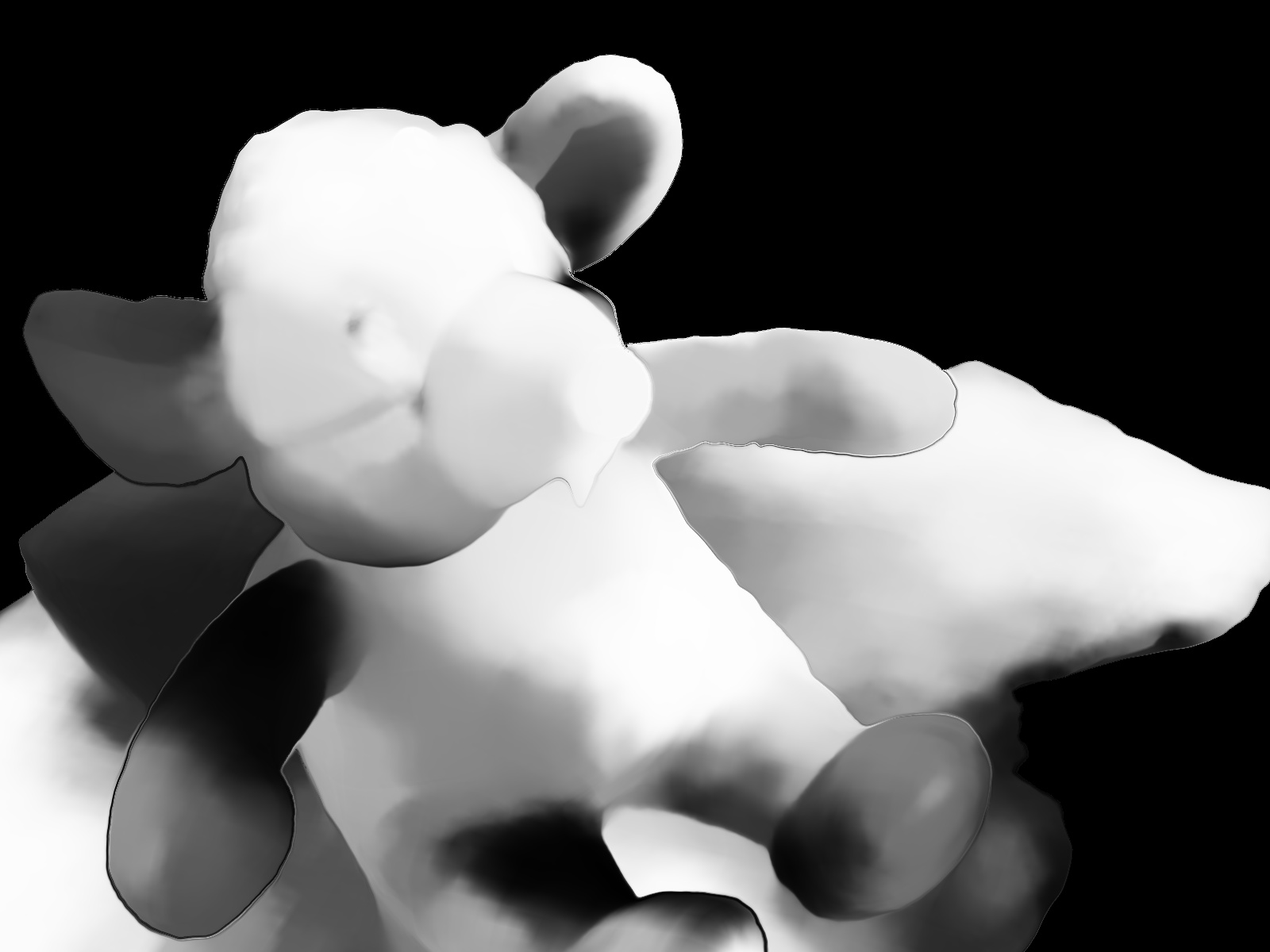}
            & \includegraphics[width=0.12\linewidth]{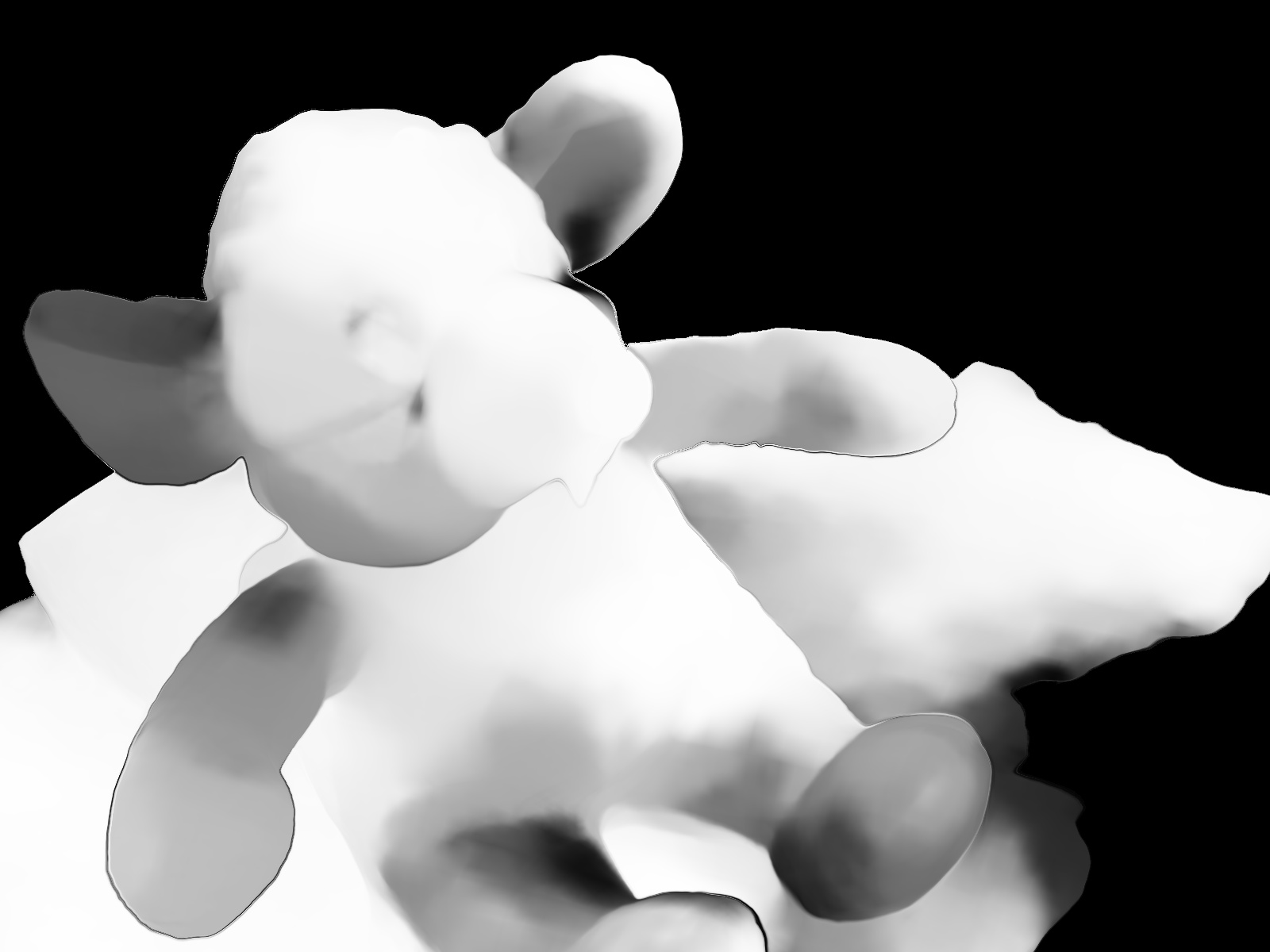} \\
            \rotatebox{90}{Albedo}
            & & \includegraphics[width=0.12\linewidth]{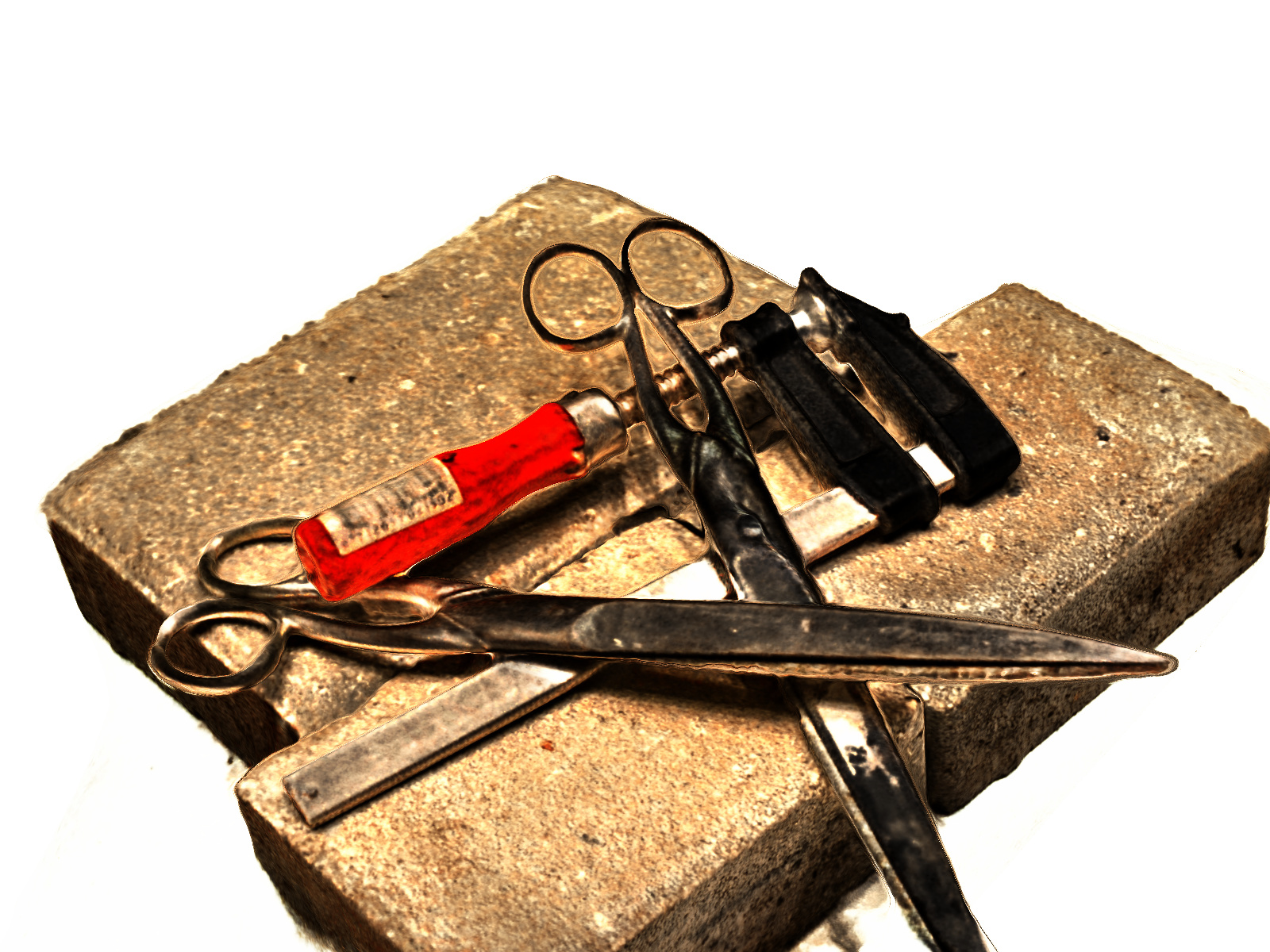}
            & \includegraphics[width=0.12\linewidth]{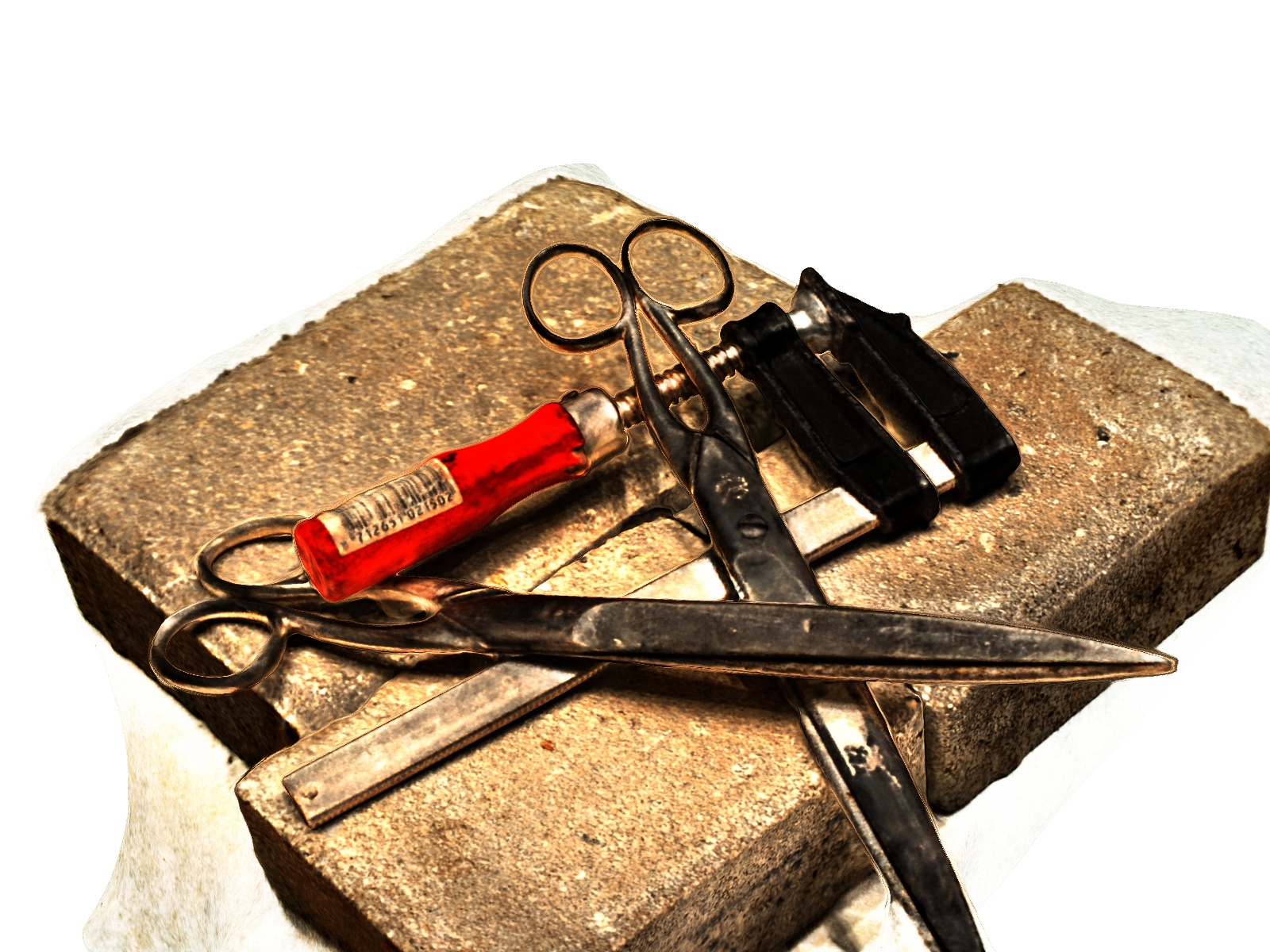}
            & & \includegraphics[width=0.12\linewidth]{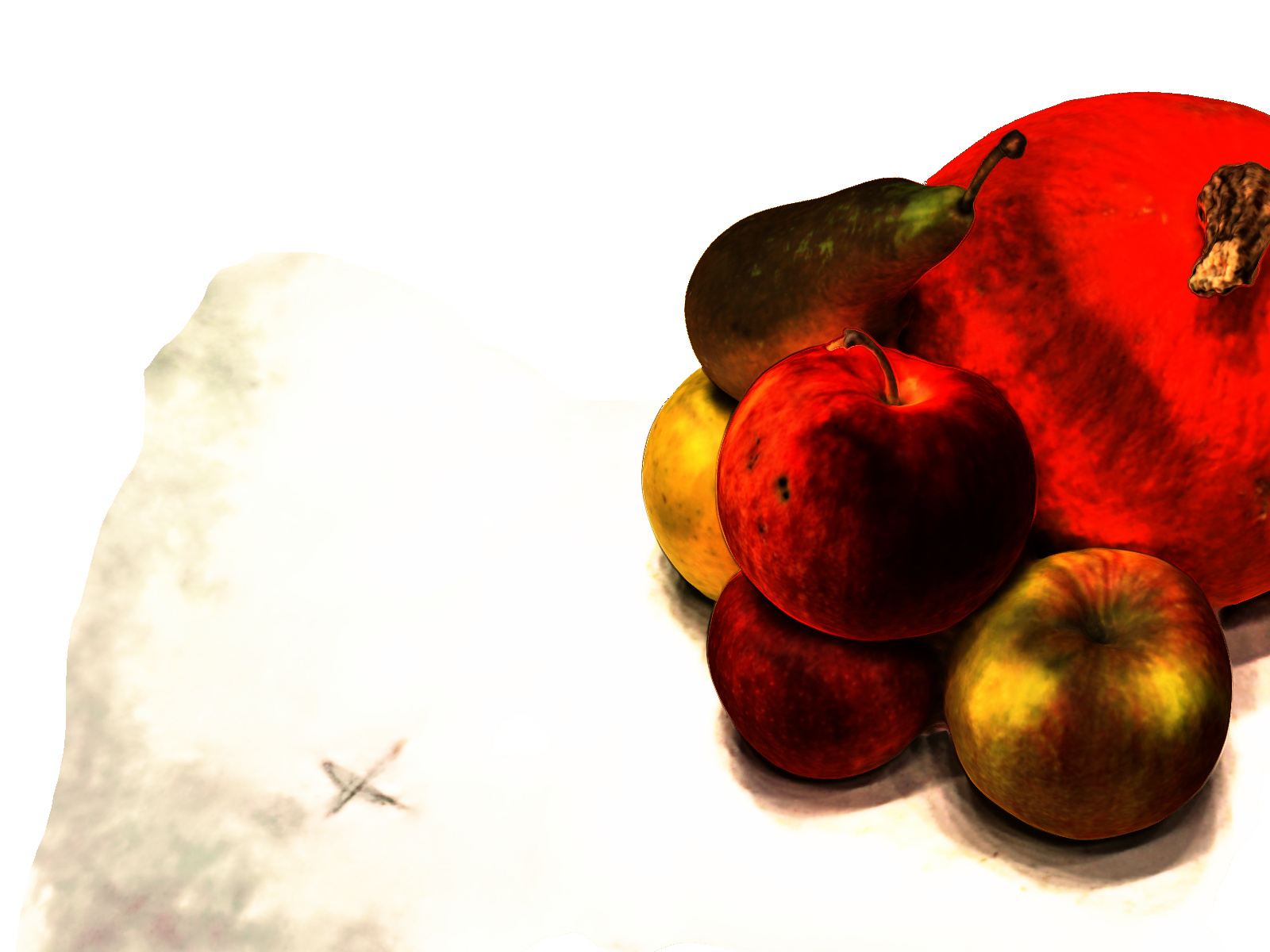}
            & \includegraphics[width=0.12\linewidth]{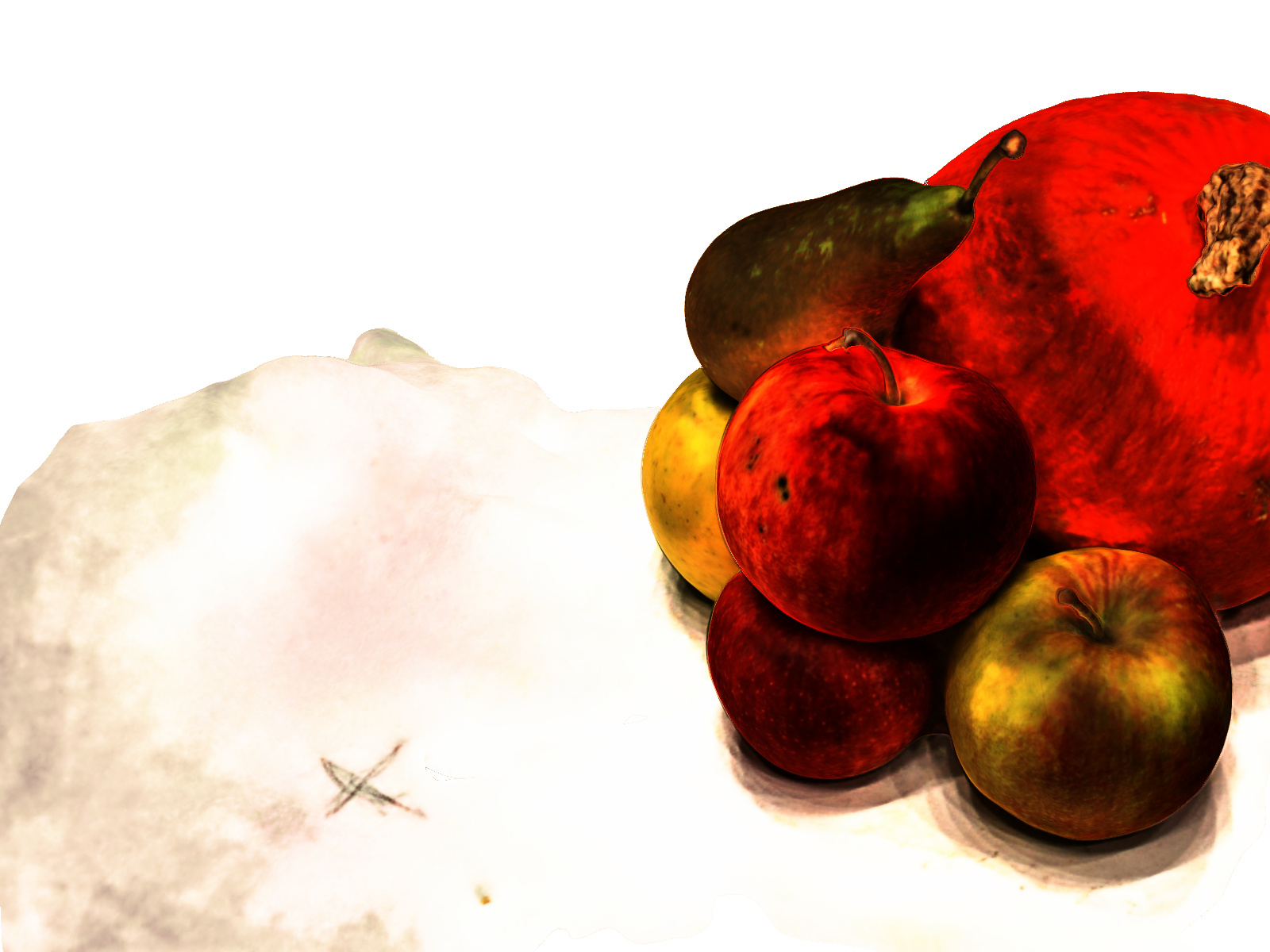}
            & & \includegraphics[width=0.12\linewidth]{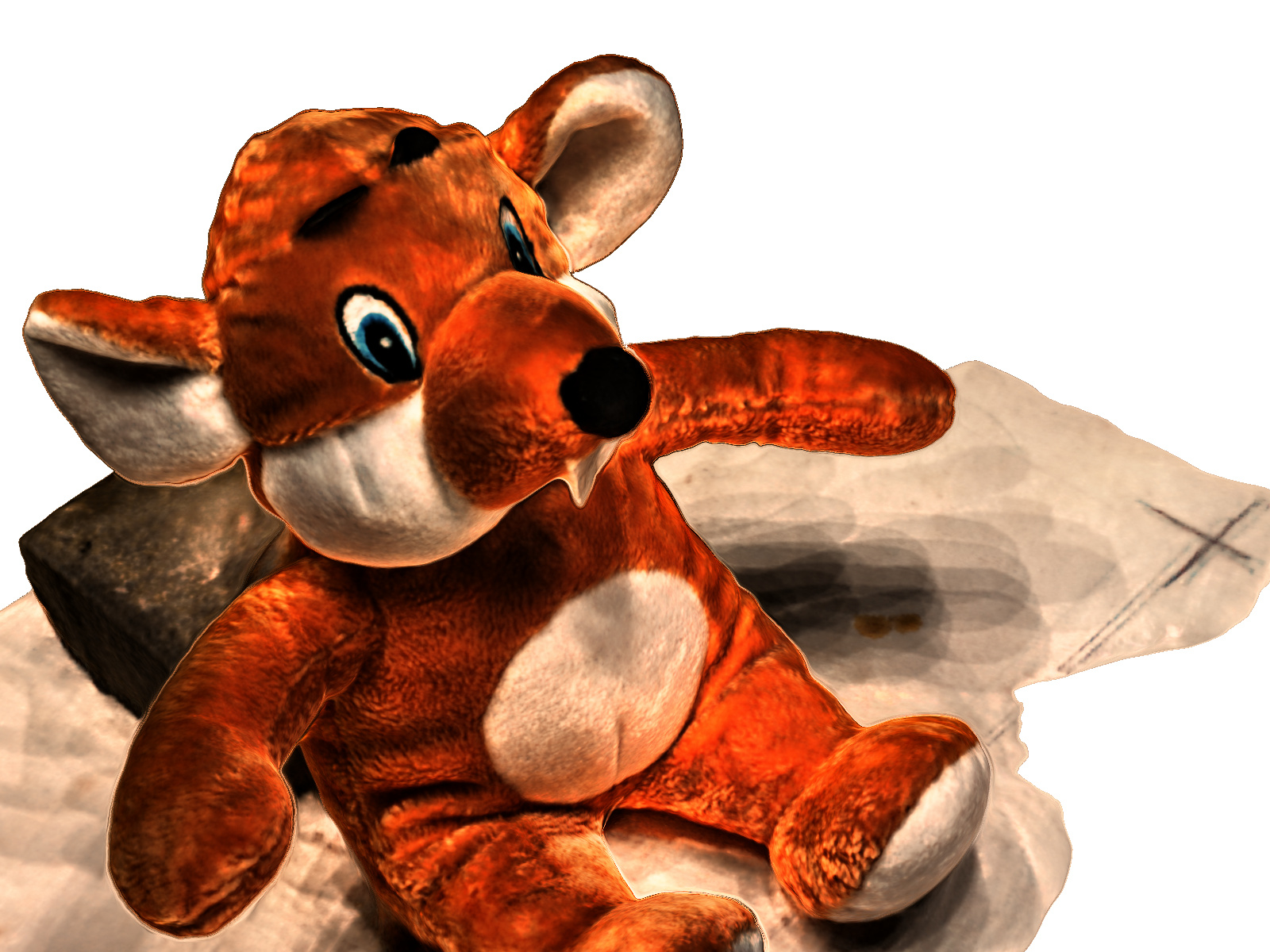}
            & \includegraphics[width=0.12\linewidth]{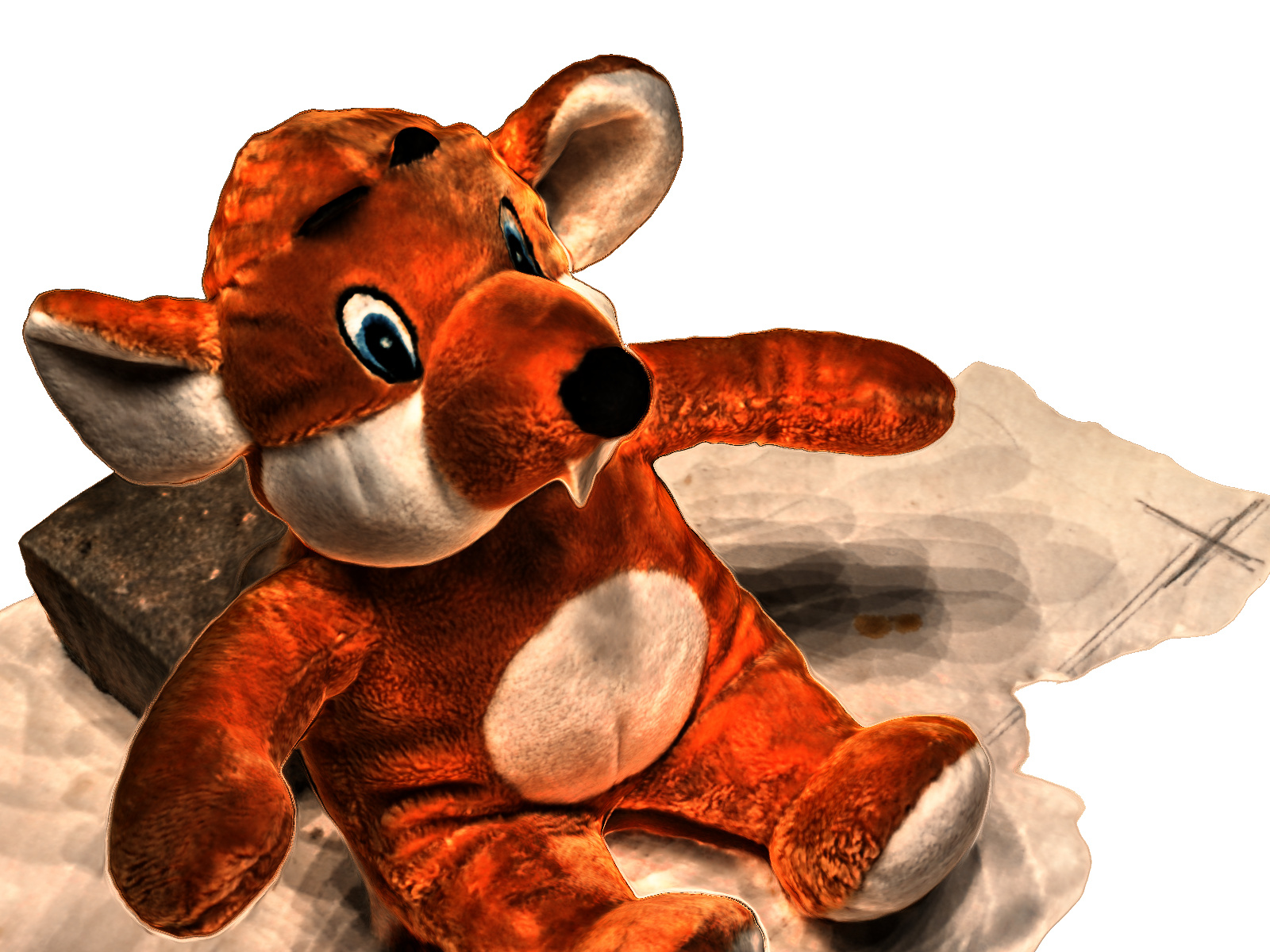} \\
            & Reference & NeILF++~\cite{zhang2023neilf++} & Ours & Reference & NeILF++~\cite{zhang2023neilf++} & Ours & Reference & NeILF++~\cite{zhang2023neilf++} & Ours \\
        \end{tabular}
    }
    \vspace{-8pt}
    \caption{\textbf{Qualitative comparisons on DTU~\cite{jensen2014dtu}.} Best viewed on a screen at full zoom.}
    \label{fig:dtu_dataset_qualitative}
    \vspace{-6mm}
\end{figure*}

\PAR{Qualitative Results.}
We further validate our material estimation performance qualitatively against NeILF++ on the NeILF++ dataset in Fig.~\ref{fig:neilfpp_dataset_qualitative} (see supplement for additional qualitative results).
Over all three challenging Mix lighting conditions, our physics-based losses improve both lighting and material estimation.
Our estimated lighting exhibits lower entropy with more concentrated sources and fewer artifacts, especially in the upper halves of environment maps.
We also eliminate fringes and patches generated on the sphere and cube by NeILF++ due to ``baked-in'' specular highlights and achieve effective separation of diffuse and specular BRDF components via our NDF-weighted Specular Loss.
Most importantly, we consistently outperform NeILF++ on albedo estimation, which is a critical challenge in inverse rendering.
In particular, our albedo estimation in Studio Mix properly recovers the helmet and visor whereas NeILF++ fails to recover any helmet details and incorrectly predicts a black albedo.
In general, our predicted albedo is---correctly---less bright than for NeILF++, with the floor, cube, and sphere all free from incorrect specular highlights and artifacts.

We compare our DTU material and geometry estimation in Fig.~\ref{fig:dtu_dataset_qualitative}.
On scan~37, we improve metallicness and roughness, accurately predicting the scissor and clamp as metal, unlike NeILF++, which creates artifacts near the scissor handle in its predicted normals, metallicness, and roughness.
On scan~63, we successfully recover fine details like the green apple stem, which NeILF++ misses in its predicted normals and materials.
On scan~105, we correctly reproduce the brick's top surface, whereas NeILF++ has RGB, normals, and albedo artifacts.
Our roughness and metallicness on the orange toy fur are also more consistent than NeILF++ for the homogeneous furry material.

\subsection{Ablation Study}
\label{sec:ablation}
We perform an ablation study of the proposed physically-based losses in Tab.~\ref{tab:pbr_loss_ablation} to individually evaluate each loss's contribution.
A qualitative analysis is provided in the supplement.
The baseline (ID 1) is NeILF++ \cite{zhang2023neilf++} without the Lambertian loss.
Adding only the Conservation of Energy Loss $\mathcal{L}_{\text{cons}}$ (ID 2) improves material estimation, with albedo PSNR increasing by 2.34 and roughness PSNR by 0.28, respectively.
Adding only the specular loss $\mathcal{L}_\text{spec}$ (ID 3) also improves material estimation over ID 1, with albedo PSNR increasing by 2.35, roughness PSNR by 0.25, and metallicness PSNR by 0.26, respectively.
In our final method (ID 4), adding both $\mathcal{L}_{\text{cons}}$ and $\mathcal{L}_\text{spec}$ achieves the highest albedo, roughness, and metallicness PSNR.
Specifically, albedo PSNR and SSIM increase by an additional 0.93 PSNR and 1.84\% over ID 3, respectively.
Although metallicness SSIM is lower for ID 2--4 compared to ID 1, metallicness PSNR is improved.
Overall, our complete method (ID 4) with both physics-based losses improves roughness by 0.35 PSNR, metallicness by 0.27 PSNR, and albedo by 3.28 PSNR over the baseline (ID 1), while maintaining on-par performance in RGB.

\begin{table}[tb]
	\caption{\textbf{Ablation of our physics-based losses.} Performance of ablated versions of PBR-NeRF and of its full version is compared on the mean PSNR
and SSIM of \textit{all} 6 scenes of the NeILF++ dataset ~\cite{zhang2023neilf++} for Disney BRDF parameters and novel views.}
	\label{tab:pbr_loss_ablation}
    \vspace{-3mm}
    \centering
    \setlength{\tabcolsep}{3pt}
	\resizebox{\linewidth}{!}{
    \begin{tabular}{l|cc|cc|cc|cc|cc}
        \specialrule{.2em}{.1em}{.1em}
        \multirow{2}{*}{ID} & \multirow{2}{*}{$\mathcal{L}_{\text{cons}}$} & \multirow{2}{*}{$\mathcal{L}_{\text{spec}}$} &
        \multicolumn{2}{c|}{RGB} & \multicolumn{2}{c|}{Roughness} & \multicolumn{2}{c|}{Metallicness} & \multicolumn{2}{c}{Albedo} \\
        & & & PSNR & SSIM & PSNR & SSIM & PSNR & SSIM & PSNR & SSIM \\ \hline
        1 & \xmark & \xmark & 31.26 & 87.05 & 22.12 & 92.20 & 21.35 & 74.98 & 16.80 & 74.04 \\
        2 & \cmark & \xmark & 31.29 & 87.09 & 22.40 & 92.31 & 21.30 & 72.42 & 19.14 & 84.30 \\
        3 & \xmark & \cmark & 31.27 & 87.05 & 22.37 & 92.32 & 21.61 & 73.58 & 19.15 & 84.98 \\
        4 & \cmark & \cmark & 31.27 & 87.05 & 22.47 & 92.31 & 21.62 & 74.08 & 20.08 & 86.82 \\
        \specialrule{.2em}{.1em}{.1em}
    \end{tabular}}
    \vspace{-6mm}
\end{table}

\section{Conclusion}
We present PBR-NeRF, a novel inverse rendering method leveraging neural fields optimized with PBR-inspired losses.
These losses provide physically valid inductive biases, enabling our neural fields to better disambiguate materials from lighting---addressing a major challenge in inverse rendering.
Thorough experiments across various benchmarks demonstrate the effectiveness of PBR-NeRF not only in material estimation and inverse rendering, but also in novel view synthesis.
In particular, we qualitatively show a compelling disentanglement of the diffuse and specular components in our estimated BRDFs.
We believe that our physics-based approach will inspire further work on PBR-driven, decomposed neural fields.

{
    \small
    \bibliographystyle{ieeenat_fullname}
    \bibliography{main}
}

\clearpage

\appendix
\maketitlesupplementary

\section{Ablation Study Qualitative Results}
\label{sec:supp:ablation_qualitative_results}
\begin{figure*}[h]
    \centering
    \small
    \resizebox{\linewidth}{!}{
        \begin{tabular}{cccccc}
            \multirow{1}{*}[0.5in]{\rotatebox[origin=c]{90}{Lighting $\dag$}}
            & \includegraphics[width=0.24\linewidth]{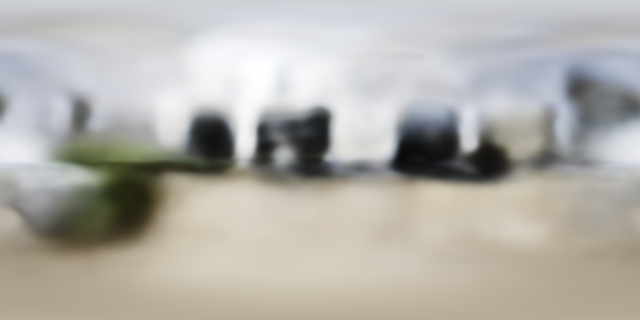}
            & \includegraphics[width=0.24\linewidth]{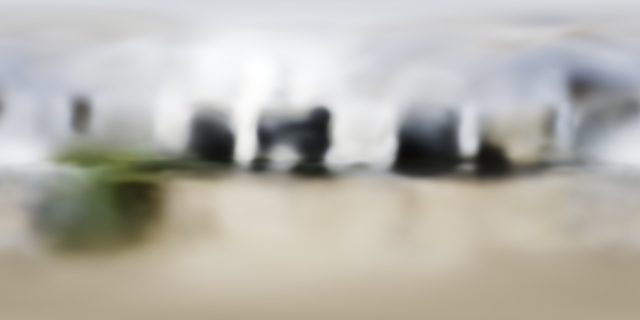}
            & \includegraphics[width=0.24\linewidth]{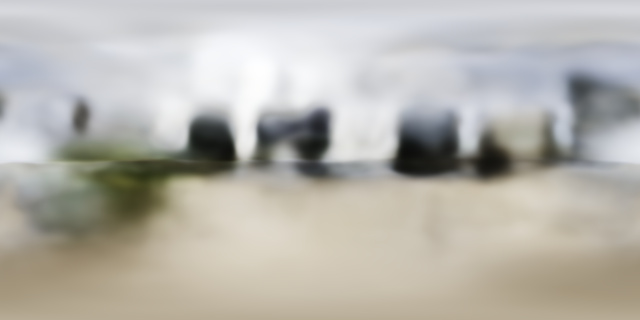}
            & \includegraphics[width=0.24\linewidth]{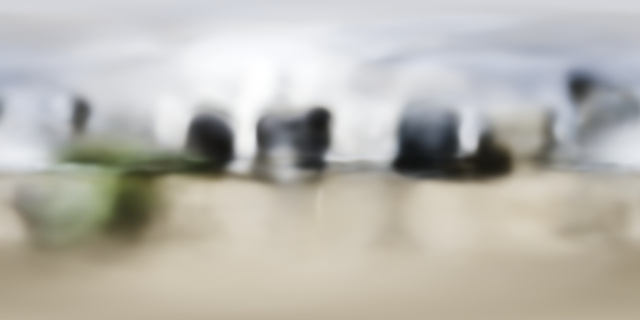}
            &
            \\
            \multirow{1}{*}[0.5in]{\rotatebox[origin=c]{90}{RGB}}
            & \includegraphics[width=0.24\linewidth]{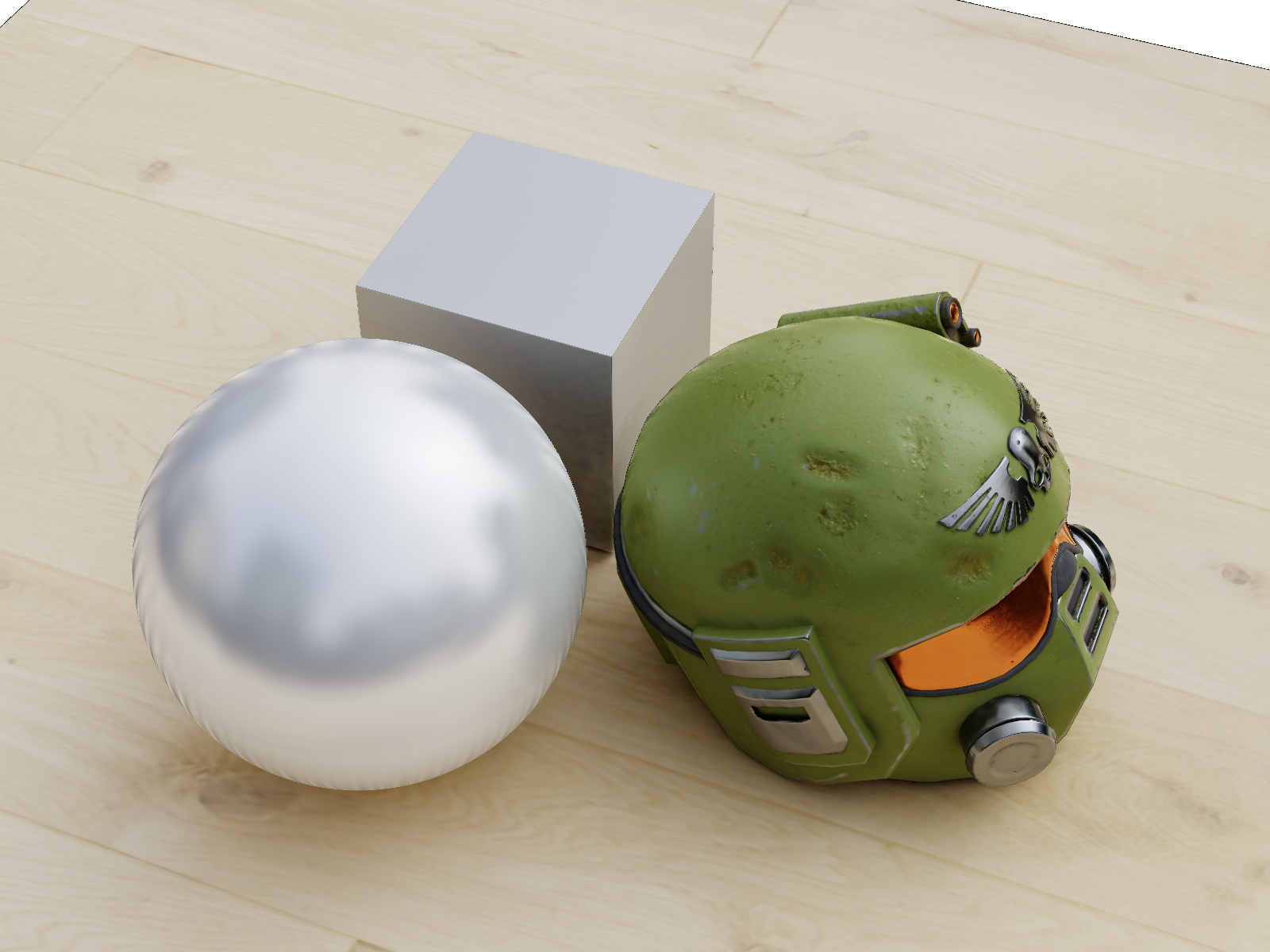}
            & \includegraphics[width=0.24\linewidth]{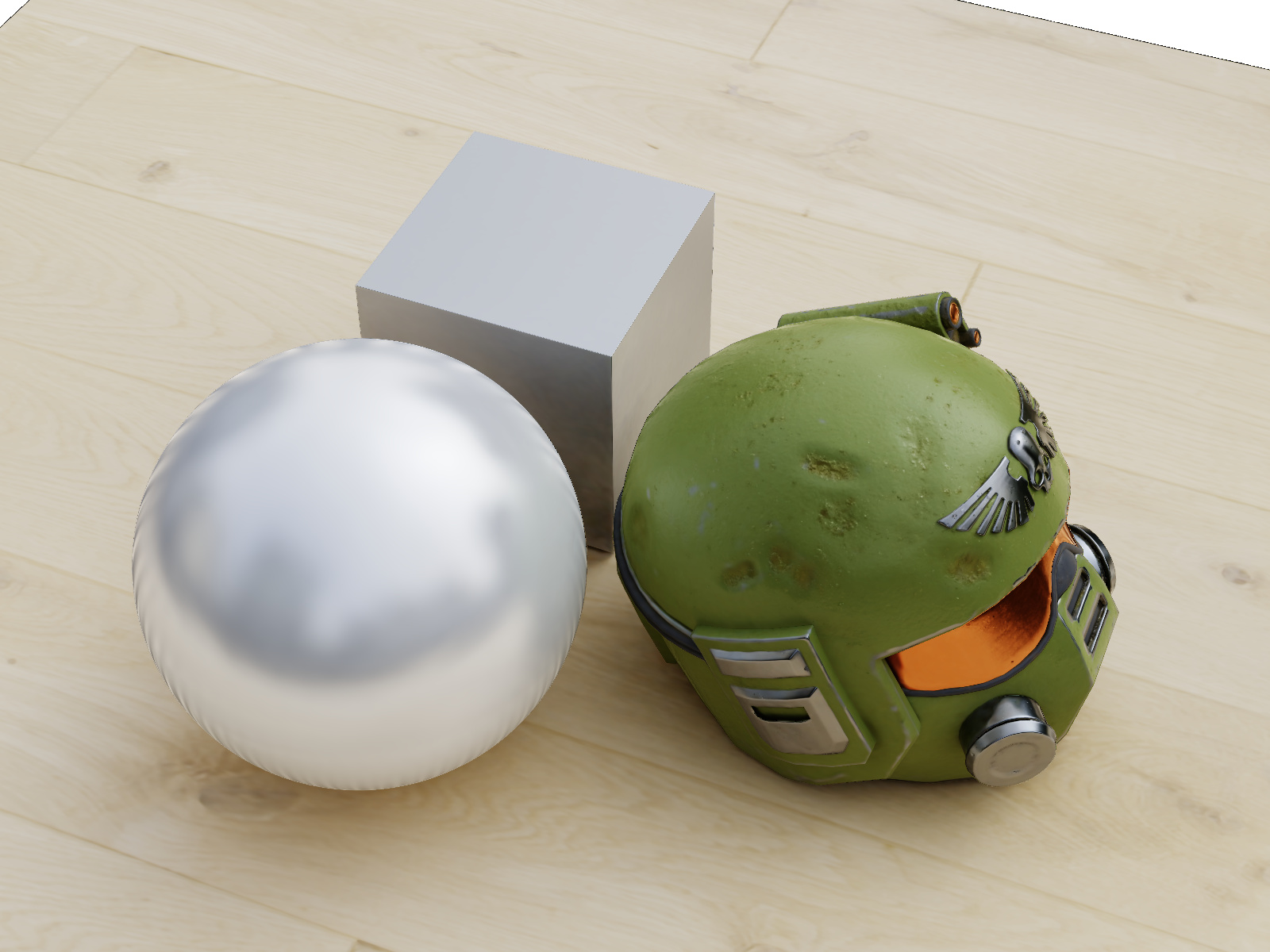}
            & \includegraphics[width=0.24\linewidth]{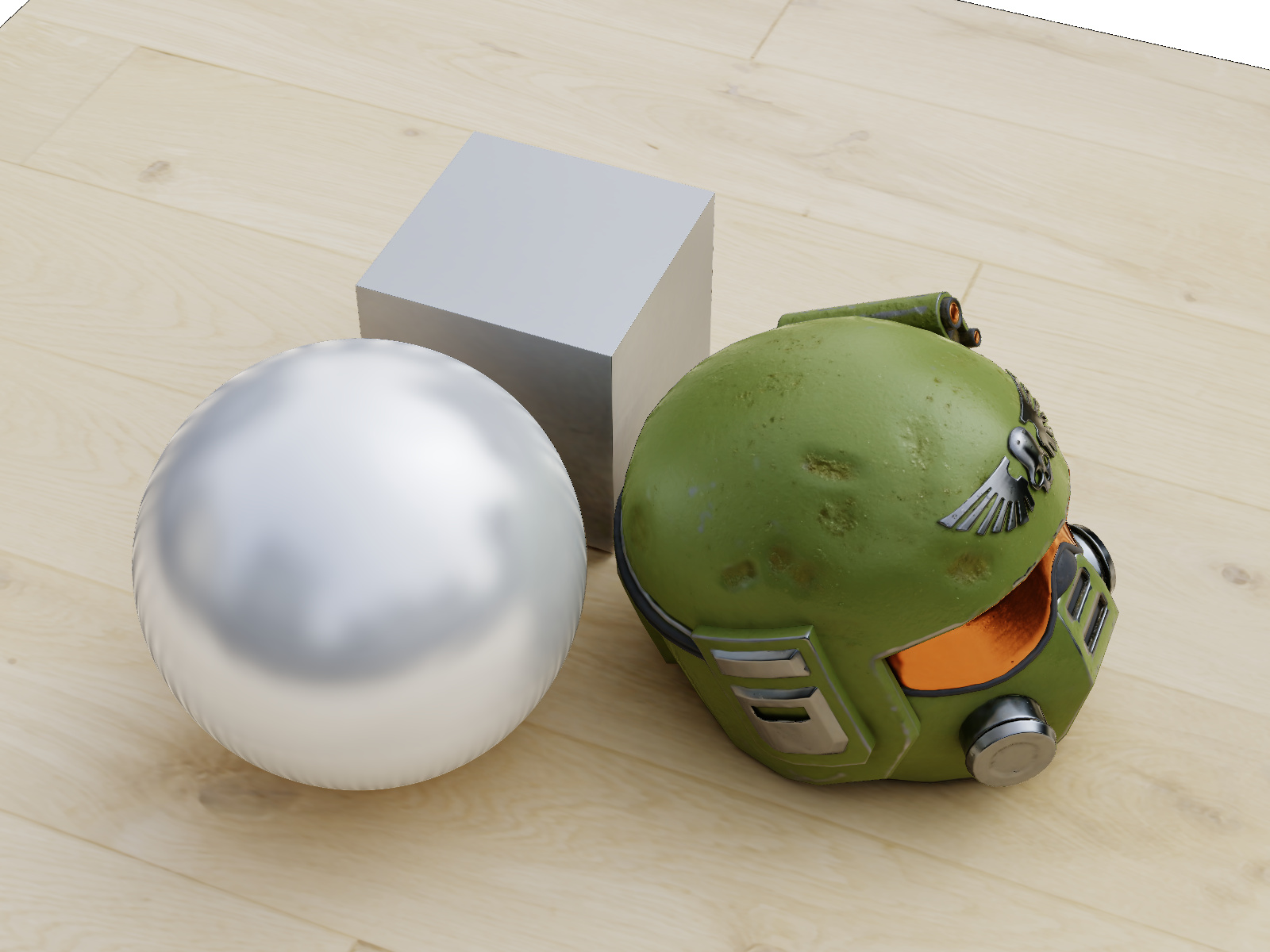}
            & \includegraphics[width=0.24\linewidth]{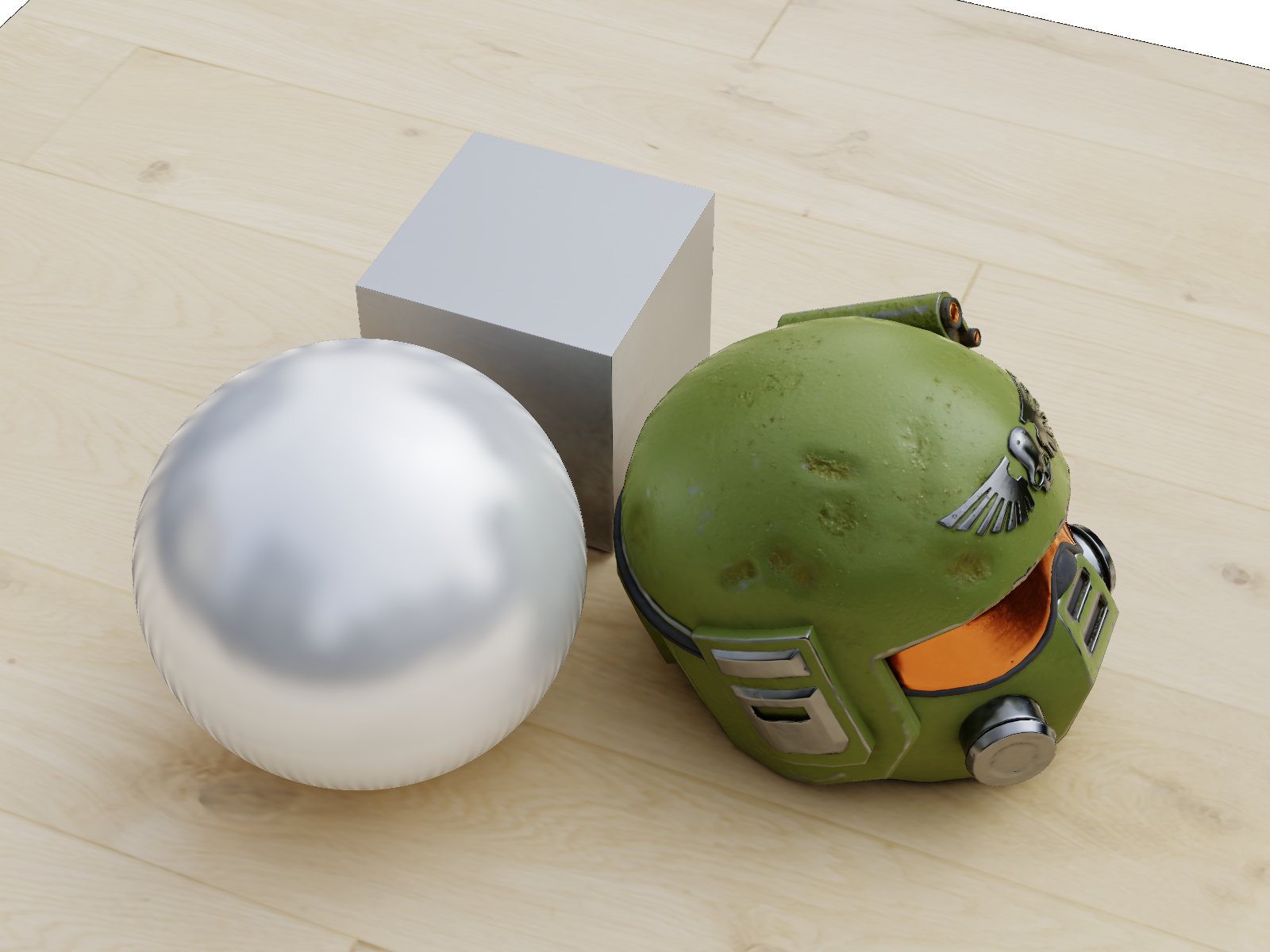}
            & \includegraphics[width=0.24\linewidth]{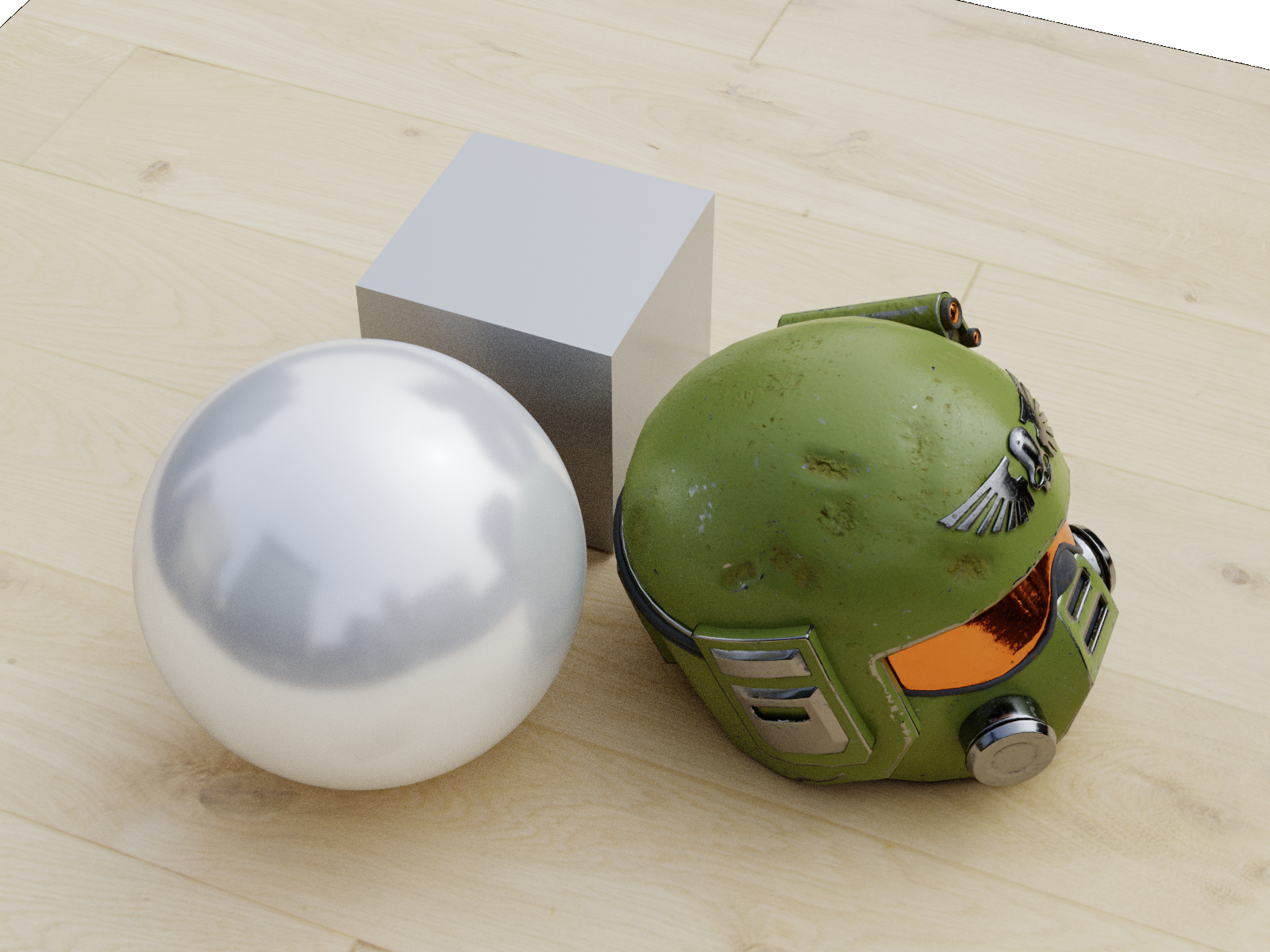}
            \\
            \multirow{1}{*}[0.5in]{\rotatebox[origin=c]{90}{Roughness}}
            & \includegraphics[width=0.24\linewidth]{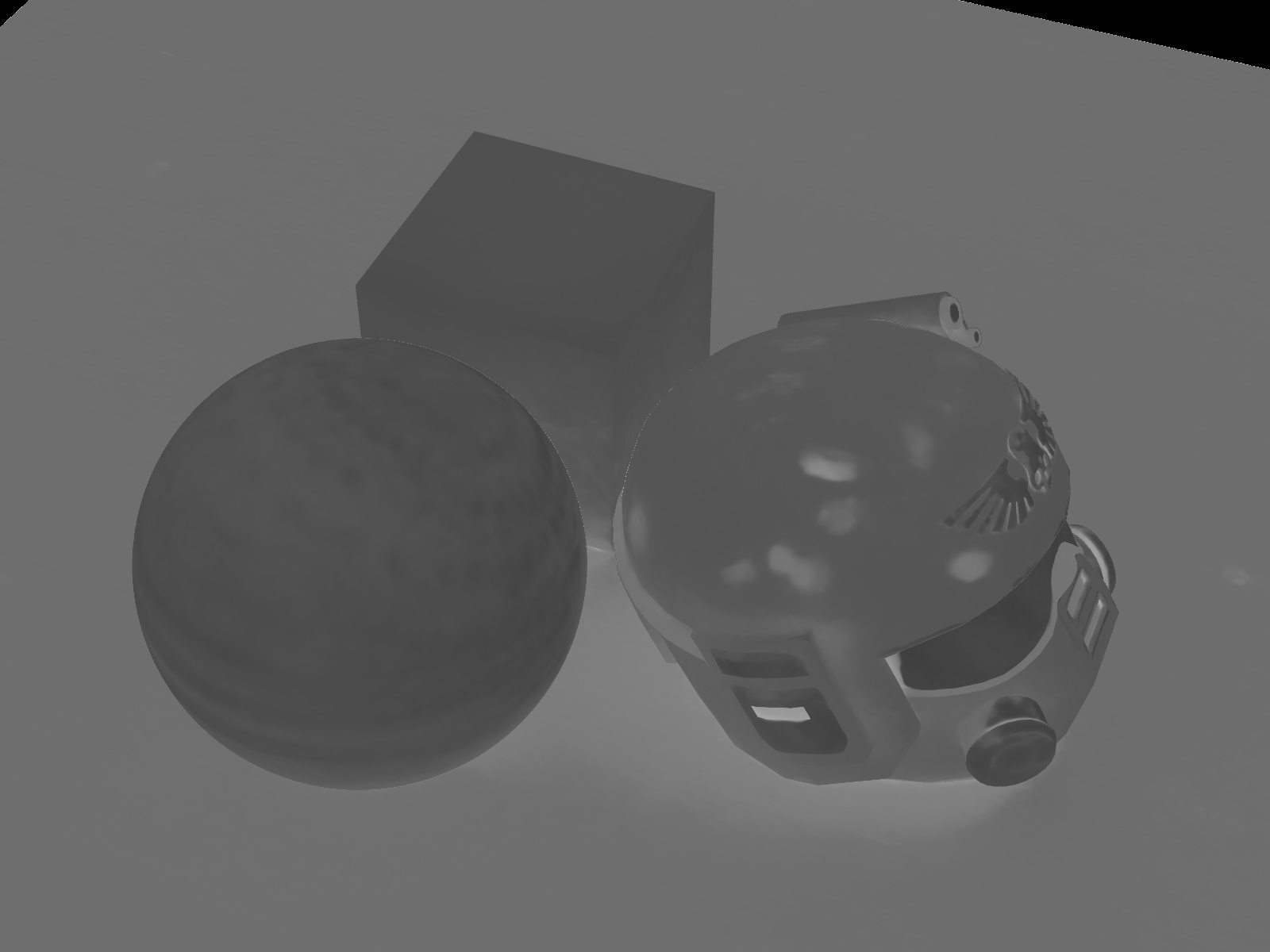}
            & \includegraphics[width=0.24\linewidth]{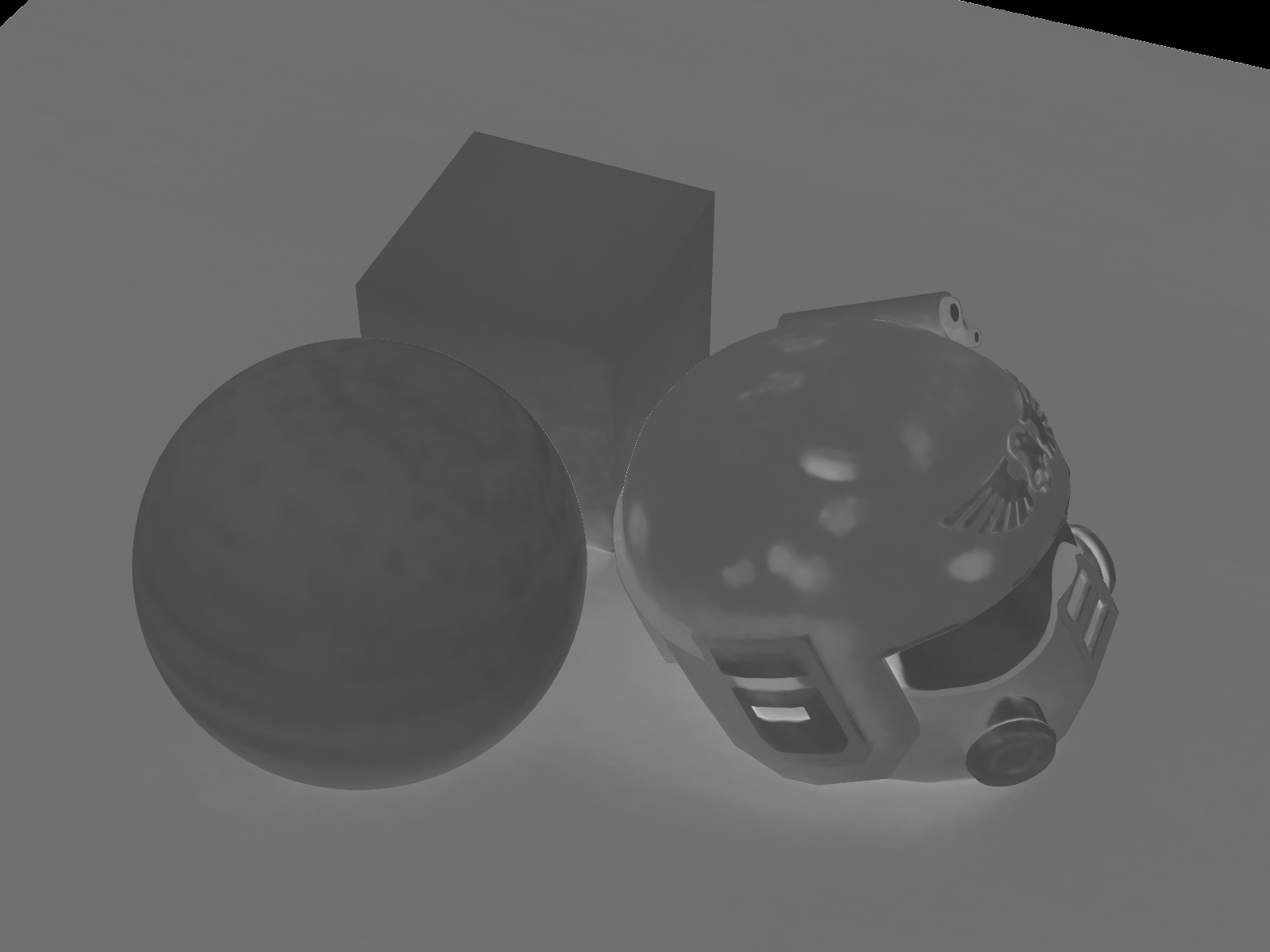}
            & \includegraphics[width=0.24\linewidth]{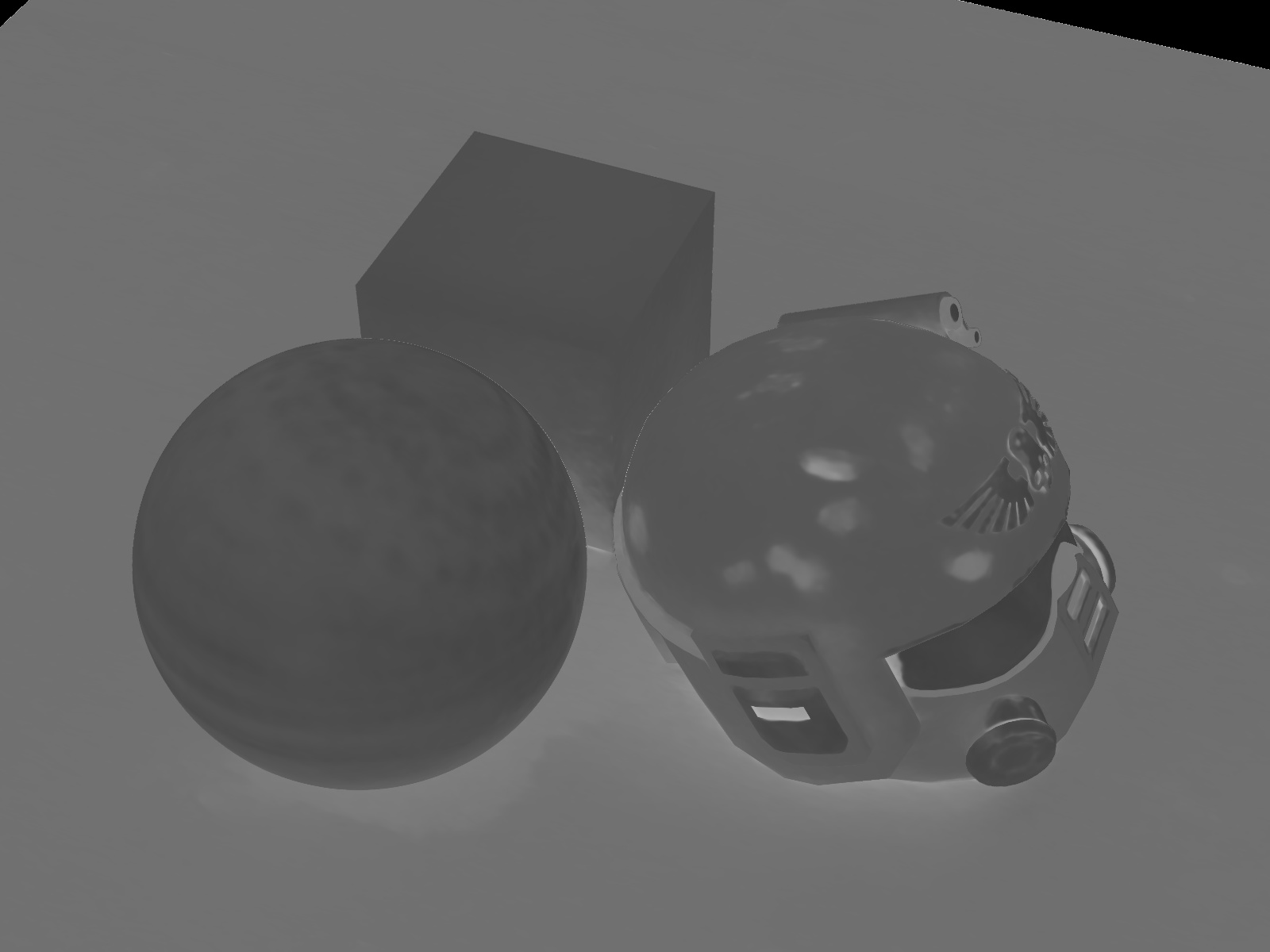}
            & \includegraphics[width=0.24\linewidth]{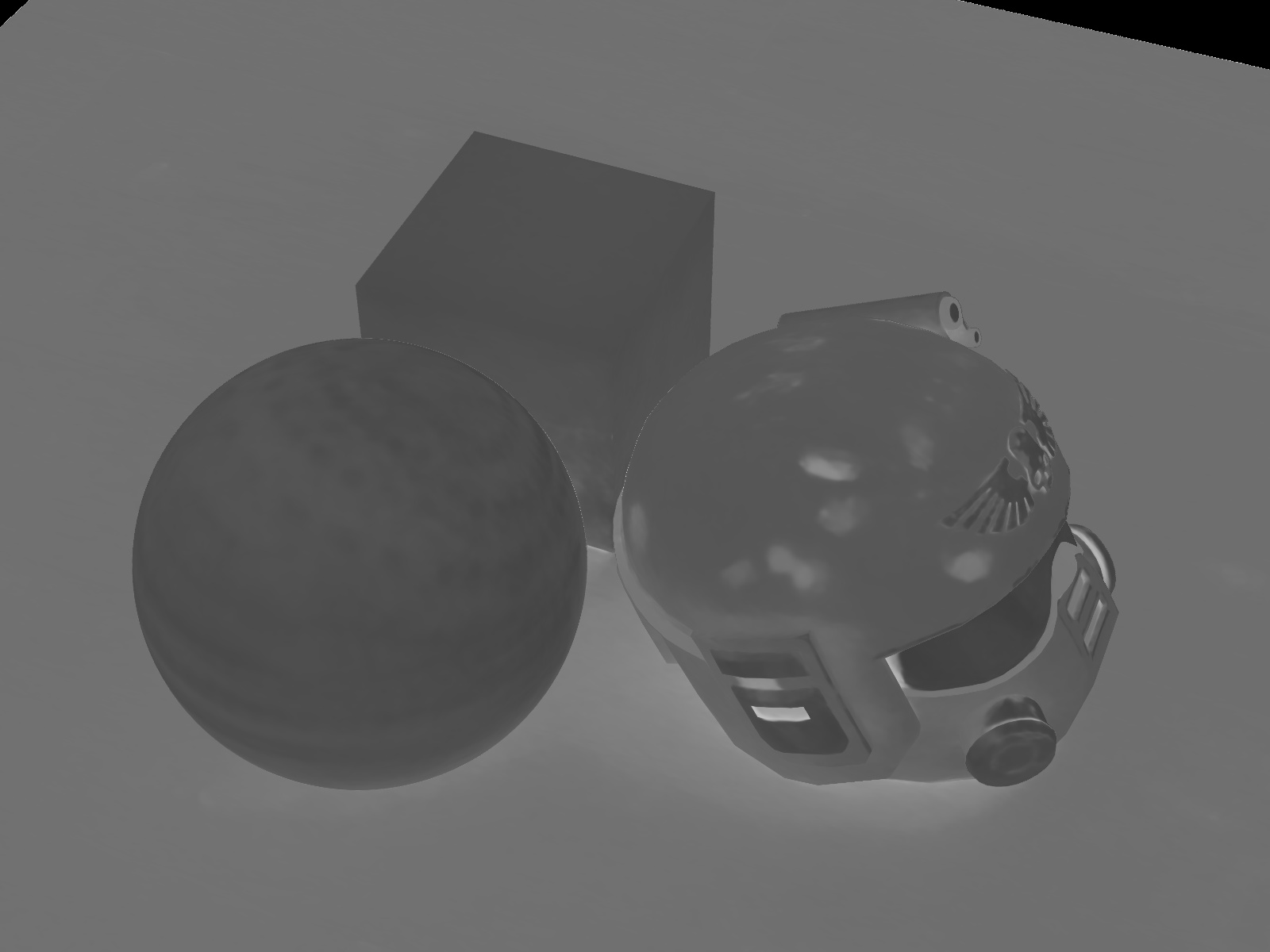}
            & \includegraphics[width=0.24\linewidth]{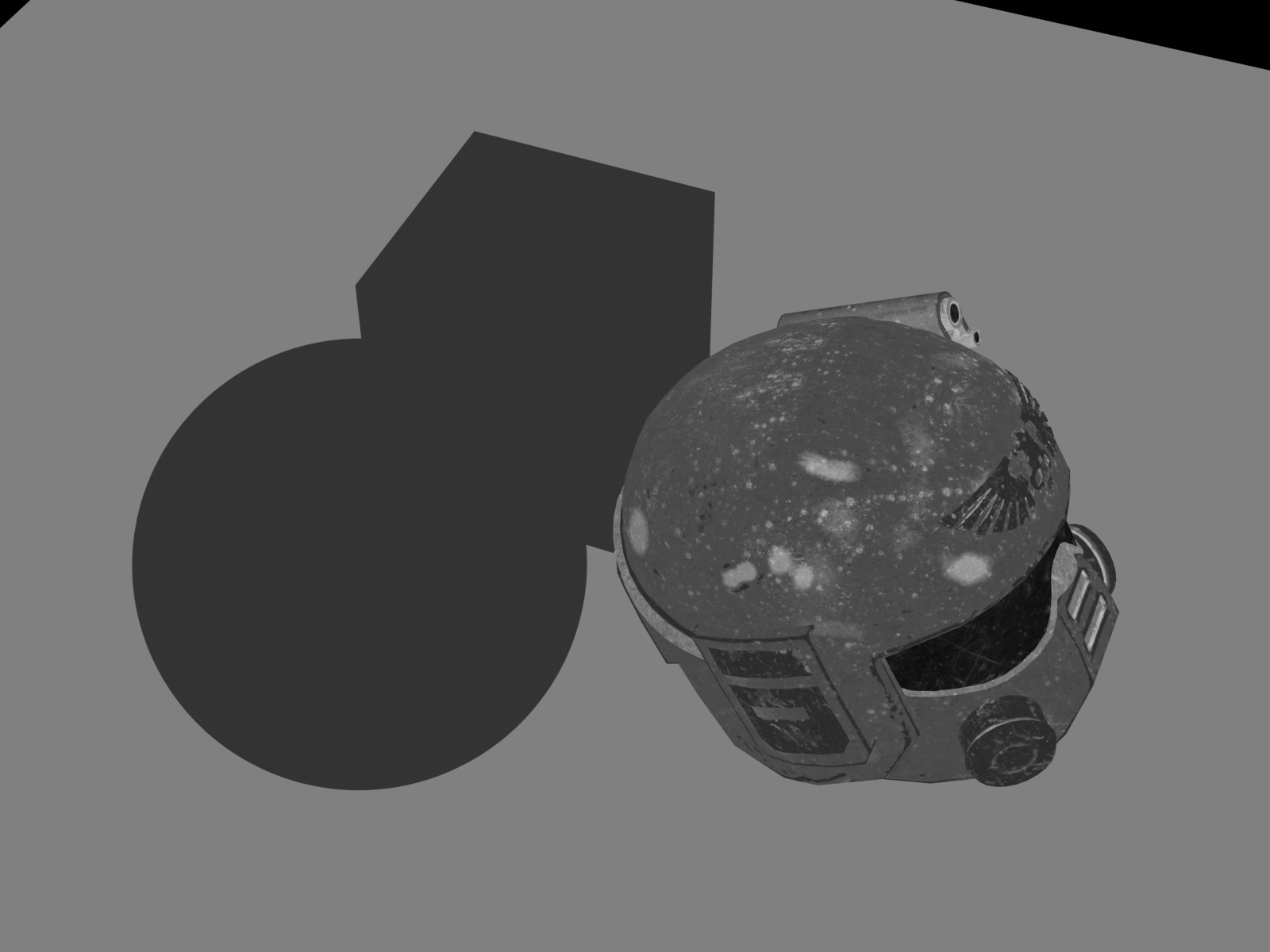}
            \\
            \multirow{1}{*}[0.75in]{\rotatebox[origin=c]{90}{Metallicness}}
            & \includegraphics[width=0.24\linewidth]{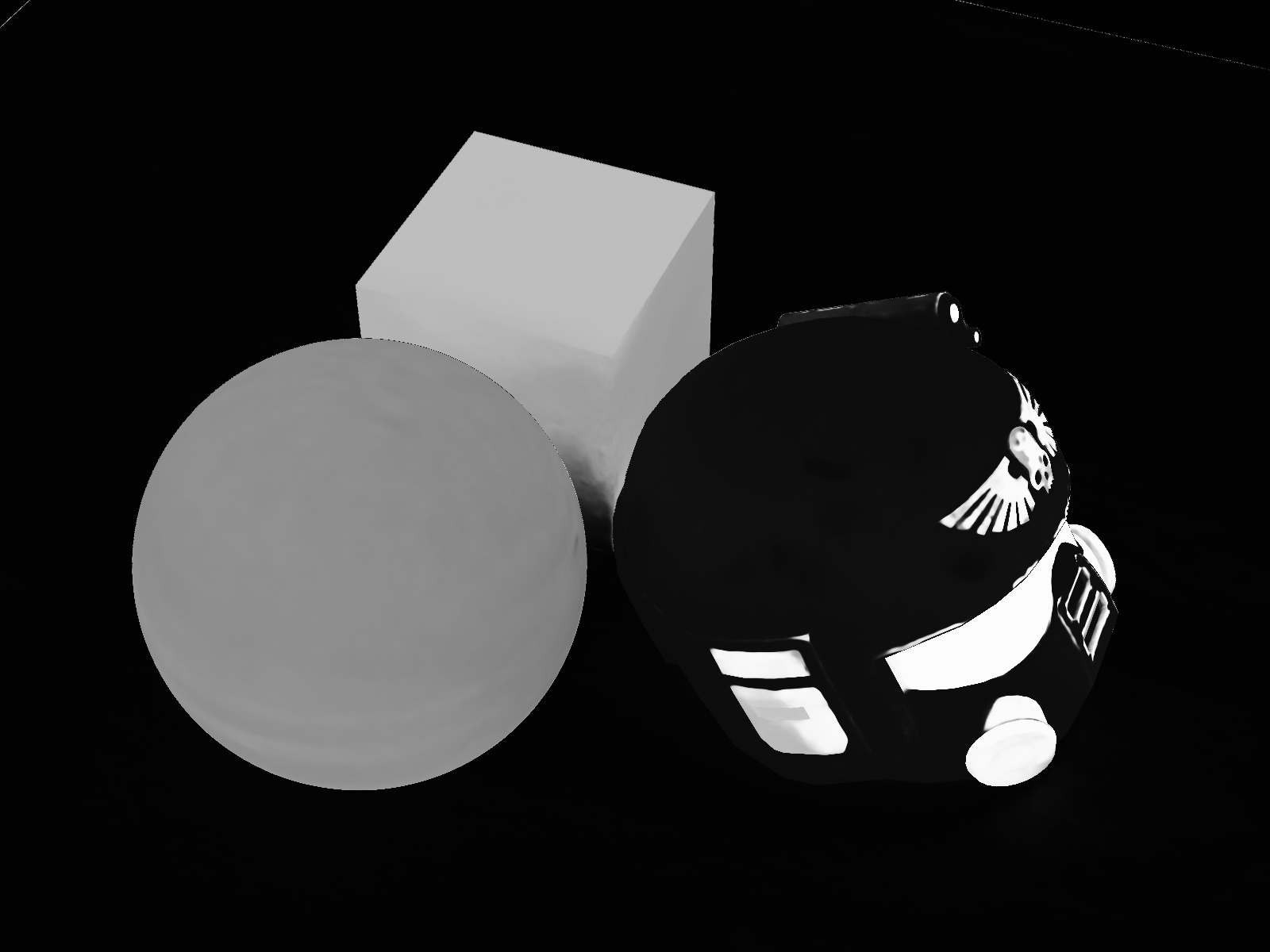}
            & \includegraphics[width=0.24\linewidth]{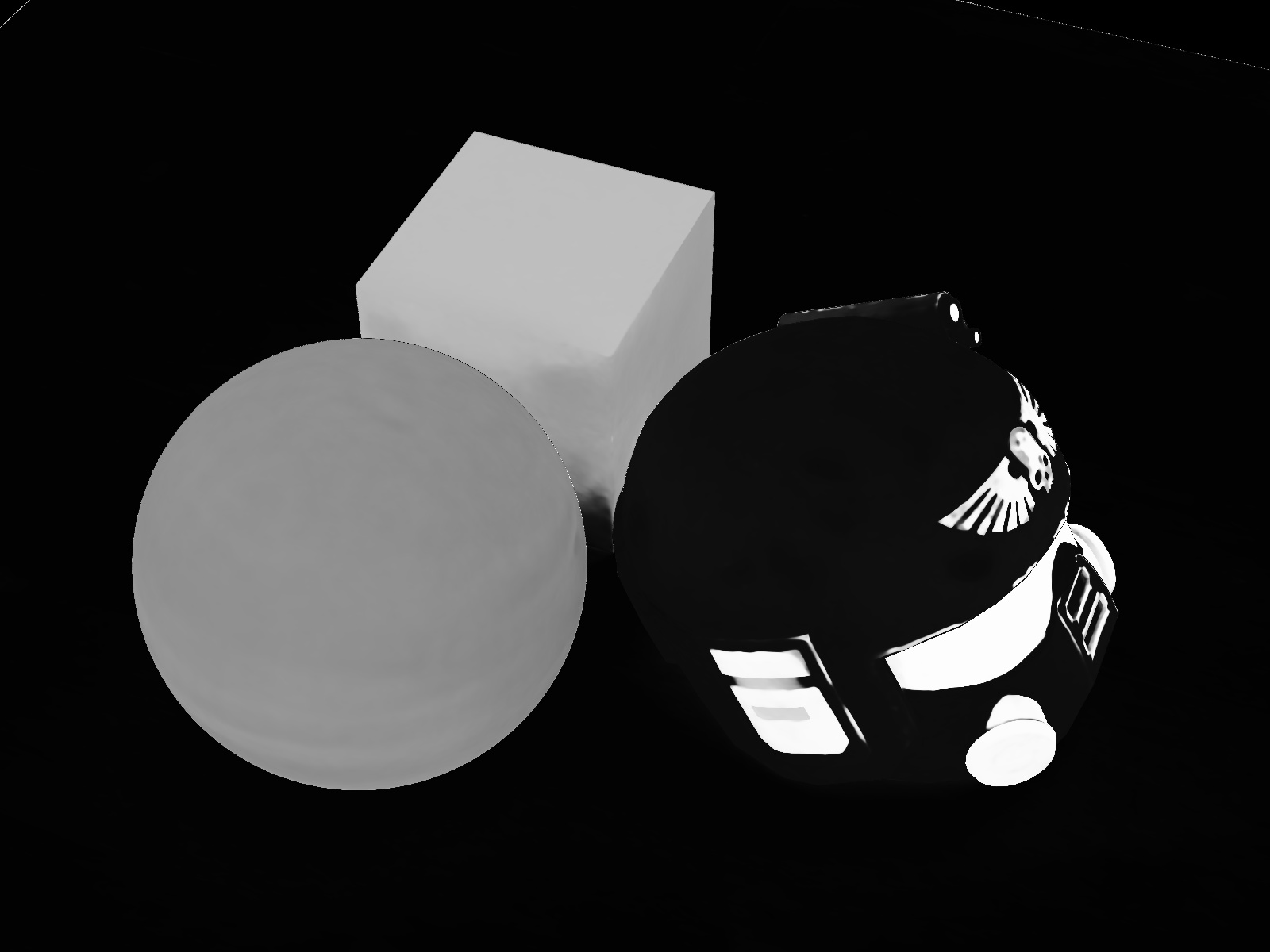}
            & \includegraphics[width=0.24\linewidth]{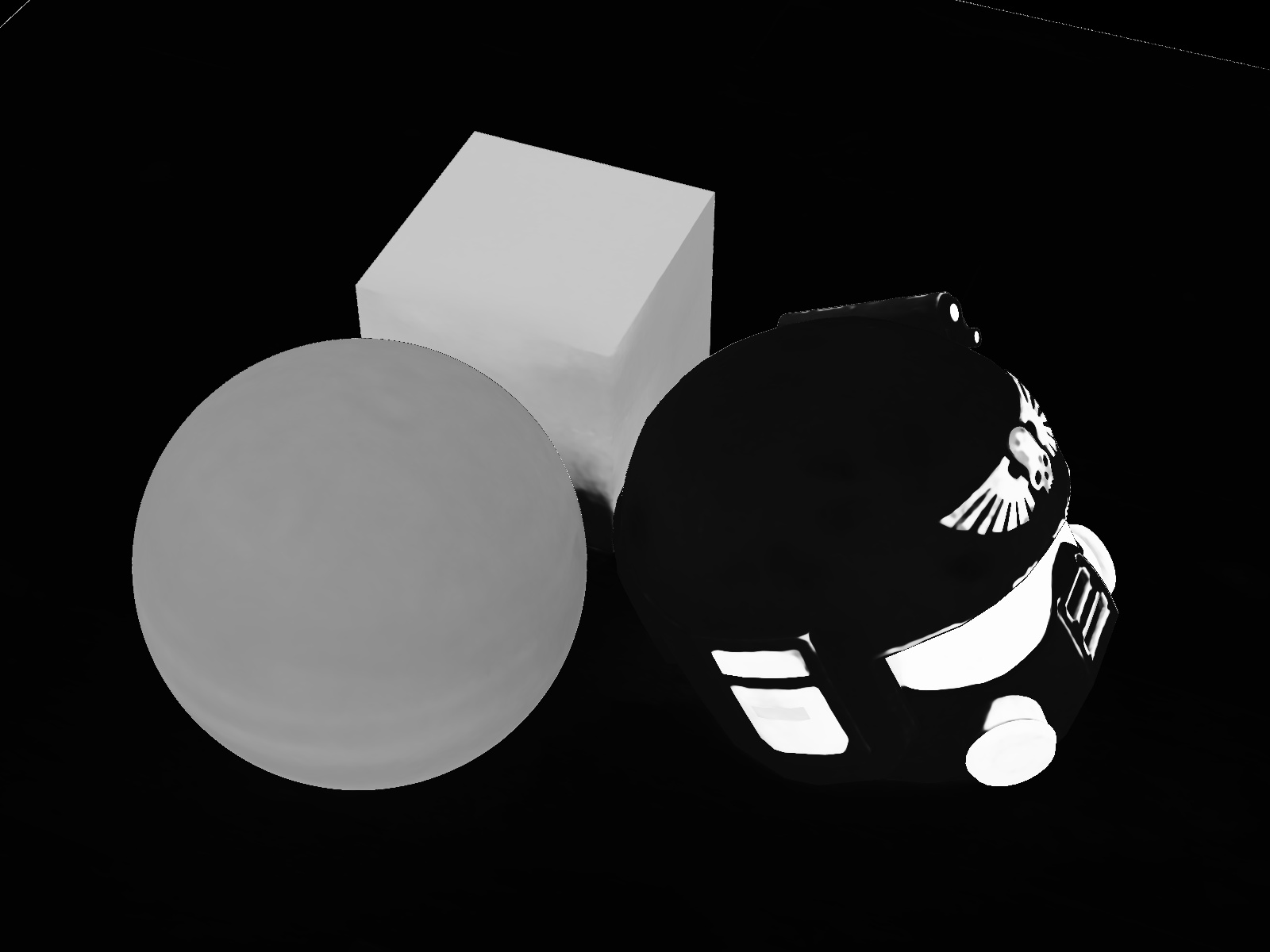}
            & \includegraphics[width=0.24\linewidth]{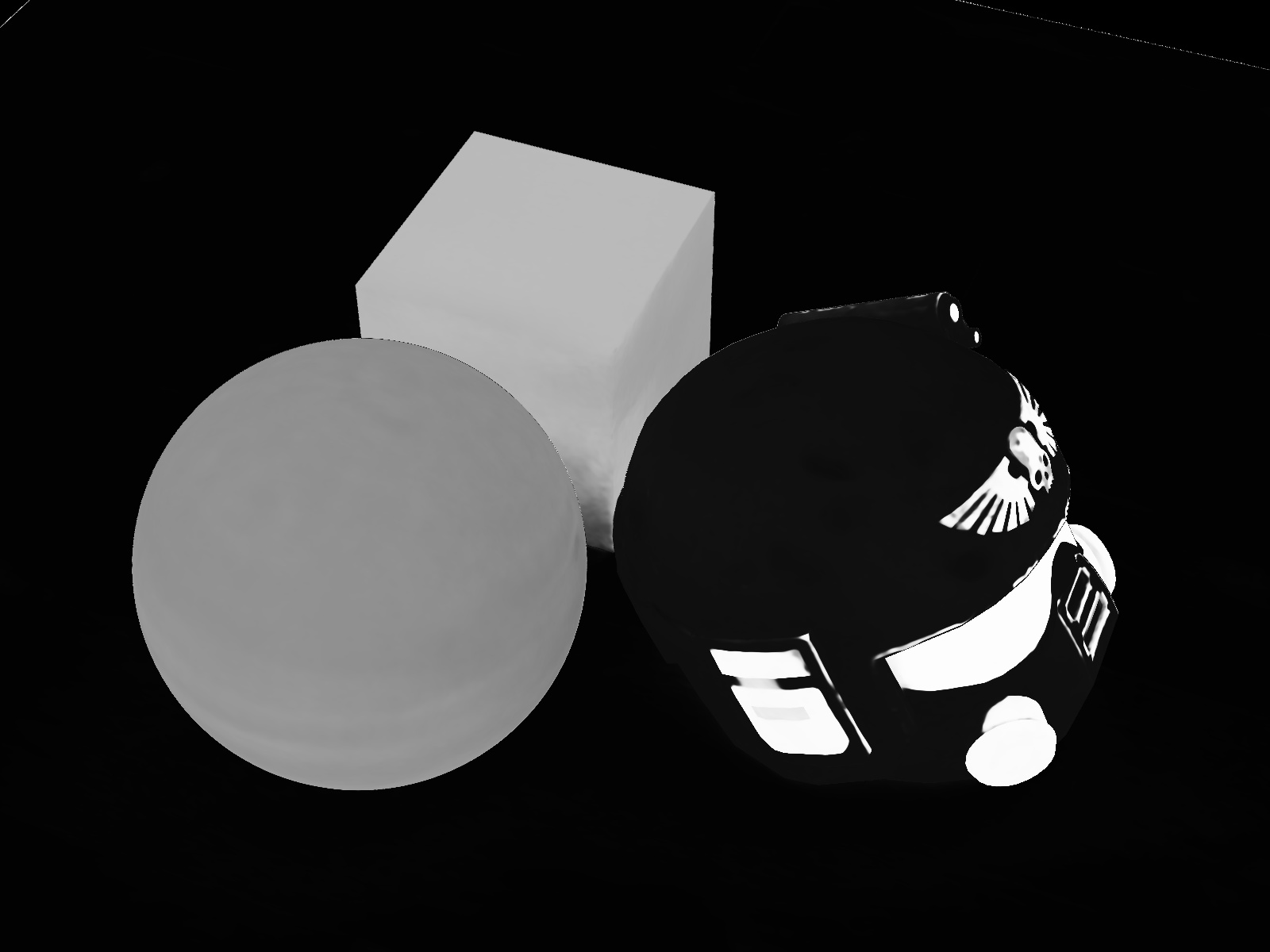}
            & \includegraphics[width=0.24\linewidth]{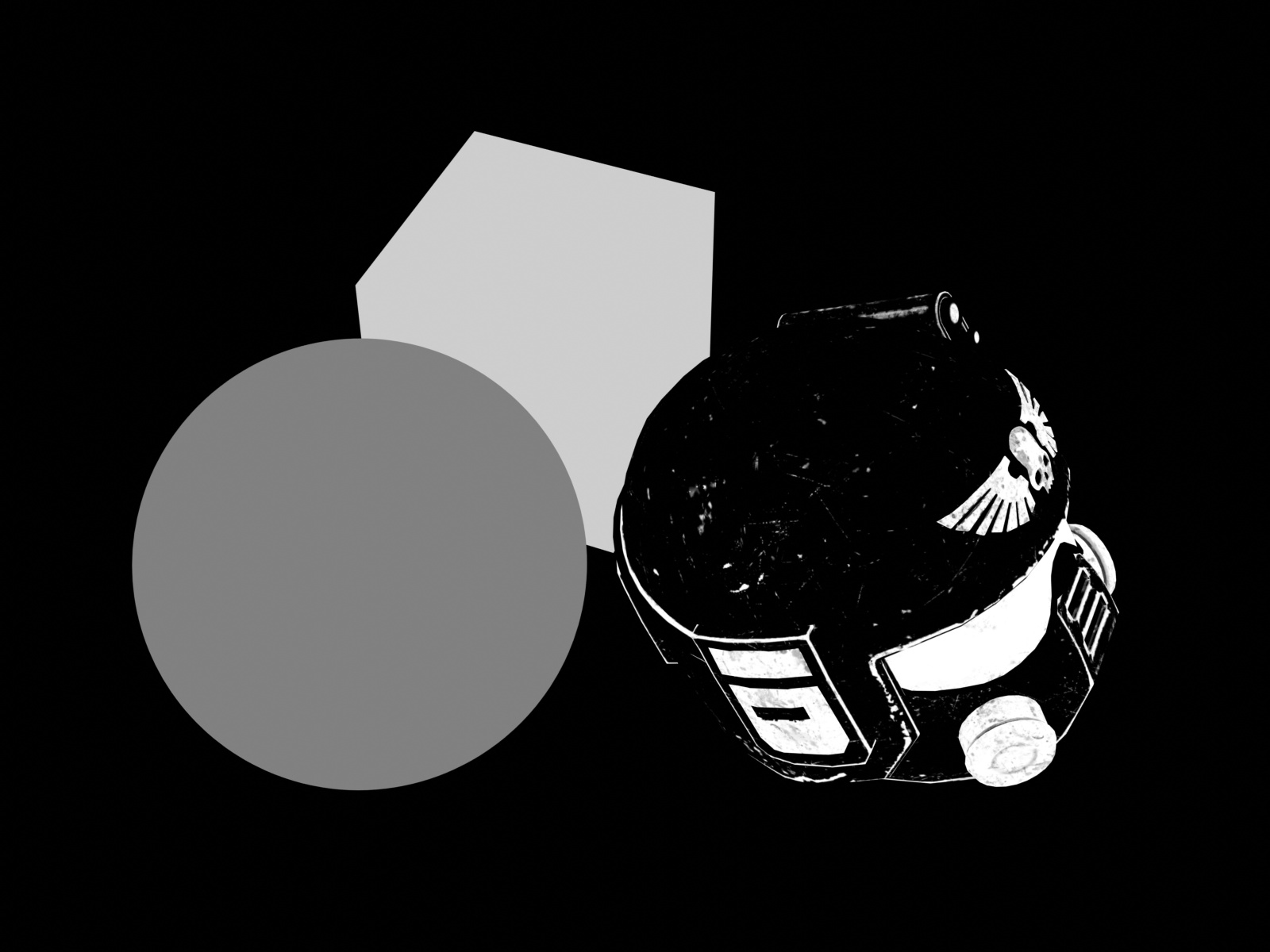}
            \\
            \multirow{1}{*}[0.65in]{\rotatebox[origin=c]{90}{Albedo}}
            & \includegraphics[width=0.24\linewidth]{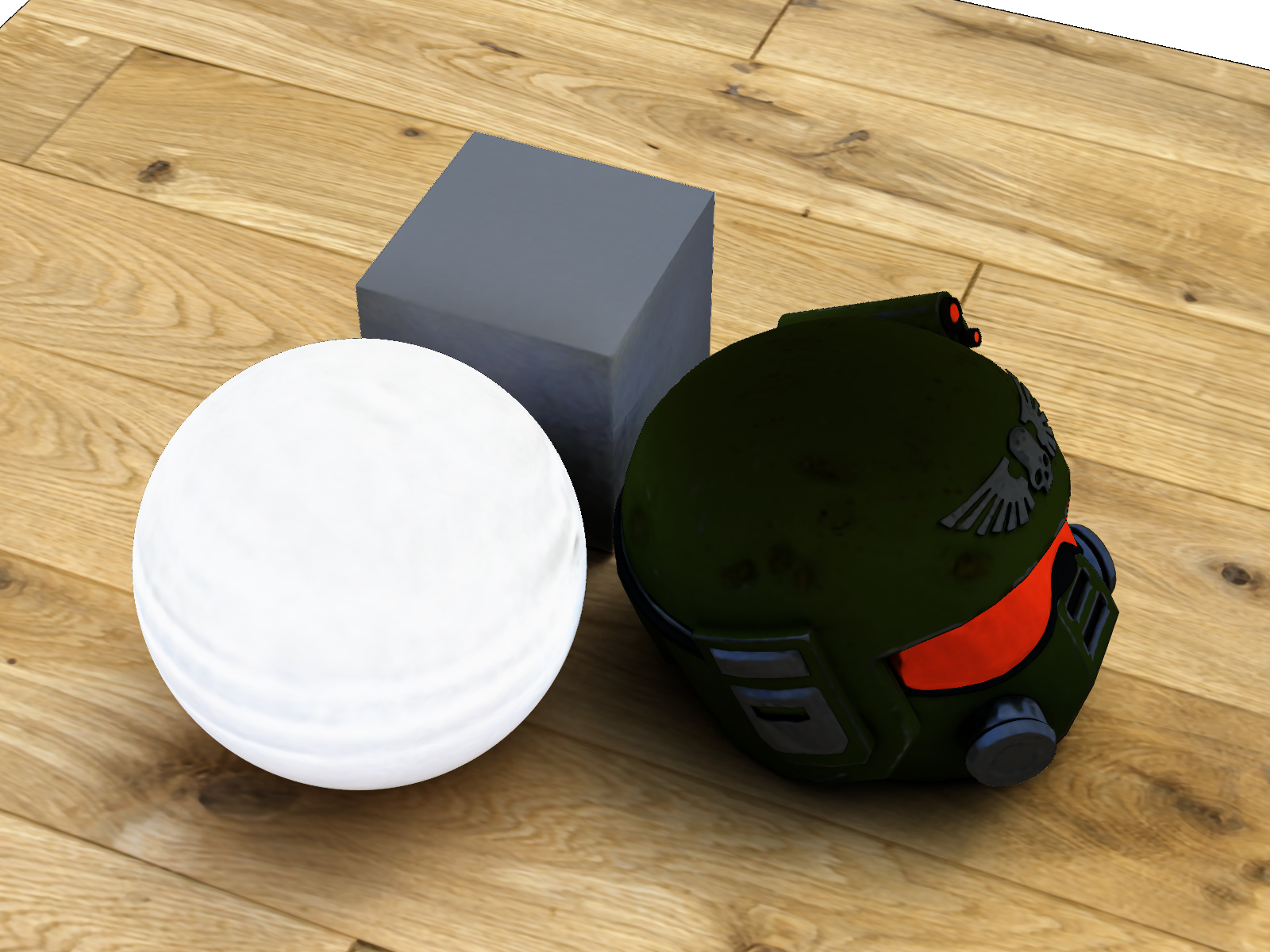}
            & \includegraphics[width=0.24\linewidth]{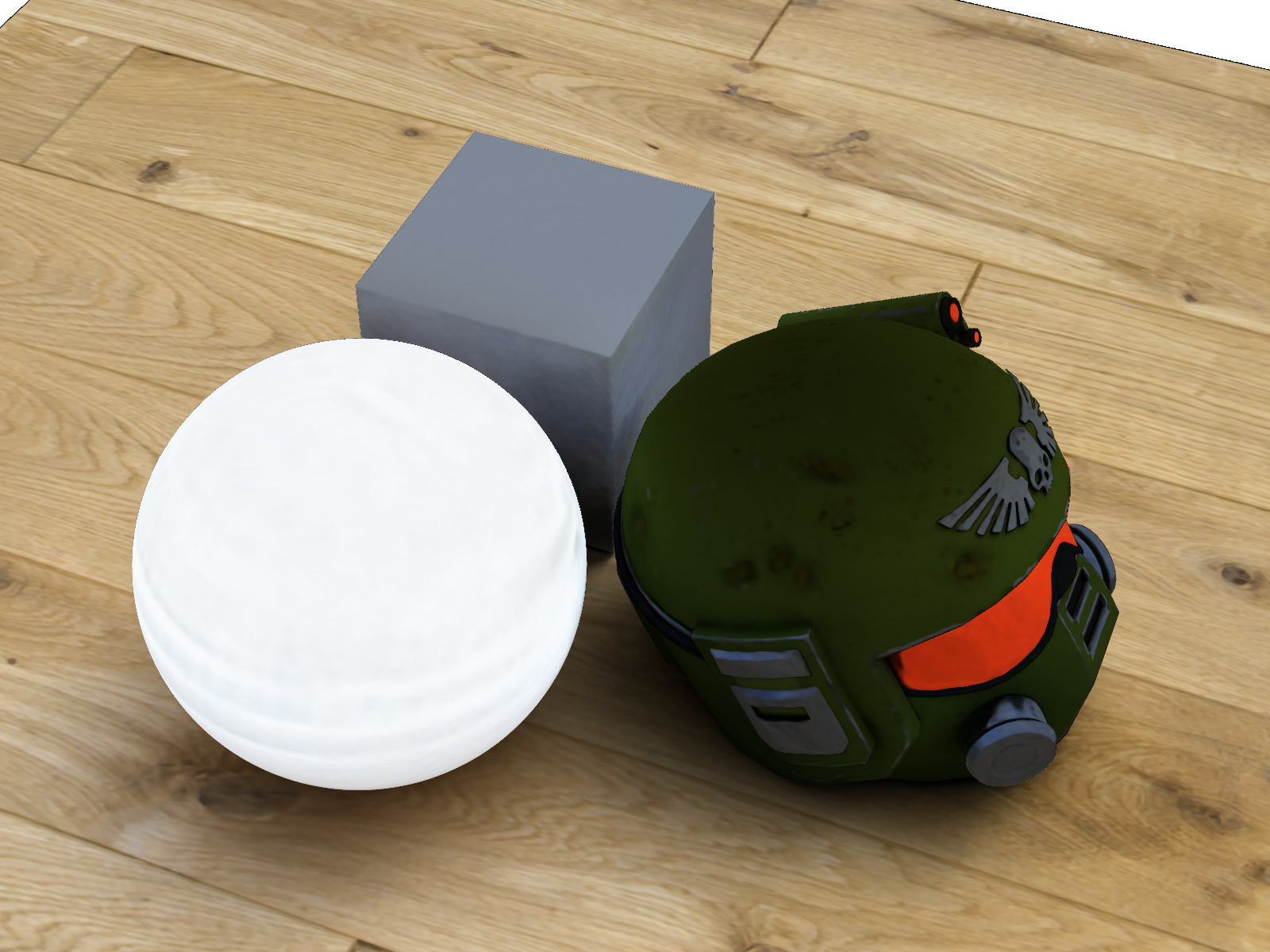}
            & \includegraphics[width=0.24\linewidth]{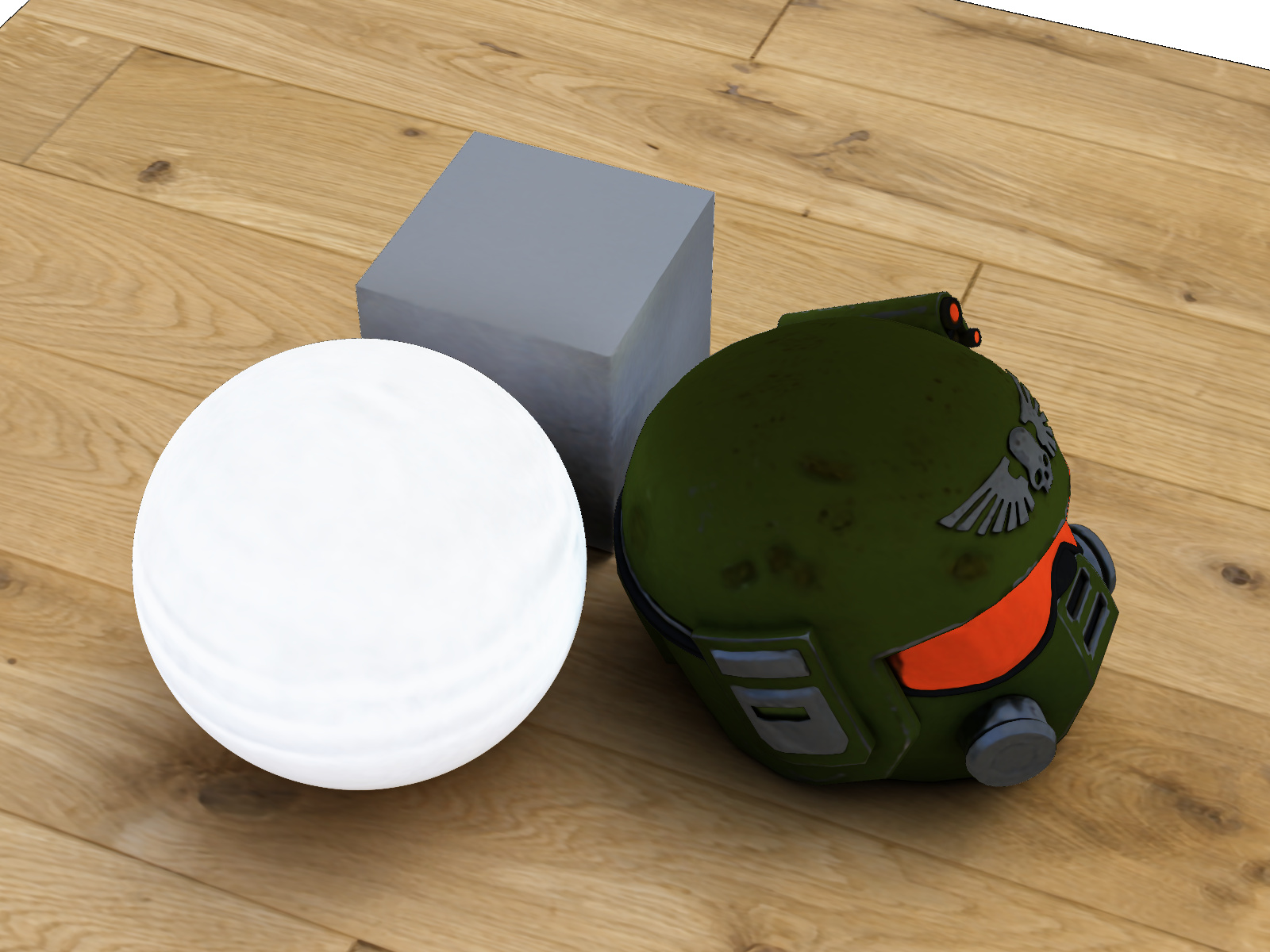}
            & \includegraphics[width=0.24\linewidth]{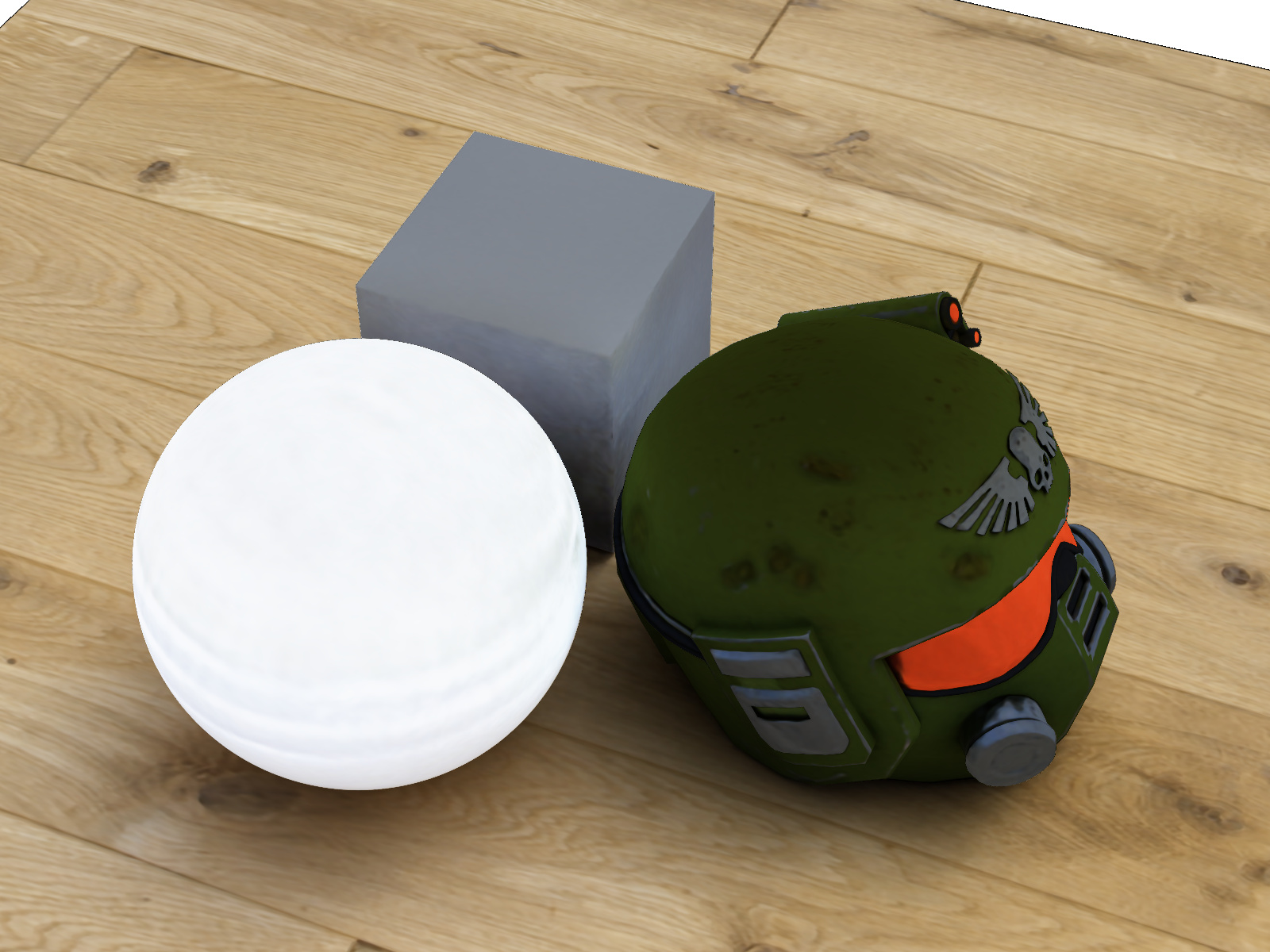}
            & \includegraphics[width=0.24\linewidth]{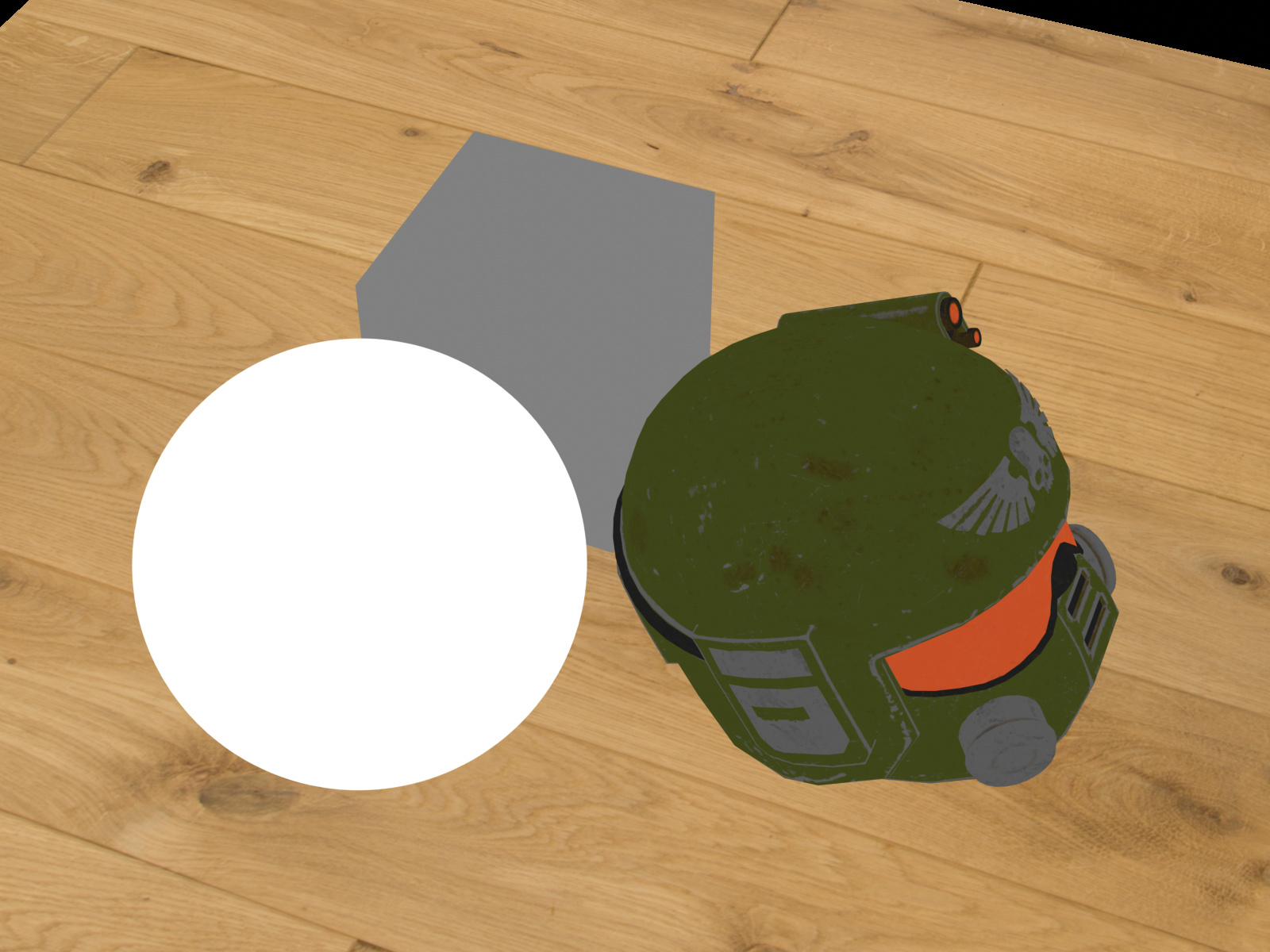}
            \\
            & Baseline (ID 1) & $\mathcal{L}_{\text{cons}}$ (ID 2) & $\mathcal{L}_{\text{spec}}$ (ID 3) & $\mathcal{L}_{\text{spec}}$ (ID 4) & Ground Truth  \\
        \end{tabular}
    }
    \vspace{-8pt}
    \caption{\textbf{Ablation of our physics-based losses.} Qualitative performance of ablated versions of PBR-NeRF and of its full version is compared on the \textit{City} illumination with predicted illumination and Disney BRDF parameter estimation on novel views. \dag: no ground-truth environment maps are provided with the dataset.}
    \label{fig:pbr_loss_ablation_qualitative}
\end{figure*}

We present qualitative results in Fig.~\ref{fig:pbr_loss_ablation_qualitative} for the ablation study of the proposed physics-based losses on the NeILF++ dataset City scene in Tab.~\ref{tab:pbr_loss_ablation} of Sec.~\ref{sec:ablation}.
To ensure a fair comparison, all ablated versions share the same hyperparameters as our proposed method described in Sec.~\ref{sec:exp_setup}, varying only the inclusion of the proposed loss terms.

The baseline (ID 1) of NeILF++ \cite{zhang2023neilf++} without the Lambertian loss, exhibits many ``baked-in'' specular highlights.
As shown in Fig.~\ref{fig:pbr_loss_ablation_qualitative}, the baseline's predicted albedo contains many specular artifacts being added to the diffuse lobe.
Specifically, there is fringing on the sphere, shadowing artifacts on the cube (particularly on its lower regions shadowed by the other two objects), and an overly bright floor due to the entanglement of the diffuse and specular lobes.
These specular effects in the albedo cause incorrect patches on the shadowed cube in the metallicness prediction and overly dark patches in the bottom left of the incident light estimation for the associated cube location.

Adding the Conservation of Energy loss (ID 2) mitigates these artifacts by enforcing energy conservation for the diffuse and specular lobes.
As a result, the lighting estimation is also improved since the estimated light does not need to compensate for the previously non-energy-conserving (overly reflective) predicted materials.
The predicted lighting is more consistent and improves the originally overly dim patches in the bottom left corner of the baseline (corresponding to the cube).
Reducing specular artifacts in the lighting also improves the material estimation, as shown by the increased albedo, roughness, and metallicness PSNR in Tab.~\ref{tab:pbr_loss_ablation}.
Qualitatively, we also see that the albedo of the helmet and cube are effectively recovered in all regions except the shadowed area.
Despite these improvements, there are still fringing artifacts on the sphere as well as shadowing effects on the ground and cube.

Adding the specular loss instead of the Conservation of Energy loss (ID 3), shows similar improvements in albedo, roughness, and metallicness over the baseline (ID 1).

Finally, incorporating both the Conservation of Energy and specular losses in our final method (ID 4) improves the albedo, roughness, and metallicness even further.
We once again highlight that using both our novel physics-based losses improves roughness by 0.35 PSNR, metallicness by 0.27 PSNR, and albedo by 3.28 PSNR over the baseline (ID 1).
Qualitatively, the albedo estimation, which is most important, is significantly improved in our final method (ID 4), especially with the green albedo of the helmet and the gray albedo of the cube.
The roughness and metallicness of the shadowed area of the cube are also recovered with fewer artifacts.
However, there are still limitations with our proposed method, which highlights directions for future improvement.
In particular, we fail to fully reproduce the specular effects in the helmet's reflective visor and metallic pieces in the predicted RGB.
This limitation is likely due to our overestimation of roughness and the missing fine details in our metallicness prediction.
Furthermore, we still exhibit limited ``baked-in'' specular effects in our predicted albedo with fringing on the sphere and incorrectly predict shadowy patches on the cube and floor.

Overall, we convincingly improve the state-of-the-art material estimation over our baseline NeILF++ \cite{zhang2023neilf++}.
While challenges remain in handling high-frequency specular effects and reflective surfaces, our method convincingly advances the state of the art in joint material and lighting estimation.

\section{Additional Qualitative Results}
\label{sec:supp:sota_qualitative_results}
\begin{figure*}[tb]
    \renewcommand{\arraystretch}{1.5}
    \centering
    \small
    \resizebox{\textwidth}{!}{
                \begin{tabular}{ccccccc}
            & & Lighting \dag & RGB & Roughness & Metallicness & Albedo \\
            \multirow{3}{*}[0.5in]{\raisebox{-1.2in}{\rotatebox[origin=c]{90}{City}}}
                        & \multirow{1}{*}[0.5in]{\rotatebox[origin=c]{90}{NeILF++}}
                        & \includegraphics[width=0.2\linewidth]{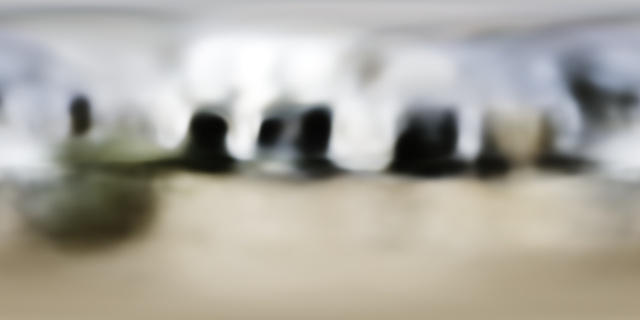}
                        & \includegraphics[width=0.2\linewidth]{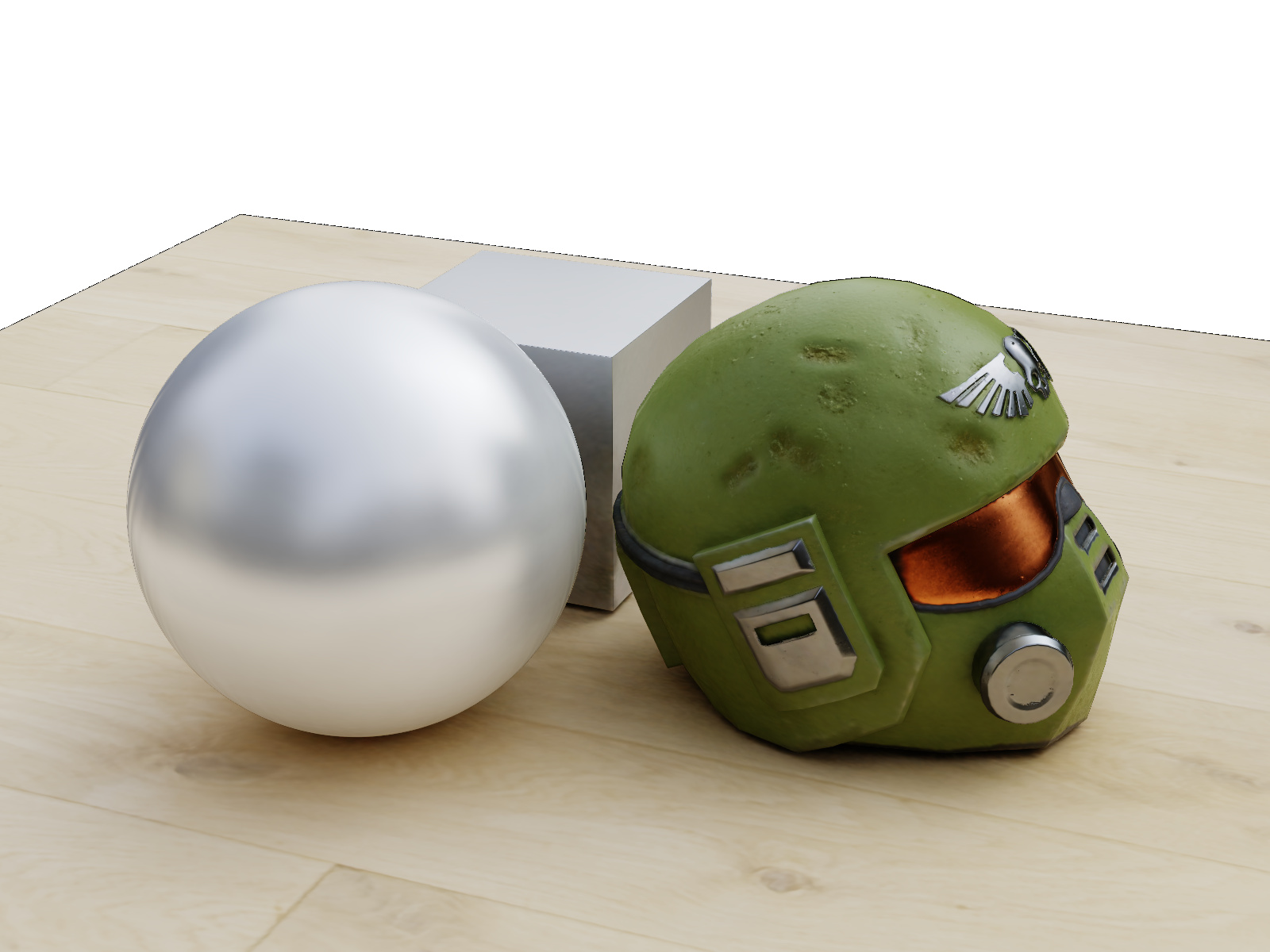}
                        & \includegraphics[width=0.2\linewidth]{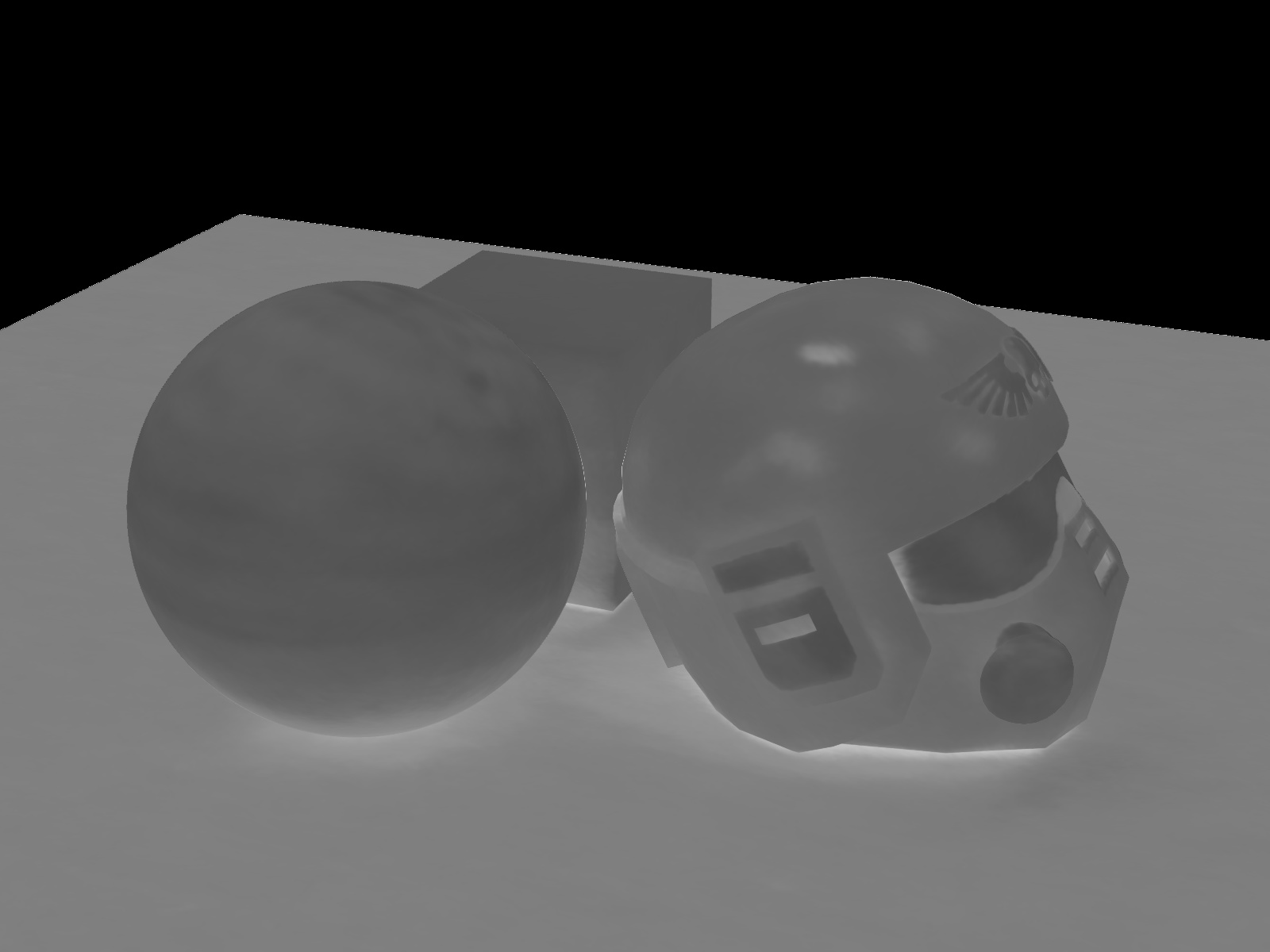}
                        & \includegraphics[width=0.2\linewidth]{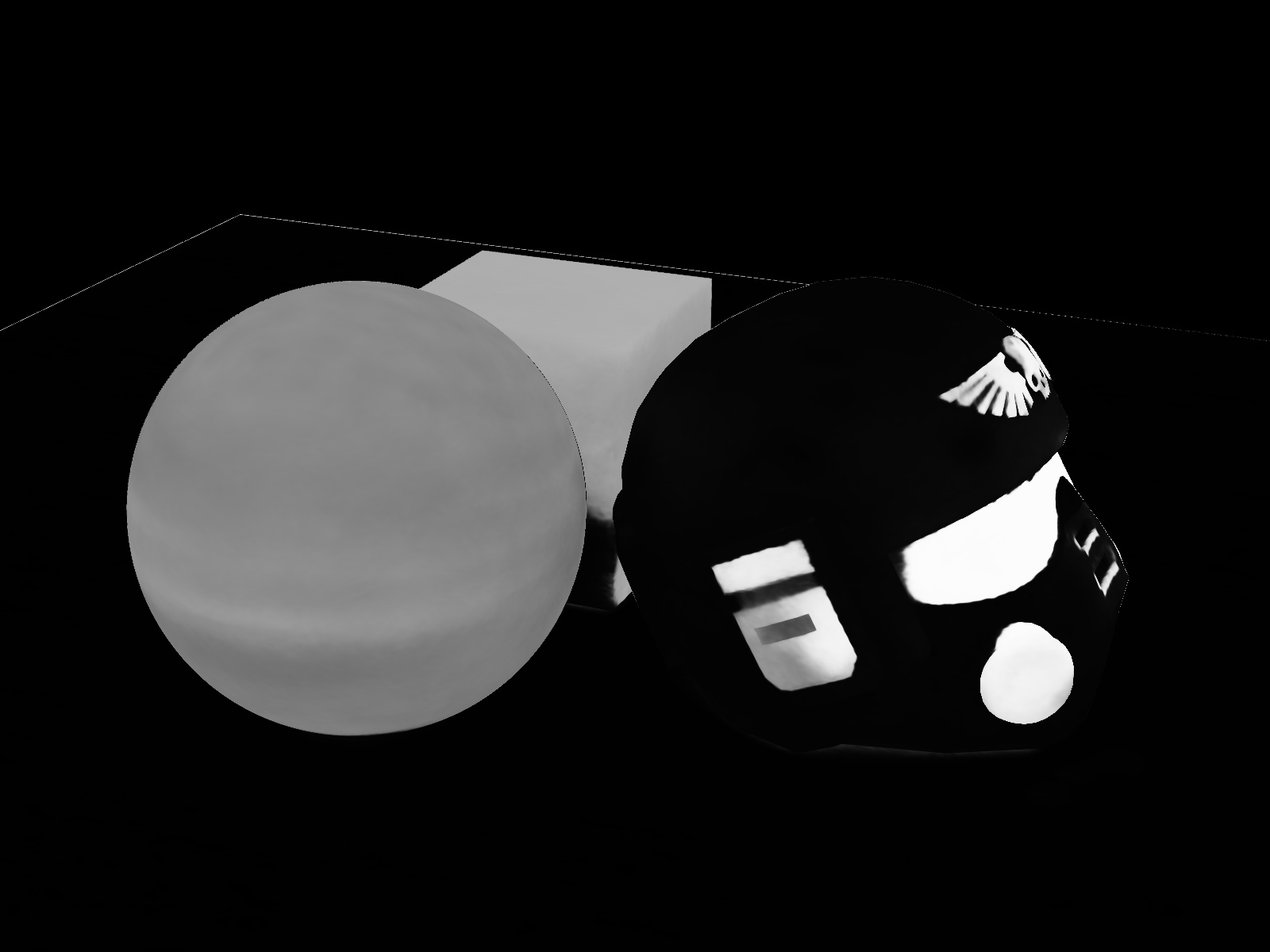}
                        & \includegraphics[width=0.2\linewidth]{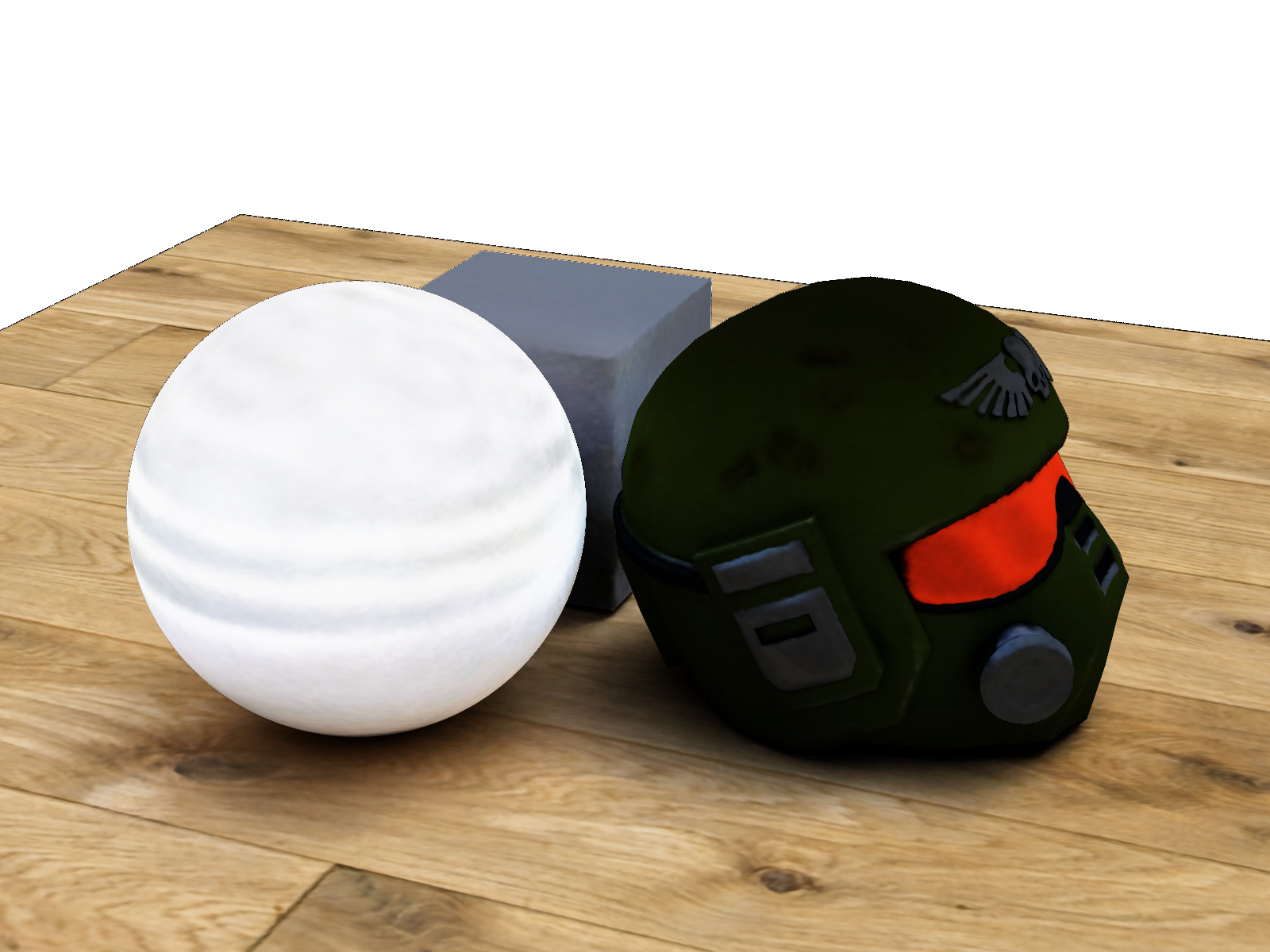} \\
                        & \multirow{1}{*}[0.5in]{\rotatebox[origin=c]{90}{Ours}}
                        & \includegraphics[width=0.2\linewidth]{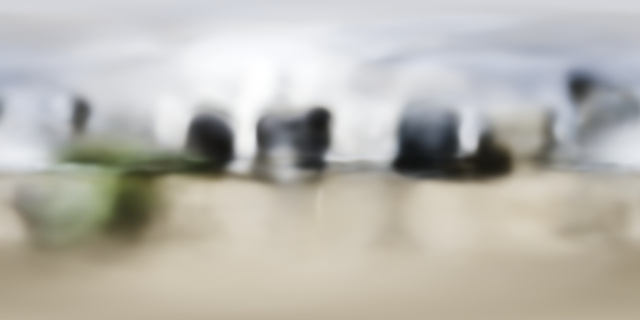}
                        & \includegraphics[width=0.2\linewidth]{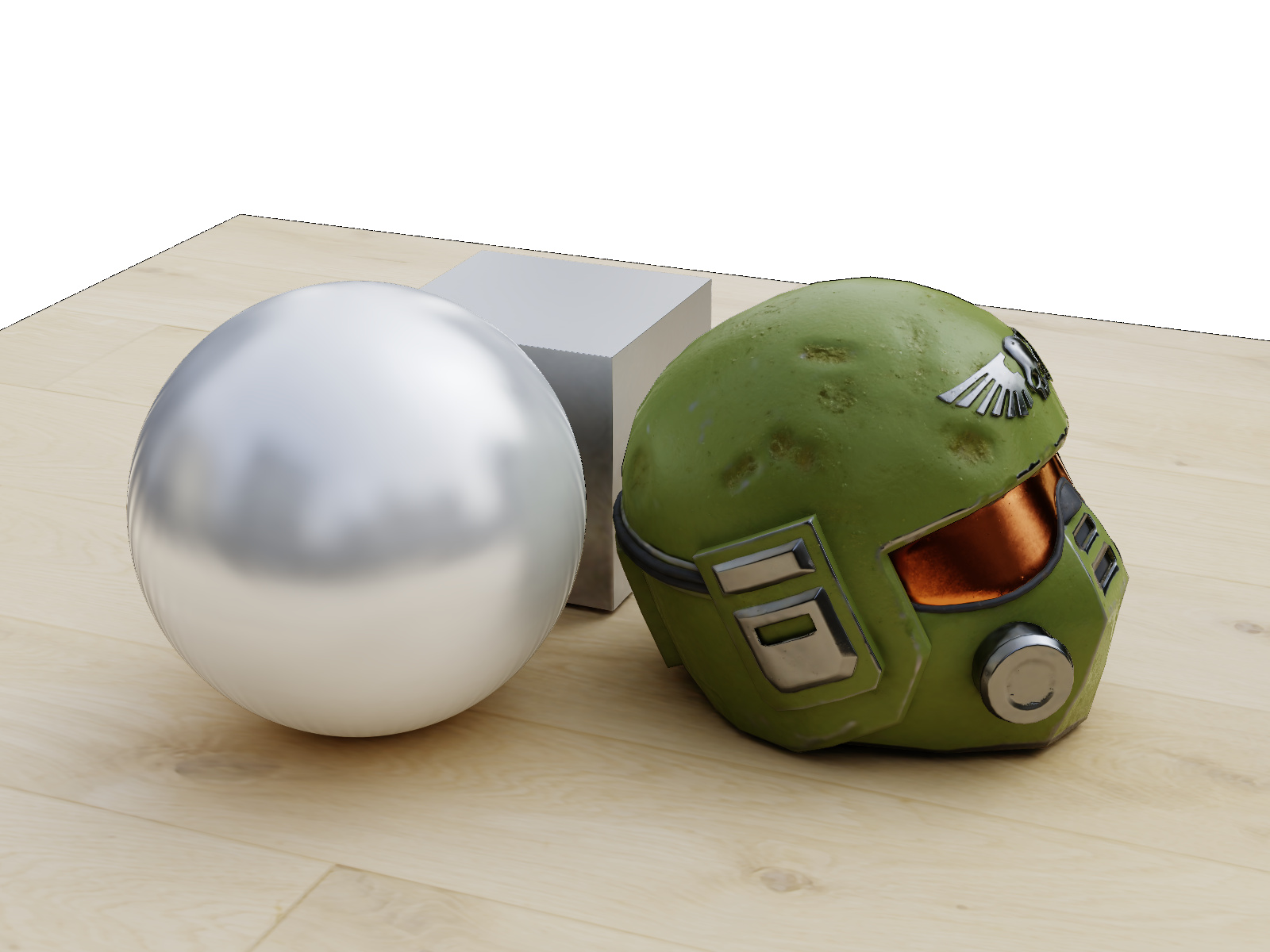}
                        & \includegraphics[width=0.2\linewidth]{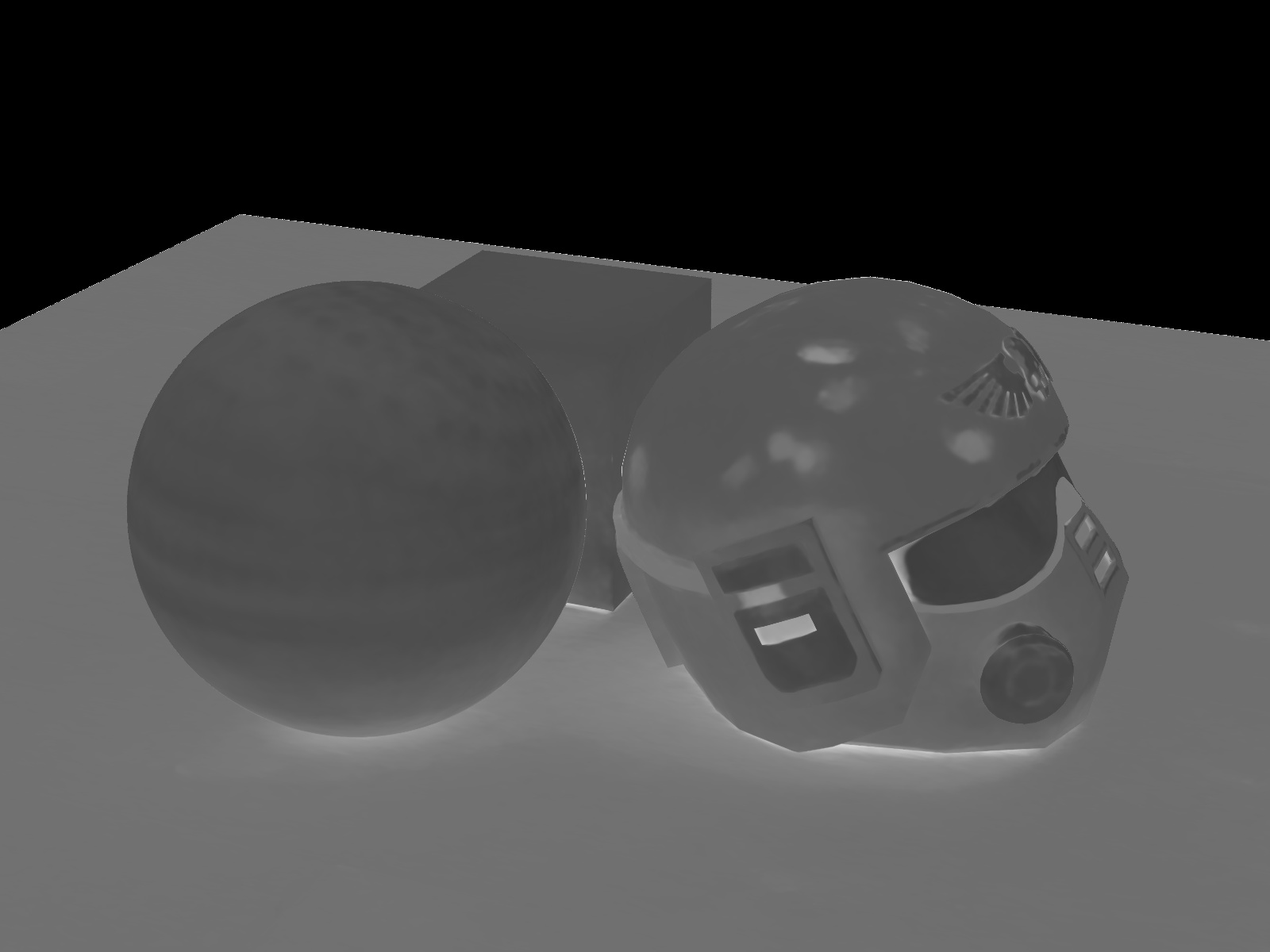}
                        & \includegraphics[width=0.2\linewidth]{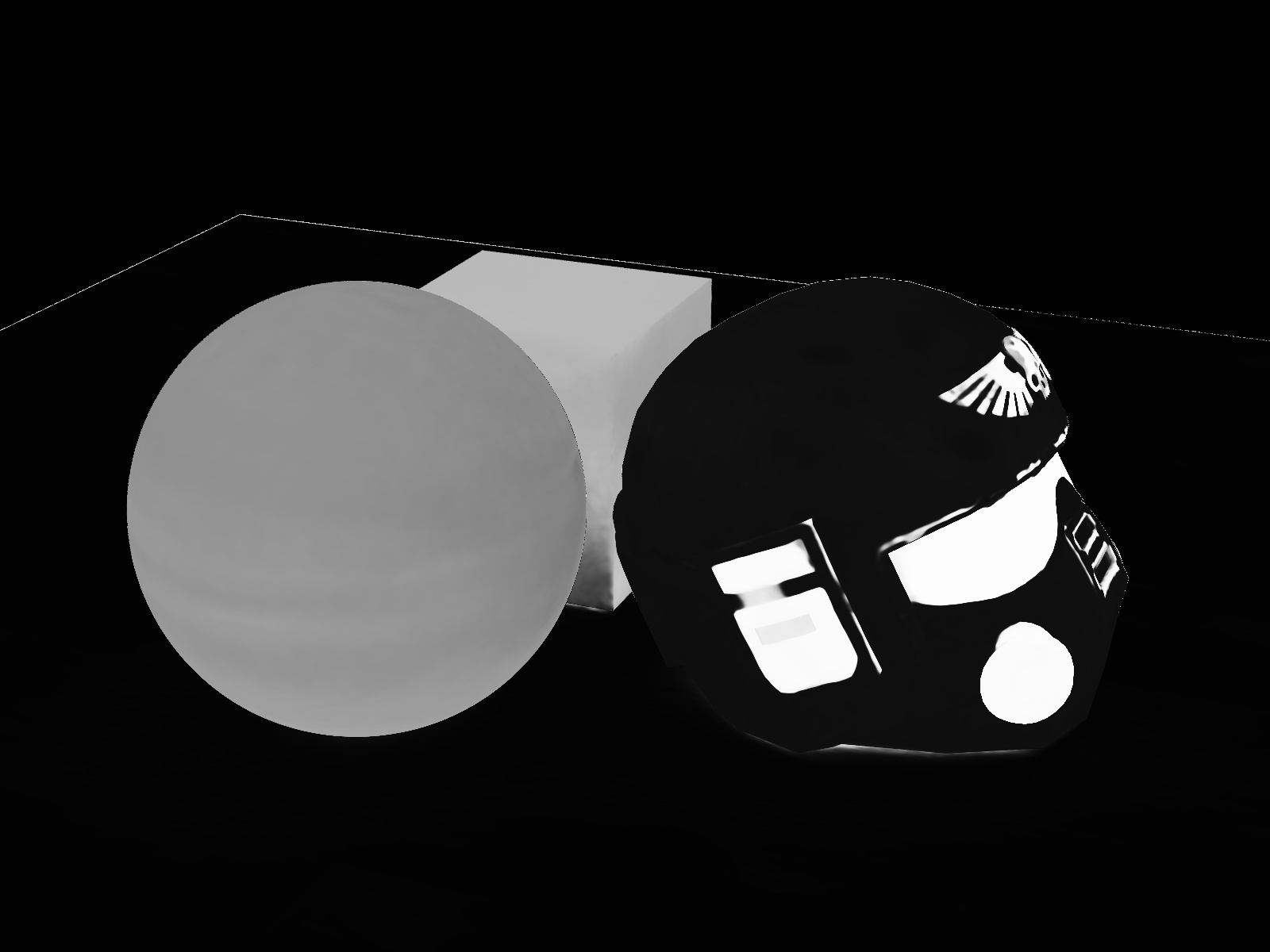}
                        & \includegraphics[width=0.2\linewidth]{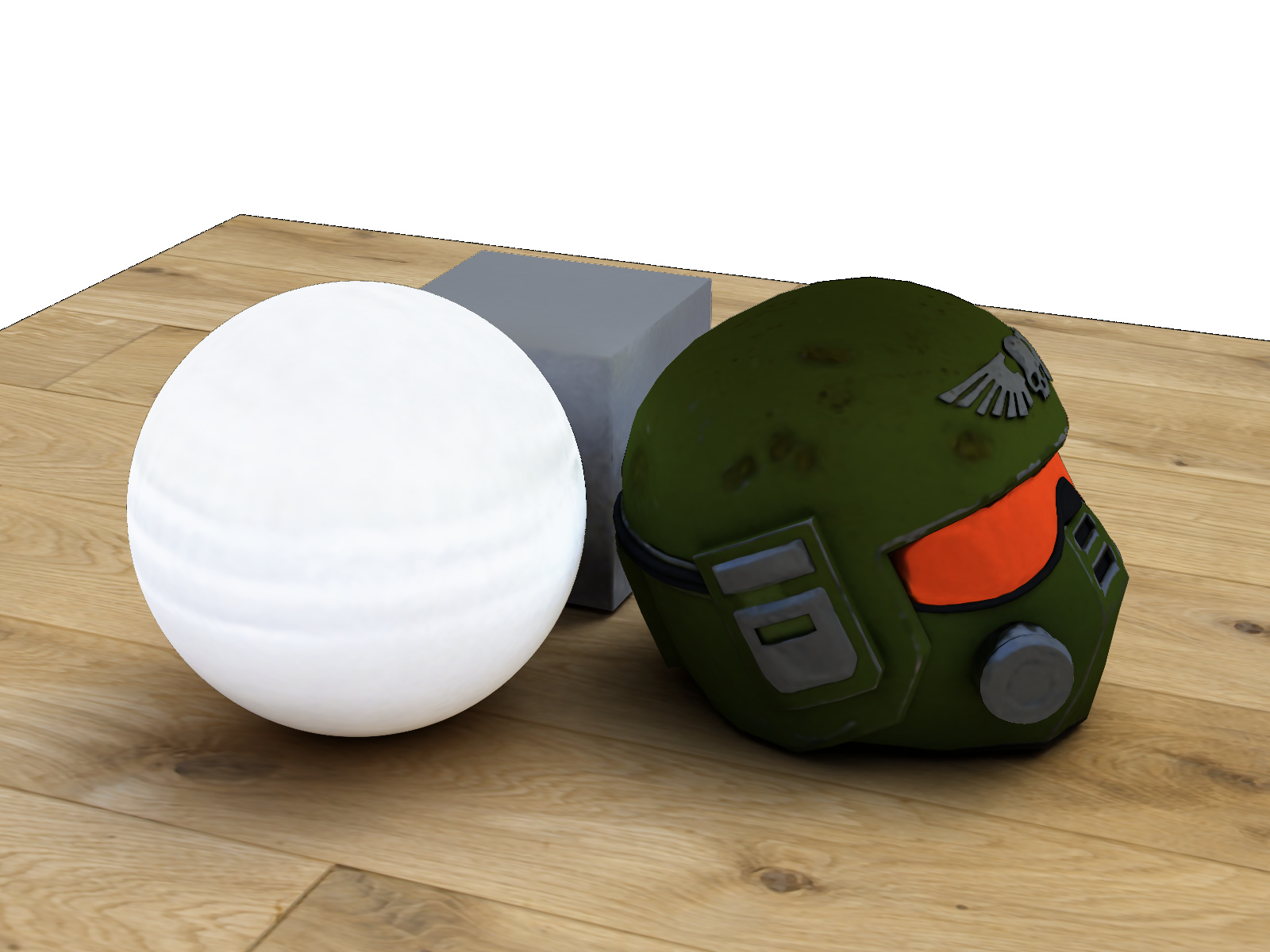} \\
                        & \multirow{1}{*}[0.5in]{\rotatebox[origin=c]{90}{Ground Truth}}
                        &
                        & \includegraphics[width=0.2\linewidth]{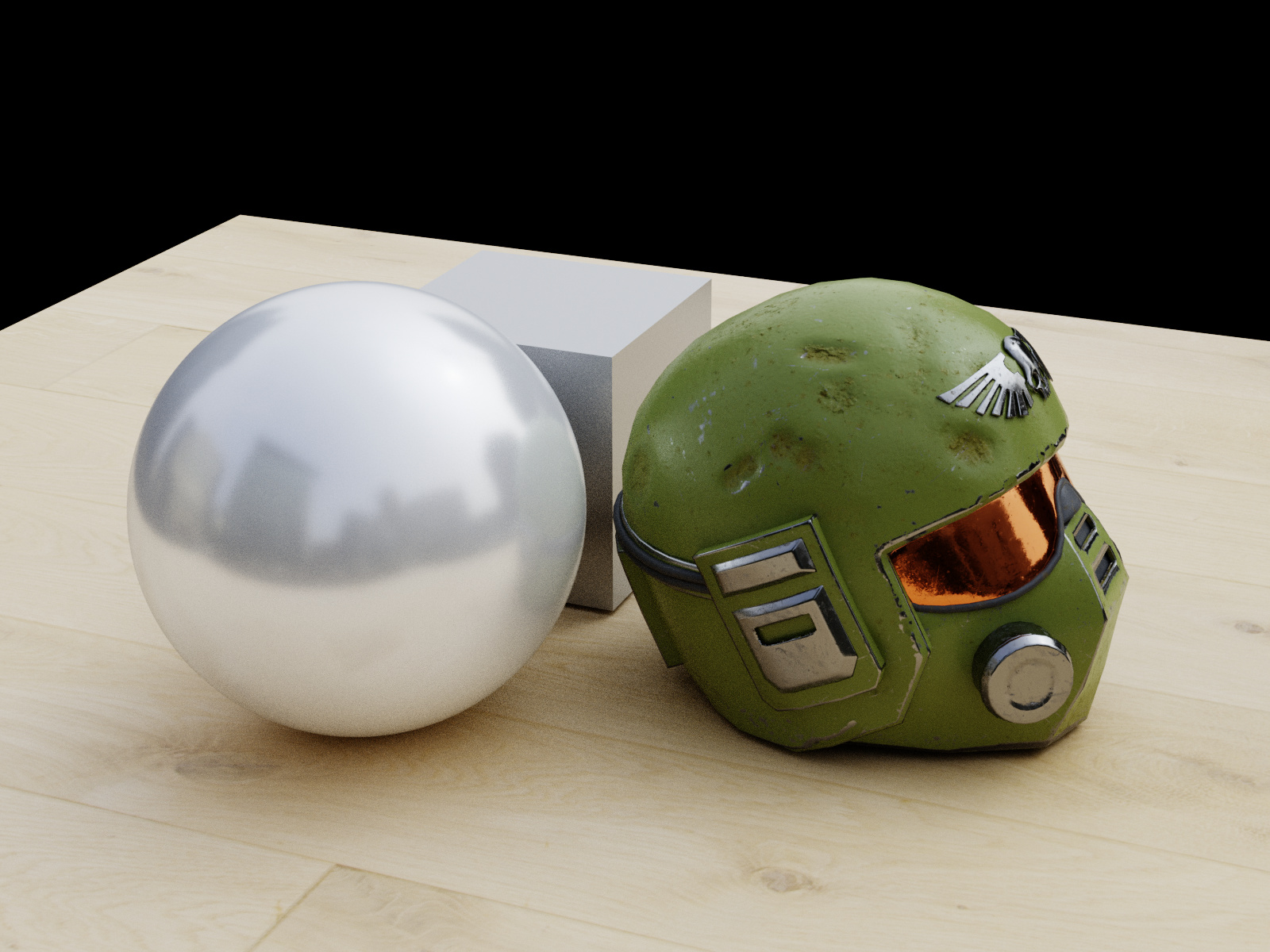}
                        & \includegraphics[width=0.2\linewidth]{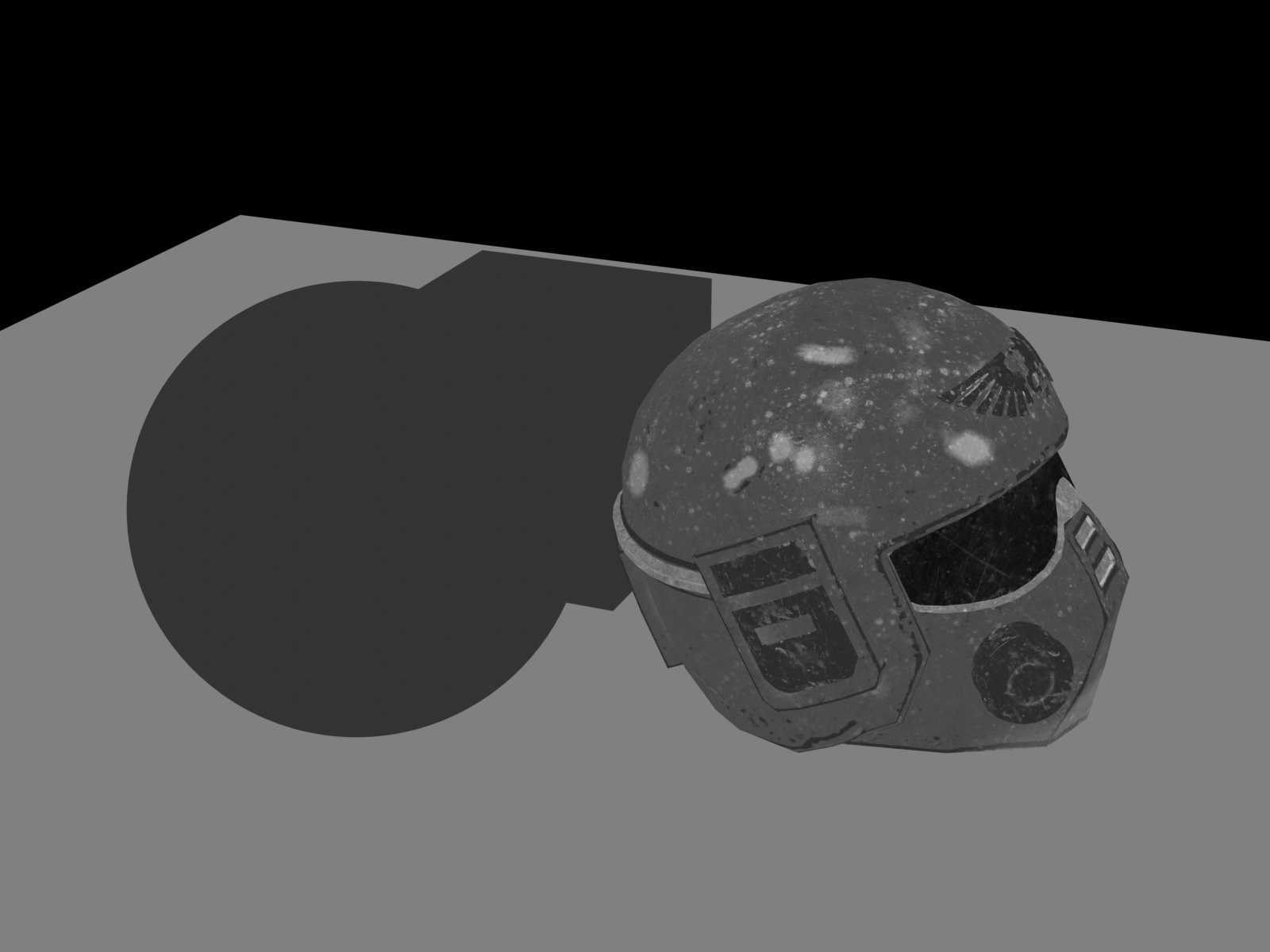}
                        & \includegraphics[width=0.2\linewidth]{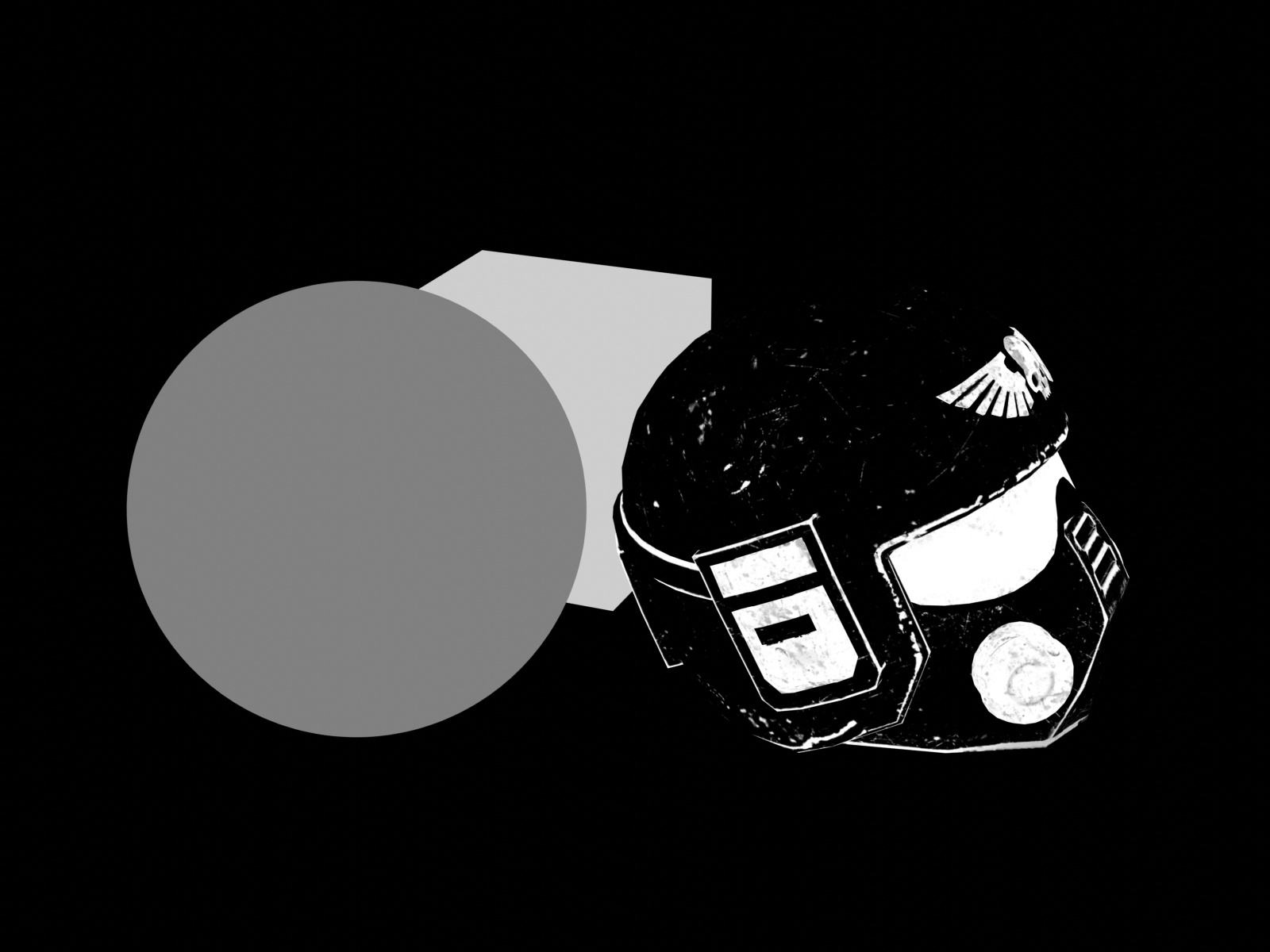}
                        & \includegraphics[width=0.2\linewidth]{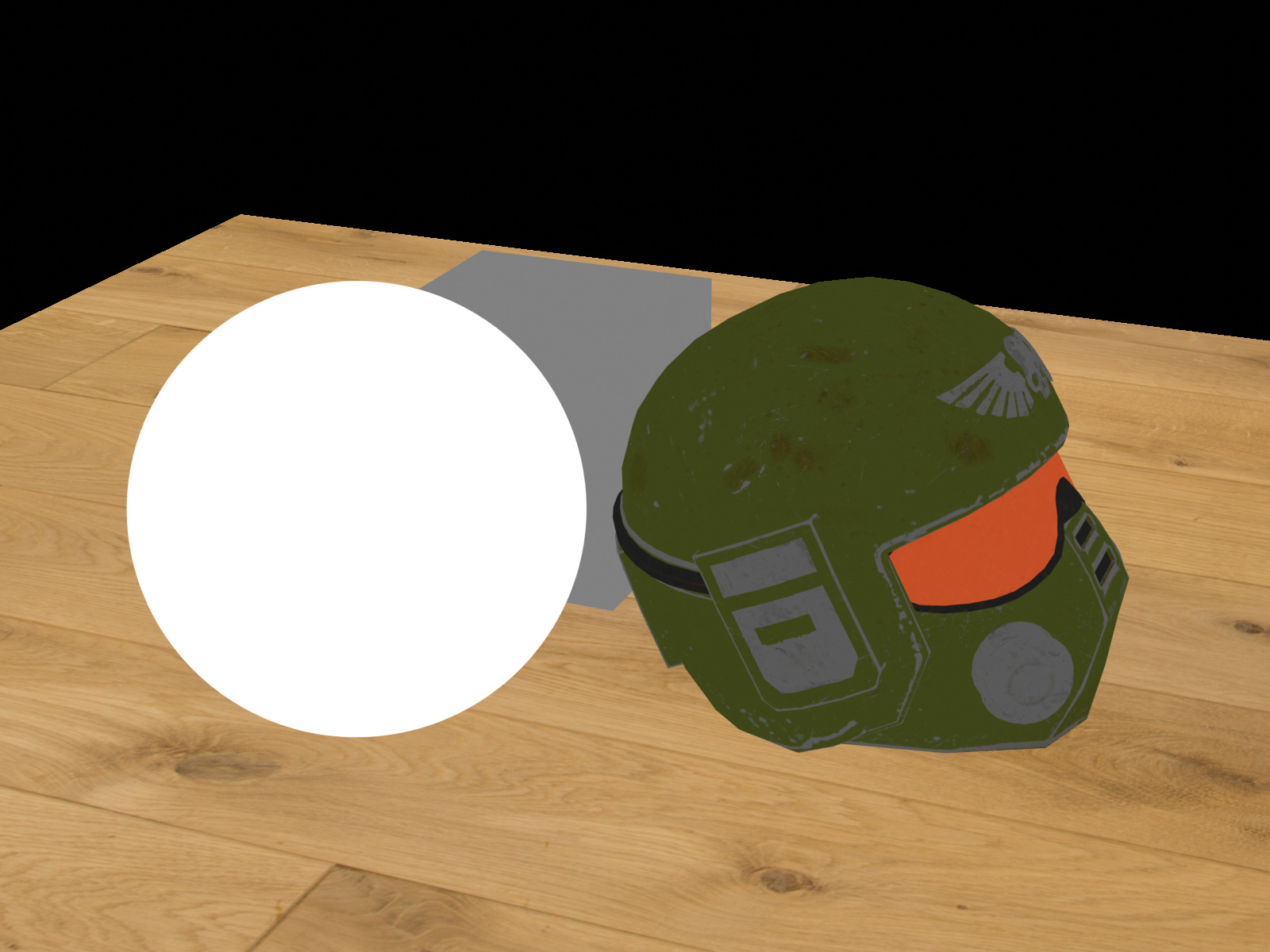} \\
            \hline & \\[-1.0em]
            \multirow{3}{*}[0.5in]{\raisebox{-1.2in}{\rotatebox[origin=c]{90}{Studio}}}
                        & \multirow{1}{*}[0.5in]{\rotatebox[origin=c]{90}{NeILF++}}
                        & \includegraphics[width=0.2\linewidth]{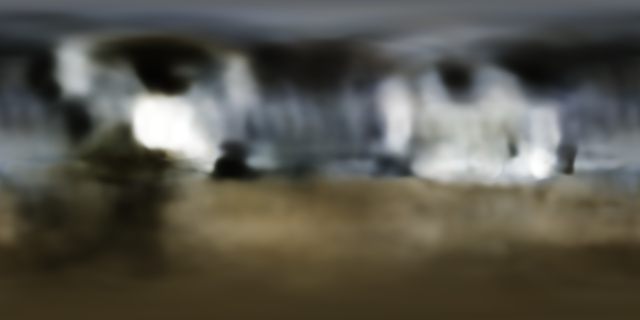}
                        & \includegraphics[width=0.2\linewidth]{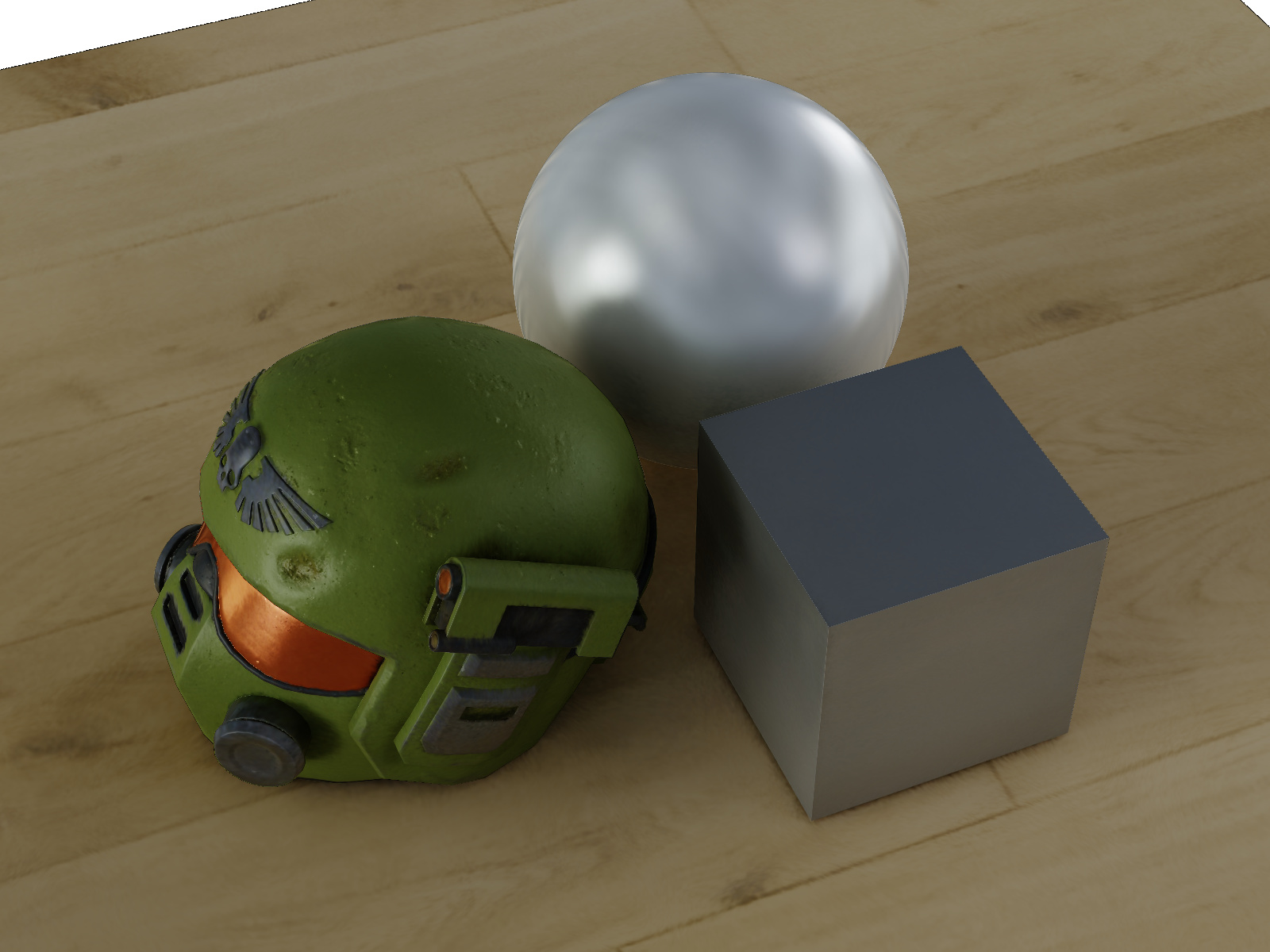}
                        & \includegraphics[width=0.2\linewidth]{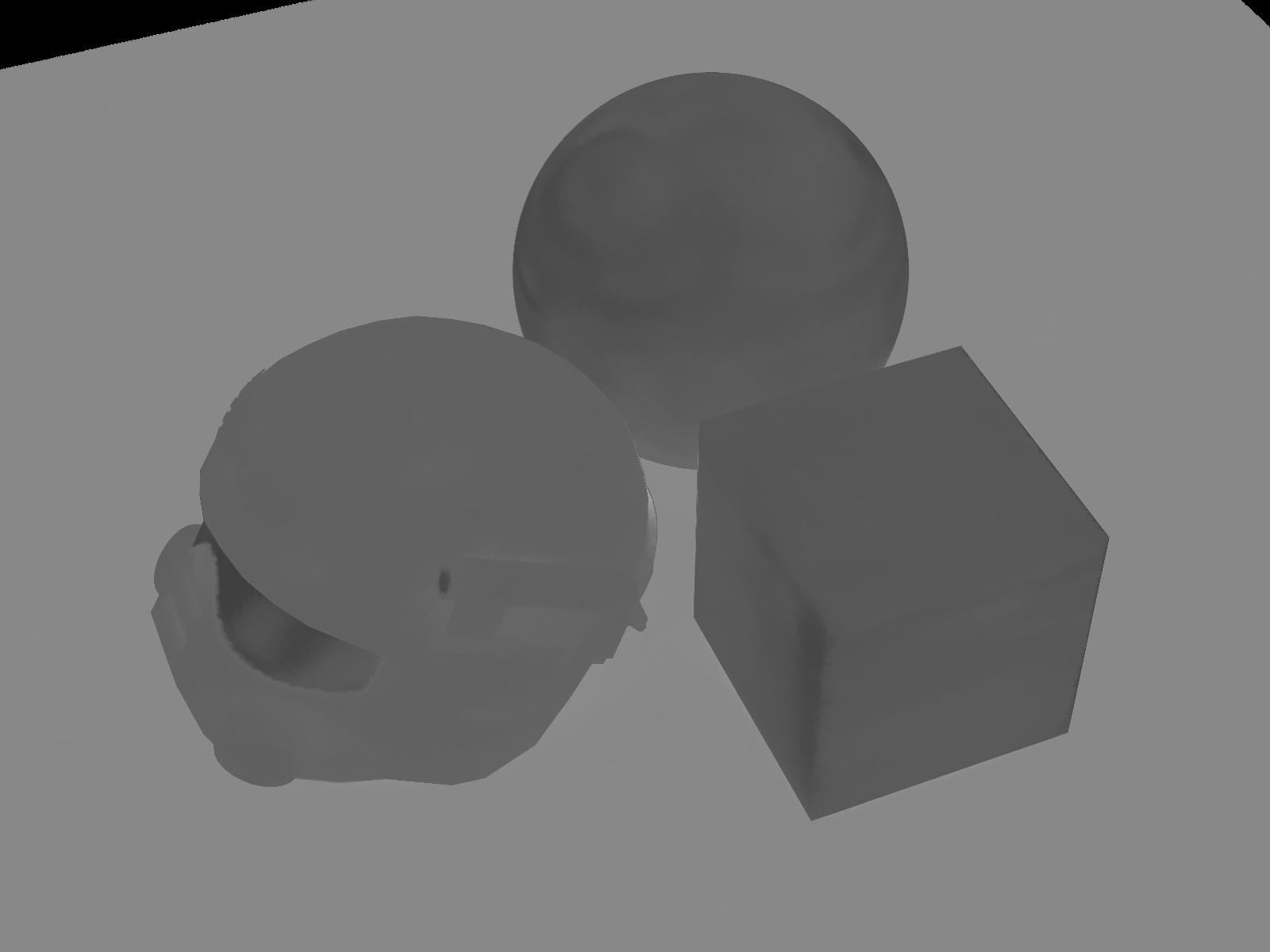}
                        & \includegraphics[width=0.2\linewidth]{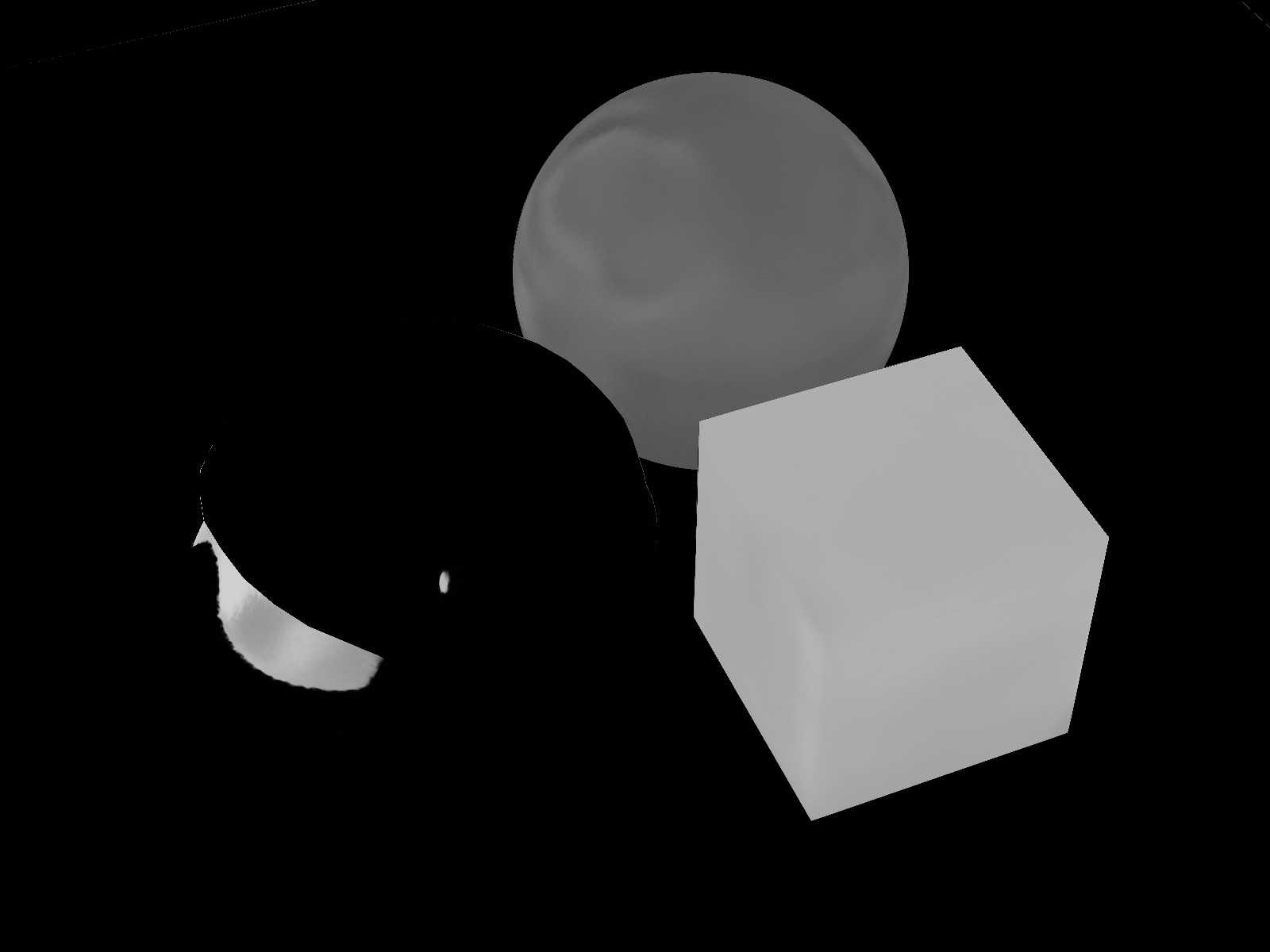}
                        & \includegraphics[width=0.2\linewidth]{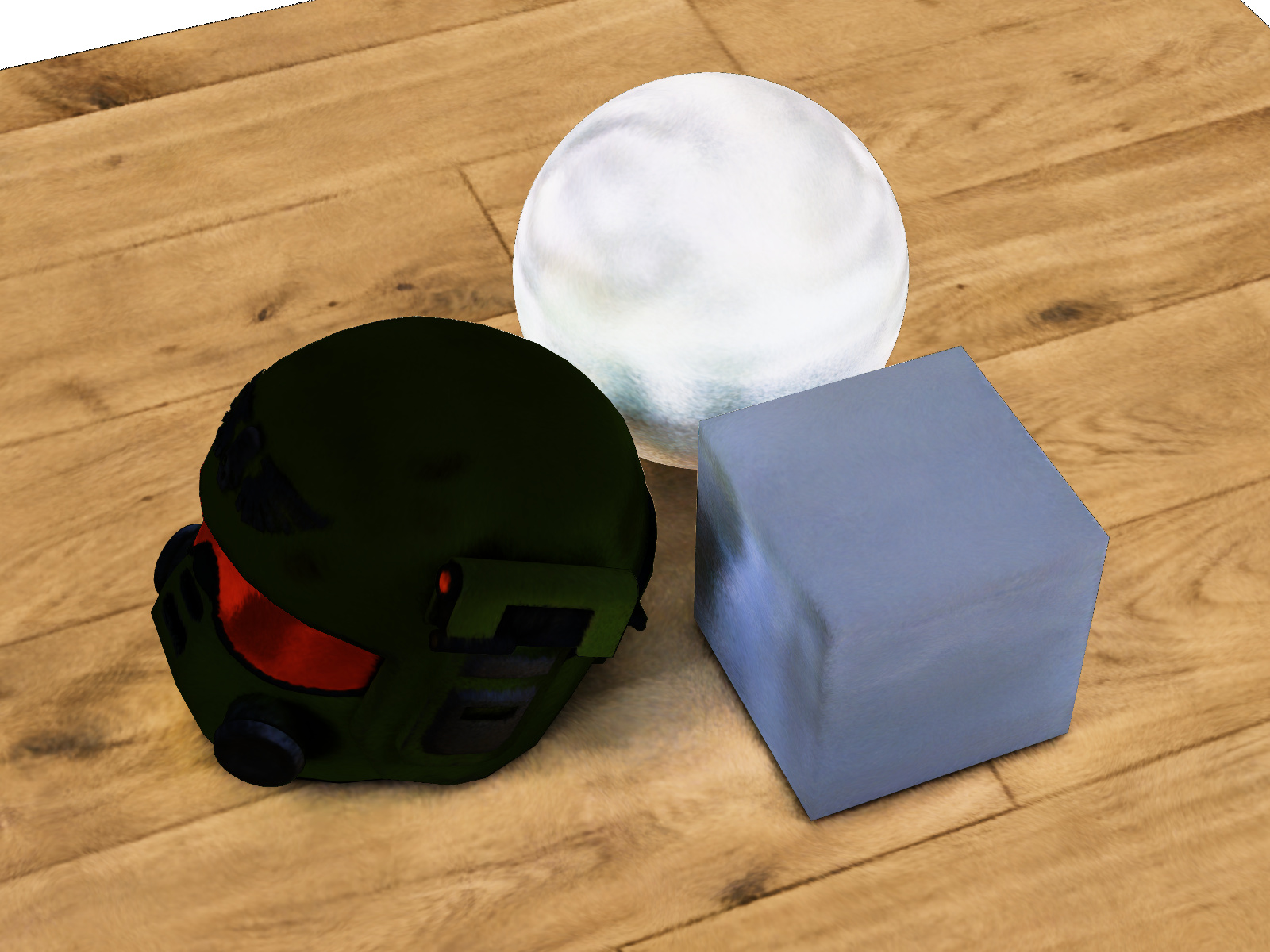} \\
                        & \multirow{1}{*}[0.5in]{\rotatebox[origin=c]{90}{Ours}}
                        & \includegraphics[width=0.2\linewidth]{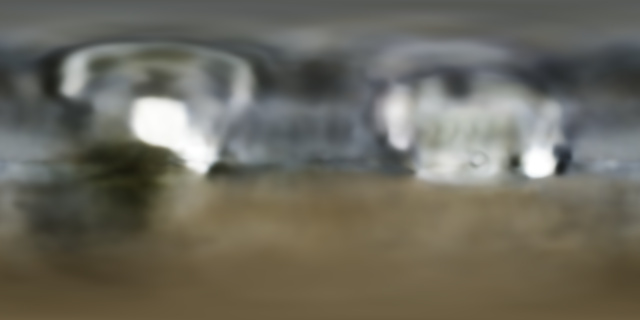}
                        & \includegraphics[width=0.2\linewidth]{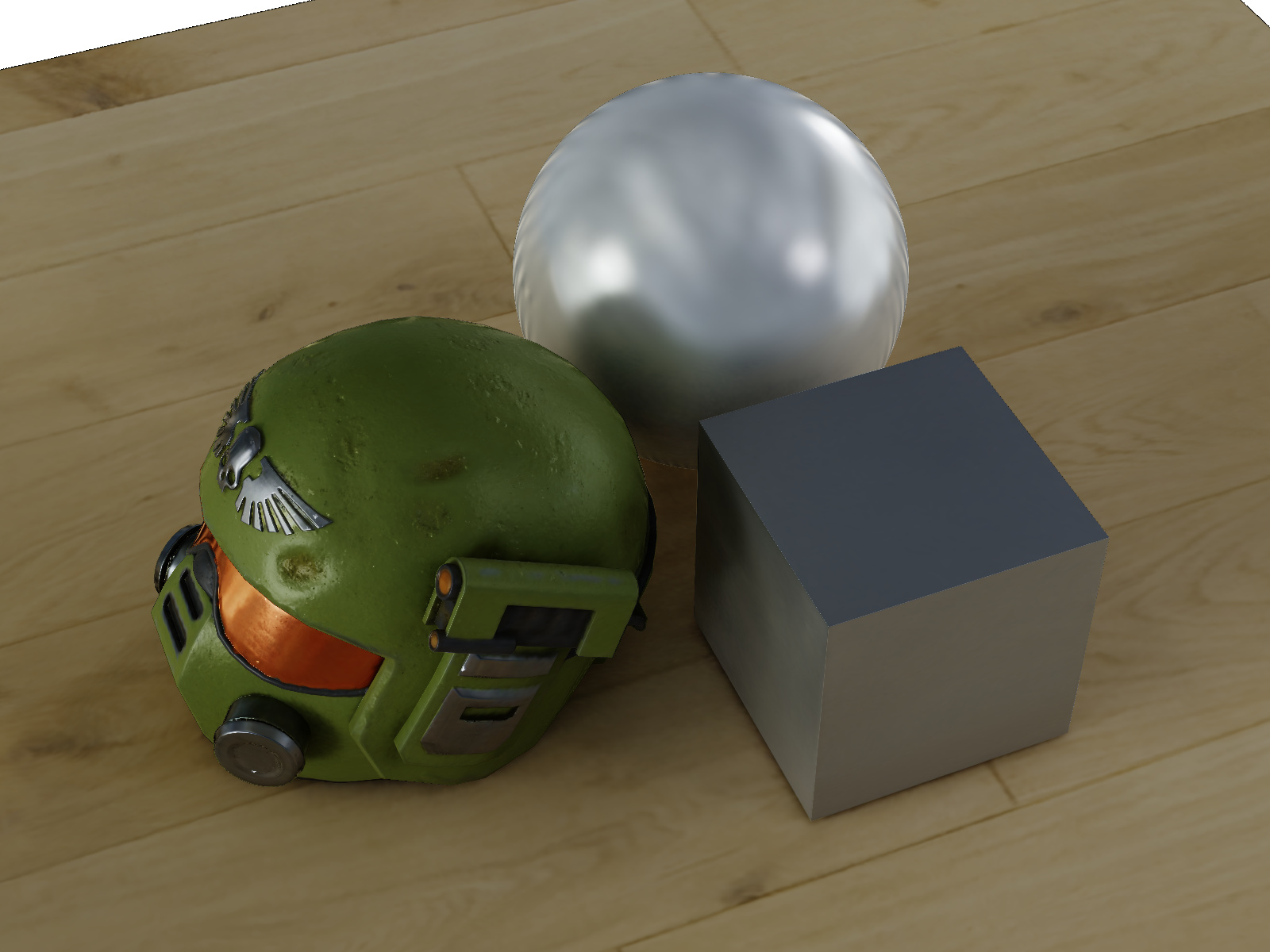}
                        & \includegraphics[width=0.2\linewidth]{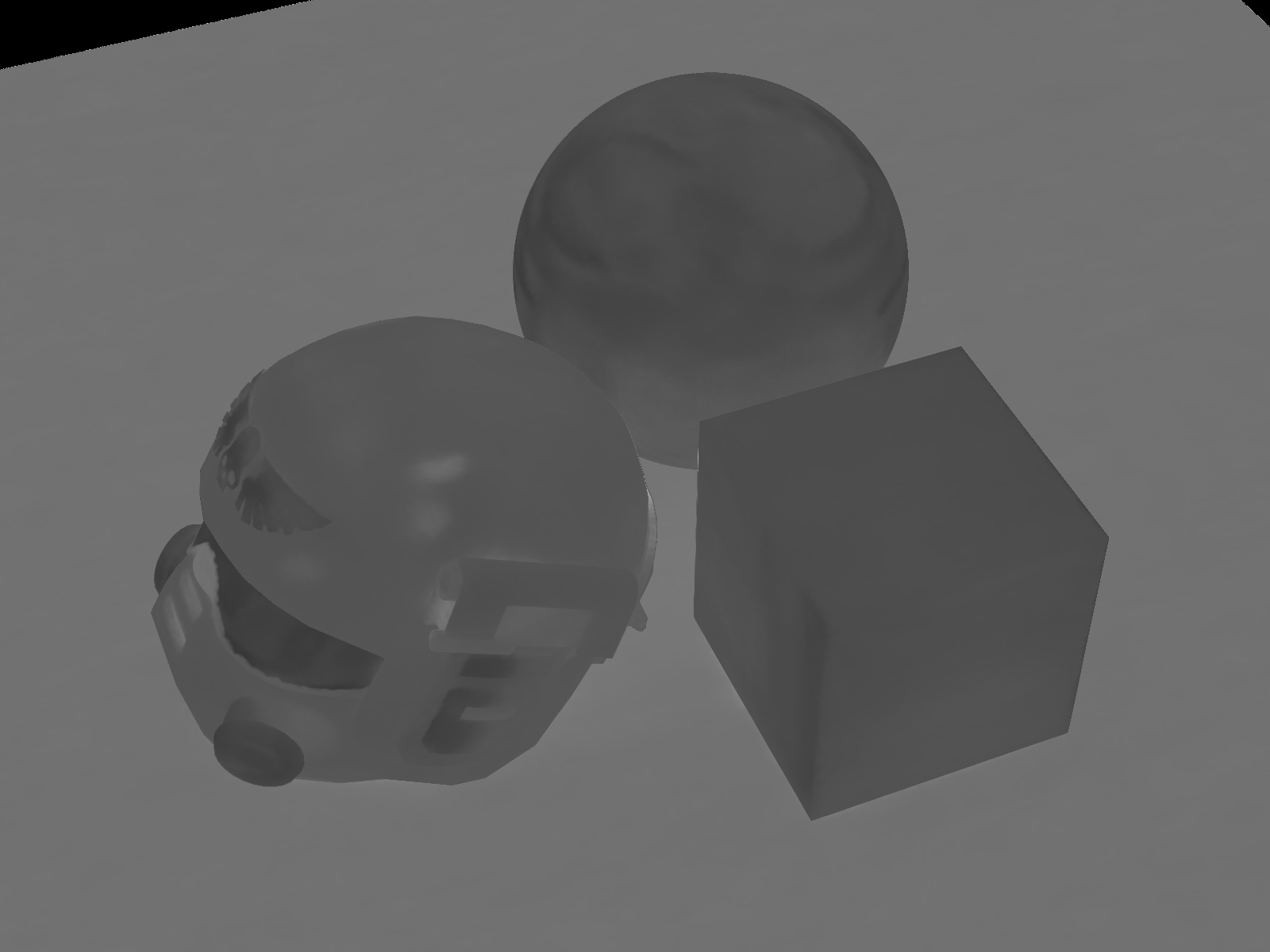}
                        & \includegraphics[width=0.2\linewidth]{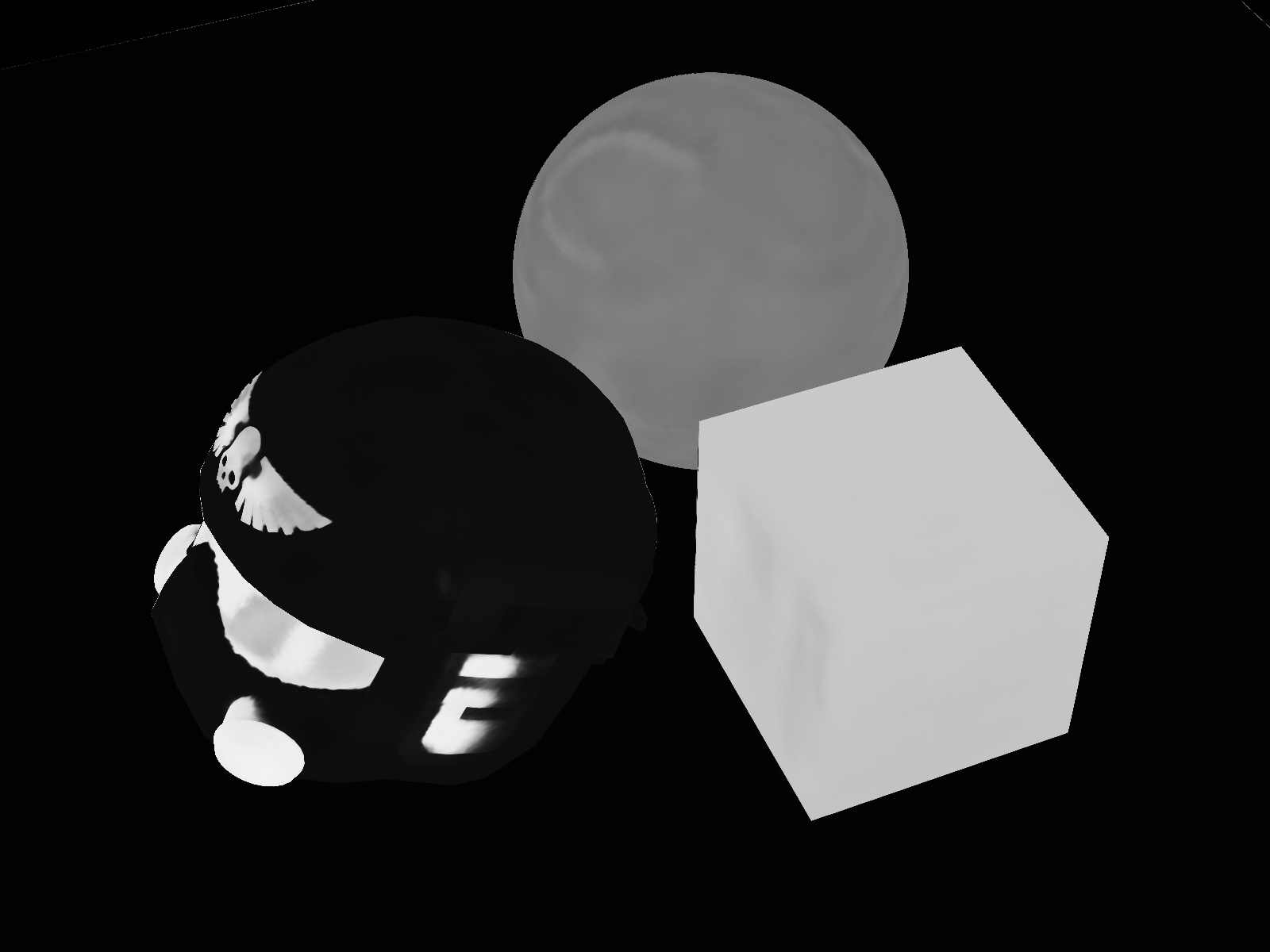}
                        & \includegraphics[width=0.2\linewidth]{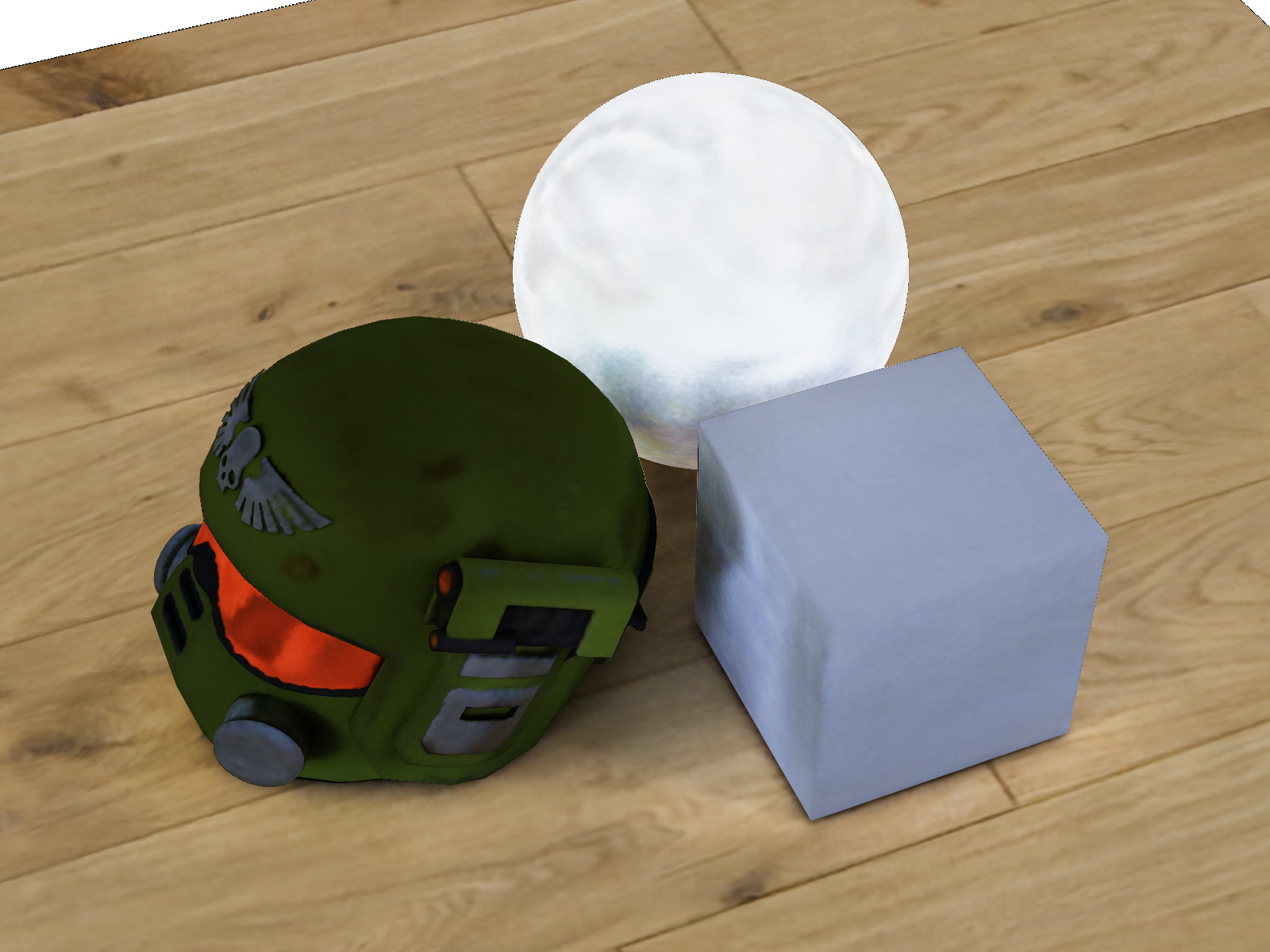} \\
                        & \multirow{1}{*}[0.5in]{\rotatebox[origin=c]{90}{Ground Truth}}
                        &
                        & \includegraphics[width=0.2\linewidth]{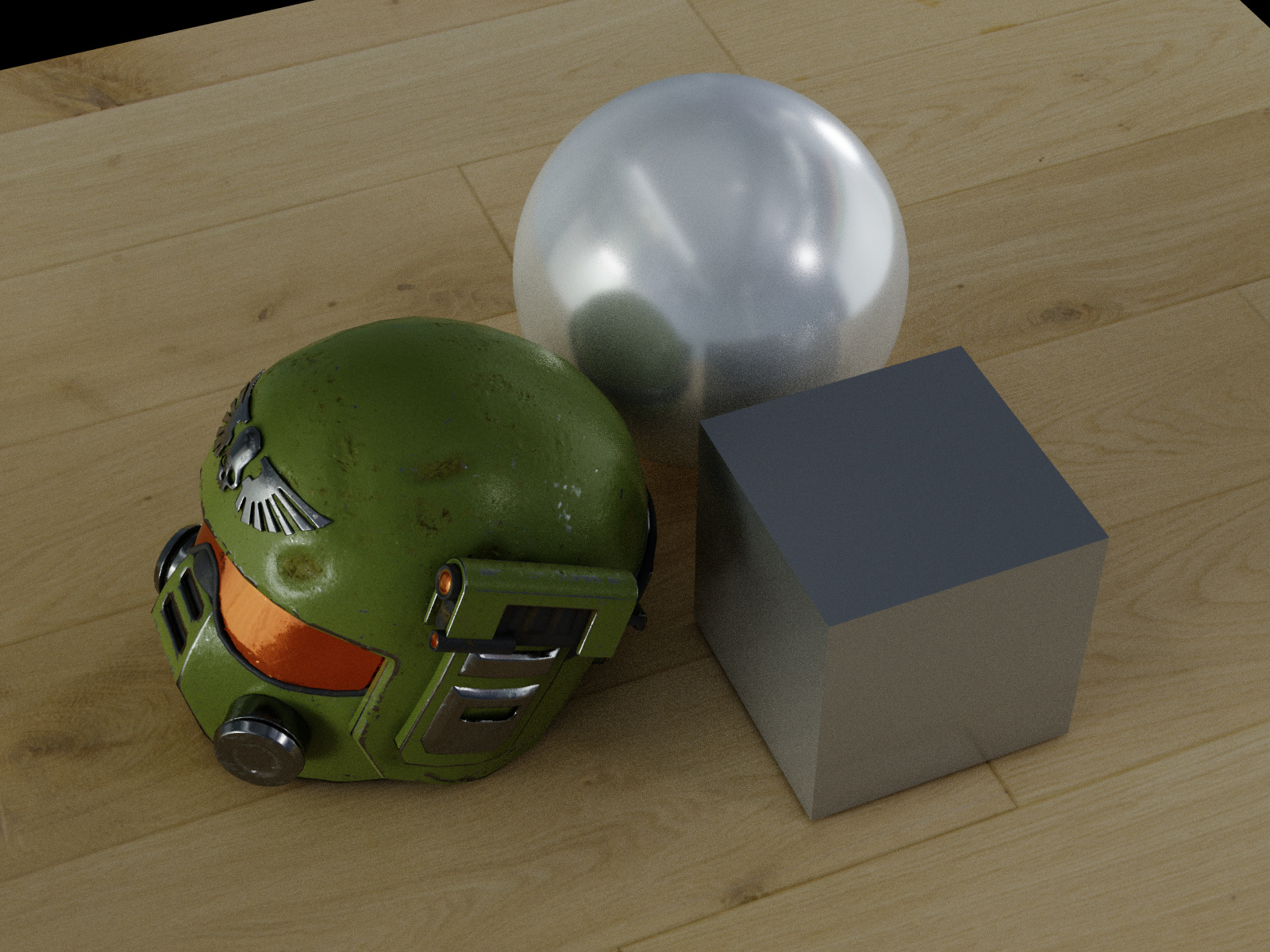}
                        & \includegraphics[width=0.2\linewidth]{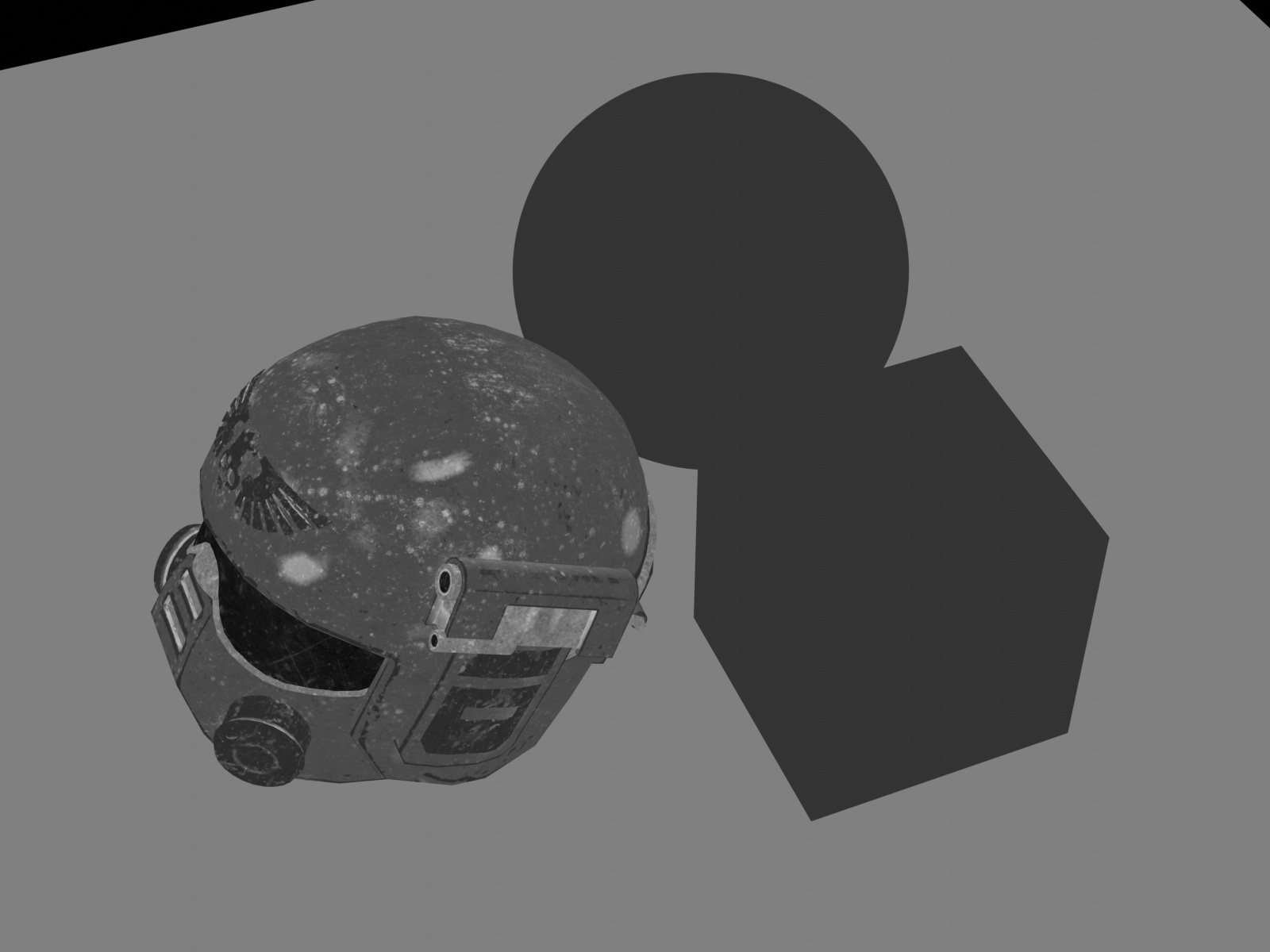}
                        & \includegraphics[width=0.2\linewidth]{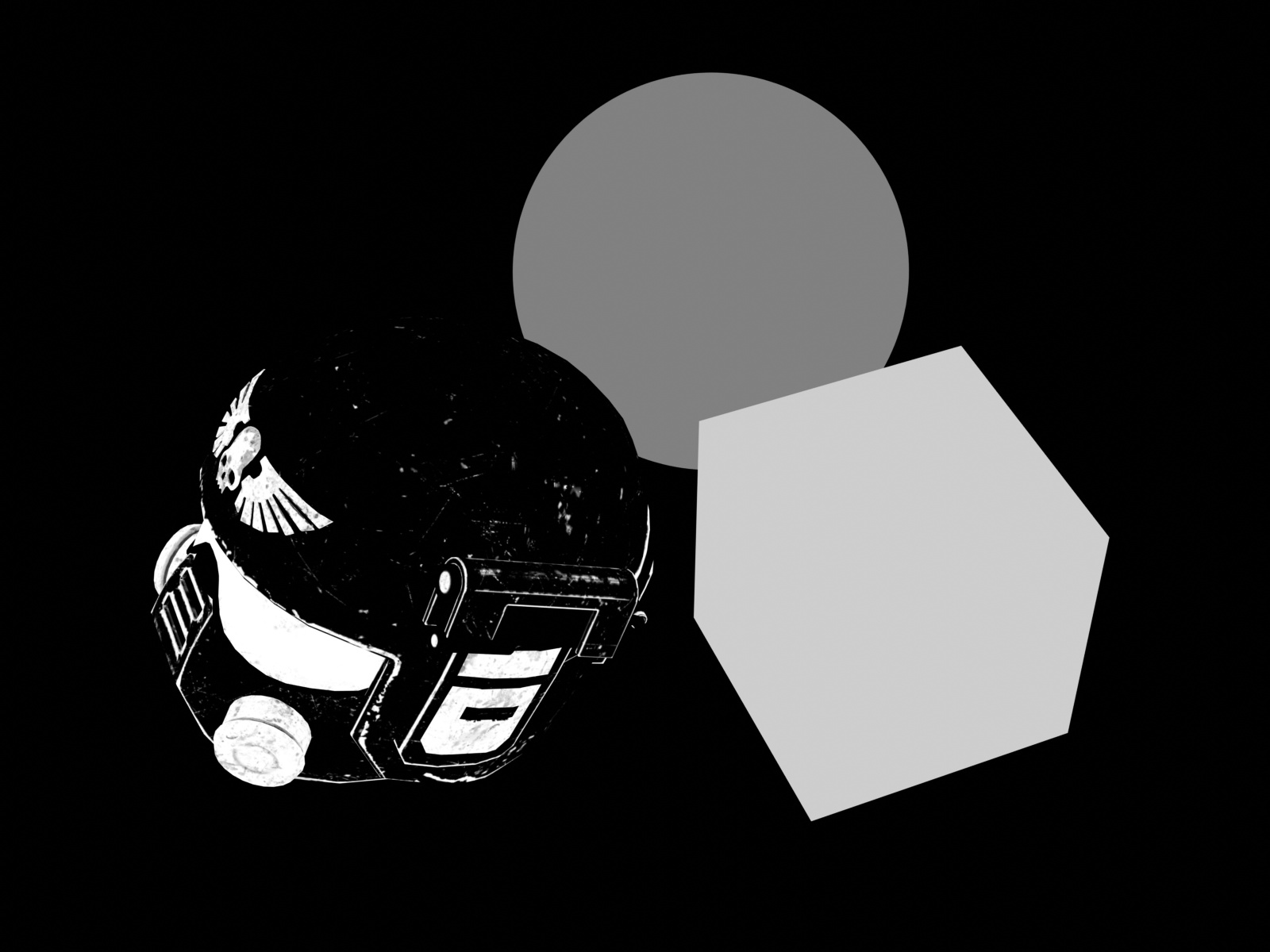}
                        & \includegraphics[width=0.2\linewidth]{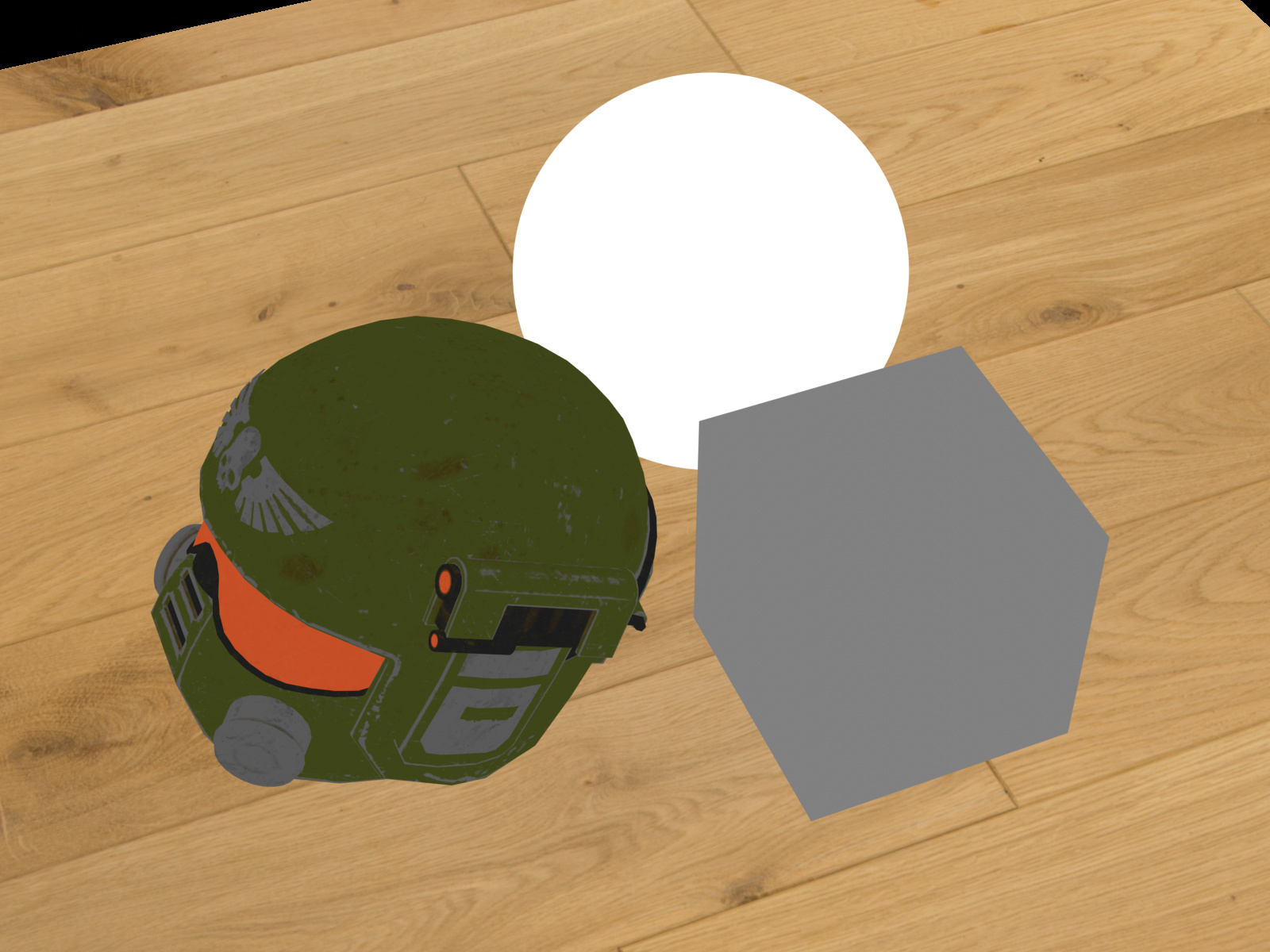} \\
            \hline & \\[-1.0em]
            \multirow{3}{*}[0.5in]{\raisebox{-1.2in}{\rotatebox[origin=c]{90}{Castel}}}
                        & \multirow{1}{*}[0.5in]{\rotatebox[origin=c]{90}{NeILF++}}
                        & \includegraphics[width=0.2\linewidth]{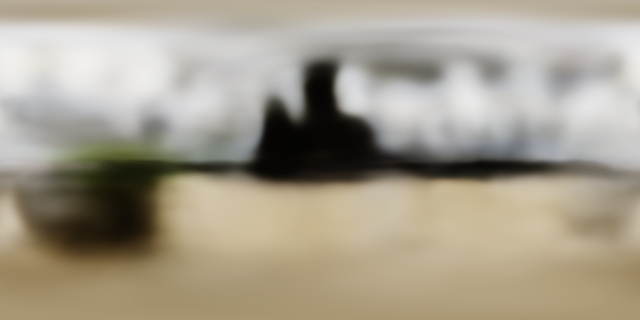}
                        & \includegraphics[width=0.2\linewidth]{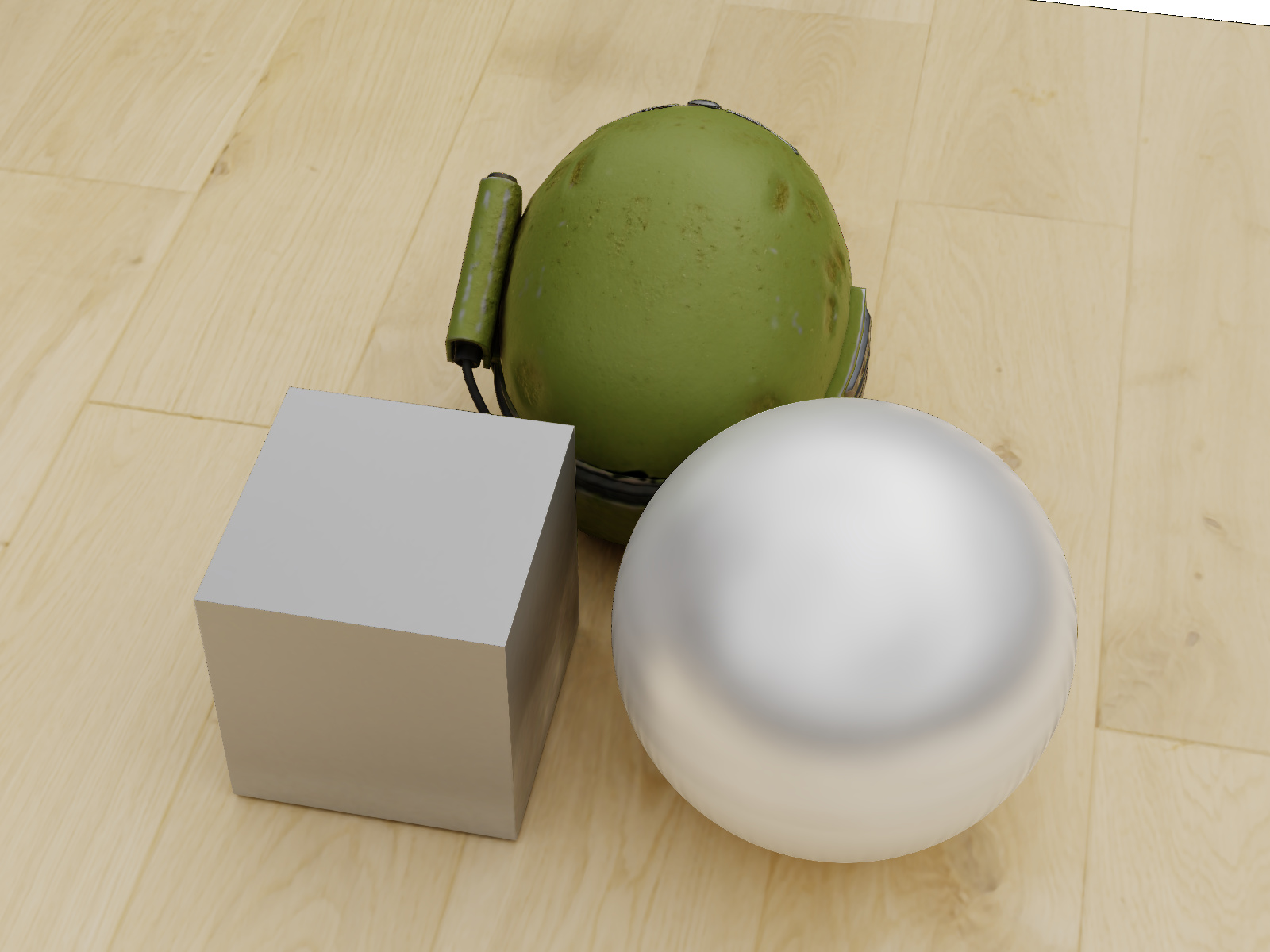}
                        & \includegraphics[width=0.2\linewidth]{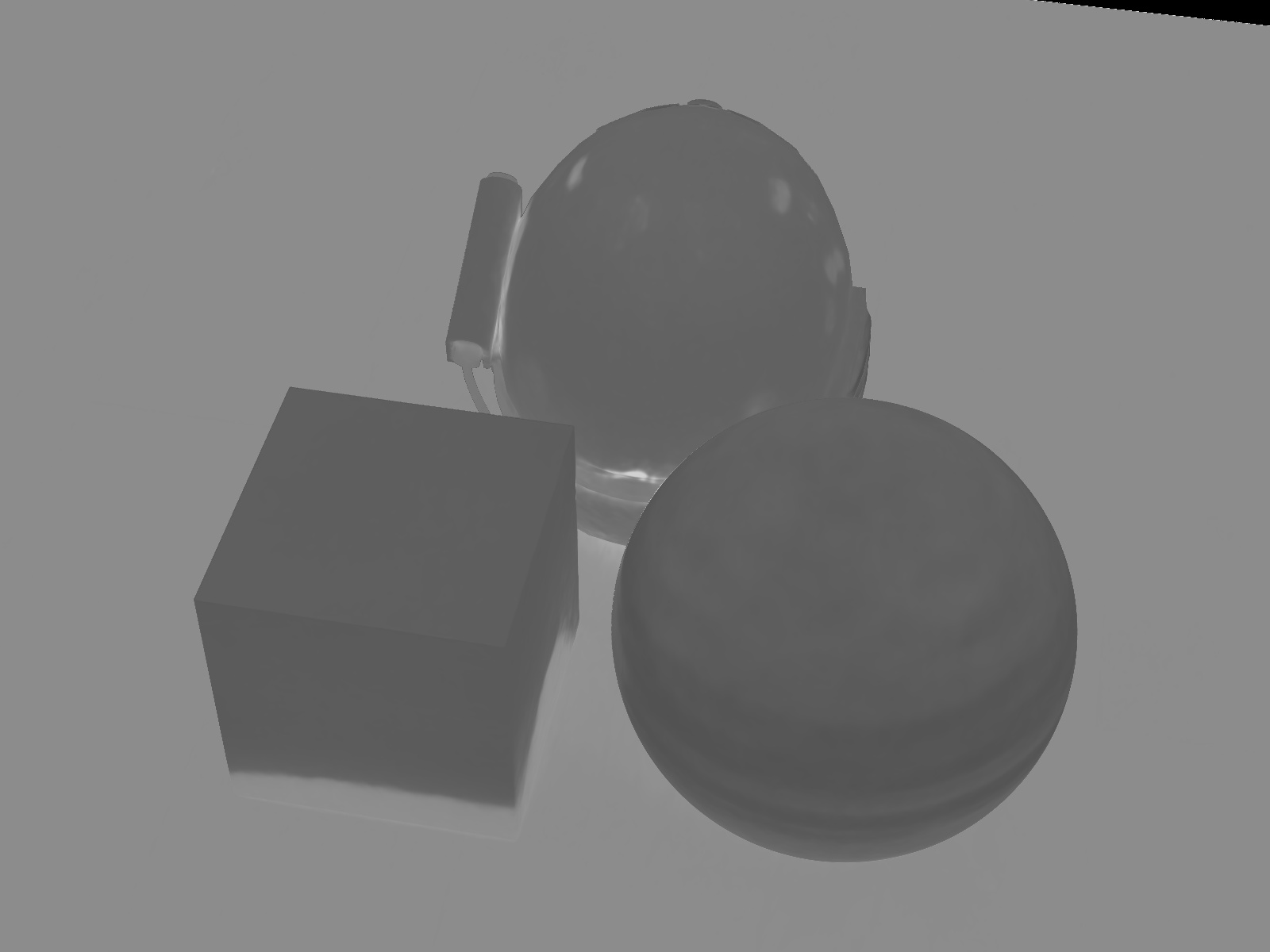}
                        & \includegraphics[width=0.2\linewidth]{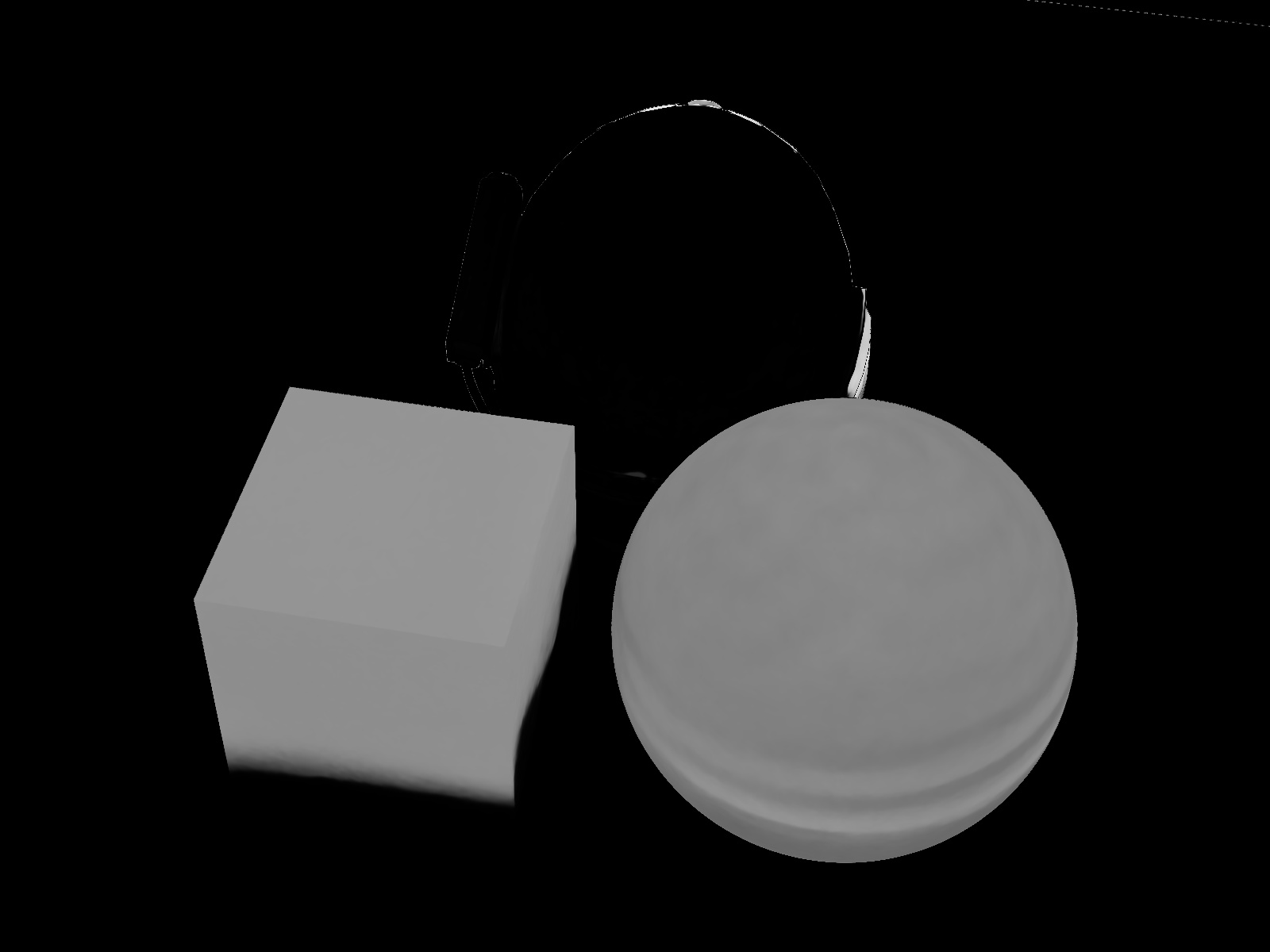}
                        & \includegraphics[width=0.2\linewidth]{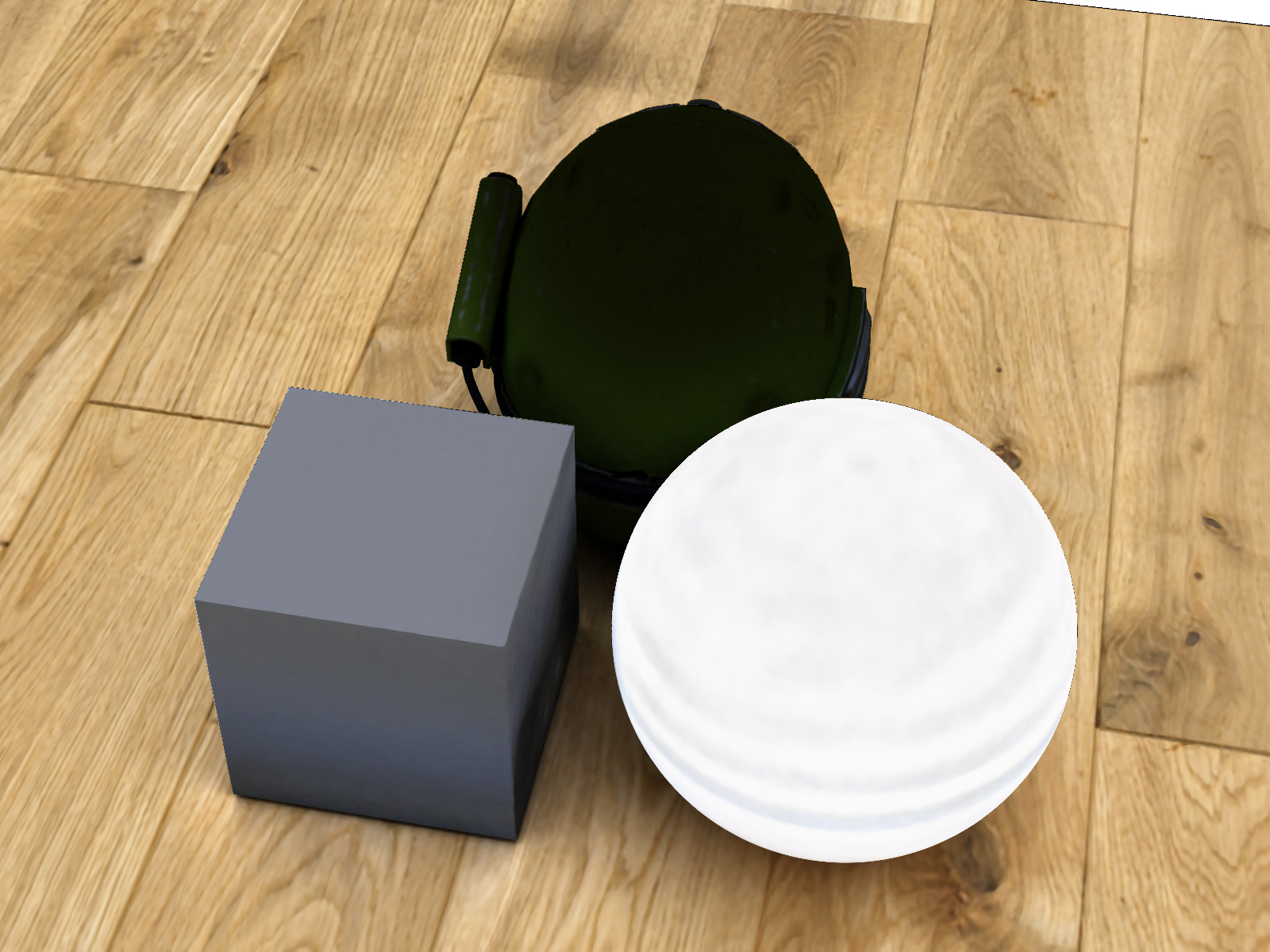} \\
                        & \multirow{1}{*}[0.5in]{\rotatebox[origin=c]{90}{Ours}}
                        & \includegraphics[width=0.2\linewidth]{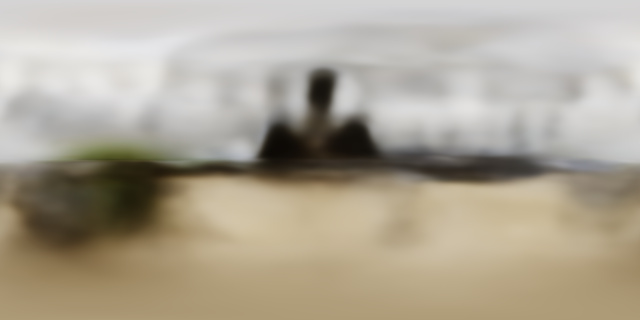}
                        & \includegraphics[width=0.2\linewidth]{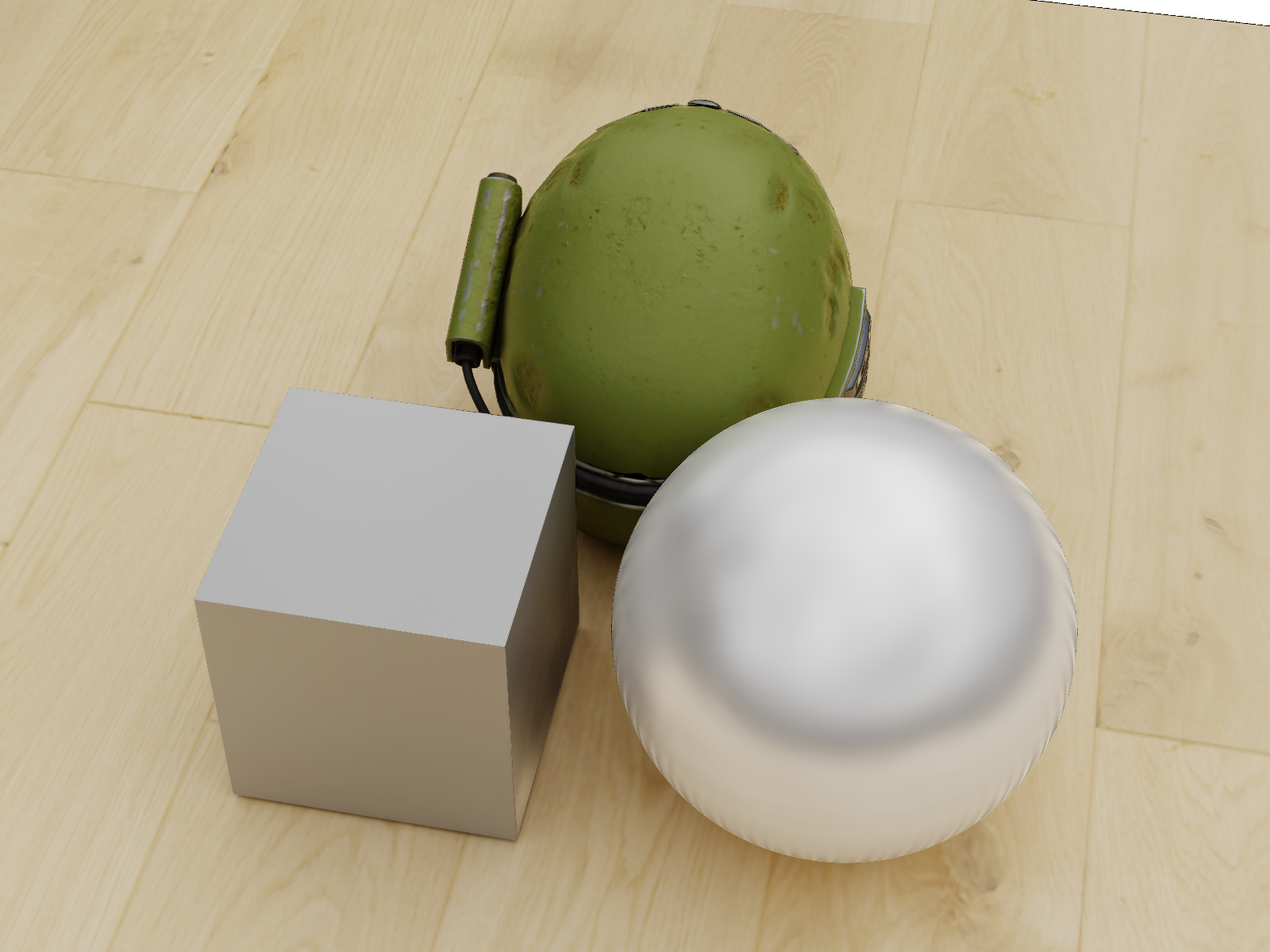}
                        & \includegraphics[width=0.2\linewidth]{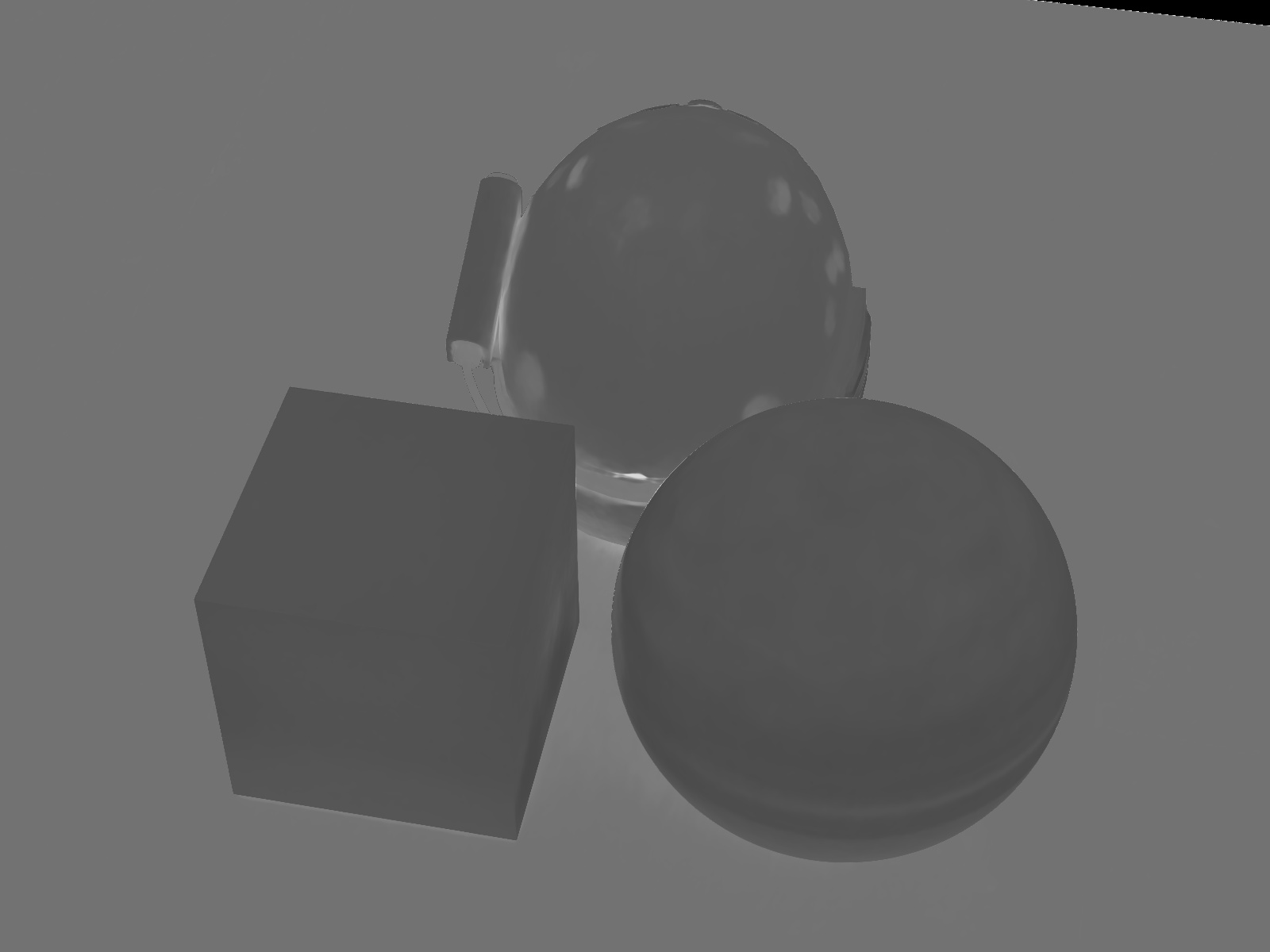}
                        & \includegraphics[width=0.2\linewidth]{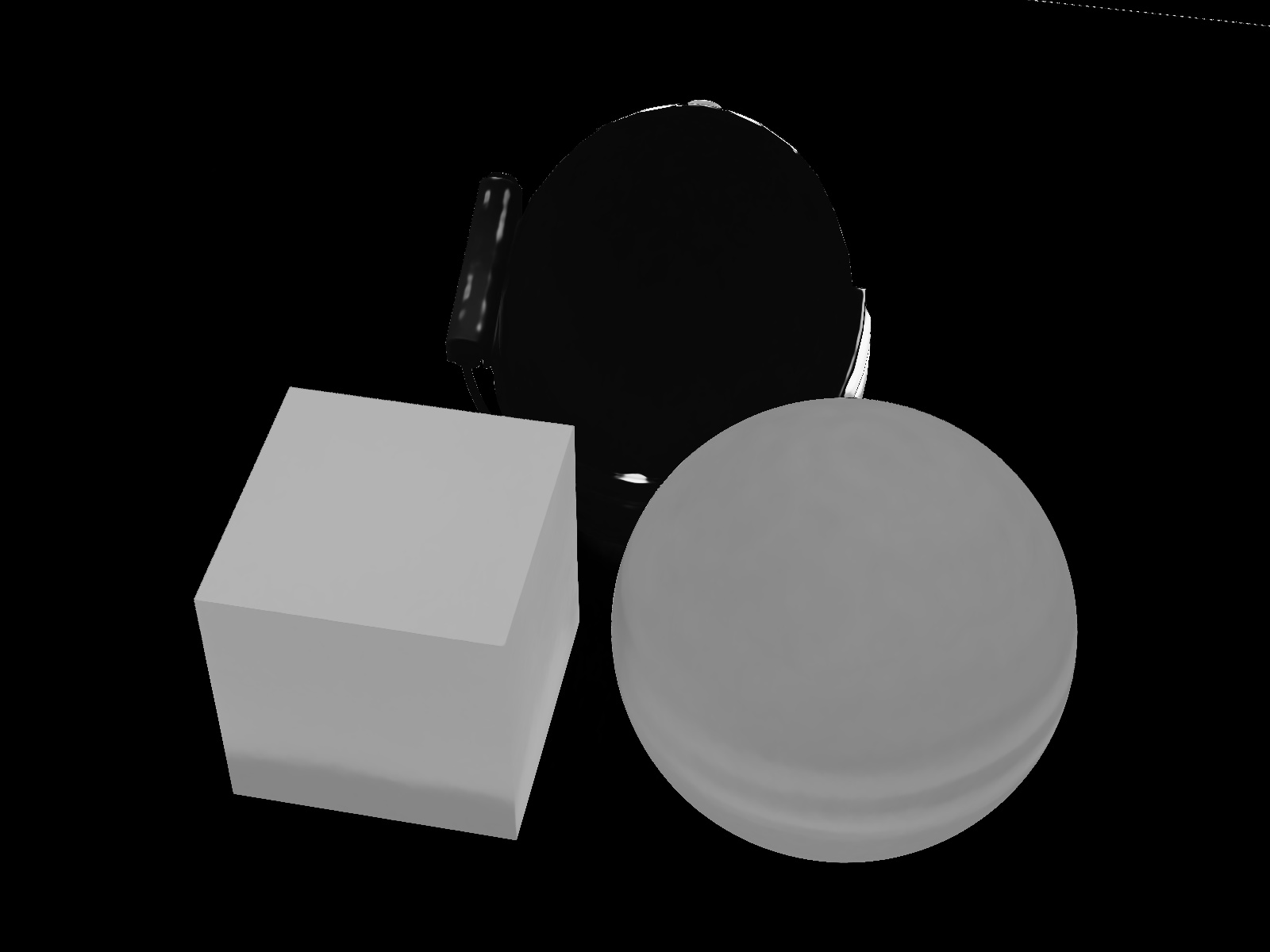}
                        & \includegraphics[width=0.2\linewidth]{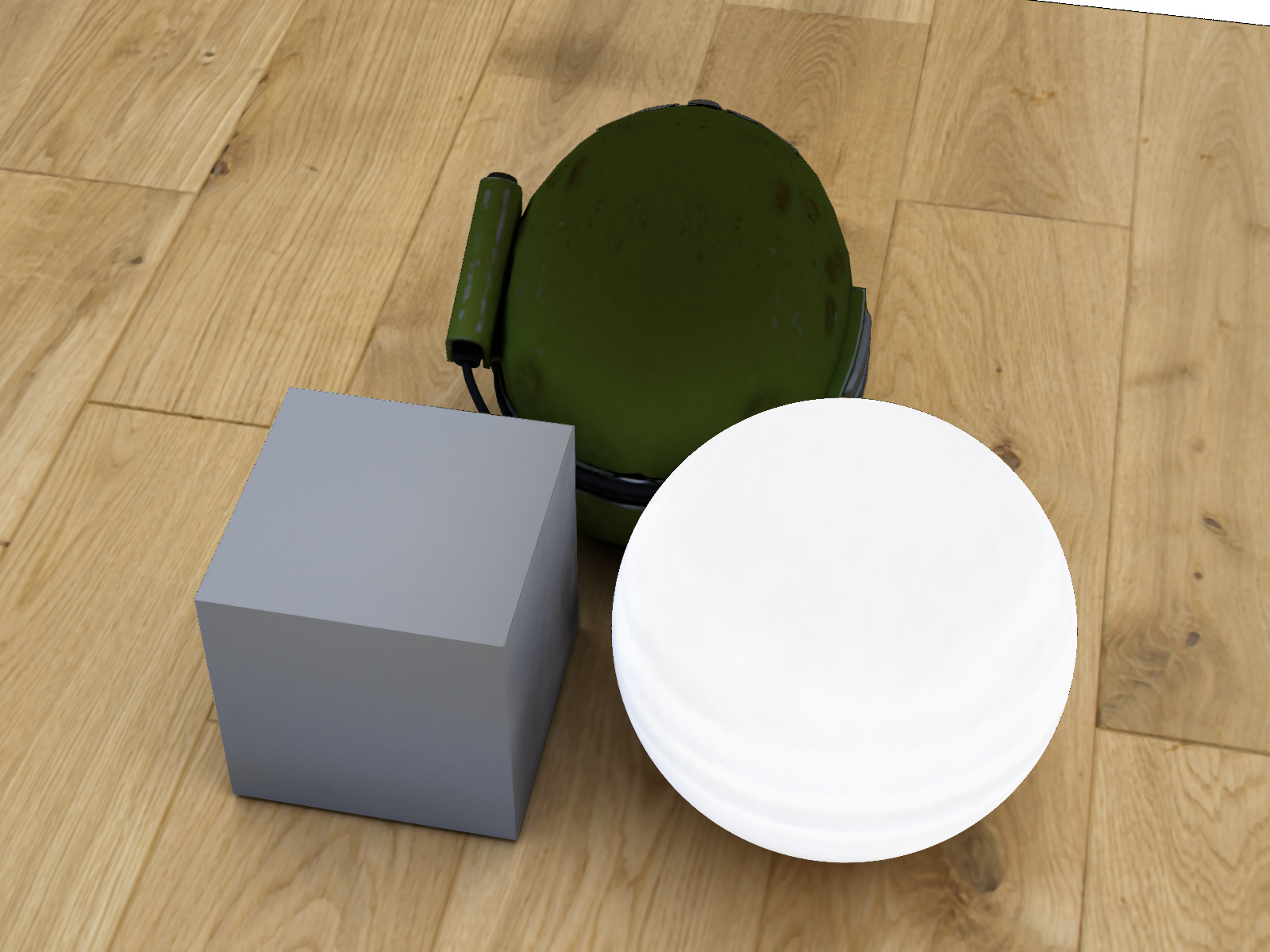} \\
                        & \multirow{1}{*}[0.5in]{\rotatebox[origin=c]{90}{Ground Truth}}
                        &
                        & \includegraphics[width=0.2\linewidth]{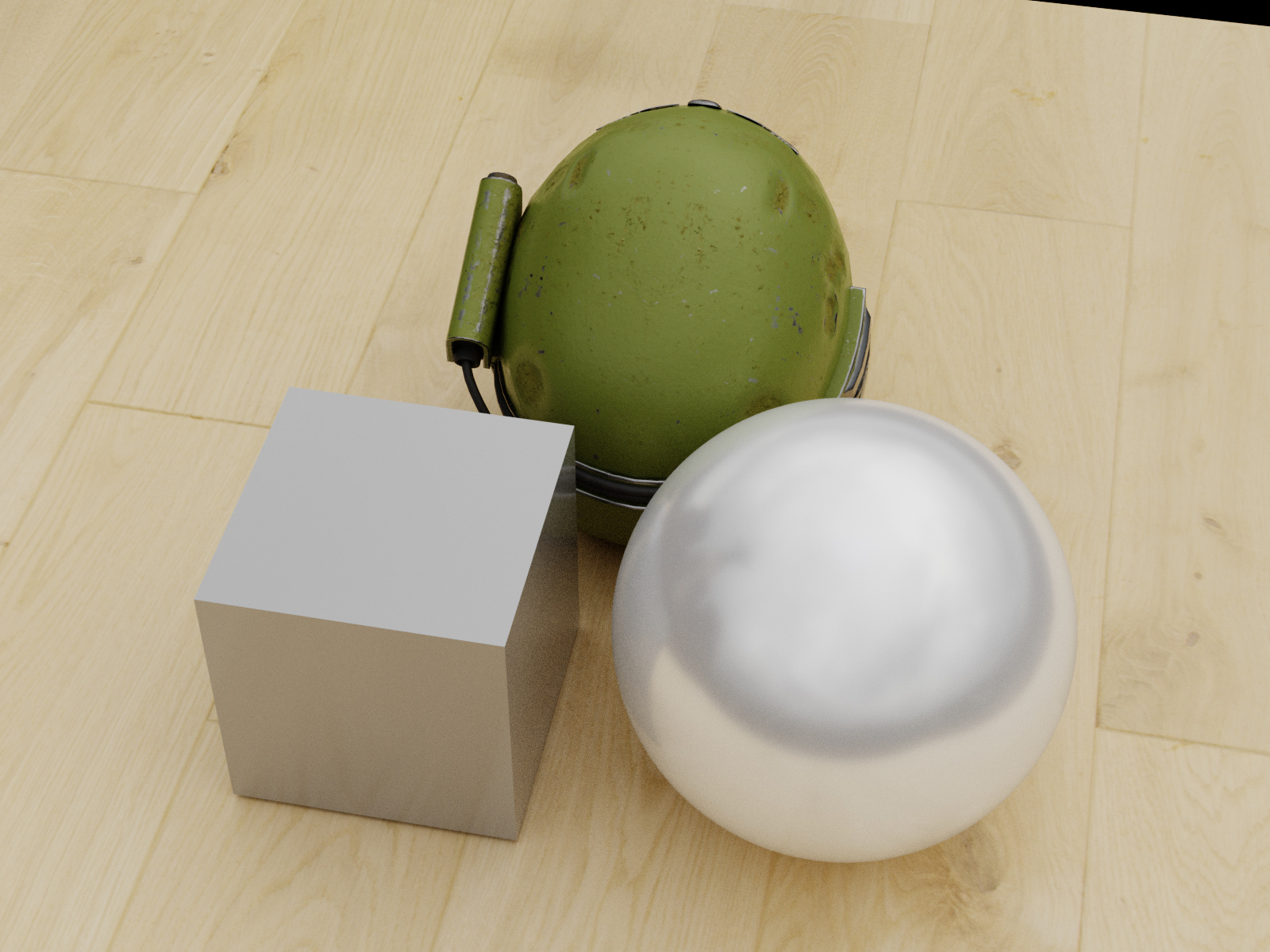}
                        & \includegraphics[width=0.2\linewidth]{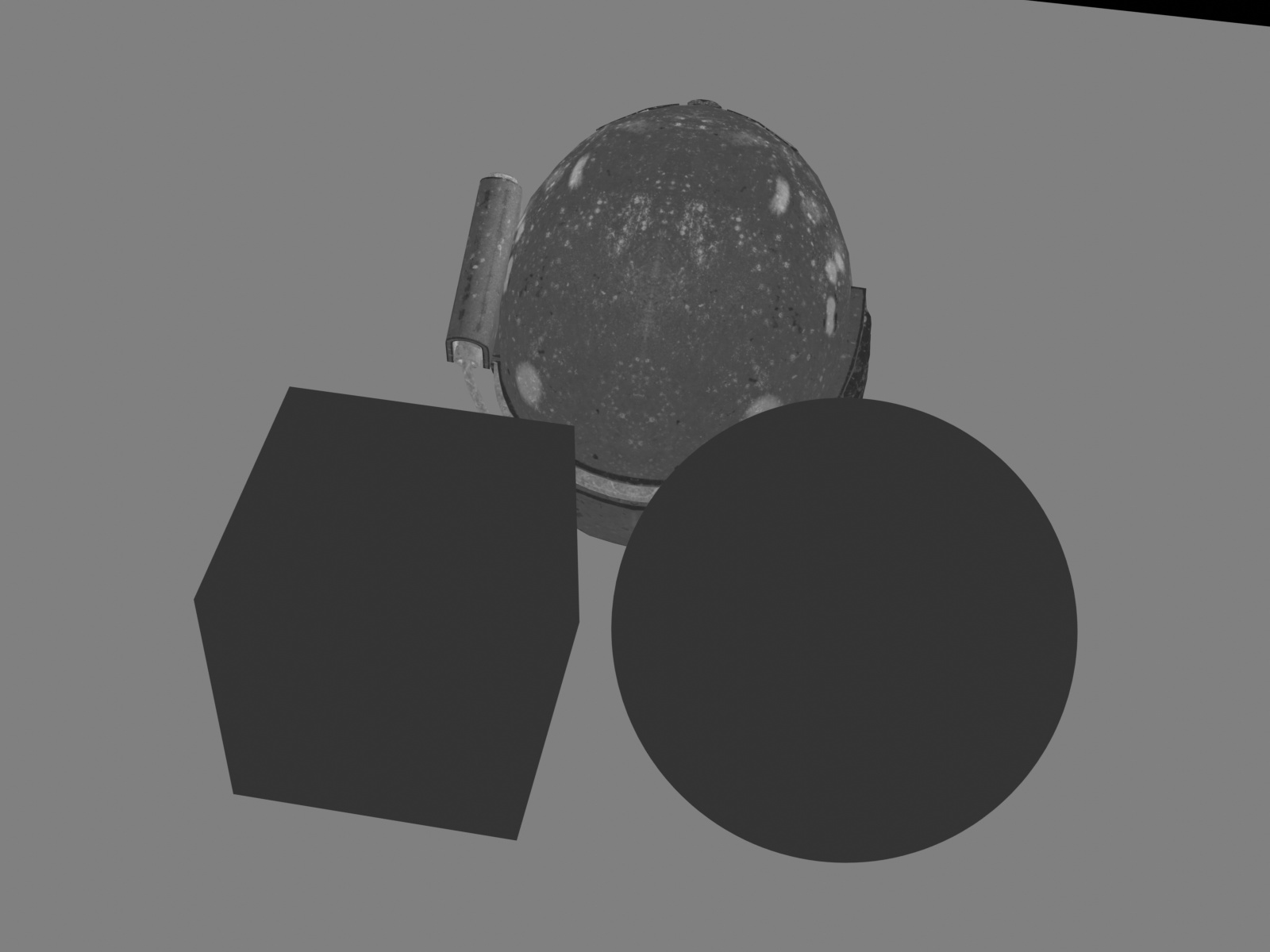}
                        & \includegraphics[width=0.2\linewidth]{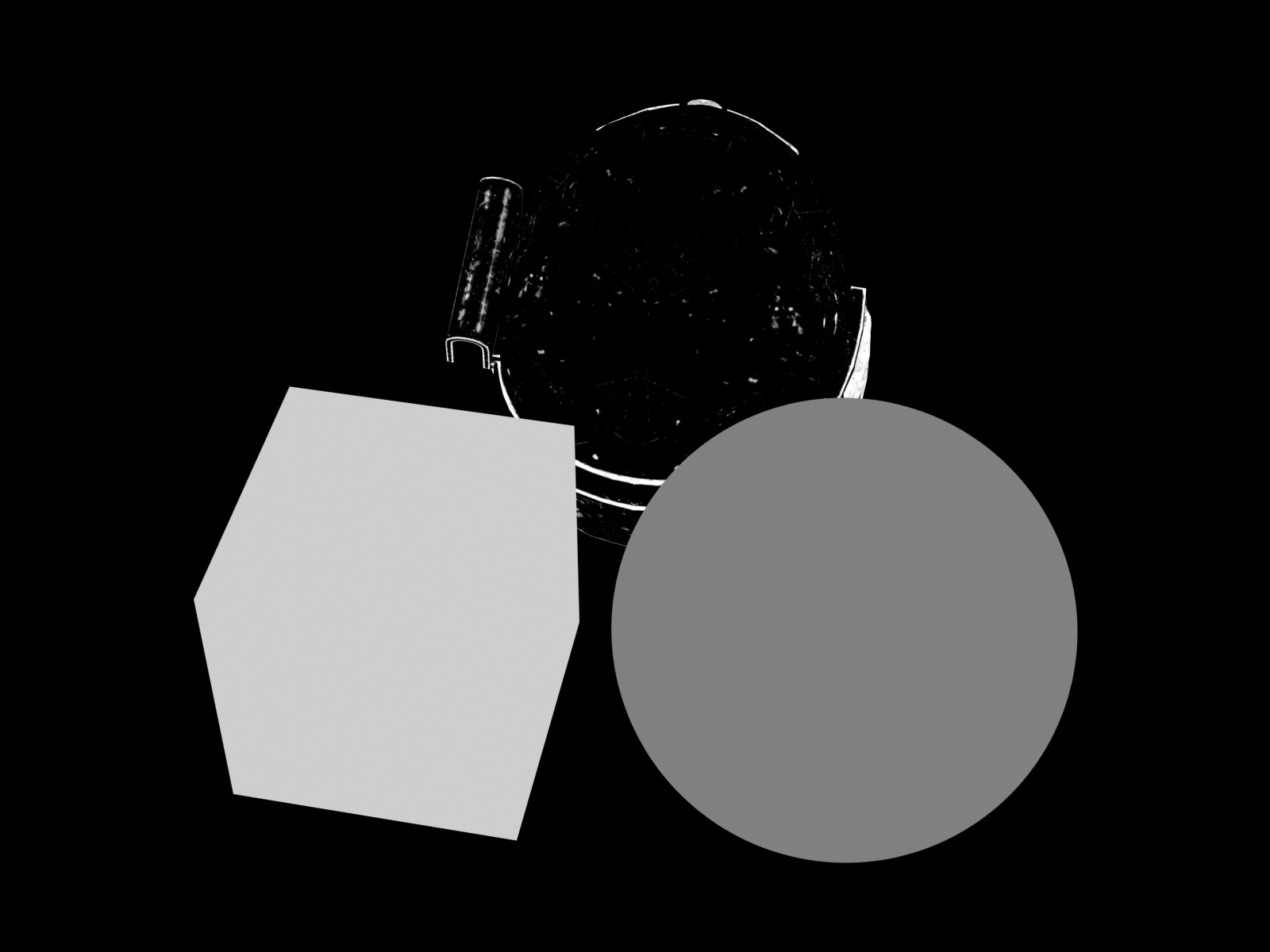}
                        & \includegraphics[width=0.2\linewidth]{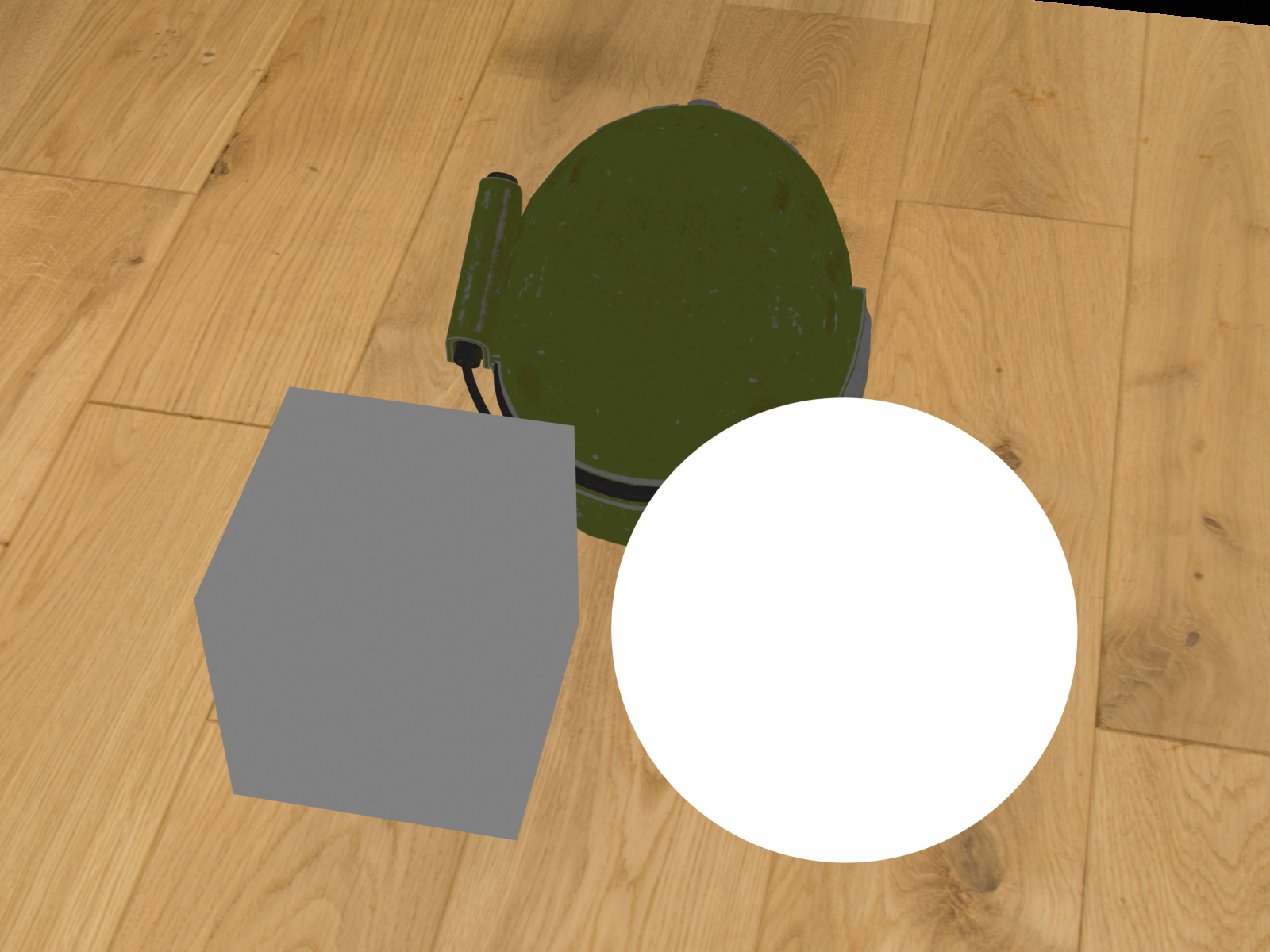} \\
            \end{tabular}
    }
    \vspace{-8pt}
    \caption{\textbf{Additional qualitative comparisons on the NeILF++ dataset~\cite{zhang2023neilf++}.} \dag: no ground-truth environment maps are provided with the dataset.}
    \label{fig:neilfpp_dataset_env_qualitative}
\end{figure*}

We present additional qualitative results for the \textit{Env} lighting subset of the NeILF++ dataset \cite{zhang2023neilf++} in Fig.~\ref{fig:neilfpp_dataset_env_qualitative} for the corresponding quantitative results in Tab.~\ref{tab:neilfpp_dataset_sota}.

Using our physics-based losses, our method improves both lighting and material estimation compared to the NeILF++ baseline.
These improvements are consistent with the trends observed in the more challenging \textit{Mix} subset shown in Fig.~\ref{fig:neilfpp_dataset_qualitative} of the main paper.
For the \textit{Env} subset, we significantly reduce artifacts in the upper halves of the estimated incident light fields and enhance albedo estimation in most scenes.

Notably, in the Studio scene, while some specular patchy effects persist on the helmet, cube, sphere, and floor, our method better captures finer details in the helmet’s metallicness and roughness, which are absent in the NeILF++ baseline.
Across the \textit{Env} subset, our roughness and metallicness predictions show considerable improvements, including the removal of shadowy patches on the cube and fringe artifacts on the sphere.

Although some limitations remain, such as missing fine-grained details on the helmet, our physics-based losses deliver consistent gains in material and lighting estimation, highlighting their effectiveness across diverse lighting conditions.

\section{Non-Energy Conserving BRDFs}
We extend the discussion in Section \ref{sec:pbr_losses} and provide additional intuition for when the Conservation of Energy property for BRDFs fails to hold. The Disney BRDF creates excess energy for unnormalized terms or when the Schlick Fresnel approximation is invalid, causing overly bright specular highlights~\cite{burley2015extending}.
Further, energy can be lost as the underlying microfacet BRDF models rough materials with \textit{single} reflection instead of multiple scattering on the microsurface.

\section{Additional Background Details}
\label{sec:supp:additional_background}
For completeness, we provide further implementation details on the NeILF++ implicit differential renderer (Sec.~\ref{sec:supplement_idr}), joint material-illumination-geometry optimization (Sec.~\ref{sec:supp:joint_optimization}), and hyperparameter sweeps (Sec.~\ref{sec:supp:hyperparam_sweep}).

\subsection{Implicit Differential Renderer (IDR)}
\label{sec:supplement_idr}
The second term of the specular BRDF $f_s$ in Equation \eqref{eq:specular_brdf} is the Fresnel term $F(\boldsymbol{\omega}_o, \boldsymbol{\omega}_h) \in \mathbb{R}^3$.
It models glossy reflection at glancing angles due to Fresnel reflection.
Following \cite{burley2012physically}, we use the Schlick Fresnel Approximation
\begin{equation}
    \label{eq:fresnel_approx}
    F(\boldsymbol{\omega}_o, \boldsymbol{\omega}_h; b, m) = F_0 + (1-F_0)(1 - \boldsymbol{\omega}_o \cdot \boldsymbol{\omega}_h)^5
\end{equation}
where
\begin{equation}
    \label{eq:fresnel_F0_approx}
    F_0 = 0.04(1-m) + bm
\end{equation}

The third term in Equation \eqref{eq:specular_brdf} is the geometry term $G(\boldsymbol{\omega}_i, \boldsymbol{\omega}_o, \mathbf{n}; r) \in \mathbb{R}$.
It models the masking and shadowing of microfacets depending on the incident and viewing direction.
\begin{equation}
    G(\boldsymbol{\omega}_i, \boldsymbol{\omega}_o, \mathbf{n}; r) = G_1(\boldsymbol{\omega}_i\cdot \mathbf{n}) G_1(\boldsymbol{\omega}_o\cdot \mathbf{n})
\end{equation}
where
\begin{equation}
    G_{\text{GGX}}(z;r) = \frac{2}{(2-r^2)z + r^2}
\end{equation}
as implemented in NeILF++~\cite{zhang2023neilf++}.

\subsection{Joint Optimization}
\label{sec:supp:joint_optimization}
We provide a more detailed description of the joint material-illumination-geometry optimization discussed in Sec.~\ref{sec:joint_mat_illum_geo_opt}.
We summarize all 3 PBR-NeRF training phases and their losses in Tab.~\ref{tab:pbrnerf_phase_losses}.
The final weightings used for each loss function term are listed in Tab.~\ref{tab:loss_term_weighting}.

\begin{table*}[!h]
    \centering
    \caption{\textbf{Summary of PBR-NeRF geometry, material, and joint training phases} with their respective losses. $\downarrow$: loss is downscaled from the previous stage. $\blacksquare$: the point cloud loss is optionally used to improve geometry SDF initialization. $\dag$: loss used when no ground-truth geometry is provided (e.g. DTU dataset). $\ast$: our novel PBR losses, namely (1) the Conservation of Energy Loss $\mathcal{L}_{\text{cons}}$ and (2) the NDF-weighted Specular Loss $\mathcal{L}_{\text{spec}}$.}
        \begin{tabular}{|l|clccc|ccccc|}
                \hline
    			\multirow{1}{*}{} &\multicolumn{5}{c|}{Geometry-based Losses} & \multicolumn{5}{c|}{Material-based Losses} \\
                     Optimization Phase& $\mathcal{L}_{\text{NeRF}}$&  $\mathcal{L}_{\text{pcd}}$ $\dag$&$\mathcal{L}_{\text{Eik}}$ $\dag$& $\mathcal{L}_{\text{Hess}}$ $\dag$& $\mathcal{L}_{\text{surf}}$ $\dag$& $\mathcal{L}_{\text{pbr}}$ & $\mathcal{L}_{\text{ref}}$ & $\mathcal{L}_{\text{smth}}$ & $\mathcal{L}_{\text{cons}}^{*}$& $\mathcal{L}_{\text{spec}}^{*}$\\ \hline
                    Geometry & \cmark &  $\blacksquare$ & \cmark & \cmark & \cmark & \xmark & \xmark & \xmark & \xmark & \xmark  \\
                    Material & \xmark &  \xmark & \xmark & \xmark & \xmark & \cmark & \cmark & \cmark & \cmark & \cmark \\
                    Joint & \cmark &  \xmark & \cmark & $\downarrow$ & $\downarrow$ & \cmark & \cmark & \cmark & \cmark & \cmark \\
                    \hline
    	\end{tabular}
    \label{tab:pbrnerf_phase_losses}
\end{table*}

\begin{table*}[!ht]
    \centering
        \caption{\textbf{PBR-NeRF loss function term weighting} for the NeILF++ and DTU datasets. \dag: the point cloud loss $\mathcal{L}_{\text{pcd}}$ is only used in the geometry phase for NeRF-SDF initialization. $\ast$: our novel PBR losses, namely (1) the Conservation of Energy Loss $\mathcal{L}_{\text{cons}}$ and (2) the NDF-weighted Specular Loss $\mathcal{L}_{\text{spec}}$.}
        \begin{tabular}{|l|ccccc|ccccc|}
            \hline
            \multirow{1}{*}{} &\multicolumn{5}{c|}{Geometry-based Losses} & \multicolumn{5}{c|}{Material-based Losses} \\
            & $\lambda_{\text{NeRF}}$ & $\lambda_\text{pcd}$ \dag & $\lambda_{\text{Eik}}$ & $\lambda_{\text{Hess}}$ & $\lambda_{\text{surf}}$ & $\lambda_{\text{pbr}}$ & $\lambda_{\text{ref}}$ & $\lambda_{\text{smth}}$ & $\lambda_{\text{cons}}^{*}$ & $\lambda_{\text{spec}}^{*}$ \\
            \hline
            NeILF++~\cite{zhang2023neilf++} & 1.0 & N/A & N/A & N/A & N/A & 1.0 & 0.1 & 0.0005 & 0.01 & 0.5 \\
            DTU~\cite{jensen2014dtu} & 1.0 & 0.1 & 0.1 & 0.001 & 0.01 & 1.0 & 0.1 & 0.0005 & 0.01 & 0.01 \\
            \hline
        \end{tabular}
        \label{tab:loss_term_weighting}
\end{table*}

\paragraph{Geometry phase.}
During the geometry phase, the NeRF SDF network is learned using the following loss.
\begin{multline}
    \label{eq:geo_loss}
    \mathcal{L}_\text{geo} = \lambda_\text{NeRF}\mathcal{L}_\text{NeRF} + \lambda_\text{pcd}\mathcal{L}_\text{pcd} + \lambda_\text{Eik}\mathcal{L}_\text{Eik} \\
    + \lambda_\text{Hess}\mathcal{L}_\text{Hess} + \lambda_\text{surf}\mathcal{L}_\text{surf}
\end{multline}

The NeRF rendering loss $\mathcal{L}_\text{NeRF}$ is identical to the original NeRF formulation.
The estimated RGB color $L_{o,\text{NeRF}}$ is rendered using the NeRF network and compared to the ground truth color $c$ using $\mathcal{L}_\text{NeRF}$ with a mean squared error.
\begin{equation}
    \label{eq:nerf_loss}
    \mathcal{L}_\text{NeRF} = ||L_{o,\text{NeRF}} - c||^2_2
\end{equation}

When ground truth geometry is provided, as in the NeILF++ dataset~\cite{zhang2023neilf++}, the estimated RGB color $L_{o,\text{NeRF}}$ is simply evaluated at the ground truth surface point using the viewing direction instead of using volume rendering.
Since we have the ground truth geometry, the other geometry priors are not needed and their corresponding losses are assigned a zero weight.
Therefore, only the NeRF rendering loss $\mathcal{L}_\text{NeRF}$ is used, with the only nonzero loss weight being  $\lambda_\text{NeRF}$.

When ground truth geometry is not provided, we use additional geometry priors encoded as loss function terms.

Following ~\cite{zhang2022critical,zhang2023neilf++}, we optimize the SDF with a point cloud loss only during the geometry phase.
The point cloud loss supervises the predicted SDF distances and normals:
\begin{equation}
    \mathcal{L}_{\text{pcd}} = |\mathbb{G}(\mathbf{x}_\text{pcd})| + \bigg(1 - \mathbf{n}_\text{pcd} \cdot \frac{\nabla_\textbf{x} \mathbb{G}(\mathbf{x}_\text{pcd})}{\| \nabla_\textbf{x} \mathbb{G}(\mathbf{x}_\text{pcd}) \|}\bigg).
\end{equation}
where $\mathbf{x}_\text{pcd}$ is a point in the point cloud, $\mathbb{G}(\mathbf{x}_\text{pcd})$ represents the SDF's predicted signed distance at $\mathbf{x}_\text{pcd}$, and $\mathbf{n}_\text{pcd}$ is the point cloud normal at $\mathbf{x}_\text{pcd}$.
This point cloud loss is only used to help initialize the SDF to a satisfactory quality and we perform full material-illumination-geometry optimization without point cloud input during the later joint phase.
This use of point clouds is similar to the use of point clouds in 3D Gaussian Splatting for initializing 3D Gaussians.

Note that the point cloud loss $\mathcal{L}_{\text{pcd}}$ is strictly optional, as indicated in Tab.~\ref{tab:pbrnerf_phase_losses}.
Following NeILF++~\cite{zhang2023neilf++}, we use the point cloud loss $\mathcal{L}_{\text{pcd}}$ on the DTU dataset to improve initial SDF quality and for a fair performance comparison.

We additionally use the Eikonal loss to penalize the SDF when the gradient at the surface point $\mathbb{G}(\mathbf{x})$ does not have a magnitude of 1.
\begin{equation}
    \mathcal{L}_\text{Eik} = \big| ||\nabla_\textbf{x} \mathbb{G}(\mathbf{x})|| - 1 \big|
\end{equation}

Furthermore, the Hessian loss penalizes rapidly changing gradient directions by minimizing the Hessian matrix norm.
\begin{equation}
    \mathcal{L}_\text{Hess} = || \mathbf{H} \mathbb{G}(\mathbf{x}) ||_1
\end{equation}

Finally, the minimal surface loss encourages compact interpolation and extrapolation of unobserved surfaces by minimizing the surface elastic energy.
\begin{equation}
    \mathcal{L}_\text{surf} = \delta_\epsilon (\mathbb{G}(\mathbf{x})))
\end{equation}
where $\delta_e$ is the regularized Dirac delta function parametrized by a sharpness $\epsilon$,
\begin{equation}
    \delta_\epsilon (z) = \frac{\epsilon \pi^{-1}}{\epsilon^2 + z^2}
\end{equation}

\paragraph{Material Phase.}
During the material phase, we train the NeILF and BRDF MLPs with the frozen NeRF SDF weights using the material phase loss:
\begin{multline}
    \label{eq:mat_loss_supp}
    \mathcal{L}_\text{mat} = \lambda_\text{pbr}\mathcal{L}_\text{pbr} + \lambda_\text{smth}\mathcal{L}_\text{smth} + \lambda_\text{ref}\mathcal{L}_\text{ref} \\
    + \lambda_\text{cons}\mathcal{L}_\text{cons} + \lambda_\text{spec}\mathcal{L}_\text{spec}
\end{multline}

The physically based rendering loss $\mathcal{L}_\text{pbr}$ supervises the estimated outgoing radiance from~\eqref{eq:neilfpp_rendering_eqn}:
\begin{equation}
    \mathcal{L}_\text{pbr} = || L_{o,\text{NeILF++}} - c||_2^2
\end{equation}

Following ~\cite{yao2022neilf,zhang2023neilf++}, we use a bilateral smoothness loss $\mathcal{L}_\text{smth}$ to encode the assumption that roughness $r$ and metallicness $m$ at surface point $x_p$ are smooth if the color of corresponding pixel $p$ has no sharp gradients:
\begin{equation}
    \mathcal{L}_\text{smth} = (||\nabla_\textbf{x} r(\mathbf{x}_p)|| + ||\nabla_\textbf{x} m(\mathbf{x}_p)||) \exp(-||\nabla_p I_p||)
\end{equation}
where $\nabla_p I_p$ is the image gradient at pixel $p$.

We also use the NeILF++ inter-reflection loss to use the NeRF SDF predicted outgoing radiance to supervise the incident light predicted by the NeILF MLP along the same ray $\boldsymbol{\omega}_i = -\boldsymbol{\omega}_o$ between two surface points $\mathbf{x}_1$ and $\mathbf{x}_2$.
\begin{equation}
    \label{eq:neilfpp_ref_loss}
    \mathcal{L}_\text{ref} = || L_{i, \text{NeILF}}(\mathbf{x}_2, \boldsymbol{\omega}_i) - L_{o, \text{NeRF}}(\mathbf{x}_1,-\boldsymbol{\omega}_i)||_1
\end{equation}

\paragraph{Joint Phase.}
During the joint optimization phase, the NeRF SDF, BRDF, and NeILF MLPs have been pre-trained by the previous geometry and material phases, allowing us to jointly optimize all networks simultaneously.
We reuse the geometry phase loss $\mathcal{L}_\text{geo}$ and material phase loss $\mathcal{L}_\text{mat}$ from \eqref{eq:geo_loss} and \eqref{eq:mat_loss_supp}, respectively, to obtain the overall joint phase loss
\begin{equation}
    \mathcal{L}_\text{joint} = \mathcal{L}_\text{geo} + \mathcal{L}_\text{mat}
\end{equation}
Note that the NeRF SDF is used twice per training sample during the Joint phase: (1) estimating $L_\text{o,NeRF}$  for $\mathcal{L}_\text{geo}$ and $\mathcal{L}_\text{ref}$; (2) sphere tracing to compute $\mathbf{x}$ and $\mathbf{n}$ for $L_\text{o,PBR}$.

\subsection{Hyperparameter Sweeps}
\label{sec:supp:hyperparam_sweep}
We now specify how we picked various hyperparameters introduced with our novel Conservation of Energy Loss~\eqref{eq:cons_loss} and NDF-weighted Specular Loss~\eqref{eq:spec_loss}.

The Conservation of Energy Loss $\mathcal{L}_\text{cons}$ does not introduce any new hyperparameters.
We increase the number of incident light directions from the default $|S_L|=128$ in NeILF++ to $|S_L|=256$ to better enforce energy conservation and separate the diffuse and specular lobes using more samples.

For the NDF-weighted Specular Loss, we use a temperature $T_\text{spec}=1$.
We include the softmax temperature $T_\text{spec}$ in our NDF-weighted Specular Loss definition~\eqref{eq:cons_loss} for full generality and to provide an option to control softmax sharpness.

To determine the loss term weightings for our novel physics-based losses, we perform a grid search with $\lambda_\text{cons} \in \{1.5, 1.0, 0.5, 0.1, 0.05, 0.01, 0.005\}$ and $\lambda_\text{spec} \in \{1.5, 1.0, 0.5, 0.1, 0.05, 0.01, 0.005\}$.

We also remove the Lambertian loss $\mathcal{L}_\text{Lam}$ used in NeILF++ by choosing $\lambda_\text{Lam}=0$ to remove the strong limitations on material estimation that the Lambertian assumption causes.

The remaining hyperparameters are identical to NeILF++~\cite{zhang2023neilf++}.

\end{document}